\documentclass{article}

\PassOptionsToPackage{numbers, compress}{natbib}

 \usepackage[eandd, final]{neurips_2026}

\usepackage[utf8]{inputenc} 

\usepackage[T1]{fontenc}    
\usepackage{hyperref}       
\usepackage{url}            
\usepackage{amsfonts}       
\usepackage{amsmath}        
\usepackage{amssymb}        
\usepackage{nicefrac}       
\usepackage{microtype}      
\usepackage{xcolor}         

\usepackage{subcaption}
\usepackage{booktabs}
\usepackage{makecell}   
\usepackage{enumitem}
\usepackage{multirow}  
\extrafloats{100}
\newcommand{\Description}[1]{}

\usepackage{pifont}
\newcommand{\cmark}{\ding{51}}%
\newcommand{\xmark}{\ding{55}}%

\usepackage{adjustbox}
\usepackage{array}
\usepackage{etoolbox}
\BeforeBeginEnvironment{tabular}{\begin{adjustbox}{max width=\textwidth}}
\AfterEndEnvironment{tabular}{\end{adjustbox}}
\newcolumntype{R}[2]{%
    >{\adjustbox{angle=#1,lap=\width-(#2)}\bgroup}%
    l%
    <{\egroup}%
}
\newcommand*\rot{\multicolumn{1}{R{45}{1em}}}
\AtBeginEnvironment{table}{\let\footnotesize\relax}
\AtBeginEnvironment{table*}{\let\footnotesize\relax}

\title{CausalCompass: Evaluating the Robustness of Time-Series Causal Discovery in Misspecified Scenarios}

\author{%
  \textbf{Huiyang Yi}$^{1}$,\ \ 
  \textbf{Xiaojian Shen}$^{2}$,\ \ 
  \textbf{Yonggang Wu}$^{1}$,\ \ 
  \textbf{Duxin Chen}$^{1,3,4}$%
  \thanks{Corresponding author.},\\
  \textbf{He Wang}$^{1}$\textbf{,}\ \   
  \textbf{Wenwu Yu}$^{1}$\\
  $^{1}$School of Mathematics, Southeast University\ \ 
  $^{2}$College of Software, Jilin University\\
  $^{3}$Jiangsu Provincial Scientific Research Center of Applied Mathematics\\
  $^{4}$State Key Laboratory of Engineering Materials for Major Infrastructure\\
  \texttt{yihuiyang@seu.edu.cn}\ \ 
  \texttt{shenxj22@mails.jlu.edu.cn}\\
  \texttt{wuyg@seu.edu.cn}\ \ 
  \texttt{chendx@seu.edu.cn}\\
  \texttt{wanghe91@seu.edu.cn}\ \ 
  \texttt{wwyu@seu.edu.cn}
}

\begin{document}

\maketitle

\begin{abstract}
Causal discovery from time series is a fundamental task in machine learning. However, its widespread adoption is hindered by a reliance on untestable causal assumptions and by the lack of robustness-oriented evaluation in existing benchmarks. To address these challenges, we propose CausalCompass, a flexible and extensible benchmark framework designed to assess the robustness of time-series causal discovery (TSCD) methods under violations of modeling assumptions. To demonstrate the practical utility of CausalCompass, we conduct extensive benchmarking of representative TSCD algorithms across eight assumption-violation scenarios. Our experimental results indicate that no single method consistently attains optimal performance across all settings. Nevertheless, in most scenarios, the methods exhibiting superior overall performance are deep learning-based approaches. We further provide hyperparameter sensitivity analyses to deepen the understanding of these findings. We additionally conduct ablation experiments to explain the strong performance of deep learning-based methods under assumption violations. We also find that NTS-NOTEARS relies on standardized preprocessing in practice, performing poorly in the vanilla setting but exhibiting strong performance after standardization. Finally, our work aims to provide a comprehensive and systematic evaluation of TSCD methods under assumption violations, thereby facilitating their broader adoption in real-world applications. The user-friendly implementation, documentation and datasets are available at \url{https://github.com/huiyang-yi/CausalCompass}.

\end{abstract}

\section{Introduction}
Many scientific applications of time-series analysis rely on understanding the underlying causal relationships~\cite{runge2023causal}. Owing to cost, risk, and ethical constraints, randomized controlled trials are frequently infeasible. Consequently, inferring causal relationships directly from purely observational time-series data, known as time-series causal discovery (TSCD), plays a critical role in addressing causal questions concerning intervention and counterfactual~\cite{peters2017elements,spirtes2001causation,pearl2009causality,pearl2018book}.

TSCD primarily encompasses constraint-based, noise-based, score-based, topology-based, Granger causality-based, and deep learning-based approaches, all of which typically rest on untestable causal assumptions~\cite{gong2024causal,peters2017elements}. Constraint-based methods (e.g., PCMCI\allowbreak~\cite{runge2019detecting}) recover causal graphs via conditional independence tests under the faithfulness assumption. Noise-based methods (e.g., VARLiNGAM\allowbreak~\cite{hyvarinen2010estimation}) impose specific assumptions on functional forms and noise distributions in structural equation models, exploiting asymmetries between causal and anticausal directions to identify causal structure. Score-based methods (e.g., DYNOTEARS~\cite{pamfil2020dynotears}) infer causal relationships by defining a scoring function to evaluate and rank candidate causal graphs. Topology-based methods (e.g., CCM~\cite{sugihara2012detecting} and TSCI~\cite{butler2024tangent}) leverage Takens’ state-space reconstruction theory~\cite{takens2006detecting} to infer causality. Granger causality (GC)-based methods (e.g., LGC~\cite{arnold2007temporal}) identify causal relationships by testing whether one time series improves the prediction of another, with classical GC being limited to linear settings~\cite{granger1969investigating}. Recently, given that the core principle of GC is highly compatible with neural networks (NNs), deep learning-based methods (e.g., cMLP~\cite{tank2021neural} and cLSTM~\cite{tank2021neural}) have been proposed, leveraging the expressive capacity of deep NNs~\cite{shen2026beatdance,zhang2026text,zhang2023generating,feng2026causal,he2024nhce,chen2025decoupling,chang2025anhp} for nonlinear Granger causal discovery~\cite{cheng2023cuts}.

Beyond the various assumptions of the aforementioned methods, causal sufficiency and no measurement error assumptions are also commonly required~\cite{peters2017elements,zhang2018causal}. In practice, real-world data rarely satisfy all such assumptions, and these assumptions are inherently difficult to test effectively, thereby constraining the application of TSCD in practical scenarios~\cite{peters2017elements}. Although certain studies have acknowledged the complexity of real-world data and developed causal discovery algorithms tailored to nonstationarity~\cite{huang2020causal,gao2025meta,ahmad2024regime,zhang2017causal2,huang2019causal,fujiwara2023causal,mameche2025spacetime,sadeghi2025causal}, latent confounders~\cite{gerhardus2020high,entner2010causal,nauta2019causal,chu2008search,huang2015identification,malinsky2018causal,xie2020generalized,cai2019triad}, mixed data~\cite{chen2025addressing,kleinberg2011logic,tsagris2018constraint,wenjuan2018mixed,zeng2022causal,cai2022causal,cui2016copula,raghu2018comparison}, measurement error~\cite{zhang2018causal,dai2022independence,yang2022causal,liu2022improving,yang2024causal,zhang2025conditional}, and missing data~\cite{cheng2023cuts,cheng2024cuts+,hyttinen2016causal,tu2019causal,gao2022missdag,zanga2023causal,wang2020causal,zanga2025federated,foraita2020causal}, when algorithms are applied to real-world data, the true data-generating mechanisms remain unknown, rendering these purpose-built algorithms still inadequately applicable. Consequently, the robustness of TSCD algorithms under violations of model assumptions is of paramount importance.

Prior research~\cite{loftus2024position} argued that the widespread application of causal models required a paradigm shift from scientific perfectionism toward scientific pragmatism, prioritizing practical utility. Building on this perspective, the work~\cite{poinsot2025position} further identified the key obstacle to transforming the research community's mindset lay in the lack of robustness assessment in current evaluation frameworks, noting that evaluations on datasets strictly adhering to all requisite assumptions severely overestimated method reliability, and underscoring that robustness evaluation was essential for the widespread adoption of causal models. 
Those studies~\cite{montagna2024assumption} and~\cite{yi2025robustness} benchmarked causal discovery algorithms in i.i.d. settings under violations of model assumptions, yet did not consider the ubiquitous time-series data in real-world applications and related TSCD works. 
Those works~\cite{glymour2019review} and~\cite{gong2024causal} respectively reviewed progress in causal discovery for i.i.d. and time-series settings, but lacked experimental support. 
Prior work~\cite{ferdous2025timegraph} benchmarked only constraint-based methods among TSCD approaches, with benchmark datasets relying solely on a limited number of manually specified structural equation models, offering no flexible parameter configurations and considering only a restricted range of assumption-violation scenarios. 
Recent work~\cite{stein2026tcdarena} benchmarked TSCD methods under varying degrees of assumption violations. However, the vanilla data considered only linear settings, precluding a fair evaluation of nonlinear methods, and it also neglected representative topology-based and deep learning-based approaches.

Our study conducts an extensive empirical evaluation of both classical and cutting-edge TSCD methods under violations of modeling assumptions. The vanilla data considered encompass both linear and nonlinear settings, ensuring a fair comparison across different approaches. We evaluate representative constraint-based, noise-based, score-based, topology-based, Granger causality-based, and deep learning-based methods, covering the broadest range of method categories considered in the existing literature and enabling a comprehensive assessment of TSCD. Notably, our work fills a critical gap in the literature by providing the first systematic evaluation of deep learning-based methods under diverse misspecified scenarios. Given their practical potential, assessing the robustness of deep learning-based approaches is of particular importance. Our contributions are summarized as follows:
\begin{itemize}[leftmargin=*]
    \item We conduct large-scale experiments on six major categories of mainstream TSCD algorithms across eight substantially different scenarios involving violations of modeling assumptions. In total, our evaluation comprises over 110,000 experimental runs across eleven distinct TSCD algorithms.

	\item Our results reveal that, overall, the best-performing methods in most scenarios are deep learning-based approaches. We further provide hyperparameter sensitivity analyses to deepen the understanding of method performance. We also conduct ablation experiments to explain the strong performance of deep learning-based methods under misspecified settings. In addition, we find that NTS-NOTEARS is sensitive to standardized preprocessing in practice. Given the robustness exhibited by deep learning-based methods under misspecified settings, conducting more in-depth investigations into these approaches holds significant value. 
    
    \item We introduce CausalCompass, released as a Python package with comprehensive documentation, which supports both linear and nonlinear vanilla models, encompasses eight misspecified scenarios, and provides unified interfaces for all eleven benchmark algorithms. The benchmark offers flexible parameter configurations and strong extensibility. Through this contribution, we aim to advance the understanding of the performance limits of TSCD algorithms, foster the development of more robust TSCD methods, and ultimately promote the broader adoption of TSCD in practice.
\end{itemize}

\section{Preliminaries}
Consider a $D$-dimensional time series denoted by $\mathbf{X} = \{\mathbf{x}_{1:T,i}\}_{i=1}^{D}$, where each sample vector 
$\mathbf{x}_t = (x_{t,1}, x_{t,2}, \ldots, x_{t,D})^\top$ 
corresponds to the observations of $D$ variables at time index 
$t \in \{1, \ldots, T\}$. Assume that the causal dependencies among these variables follow 
a structural causal model $\mathcal{M}$~\cite{pearl2009causality} of the form: 
\begin{equation}
x_{t,i} = f_i\!\big(Pa(x_{t,i}),\, u_{t,i}\big), 
\quad \forall i = 1, \ldots, D,
\end{equation}
where $Pa(x_{t,i})$ denote the parents of $x_{t,i}$, $f_i: \mathbb{R}^{\left|Pa(x_{t,i})\right|+1} \rightarrow \mathbb{R}$ is the causal structure function, and $u_{t,i}$ represents the independent noise variable. 

The task of TSCD is to recover the underlying temporal dependency structure among variables~\cite{peters2017elements}. In general, two types of causal graphs 
can be inferred: the summary causal graph $\mathcal{G}_{\mathrm{SCG}}$ and the window causal graph $\mathcal{G}_{\mathrm{WCG}}$~\cite{gong2024causal}. 

For the summary causal graph, each vertex corresponds to one temporal 
variable. Formally, the vertex set is 
$\{X_1, \ldots, X_D\}$, where 
$X_i = \{x_{t,i}\}_{t=1}^T$ denotes the temporal 
process of the $i$-th variable. An edge $X_p \!\rightarrow\! X_q$ is included in 
$\mathcal{G}_{\mathrm{SCG}}$ if and only if there exists at least one time index $t$ 
and a lag $\tau$ such that $x_{t-\tau,p}$ has a causal influence on $x_{t,q}$, 
where $\tau \ge 0$ is allowed for $p \neq q$ and $\tau > 0$ is required for $p = q$~\cite{assaad2022survey}.

For the window causal graph, we consider a fixed 
temporal window of size $\tau_{max}$. The vertex set contains all variable-time 
instances within this window, that is,
$\{x_{t,i}, x_{t+1,i}, \ldots, x_{t+\tau_{max},i}\}$ for each variable 
$i \in \{1,\ldots,D\}$. A directed 
edge $x_{t-l,p} \!\rightarrow\! x_{t,q}$ is present in 
$\mathcal{G}_{\mathrm{WCG}}$ if and only if $x_{t-l,p}$ has a causal influence on $x_{t,q}$, where $0 \le l \le \tau_{max}$ is allowed for $p \neq q$ and $0 < l \le \tau_{max}$ is required for $p = q$~\cite{assaad2022survey}.

\begin{table*}[t]
    \centering
    \caption{Summary of the assumptions associated with each algorithm and the types of causal graphs they are designed to recover. Each cell indicates whether an algorithm explicitly supports (\cmark) or does not support (\xmark) the corresponding condition. ``Sum.'' and ``Win.'' denote the summary causal graph and the window causal graph, respectively. The table format is adapted from~\cite{montagna2024assumption}.}
    \label{tab:methods_assumptions}
	\setlength{\tabcolsep}{6.5pt}
    \begin{tabular}{lccccccccccc}
        & \rot{VAR} & \rot{LGC} & \rot{VARLiNGAM} & \rot{PCMCI} & \rot{DYNOTEARS} & \rot{NTS-NOTEARS} & \rot{TSCI} & \rot{cMLP} & \rot{cLSTM} & \rot{CUTS} & \rot{CUTS+} \\
        \midrule
        Gaussian noise           & \cmark & \cmark & \xmark & \cmark & \cmark & \cmark & \xmark & \cmark & \cmark & \cmark & \cmark \\
        Linear mechanisms        & \cmark & \cmark & \cmark & \cmark & \cmark & \xmark & \xmark & \xmark & \xmark & \xmark & \xmark \\
        Nonlinear mechanisms     & \xmark & \xmark & \xmark & \cmark & \xmark & \cmark & \cmark & \cmark & \cmark & \cmark & \cmark \\
        Latent confounders       & \xmark & \xmark & \xmark & \xmark & \xmark & \xmark & \xmark & \xmark & \xmark & \xmark & \xmark \\
        Measurement error        & \xmark & \xmark & \xmark & \xmark & \xmark & \xmark & \xmark & \xmark & \xmark & \xmark & \xmark \\
        Nonstationary effects    & \xmark & \xmark & \xmark & \xmark & \xmark & \xmark & \xmark & \xmark & \xmark & \xmark & \xmark \\
        Missing mechanisms       & \xmark & \xmark & \xmark & \xmark & \xmark & \xmark & \xmark & \xmark & \xmark & \cmark & \cmark \\
        Mixed data effects       & \xmark & \xmark & \xmark & \xmark & \xmark & \xmark & \xmark & \xmark & \xmark & \xmark & \xmark \\
        Z-score standardization  & \xmark & \xmark & \xmark & \xmark & \xmark & \xmark & \xmark & \xmark & \xmark & \xmark & \xmark \\
        Min--max normalization   & \xmark & \xmark & \xmark & \xmark & \xmark & \xmark & \xmark & \xmark & \xmark & \xmark & \xmark \\
        Trend and seasonality    & \xmark & \xmark & \xmark & \xmark & \xmark & \xmark & \xmark & \xmark & \xmark & \xmark & \xmark \\
        \midrule
        Output & Sum. & Sum. & Win. & Win. & Win. & Win. & Sum. & Win. & Sum. & Win. & Sum. \\
        \bottomrule
    \end{tabular}
\end{table*}

\section{Experimental design}
\label{sec:experimental}
In this section, we introduce the vanilla model, the assumption-violation scenarios, the evaluated causal discovery methods, and the evaluation metrics.

\subsection{Vanilla model}
\label{sec:vanilla model}
Most causal discovery methods rely on unverifiable assumptions, and this work examines their performance under assumption violations. We therefore begin by introducing the vanilla models for both linear and nonlinear settings, followed by the various assumption-violation scenarios.

\textbf{Linear vanilla model.} In linear settings, we consider a vector autoregressive (VAR) model, following the settings of~\cite{tank2021neural}:
\begin{equation}
\mathbf{x}_t=\sum_{l=1}^{\tau_{max}} \boldsymbol{A}_l \mathbf{x}_{t-l}+\mathbf{u}_t,
\label{eq:var}
\end{equation}
where $\boldsymbol{A}_l$ denotes the sparse coefficient matrix for time lag $l$, $\tau_{max}$ is the maximum lag, and $\mathbf{u}_t = (u_{t,1}, u_{t,2}, \ldots, u_{t,D})^\top$ is a jointly independent zero-mean Gaussian noise vector with covariance matrix $\Omega = \mathrm{cov}(\mathbf{u}_t) = \mathrm{diag}(\sigma_1^2, \ldots, \sigma_D^2)$. We refer to this setting as the linear vanilla model, since it aligns with the assumptions adopted by the majority of linear benchmark methods (see Table~\ref{tab:methods_assumptions}).

\textbf{Nonlinear vanilla model.} In nonlinear settings, we consider the Lorenz-96 model, following the settings of~\cite{tank2021neural}: 
\begin{equation}
\frac{\partial x_{t, i}}{\partial t}=-x_{t, i-1}\left(x_{t, i-2}-x_{t, i+1}\right)-x_{t, i}+F, \quad  1 \leq i \leq D
\label{eq:lorenz}
\end{equation}
where $x_{t, 0}=x_{t, D}$, $x_{t, -1}=x_{t, D-1}$, $x_{t, D+1}=x_{t, 1}$, and $F$ denotes the magnitude of the external forcing (a larger $F$ corresponds to a more chaotic system~\cite{karimi2010extensive}).
We refer to this setting as the nonlinear vanilla model, since it aligns with the assumptions adopted by the majority of nonlinear benchmark methods (see Table~\ref{tab:methods_assumptions}).

To eliminate the potential influence of Gaussian noise in the vanilla model on the experimental results, we also consider the vanilla model with non-Gaussian noise (see Appendix~\ref{app_non_gaussian}).

\subsection{Model assumption violation scenarios}\label{sec:misspecified}

We define eight distinct assumption-violation scenarios, each of which can be applied to both the linear vanilla and nonlinear vanilla model to generate datasets that exhibit specific assumption violations.

\textbf{Measurement error model.} 
Causal discovery algorithms commonly assume that variables are observed without measurement error~\cite{zhang2018causal,dai2022independence,scheines2017measurement}. However, measurement noise is almost inevitable during real-world data acquisition. 
To account for this effect, we define the observation mechanism as
\begin{equation}
\hat{X}_i = X_i + \epsilon_i, \quad \forall\, i = 1, \ldots, D,
\end{equation}
where $X_i = \{x_{t,i}\}_{t=1}^T$ is generated from the linear vanilla model~\eqref{eq:var} or nonlinear vanilla model \eqref{eq:lorenz}. $\epsilon_i$ denotes a zero-mean Gaussian noise term independent of $X_i$, with variance given by $\mathrm{Var}(\epsilon_i) = \alpha \cdot \mathrm{Var}(X_i)$ for $\alpha \in \{0.4, 0.6, 0.8, 1.0, 1.2\}$.

\textbf{Nonstationary model.}
Most TSCD methods typically rest on the assumption of stationarity, namely that the underlying structural causal model remains invariant over time~\cite{gao2025meta,ahmad2024regime,huang2020causal,kang2024identifying,kang2026spatio}. However, real-world data often exhibit pronounced nonstationary behavior. Following the setting in~\cite{huang2020causal}, the form of nonstationarity we consider refers to scenarios in which the underlying causal generative process remains fixed, while the variance of the noise terms evolves smoothly over time. A detailed description of the nonstationary data-generation mechanism is provided in Appendix~\ref{app_assumption}. In Appendix~\ref{app_non_coeff}, we additionally investigate a nonstationary setting where the coefficients of the linear vanilla model are allowed to vary over time. The main experimental conclusions presented in Section~\ref{sec:results} remain consistent regardless of the form of nonstationarity considered.

\textbf{Latent confounders model.} 
The causal sufficiency assumption ensures that no unobserved confounders exist between any pair of observed variables~\cite{peters2017elements}. However, in real-world applications, it is often impossible to guarantee that all relevant factors have been measured~\cite{montagna2024assumption}. Let $\mathbf{L}=\{L_i\}_{i=1}^{D} $ denote a collection of latent confounding variables. For each pair of observed nodes, we sample a Bernoulli random variable $C \sim \mathrm{Bernoulli}(\zeta)$. Whenever $C = 1$, a variable from $\mathbf{L}$ is randomly selected to serve as a latent common cause. In the linear setting, we introduce cross-lag confounders~\cite{malinsky2018causal}, where the latent confounders $\mathbf{L}$ follow a VAR process. In the nonlinear setting, we introduce contemporaneous confounders~\cite{malinsky2018causal}, where $\mathbf{L}$ evolves according to the Lorenz-96 mechanism. The parameter $\zeta \in \{0.1, 0.3, 0.5, 0.7, 0.9\}$ controls the number of confounding links present in the causal graph.

\textbf{Z-score standardization model.}
The work~\cite{reisach2021beware} observed that in i.i.d. causal discovery settings, methods based on continuous optimization, such as NOTEARS~\cite{zheng2018dags} and GOLEM~\cite{ng2020role}, exhibit a substantial performance drop when applied to standardized data. This behavior is attributed to the fact that these methods exploit high variance-sortability in the data, which is removed by standardization~\cite{reisach2021beware}. As argued in~\cite{reisach2021beware}, standardization can therefore serve as a diagnostic tool to assess whether such methods genuinely recover causal structure, rather than relying on variance-sortability shortcuts. In i.i.d. data, standardization is typically performed across samples, whereas in time-series data it is applied across the temporal dimension, and these two procedures are inherently different~\cite{gao2025meta}. Consequently, the conclusions of~\cite{reisach2021beware} cannot be directly extrapolated to time-series data. To date, no studies have explicitly pointed out how standardization influences TSCD, even though many TSCD methods (NTS-NOTEARS~\cite{sun2021nts}, CUTS~\cite{cheng2023cuts}, CUTS+~\cite{cheng2024cuts+}) in practice directly adopt standardized data and report only the results under such preprocessing. This gap highlights the importance of understanding the role of standardization in TSCD. A detailed description of the data-generation mechanism is provided in Appendix~\ref{app_assumption}. 

\textbf{Mixed data model.}
Most causal discovery methods assume that time-series data consist solely of continuous-valued variables~\cite{chen2025addressing}. However, in many real-world scenarios, the collected data typically contain both discrete and continuous variables, a setting commonly referred to as mixed data~\cite{chen2025addressing,tsagris2018constraint,kleinberg2011logic,wenjuan2018mixed,zeng2022causal,cui2016copula}. Thus, ensuring that algorithms maintain robustness in the presence of mixed data is of considerable importance. 
We follow the procedure in~\cite{chen2025addressing} to generate mixed data by applying min--max normalization to the continuous-valued data produced by~\eqref{eq:var} and~\eqref{eq:lorenz}, thereby mapping them into the range $[0,1]$ as $\tilde{X}_i = \mathrm{min\text{--}max}(X_i)$. We then randomly select $D \times \beta$ variables for discretization, where $\beta \in \{0.1, 0.3, 0.5, 0.7, 0.9\}$. The discrete variables are generated as follows:
\begin{equation}
x_{t,i}^{\mathrm{DV}}= \begin{cases}1 & \text { if } \tilde{x}_{t,i}>0.5 \\ 0 & \text { if } \tilde{x}_{t,i} \leq 0.5\end{cases}
\end{equation}
Here, $x_{t,i}^{\mathrm{DV}}$ denotes the value of the $i$-th variable at time $t$ after discretization. 

\textbf{Min--max normalization model.} 
In TSCD, it remains unclear which normalization strategy is appropriate~\cite{gao2025meta}. Nevertheless, the existing work~\cite{chen2025addressing} adopts min--max normalization when generating mixed data, which motivates the need to evaluate the robustness of causal discovery methods under normalization settings. A detailed description of min--max normalization model is provided in Appendix~\ref{app_assumption}. It is worth noting that, although Z-score standardization and min--max normalization do not explicitly violate the causal assumptions of certain methods, we nonetheless include them as special scenarios in this work given their impact on practice.

\textbf{Missing model.} 
Missing data are pervasive in real-world time-series applications, and performing causal inference directly in the presence of such gaps can lead to substantial estimation errors~\cite{cheng2023cuts,cheng2024cuts+,runge2018causal,hyttinen2016causal}. It is therefore crucial that causal discovery algorithms be capable of handling incomplete observations. In this work, we adopt the Missing Completely At Random (MCAR) mechanism~\cite{tu2019causal}, where each entry is independently removed according to a Bernoulli distribution with probability~$\gamma \in \{0.1, 0.2, 0.3, 0.4, 0.5\}$. Since most benchmark methods considered in our study cannot operate directly on datasets containing missing values, we impute the missing entries using zero-order hold (ZOH) interpolation~\cite{cheng2023cuts,cheng2024cuts+} before applying causal discovery. To rule out the possibility that the choice of imputation strategy influences our findings, we further evaluate all benchmark methods under Gaussian process interpolation and linear interpolation in Appendix~\ref{app_imputation}. The results demonstrate that the choice of imputation strategy does not affect the main experimental conclusions in Section~\ref{sec:results}.

\textbf{Trend and seasonality model.}
Real-world time series frequently exhibit deterministic trends and periodic seasonal fluctuations~\cite{cleveland6seasonal,chatfield2019analysis}. 
Following the setup in~\cite{ferdous2025timegraph}, we augment the vanilla data-generating process by 
adding a trend component together with a seasonal component. We provide a detailed description of the data-generation mechanism in Appendix~\ref{app_assumption}.

\subsection{Data generation}\label{sec:data generation}

We generate datasets for each scenario using the number of nodes $D \in \{10, 15\}$, time-series lengths $T \in \{500, 1000\}$, and forcing constants $F \in \{10, 40\}$. For each experimental configuration and modeling setting, we further sample data under $5$ different random seeds to ensure statistical reliability. In our benchmark, the severity of model assumption violations is controlled through dedicated data-generation parameters, and the detailed selection procedure is described in Appendix~\ref{app_violation_severity}. Minimal Responsible AI metadata for the released benchmark datasets are provided in the accompanying Croissant file.

\subsection{Methods}
\label{sec:methods}

We evaluate $11$ well-established TSCD algorithms, covering Granger causality-based 
(VAR~\cite{assaad2022survey}\allowbreak,
LGC~\cite{arnold2007temporal}), 
constraint-based (PCMCI~\cite{runge2019detecting}\allowbreak),
noise-based (VARLiNGAM~\cite{hyvarinen2010estimation}\allowbreak),
score-based (DYNO\-TEARS~\cite{pamfil2020dynotears}\allowbreak, 
NTS\allowbreak-NOTEARS~\cite{sun2021nts}), 
topology-based (TSCI~\cite{butler2024tangent}\allowbreak), 
and deep learning-based (cMLP~\cite{tank2021neural}\allowbreak, 
cLSTM~\cite{tank2021neural}\allowbreak, 
CUTS~\cite{cheng2023cuts}\allowbreak, 
CUTS+~\cite{cheng2024cuts+}) approaches. To the best of our knowledge, the breadth of benchmark methods considered in this study constitutes one of the most comprehensive collections of methods in current robustness benchmarking research~\cite{ferdous2025timegraph}. For a more detailed
introduction to the various methods, see the Appendix~\ref{app_methods}. A detailed description of the hyperparameter search ranges is provided in Appendix~\ref{app_hyper_settings}.

\subsection{Evaluation metrics}
\label{sec:evaluation metrics}
Following the settings in~\cite{herdeanu2025causaldynamics,chen2025addressing,marcinkevivcs2021interpretable}, we evaluate performance using areas under receiver operating characteristic (AUROC~\cite{bradley1997use}) and precision–recall (AUPRC~\cite{davis2006relationship}) curves. Consistent with prior studies~\cite{marcinkevivcs2021interpretable,stein2025causalrivers}, our evaluation considers only the off-diagonal entries of the adjacency matrix, excluding self-causal links that are typically the easiest to infer. We report results under the configuration achieving the best AUPRC. Higher AUROC and AUPRC values indicate more accurate recovery of the target causal graph. We additionally report the normalized Structural Hamming Distance (NSHD) as an evaluation metric. A lower NSHD indicates more accurate recovery of the true causal graph. A detailed definition of NSHD is provided in Appendix~\ref{app:nshd}.

\section{Critical experimental results and insights}
\label{sec:results}
\begin{figure*}[t]
     \centering
     \begin{subfigure}[b]{0.49\textwidth}
         \centering
        \includegraphics[width=\textwidth]{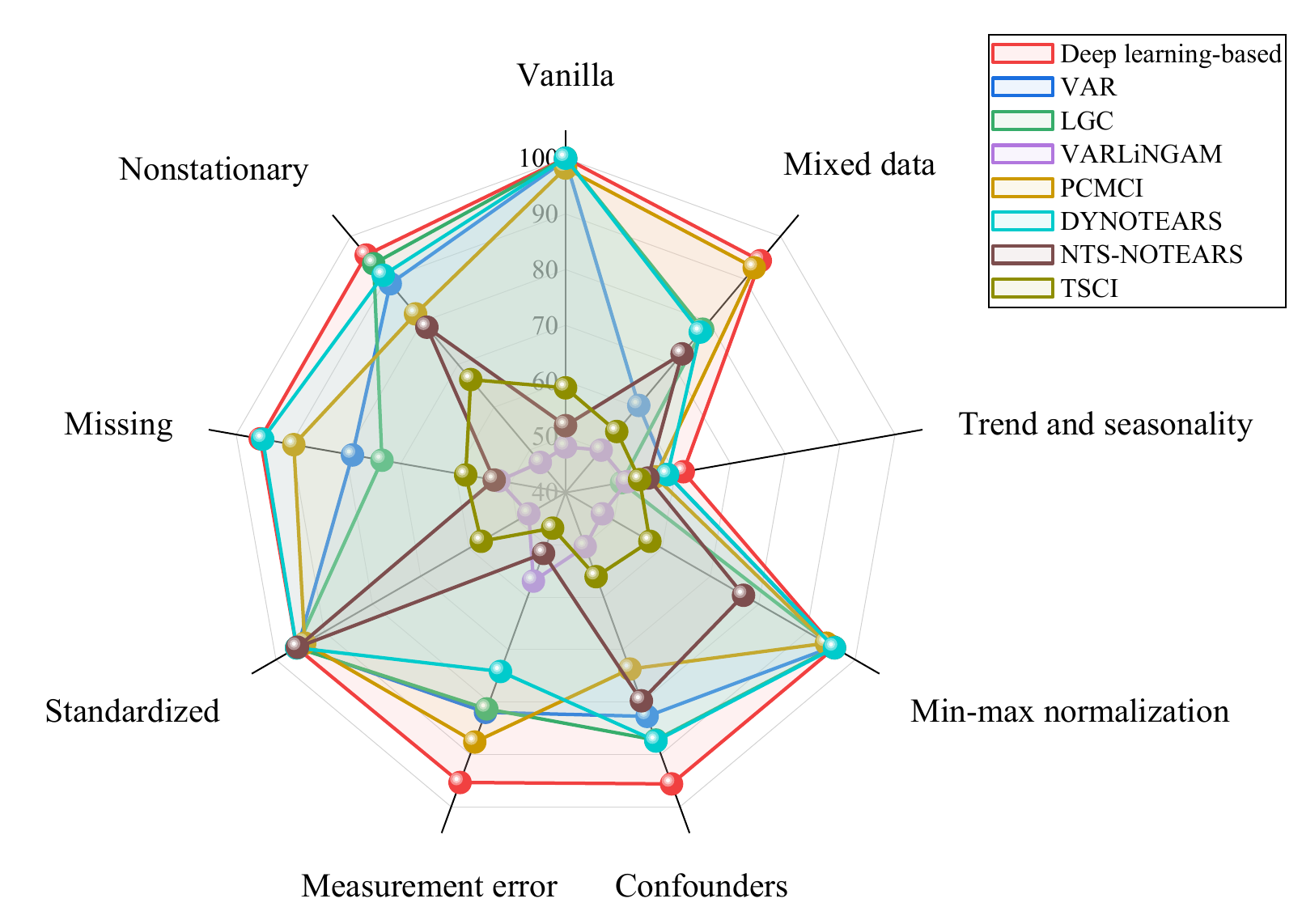}
         \caption{AUROC for linear 10-node case with $T = 1000$.}
         \label{fig:linear_10_1000_auroc}
     \end{subfigure}%
     \hfill  
     \begin{subfigure}[b]{0.49\textwidth}
         \centering
         \includegraphics[width=\textwidth]{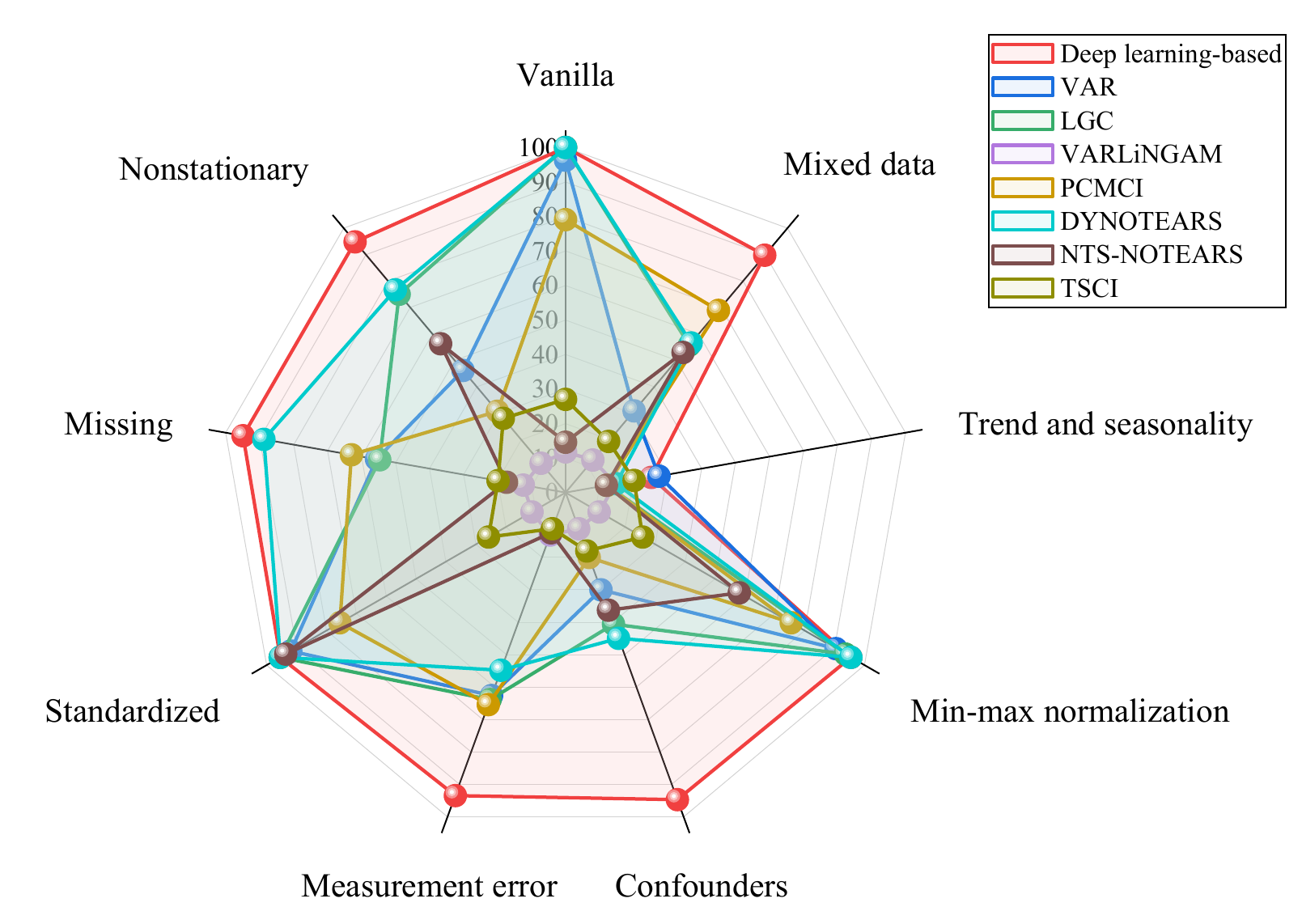}
         \caption{AUPRC for linear 10-node case with $T = 1000$.}
         \label{fig:linear_10_1000_auprc}
     \end{subfigure}

     \medskip  

     \begin{subfigure}[b]{0.49\textwidth}
         \centering
        \includegraphics[width=\textwidth]{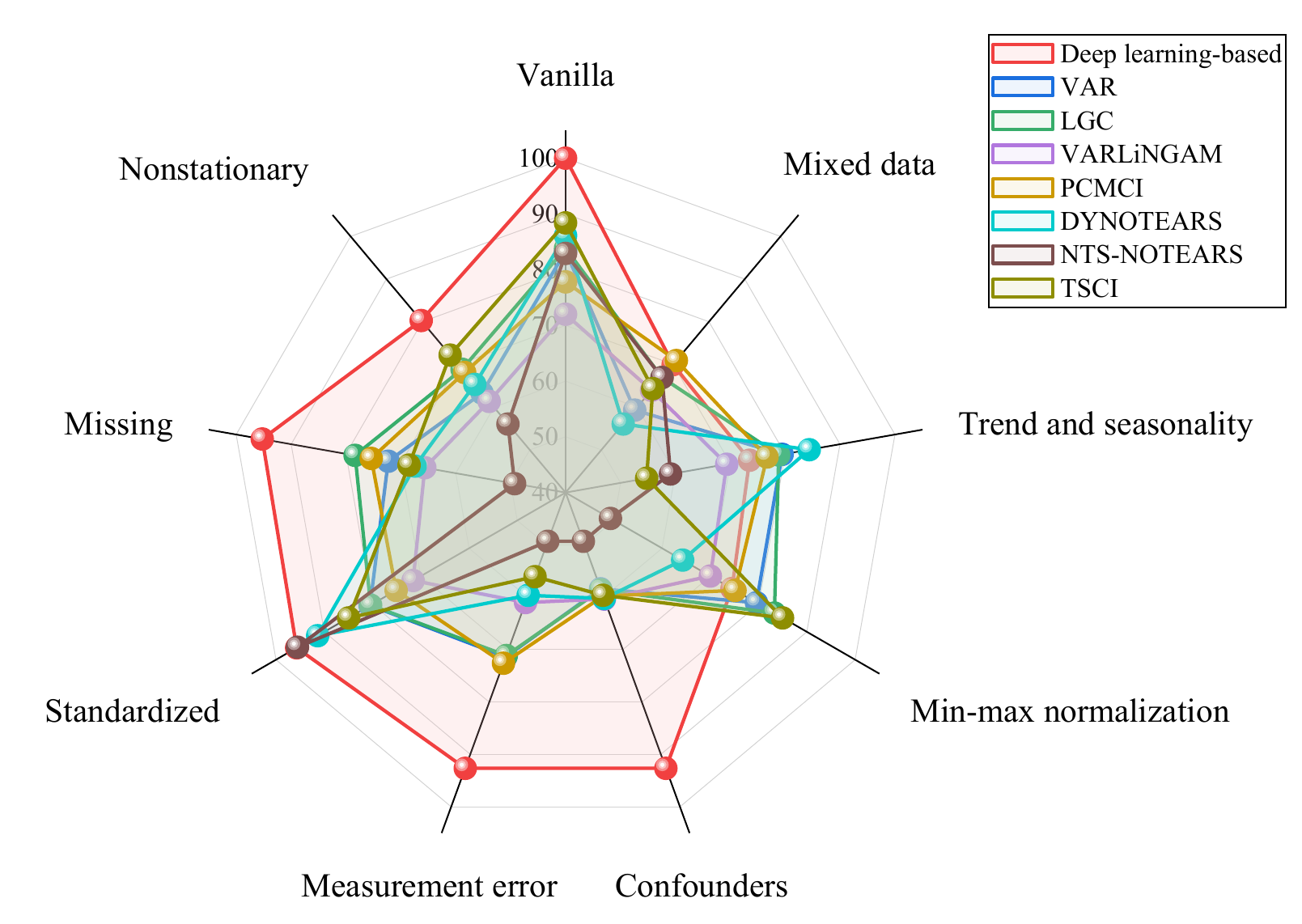}
        \caption{AUROC for nonlinear 10-node case with $T = 1000$ and $F = 10$.}
         \label{fig:nonlinear_10_1000_f10_auroc}
     \end{subfigure}%
     \hfill
     \begin{subfigure}[b]{0.49\textwidth}
         \centering
         \includegraphics[width=\textwidth]{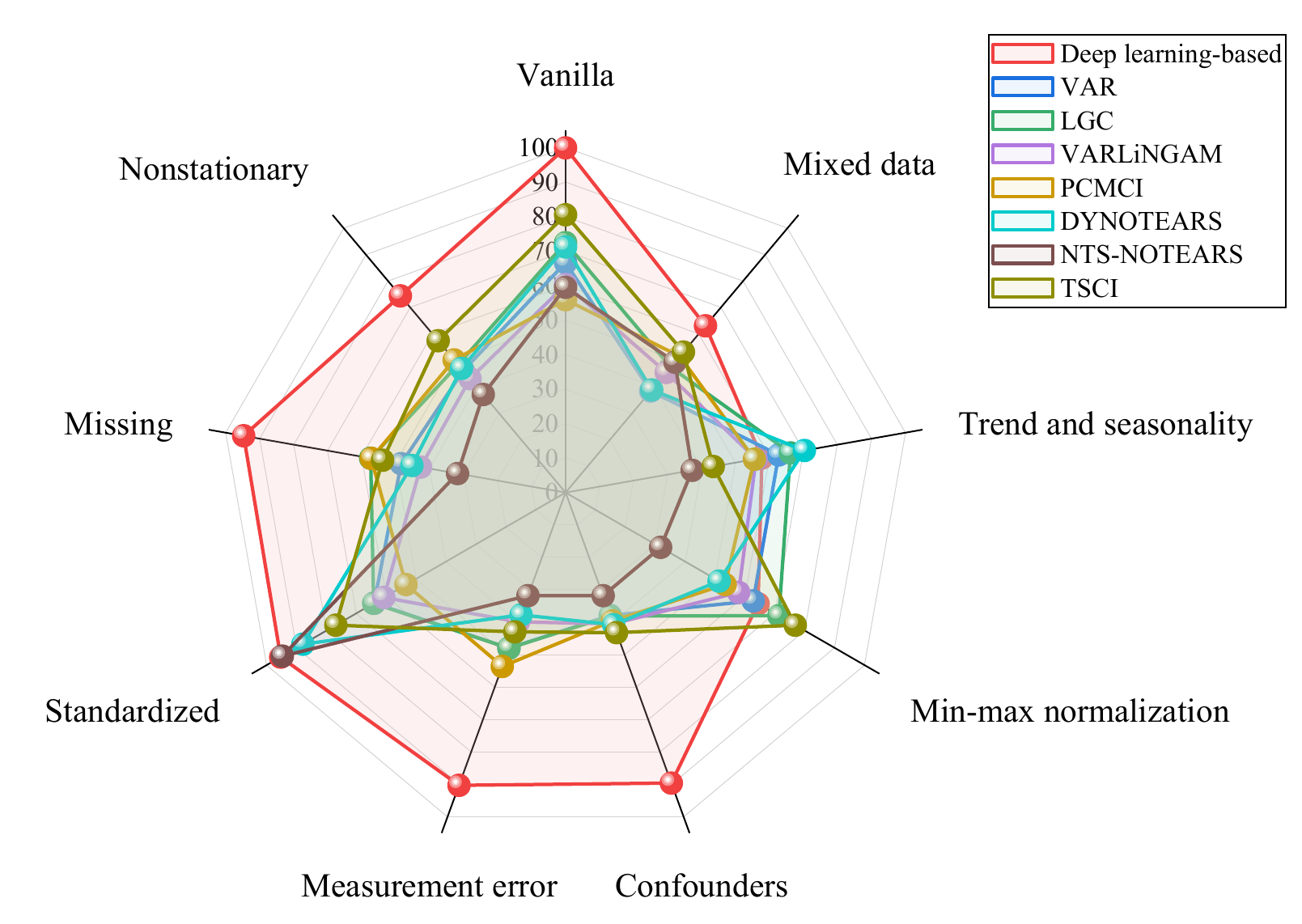}
         \caption{AUPRC for nonlinear 10-node case with $T = 1000$ and $F = 10$.}
         \label{fig:nonlinear_10_1000_f10_auprc}
     \end{subfigure}

     \medskip  

     \begin{subfigure}[b]{0.49\textwidth}
         \centering
        \includegraphics[width=\textwidth]{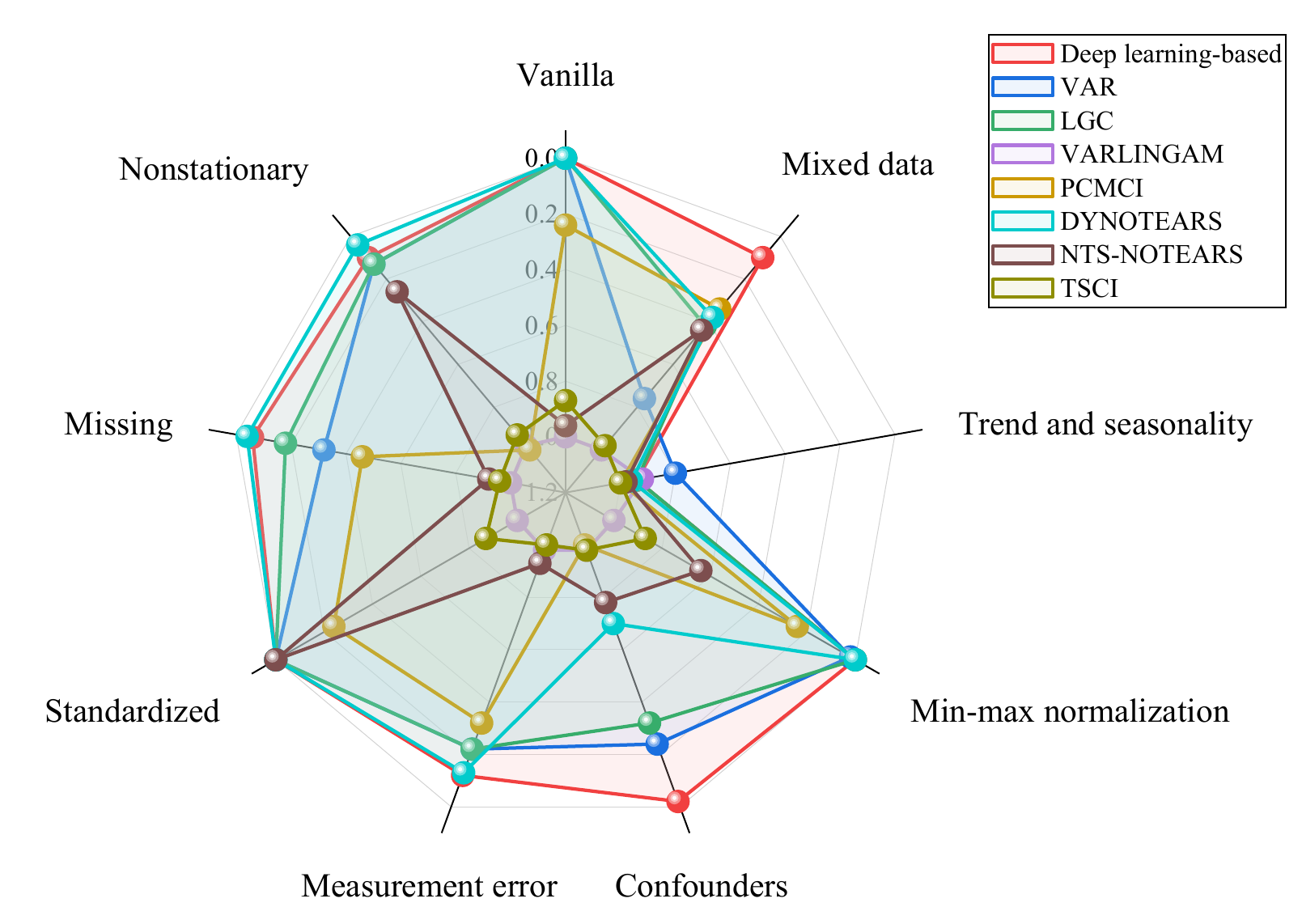}
        \caption{NSHD for linear 10-node case with $T = 1000$.}
         \label{fig:linear_10_1000_nshd}
     \end{subfigure}%
     \hfill
     \begin{subfigure}[b]{0.49\textwidth}
         \centering
         \includegraphics[width=\textwidth]{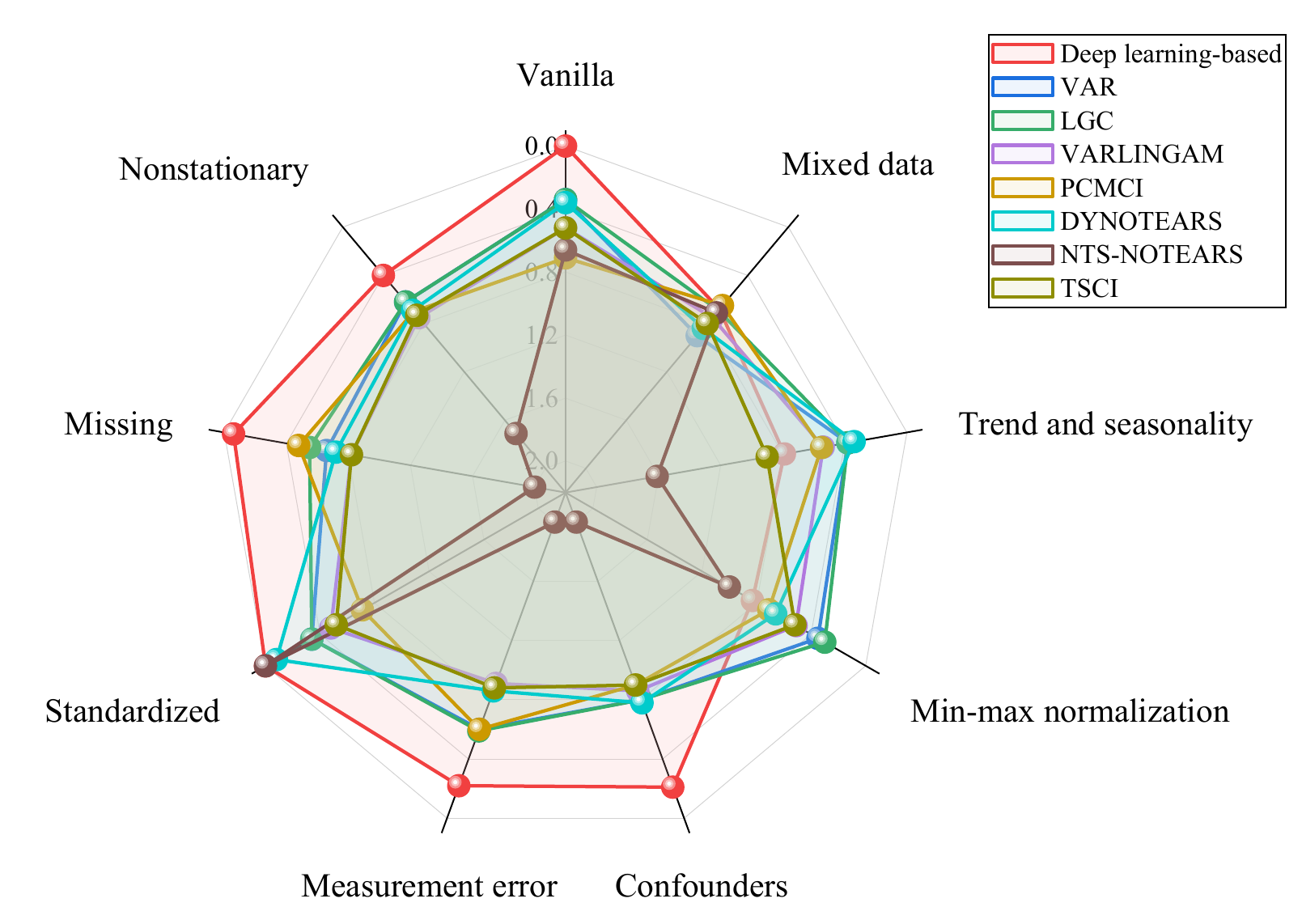}
         \caption{NSHD for nonlinear 10-node case with $F = 10$.}
         \label{fig:nonlinear_10_1000_f10_nshd}
     \end{subfigure}

\Description{Four radar charts comparing causal discovery methods on 10-node networks with T=1000 across 9 scenarios. Superior performance is predominantly achieved by deep learning-based approaches.}
\caption{Experimental results under the linear and nonlinear settings across the vanilla scenario and eight assumption violation scenarios. AUROC, AUPRC and NSHD are evaluated over 5 trials for the 10-node case with $T = 1000$. For the deep learning-based methods, we present only the optimal results.}
\label{fig:experiments_10_1000_f10}
\end{figure*}

In this section, we first present the experimental results under model assumption violations introduced in Section~\ref{sec:misspecified} and draw conclusions by comparing them with the results obtained in the vanilla scenario. We further provide sensitivity analyses of various methods with respect to hyperparameters (Section~\ref{sec:sensitivity}) to ensure a comprehensive understanding of their performance characteristics. We also conduct ablation experiments (Section~\ref{sec:ablation}) to explain the strong performance of deep learning-based methods under assumption violations. 
Due to space constraints, the main text primarily discusses the linear 10-node case with $T=1000$ and the nonlinear 10-node case with $T=1000$ and $F=10$ (Table~\ref{tab:linear-10-1000-1},~\ref{tab:linear-10-1000-2},~\ref{tab:nshd-linear-10-1000-1},~\ref{tab:nshd-linear-10-1000-2},~\ref{tab:nonlinear-10-1000-f10-1},~\ref{tab:nonlinear-10-1000-f10-2},~\ref{tab:nshd-nonlinear-10-1000-f10-1},~\ref{tab:nshd-nonlinear-10-1000-f10-2}), while similar conclusions hold for different numbers of nodes, time series lengths, and external forcing intensities (see Appendix~\ref{app_table_results_best} and~\ref{app_fig_results_best}). For a concise and intuitive presentation of the results, Figure~\ref{fig:experiments_10_1000_f10} summarizes the outcomes for these two settings. For each scenario, we generate datasets using $5$ different random seeds and report the mean and standard deviation of evaluation metrics across these $5$ trials. To ensure fair comparison among methods, we determine each method's hyperparameters as optimal values relative to the specific dataset. Recognizing that optimal hyperparameters are typically unknown in practical applications~\cite{machlanski2024robustness}, we additionally select for each causal discovery method a single hyperparameter configuration that maximizes average performance across all scenarios (see Appendix~\ref{app_table_results_avg_scen} and~\ref{app_fig_results_avg_scen}). Since the principal conclusions drawn from these two hyperparameter selection strategies are essentially identical, we discuss only the results obtained under optimal hyperparameters in the main text. We also compare the computational resources of the benchmark methods (see Appendix~\ref{app_computational_cost}).

\subsection{Current methods' performance in misspecified scenarios}
Our experiments demonstrate that no single method achieves optimal performance across all scenarios. Nevertheless, in most scenarios, the methods exhibiting superior overall performance are deep learning-based approaches.
Our paper does not claim that every deep learning-based method outperforms every method of another type. Instead, we compare the best-performing methods across different method types. 
In this work, robustness refers to a model’s ability to recover the true causal graph under violations of modeling assumptions, consistent with the interpretations in~\cite{montagna2024assumption},~\cite{stein2026tcdarena} and~\cite{yi2025robustness}. We quantify the degree of robustness through the values of the evaluation metrics.

\textbf{Latent confounders, measurement error, missing, mixed data, and nonstationary models.}
Figure~\ref{fig:experiments_10_1000_f10} shows that under confounded ($\zeta=0.5$), measurement error ($\alpha=1.2$), mixed data ($\beta=0.5$), missing ($\gamma=0.4$), and nonstationary scenarios (linear: $m=1$, $\nu=1$; nonlinear with $F=10$: $m=2.5$, $\nu=2.0$; nonlinear with $F=40$: $m=3.5$, $\nu=2.0$), the best-performing methods are predominantly deep learning-based approaches. The severity of model assumption violations can be flexibly controlled through the corresponding parameters. In the main text, we report results under a representative parameter setting, and we additionally consider experiments under varying levels of measurement error (see Appendix~\ref{app_mea}), nonstationarity (see Appendix~\ref{app_non}), and confounding (see Appendix~\ref{app_confounder}). The results indicate that, overall, method performance generally degrades as the severity of model assumption violations increases, while the best-performing methods are predominantly deep learning-based approaches.

\textbf{Z-score standardization model.}
In the linear settings of Table~\ref{tab:linear-10-1000-2} and the nonlinear settings of Table~\ref{tab:nonlinear-10-1000-f10-2}, we observe that NTS-NOTEARS and CUTS exhibit notable performance improvements under standardization. In particular, although NTS-NOTEARS performs poorly in the vanilla setting, its performance improves substantially once standardization is applied. 
DYNOTEARS shows improved performance in the nonlinear settings reported in Table~\ref{tab:nonlinear-10-1000-f10-2}, while its performance remains unchanged in the linear $10$-node case with $T=1000$ and degrades in the remaining linear settings. In the nonlinear $F=40$ scenario of Table~\ref{tab:nonlinear-10-1000-f40-2}, among the deep learning-based methods, cMLP, CUTS, and CUTS+ all exhibit performance improvements, while cLSTM shows performance degradation. Apart from these cases, the performance of most methods remains largely unchanged under standardization. It is particularly noteworthy that continuous optimization methods in the time-series settings (DYNOTEARS, NTS-NOTEARS) do not consistently exhibit the performance degradation observed for continuous optimization methods in the i.i.d. settings. Instead, they may even exhibit performance improvements. This observation helps explain why the original implementation of NTS-NOTEARS adopts standardized preprocessing, and our results further highlight its reliance on standardization. Nevertheless, not all time-series data are standardized in real-world applications, making it crucial to understand the impact of standardization on continuous optimization methods.

\textbf{Min--max normalization model.}
In the linear settings of Table~\ref{tab:linear-10-1000-1}, NTS-NOTEARS exhibits performance improvements under min--max normalization, whereas both CUTS and CUTS+ show pronounced performance degradation. In the nonlinear settings of Table~\ref{tab:nonlinear-10-1000-f10-1}, all deep learning-based methods (cMLP, cLSTM, CUTS, and CUTS+) experience substantial performance declines, and both DYNOTEARS and NTS-NOTEARS also deteriorate under the $F=10$ condition. Apart from these cases, the performance of most methods remains largely unchanged under min--max normalization. Overall, under min--max normalization, we observe that the performance of Granger causality-based, constraint-based, noise-based, and topology-based methods remains largely unchanged, whereas the performance of deep learning-based and score-based methods can vary across certain settings.

\textbf{Trend and seasonality model.}
We generate trend and seasonality data with $\rho=0.01$, $\eta=0.5$, and $P=12$. In the linear settings of Table~\ref{tab:linear-10-1000-1}, the performance of all methods deteriorates markedly under the influence of trend and seasonality. In the nonlinear settings of Table~\ref{tab:nonlinear-10-1000-f10-1}, the performance of nonlinear methods generally deteriorates substantially, whereas the performance of linear methods remains largely unchanged. Additional experiments with stronger trend and seasonal components ($\rho=0.02$, $\eta=12$) in the nonlinear setting are provided in Appendix~\ref{app_stronger_trend_seasonality}.

\subsection{Hyperparameter sensitivity analysis}
\label{sec:sensitivity}
\noindent \textbf{Motivations.} In real-world applications, the ground truth graph is typically unknown, which makes principled hyperparameter selection difficult. Consequently, investigating the performance of different methods across a range of hyperparameter configurations is of crucial importance.

\noindent \textbf{Results.} For each scenario, we report the mean and standard deviation of the evaluation metrics for each method under different hyperparameter settings (see Appendix~\ref{app_table_results_avg_hyper} and~\ref{app_fig_results_avg_hyper}). The mean reflects the overall performance of a method, while the standard deviation quantifies its sensitivity to hyperparameter choices. The results indicate that deep learning-based methods continue to exhibit superior overall performance (mean values). However, in linear settings, deep learning-based methods tend to show relatively high standard deviations, whereas in nonlinear settings, their standard deviations are generally much lower. This suggests that deep learning-based methods require more careful hyperparameter tuning in linear scenarios, while their performance is comparatively more stable in nonlinear settings. Although these findings are influenced by the chosen hyperparameter search space, our configurations follow the recommendations or default settings in prior work and involve an exhaustive hyperparameter exploration (see Appendix~\ref{app_hyper_settings}). Consequently, our results remain instructive for real-world applications of TSCD.

\subsection{cMLP ablation analysis}
\label{sec:ablation}
\noindent \textbf{Motivations.} To clarify why deep learning-based methods tend to achieve stronger performance under violations of modeling assumptions, we select cMLP as a representative deep learning-based method and conduct targeted ablation experiments. Specifically, we consider two ablations: cMLP-linear, which removes nonlinear expressive power, and cMLP (no sparsity), which removes the explicit sparsity constraint. Detailed ablation definitions and the corresponding summary table are provided in Appendix~\ref{app_ablation}.

\noindent \textbf{Results.} Table~\ref{tab:combined_ablation_summary} shows that cMLP-linear remains competitive in the linear setting, but its performance drops substantially in the nonlinear setting. This indicates that the nonlinear expressive power of cMLP is particularly important when the underlying causal dynamics are nonlinear. In contrast, cMLP (no sparsity) underperforms the original cMLP in both the linear and nonlinear settings, with a more pronounced degradation in AUPRC, indicating that hierarchical sparsity regularization plays an important role in suppressing spurious edges and improving causal edge selection. Overall, removing either nonlinear expressive power or hierarchical sparsity regularization causes cMLP to lose its performance advantage over other types of methods under model-assumption violations.

\subsection{Summary and implications for practice}\label{sec:Summary}
Based on the summarized results across all scenarios and configurations (see Appendix~\ref{app_summary}), as well as results obtained with dataset-specific optimal hyperparameters (see Appendix~\ref{app_table_results_best} and~\ref{app_fig_results_best}), hyperparameters selected by average performance (see Appendix~\ref{app_table_results_avg_scen} and~\ref{app_fig_results_avg_scen}), different levels of assumption violations (see Appendix~\ref{app_mea},~\ref{app_non}, and~\ref{app_confounder}), real-world datasets (see Appendix~\ref{app_real_datasets}), different time steps (see Appendix~\ref{app_time_steps}), different conditional independence tests (see Appendix~\ref{app_ci_tests}), and non-Gaussian noise (see Appendix~\ref{app_non_gaussian}), we observe that deep learning-based methods exhibit superior overall performance in most scenarios. 
Hyperparameter sensitivity analyses (Section~\ref{sec:sensitivity}) further indicate that deep learning methods exhibit higher sensitivity to hyperparameter choices in linear settings, whereas their performance is comparatively more stable in nonlinear settings. 
In practice, violations of model assumptions are inevitable, rendering the robustness of methods critically important. 
The robustness exhibited by deep learning-based methods in most scenarios suggests their potential for practical applications.
We further observe that NTS-NOTEARS relies on standardized preprocessing, performing poorly in the vanilla setting but achieving strong performance after standardization. Existing work~\cite{reisach2021beware,ng2024structure} has provided theoretical explanations for the performance degradation of continuous optimization methods under standardization in i.i.d. settings. In contrast, our empirical results reveal that, in time-series settings, the continuous optimization method NTS-NOTEARS exhibits performance gains under standardization.
Establishing a rigorous theoretical explanation for this phenomenon is an important direction for future research. Our findings also suggest that TSCD methods respond differently to Z-score standardization and min--max normalization. We recommend reporting results under vanilla, standardized, and min--max normalized settings when developing TSCD algorithms, as this practice can provide practitioners with clearer insights into the effects of preprocessing choices and ultimately facilitate the broader adoption of TSCD methods.

\section{Limitations}
\label{sec:limitations}

Our main evaluation uses data-type-dependent learning-rate search ranges for cMLP and cLSTM and does not apply a uniform threshold-search protocol to all methods. Appendix~\ref{app_unified_protocol} provides supplementary results and discussion using a unified threshold-search protocol and learning-rate ranges shared across linear and nonlinear settings. These results support our conclusion under the original learning-rate settings that, in most scenarios, the methods exhibiting superior overall performance are deep learning-based approaches. Under this unified automatic threshold-search protocol, the sensitivity of NTS-NOTEARS to standardization is observed only in the evaluated linear setting. In practical applications, however, the ground-truth causal graph is typically unavailable, preventing automatic threshold selection by minimizing NSHD against the ground truth. Our observations of the sensitivity of NTS-NOTEARS to standardization in both linear and nonlinear settings under manually specified thresholds therefore remain practically relevant.

\section{Broader impacts}
\label{sec:broader_impacts}

Superior robustness in benchmark experiments does not imply reliable deployment in all real-world settings, nor should it be interpreted as a general recommendation of deep learning-based methods for high-stakes applications. Over-reliance on a causal graph produced by a single method may lead to erroneous decisions. In high-stakes applications, the results of TSCD methods should be validated by domain experts, and their sensitivity should be assessed across different methods, hyperparameters, and data preprocessing settings. A causal structure obtained from a single estimate should not be treated as definitive. Decisions with significant consequences for individuals or groups should retain human oversight and independent review, while accounting for data bias, distribution shifts, and potentially unfair impacts across different groups. Without sufficient real-world validation, reliable uncertainty quantification, and appropriate human oversight, these methods should not be used directly for automated decision-making in high-stakes applications.

\section{Conclusion}
\label{sec:conclusion}
This work evaluates the performance of eleven mainstream time-series causal discovery methods under eight misspecified scenarios. The methods considered include constraint-based, noise-based, score-based, topology-based, Granger causality-based, and deep learning-based approaches. Our experimental results demonstrate that no single method achieves optimal performance across all scenarios. Nevertheless, in most scenarios, the methods exhibiting superior overall performance are deep learning-based approaches. 
We further provide hyperparameter sensitivity analyses to gain deeper insight into method performance. We also conduct ablation experiments to explain the strong performance of deep learning-based methods under misspecified settings. In particular, we find that NTS-NOTEARS relies on standardized preprocessing in practice, performing poorly in the vanilla setting but exhibiting strong performance after standardization is applied. Given the robustness displayed by deep learning-based methods in our benchmark, further in-depth investigation of these approaches is of substantial importance. To facilitate reproducible and fair evaluation, we release CausalCompass, a flexible and extensible Python package designed for long-term applicability. Our benchmark evaluates TSCD algorithms across diverse scenarios that reflect realistic data complexities, thereby improving understanding of their practical applicability, supporting more informed decision-making by practitioners, and ultimately promoting the broader adoption of TSCD methods. Finally, while this work focuses on TSCD algorithms, assessing the robustness of event sequence causal discovery in misspecified scenarios also represents an important direction for future research.

\begin{ack}
This work was supported by the Jiangsu Provincial Scientific Research
Center of Applied Mathematics under Grant No. BK20233002,
the National Key R\&D Program of China under Grant No. G2025YFF0524100,
the Fundamental Research Funds for the Central Universities,
the Natural Science Foundation of Jiangsu Province of China under
Grant Nos. BK20253020 and BK20253018,
the CCF-Tencent Rhino-Bird Open Research Fund,
and the National Natural Science Foundation of China under
Grant Nos. 62233004, 62273090, 62673129, and T2541017. 
This work was partially supported by SEU Kunpeng\&Ascend Center of Cultivation.
The authors declare no competing interests.
\end{ack}

\bibliographystyle{plainnat}
\bibliography{sample-base}

\newpage

\appendix
\tableofcontents  
\newpage

\section{Benchmark methods}
\label{app_methods}

\subsection{VAR}

A standard approach to Granger causal discovery in multivariate time series is based on the VAR model~\cite{arnold2007temporal}. If the lagged terms of a source variable receive non-negligible coefficients in the regression equation of a target variable, the source is considered to Granger-cause the target. Conversely, if all corresponding lagged coefficients are effectively zero, no Granger causal influence is inferred~\cite{shojaie2022granger}. In practice, a threshold hyperparameter is introduced to filter the estimated coefficients before constructing the Granger causal graph. We refer to a publicly available VAR implementation from \url{https://github.com/cloud36/graphical_granger_methods}.

\subsection{LGC}
Lasso Granger Causality (LGC~\cite{arnold2007temporal}) is a linear Granger causality approach that augments the VAR model with a Lasso penalty term, enabling the identification of sparse temporal causal relationships. We use a publicly available implementation of LGC from \url{https://github.com/cloud36/graphical_granger_methods}.

\subsection{VARLiNGAM}

VARLiNGAM~\cite{hyvarinen2010estimation} is a temporal extension of LiNGAM~\cite{shimizu2006linear}. It first fits a vector autoregressive model via least squares estimation and then applies LiNGAM to the resulting residuals to recover the instantaneous causal graph. Subsequently, the instantaneous causal structure is used to reparameterize the lagged causal coefficients of the vector autoregressive model, yielding a structural vector autoregressive (SVAR) model. We use the implementation of the VARLiNGAM algorithm described in~\cite{assaad2022survey}, available at \url{https://github.com/ckassaad/causal_discovery_for_time_series}.

\subsection{PCMCI}

PCMCI~\cite{runge2019detecting} is a constraint-based TSCD that extends the PC~\cite{spirtes1991algorithm} algorithm to multivariate time series. 
It first performs variable selection to reduce the dimensionality of the conditioning sets, and then employs the momentary conditional independence (MCI) test to address the inflated false positive rate induced by strong autocorrelations. 
The method outputs a window causal graph representing time-lagged causal dependencies. 
Although PCMCI supports both linear and nonlinear conditional independence (CI) tests, we use the linear CI test in our main experiments, following the design choice adopted in large-scale TSCD benchmarks such as TCD-Arena~\cite{stein2026tcdarena} and CausalDynamics~\cite{herdeanu2025causaldynamics}. This choice is primarily motivated by practical considerations: nonlinear CI tests are substantially more computationally expensive, and existing benchmark~\cite{stein2026tcdarena} evidence suggests that, although they can improve the performance of PCMCI+~\cite{runge2020discovering} relative to standard PCMCI in certain nonlinear settings, the gain is limited compared with the additional cost and still does not allow it to outperform other classes of methods. Therefore, given the scale of our experiments, we use the linear CI test as the default configuration and focus our main evaluation on PCMCI only. Results and discussion for PCMCI and PCMCI+ using nonlinear conditional independence tests are provided in Appendix~\ref{app_ci_tests}.
We use the implementation of the PCMCI algorithm
in Tigramite python package, available at \url{https://github.com/jakobrunge/tigramite}.

\subsection{DYNOTEARS}
DYNOTEARS~\cite{pamfil2020dynotears} is a score-based temporal extension of NOTEARS\allowbreak~\cite{zheng2018dags} that casts causal discovery as a continuous constrained optimization problem. It simultaneously estimates instantaneous and lagged causal effects in linear settings and produces a window causal graph. A smooth acyclicity constraint is imposed to guarantee that the instantaneous causal structure forms a directed acyclic graph (DAG). We use the implementation of the DYNOTEARS algorithm
in the CausalNex python package, available at \url{https://github.com/mckinsey/causalnex}.

\subsection{NTS-NOTEARS}
NTS-NOTEARS~\cite{sun2021nts} is a score-based nonlinear extension of DYNO\-TEARS. To address the limitation of multilayer perceptrons in exploiting temporal ordering information, it adopts one-dimensional convolutional neural networks to model nonlinear dependencies in time series. The method jointly estimates instantaneous and lagged causal effects, adopts an optimization framework similar to DYNOTEARS, supports the incorporation of prior knowledge as additional constraints, and outputs a window causal graph. DYNOTEA\-RS and NTS-NOTEARS are also referred to as continuous optimization based methods, which constitute a subclass of score-based TSCD approaches~\cite{gong2024causal}. We use the implementation of the NTS-NOTEARS algorithm provided by the authors, available at \url{https://github.com/xiangyu-sun-789/NTS-NOTEARS}.

\subsection{TSCI}
Tangent Space Causal Inference (TSCI~\cite{butler2024tangent}) is based on the idea that nonlinear dynamical systems can be locally approximated by linear dynamics in sufficiently small neighborhoods. Leveraging this property, TSCI conducts causal inference by analyzing the tangent spaces of the underlying data manifold. Concretely, it models the dynamics of each variable as a continuous vector field, which can be learned using neural ordinary differential equations~\cite{chen2018neural} or Gaussian processes. Causal relationships are then assessed by applying Convergent Cross Mapping (CCM~\cite{sugihara2012detecting}) in local tangent spaces, evaluating whether the reconstructed state space of one variable can reliably predict another. This approach is designed for causal discovery in deterministic dynamical systems. We use the implementation of the TSCI algorithm provided by the authors, available at \url{https://github.com/KurtButler/tangentspaces}.

\subsection{NGC}
Neural Granger Causality (NGC~\cite{tank2021neural}) refers to a class of deep learning-based nonlinear Granger causal discovery methods that capture multivariate time series dependencies through neural architectures with sparsified input structures. The framework can be implemented using component-wise MLP (cMLP~\cite{tank2021neural}) or component-wise LSTM (cLSTM~\cite{tank2021neural}), and identifies Granger causality by imposing structural sparsity penalties. While the cMLP yields a window causal graph, the cLSTM is restricted to producing a summary causal graph. We use the implementations of the cMLP and cLSTM algorithms provided by the authors, available at \url{https://github.com/iancovert/Neural-GC}. 

\subsection{CUTS}
CUTS~\cite{cheng2023cuts} is a deep learning-based TSCD method tailored for irregular time series data. It employs an iterative two-module framework that jointly performs data imputation and causal structure learning in a mutually enhancing manner. Specifically, the first stage leverages a delayed supervision graph neural network to infer latent data from high-dimensional, irregularly sampled time series with complex distributions. The second stage applies sparsity-inducing regularization to the imputed data in order to recover causal relationships between variables. In the missing-data setting, CUTS is run using its native framework without external imputation. We use the implementation of the CUTS algorithm provided by the authors, available at \url{https://github.com/jarrycyx/UNN}. 

\subsection{CUTS+}
CUTS+ extends the CUTS framework by integrating a two-stage coarse-to-fine search strategy with a message-passing graph neural network. This design allows the method to handle causal discovery in high-dimensional time series with irregular sampling. In the first stage, lightweight Granger causality tests are used to preselect candidate parents for each variable, substantially reducing the search space. In the second stage, a graph neural network jointly imputes missing or unevenly sampled observations and learns a sparse causal graph through a penalized reconstruction loss with temporal encoding. The alternating optimization between imputation and structure learning enables CUTS+ to remain scalable while maintaining strong performance under missing-data scenarios. In the missing-data setting, CUTS+ is run using its native framework without external imputation. We use the implementation of the CUTS+ algorithm provided by the authors, available at \url{https://github.com/jarrycyx/UNN}.

\clearpage

\section{Detailed description of assumption violations}
\label{app_assumption}

\textbf{Nonstationary model.}
The observed nonstationary data are generated as:
\begin{equation}
x_{t,i}^{\mathrm{NS}} = f_i\!\big(Pa(x_{t,i})) + \omega_{t,i} u_{t,i}, \quad \forall i = 1, \ldots, D,
\label{eq:nonstat}
\end{equation}
where $f_i(\cdot)$ represents the deterministic causal mechanism for variable $i$ (either the VAR dynamics in~\eqref{eq:var} or the Lorenz-96 dynamics in~\eqref{eq:lorenz}), $u_{t,i} \sim \mathcal{N}(0, \sigma_i^2)$ is the base-level independent noise, and $\omega_{t,i}$ is the time-varying scaling parameter generated by sampling from a Gaussian process. Specifically, we sample $\log \boldsymbol{\omega}_i = (\log \omega_{1,i}, \ldots, \log \omega_{T,i})^\top \sim \mathcal{GP}(m \cdot \mathbf{1}_T, \nu^2 K(\mathbf{t}, \mathbf{t}))$, where $\mathbf{1}_T$ denotes a $T$-dimensional vector of ones, $m$ controls the average log-scale noise level, $\nu$ controls the magnitude of variance fluctuations, and $K(\mathbf{t}, \mathbf{t}) \in \mathbb{R}^{T \times T}$ is the covariance matrix with entries $(K)_{ij} = \exp\!\left(-\frac{(t_i - t_j)^2}{2\ell^2}\right)$ (squared exponential kernel), where $\ell$ is the kernel width. The actual scaling parameter is computed as $\omega_{t,i} = \exp(\log \omega_{t,i})$ to ensure positivity.

\textbf{Z-score standardization model.}
The data-generation mechanism considered is:
\begin{equation}
\overline{X_i}=\frac{X_i-\mathbb{E}[X_i]}{\sqrt{\operatorname{Var}(X_i)}}, \quad \forall i=1, \ldots, D,
\end{equation}
 where $\mathbb{E}[X_i]$ and $\mathrm{Var}(X_i)$ are the mean and variance of $X_i$, respectively. The standardized data are used as the algorithm inputs, while the ground-truth causal graph remains identical to that of the original vanilla data.

\textbf{Min--max normalization model.} 
In this work, the data-generation mechanism we consider is:
\begin{equation}
\tilde{X}_i = \frac{X_i - \min(X_i)}{\max(X_i) - \min(X_i)}, 
\quad \forall\, i = 1, \ldots, D,
\end{equation}
where $\tilde{X}_i$ denotes the normalized temporal process of the $i$-th variable.

\textbf{Trend and seasonality model.}
The observed process is defined as:
\begin{equation}
x_{t,i}^{\mathrm{TS}}
= x_{t,i} + \mathrm{trend}_{t,i} + \mathrm{season}_{t,i},
\qquad 1 \le i \le D ,
\end{equation}
where $x_{t,i}$ denotes the value generated by the linear vanilla model~\eqref{eq:var} 
or nonlinear vanilla model~\eqref{eq:lorenz}. The trend component introduces a variable-specific linear drift:
\begin{equation}
\mathrm{trend}_{t,i} = \frac{\rho t i}{2},
\end{equation}
where $\rho$ controls the strength of the trend. The seasonal component incorporates two harmonics with variable-specific phase shifts:
\begin{equation}
\mathrm{season}_{t,i}
= \eta \sin\!\left( \frac{2\pi t}{P} + \phi_i \right)
+ \frac{\eta}{2}\cos\!\left( \frac{4\pi t}{P} + \phi_i \right),
\end{equation}
where $P$ denotes the seasonal period, $\eta$ specifies the overall seasonal magnitude, and $\phi_i=\frac{2\pi i}{D}$ induces distinct phase shifts across variables.

\section{Selection of violation severity levels}
\label{app_violation_severity}
The severity of each model assumption violation in our benchmark is controlled through dedicated data-generation parameters. Inspired by the design philosophy of robustness benchmarks such as TCD-Arena~\cite{stein2026tcdarena}, we adopt the following principled procedure for selecting the range of violation severity for each category (excluding z-score and min--max normalization, which do not involve a continuous severity parameter). We first identify the minimum severity level at which the VAR method exhibits only a mild performance degradation compared to the vanilla setting. We then determine the maximum severity level at which the VAR method approaches chance-level performance. Finally, the data-generation parameters used in our experiments are sampled from the interval between these two extremes. This strategy ensures that the benchmark covers a meaningful and interpretable range of violation strengths, from near-benign to highly challenging.

\clearpage

\section{Summary of methods' performance}
\label{app_summary}

In Table~\ref{tab:overall_summary}, we present a comprehensive summary of methods' performance across diverse scenarios and configurations. The results indicate that deep learning-based approaches demonstrate superior performance, with CUTS+ achieving the best overall performance.

\begin{table}[!htb]
\centering
\caption{Summary of methods' performances across all scenarios and configurations. The reported results are the mean and standard deviation of the metrics across different time series lengths, external forcing intensities, vanilla scenarios and misspecified scenarios.}
\label{tab:overall_summary}
\begin{tabular}{@{}lccc@{}}
\toprule
\textbf{Method} & \textbf{$D$ (Nodes)} & \textbf{AUROC} & \textbf{AUPRC} \\ \midrule
\multirow{2}{*}{VAR} & 10 & 71.73$\pm$14.55 & 48.91$\pm$17.62 \\
 & 15 & 72.17$\pm$12.91 & 40.74$\pm$17.99 \\
\cmidrule(lr){2-4}
\multirow{2}{*}{LGC} & 10 & 74.49$\pm$15.09 & 55.00$\pm$20.04 \\
 & 15 & 72.52$\pm$13.44 & 45.50$\pm$19.83 \\
\cmidrule(lr){2-4}
\multirow{2}{*}{VARLiNGAM} & 10 & 59.34$\pm$9.67 & 34.91$\pm$17.88 \\
 & 15 & 59.01$\pm$8.86 & 27.99$\pm$12.53 \\
\cmidrule(lr){2-4}
\multirow{2}{*}{PCMCI} & 10 & 74.60$\pm$13.91 & 51.13$\pm$16.93 \\
 & 15 & 74.23$\pm$12.77 & 44.58$\pm$17.52 \\
\cmidrule(lr){2-4}
\multirow{2}{*}{DYNOTEARS} & 10 & 73.90$\pm$16.60 & 56.14$\pm$22.91 \\
 & 15 & 73.34$\pm$14.64 & 46.46$\pm$22.23 \\
\cmidrule(lr){2-4}
\multirow{2}{*}{NTS-NOTEARS} & 10 & 62.68$\pm$17.23 & 42.50$\pm$22.96 \\
 & 15 & 59.92$\pm$15.68 & 30.78$\pm$21.51 \\
\cmidrule(lr){2-4}
\multirow{2}{*}{TSCI} & 10 & 63.30$\pm$12.72 & 42.76$\pm$19.76 \\
 & 15 & 63.01$\pm$14.07 & 35.35$\pm$19.50 \\
\cmidrule(lr){2-4}
\multirow{2}{*}{cMLP} & 10 & 74.62$\pm$21.08 & 63.66$\pm$27.41 \\
 & 15 & 73.44$\pm$19.47 & 52.85$\pm$29.81 \\
\cmidrule(lr){2-4}
\multirow{2}{*}{cLSTM} & 10 & 81.61$\pm$19.76 & 73.91$\pm$26.04 \\
 & 15 & 80.05$\pm$18.68 & 63.18$\pm$29.06 \\
\cmidrule(lr){2-4}
\multirow{2}{*}{CUTS} & 10 & 72.40$\pm$19.64 & 57.75$\pm$25.45 \\
 & 15 & 69.55$\pm$18.32 & 46.64$\pm$26.85 \\
\cmidrule(lr){2-4}
\multirow{2}{*}{CUTS+} & 10 & \textbf{84.10$\pm$17.92} & \textbf{75.07$\pm$24.98} \\
 & 15 & \textbf{82.53$\pm$16.87} & \textbf{67.17$\pm$27.26} \\
\bottomrule
\end{tabular}
\end{table}

\section{Hyperparameter search ranges}
\label{app_hyper_settings}

In practice, the underlying causal structure of real-world datasets is typically unknown, making it difficult to determine optimal hyperparameters for different algorithms. 
To ensure a fair comparison among methods, we follow the default or recommended hyperparameter settings provided in the official code repositories of the benchmark methods and further tune the hyperparameters to achieve optimal performance wherever possible~\cite{chen2025addressing,cheng2023cuts,cheng2024cuts+,tank2021neural,sun2021nts,runge2019detecting,assaad2022survey,butler2024tangent}. 
The hyperparameter search ranges are summarized in Table~\ref{tab:hyperparams}. The meanings of hyperparameters are as follows:

\begin{itemize}
    \item $\tau_{max}$: pre-defined maximum lag for capturing temporal dependencies in time series;
    
    \item $\alpha$: significance level for statistical tests (used in PCMCI);
    
    \item $threshold$: threshold value for pruning weak causal connections (used in VAR, LGC, and VARLiNGAM);
    
    \item $w_{thre}$: threshold for edge weights to determine the presence of causal relationships (used in DYNOTEARS and NTS-NOTEARS);
    
    \item $\lambda$, $\lambda_1$, $\lambda_2$, $\lambda_a$, $\lambda_w$: regularization parameters;
    
    \item $\theta$, $fnn\_tol$: specific hyperparameters of TSCI;
    
    \item $input\ step$, $batch\ size$, $weight\ decay$: hyperparameters of CUTS and CUTS+;
    
    \item $lr$: learning rate.
\end{itemize}

\begin{table*}[t]
\centering
\caption{Hyperparameter settings of benchmark methods.}
\label{tab:hyperparams}
\small
\setlength{\tabcolsep}{3.5pt}
\begin{tabular}{l|c|c}
\toprule
Method & Linear Settings & Nonlinear Settings \\
\midrule
VAR & \multicolumn{2}{c}{$\tau_{max} = \{3, 5\}$, $threshold = \{0, 0.01, 0.05, 0.1, 0.3\}$} \\
\midrule
LGC & \multicolumn{2}{c}{$\tau_{max} = \{3, 5\}$, $threshold = \{0, 0.01, 0.05, 0.1, 0.3\}$} \\
\midrule
VARLiNGAM & \multicolumn{2}{c}{$\tau_{max} = \{3, 5\}$, $threshold = \{0, 0.01, 0.05, 0.1, 0.3\}$} \\
\midrule
PCMCI & \multicolumn{2}{c}{$\tau_{max} = \{3, 5\}$, $\alpha = \{0.01, 0.05, 0.1\}$} \\
\midrule
DYNOTEARS & \multicolumn{2}{c}{$\tau_{max} = \{3, 5\}$, $w_{thre} = \{0.01, 0.05, 0.1, 0.3\}$, $\lambda_a = \{0.001, 0.01, 0.1\}$, $\lambda_w = \{0.001, 0.005, 0.01\}$} \\
\midrule
NTS-NOTEARS & \makecell[l]{$\tau_{max} = \{3\}$, $w_{thre} = \{0.01, 0.05, 0.1, 0.3, 0.5\}$, \\ $\lambda_1 = \{0.001, 0.01, 0.1\}$, $\lambda_2 = \{0.001, 0.005, 0.01\}$} & \makecell[l]{$\tau_{max} = \{1\}$, $w_{thre} = \{0.1, 0.3, 0.5\}$, \\ $\lambda_1 = \{(0.001, 0.1)\}$, $\lambda_2 = \{0.01\}$} \\
\midrule
TSCI & \multicolumn{2}{c}{$\theta = \{0.4, 0.5, 0.6\}$, $fnn\_tol = \{0.005, 0.01\}$} \\
\midrule
cMLP & $\lambda = \{1\text{e-}4, 5\text{e-}3, 0.05\}$, $lr = \{0.01, 0.1\}$ & $\lambda = \{1\text{e-}4, 5\text{e-}3, 0.05\}$, $lr = \{5\text{e-}4, 0.001\}$ \\
\midrule
cLSTM & $\lambda = \{1\text{e-}4, 5\text{e-}3, 0.05\}$, $lr = \{0.01, 0.1\}$ & $\lambda = \{1\text{e-}4, 5\text{e-}3, 0.05\}$, $lr = \{5\text{e-}4, 0.001\}$ \\
\midrule
CUTS & \multicolumn{2}{c}{$input\ step = \{1, 3, 5, 10\}$, $batch\ size = \{32, 128\}$, $weight\ decay = \{0, 0.001, 0.003\}$} \\
\midrule
CUTS+ & \multicolumn{2}{c}{$input\ step = \{1, 3, 5, 10\}$, $batch\ size = \{32, 128\}$, $weight\ decay = \{0, 0.001, 0.003\}$} \\
\bottomrule
\end{tabular}
\end{table*}

\section{Computational cost comparison}
\label{app_computational_cost}

All experiments were conducted on a Linux server equipped with two NVIDIA RTX A6000 GPUs (49 GB memory per GPU; driver version 550.78), Intel Xeon Gold 6326 CPUs at 2.90 GHz, and 59 GiB of system memory. Unless otherwise stated, all experiments were run on this hardware configuration.

Table~\ref{tab:computational_cost} reports the AUROC, AUPRC,
average runtime, peak CPU memory, peak GPU memory, and average
GPU utilization of 11 methods under the mixed-data setting
with 10 nodes and 1,000 samples generated by the VAR model.
In Table~\ref{tab:computational_cost}, ``---'' indicates that
the method uses only the CPU. Therefore, GPU memory and
utilization are not applicable. The results show that deep learning-based methods require longer runtimes and greater computational resources to achieve superior performance. This systematic comparison helps practitioners make informed trade-offs between performance and computational efficiency.

\begin{table*}[!htbp]
    \centering
    \caption{Performance and computational resource comparison.}
    \label{tab:computational_cost}
    \small
    \setlength{\tabcolsep}{3pt}
    \begin{adjustbox}{max width=\textwidth}
    \begin{tabular}{lcccccc}
        \toprule
        Method
        & \makecell{AUROC\\(\%)}
        & \makecell{AUPRC\\(\%)}
        & \makecell{Runtime\\(s)}
        & \makecell{Peak CPU\\memory (MiB)}
        & \makecell{Peak GPU\\memory (MiB)}
        & \makecell{Mean GPU\\utilization (\%)} \\
        \midrule
        VAR
        & 62.0 & 32.4 & 1.17 & 603.19 & --- & --- \\

        LGC
        & 81.2 & 58.4 & 1.41 & 593.85 & --- & --- \\

        VARLiNGAM
        & 50.7 & 12.9 & 1.71 & 600.63 & --- & --- \\

        PCMCI
        & 96.7 & 72.2 & 6.74 & 594.68 & --- & --- \\

        DYNOTEARS
        & 80.5 & 59.3 & 2.07 & 595.91 & --- & --- \\

        NTS-NOTEARS
        & 75.0 & 55.5 & 26.24 & 1173.44 & 372.00 & 3.41 \\

        TSCI
        & 55.4 & 20.3 & 31.86 & 777.88 & --- & --- \\

        cMLP
        & 98.3 & \textbf{94.1} & 208.05
        & 1219.15 & 322.00 & 10.87 \\

        cLSTM
        & \textbf{98.5} & 90.8 & 264.58
        & 1222.67 & 732.00 & 39.39 \\

        CUTS
        & 51.6 & 21.5 & 127.29
        & 1455.96 & 354.00 & 8.80 \\

        CUTS+
        & 74.5 & 45.8 & 167.58
        & 1531.85 & 354.00 & 8.56 \\
        \bottomrule
    \end{tabular}
    \end{adjustbox}
\end{table*}

\section{Definition of NSHD}
\label{app:nshd}
Following TCD-Arena~\cite{stein2026tcdarena}, we use the normalized Structural Hamming Distance (NSHD) to measure the discrepancy between an estimated causal graph and the ground-truth graph. Specifically, for a thresholded estimate $\hat{G}_{\tau}$, NSHD is defined as
\begin{equation}
\mathrm{NSHD}
=
\min_{\tau \in \mathcal{T}}
\frac{\mathrm{SHD}(G, \hat{G}_{\tau})}{|A_G|},
\end{equation}
where $G$ denotes the ground-truth graph, $|A_G|$ is the number of edges in the ground-truth graph, and $\mathcal{T}$ is the set of decision thresholds. It is worth noting that in~\cite{jiralerspong2024efficient}, NSHD is defined as SHD divided by the square of the number of nodes in the ground-truth graph, whereas in TCD-Arena~\cite{stein2026tcdarena} and our definition, we normalize by the number of edges instead. This choice avoids excessively small metric values and makes the differences across methods more discernible.

In the main text, we report NSHD results for the representative linear 10-node setting with $T=1000$ and the nonlinear 10-node setting with $T=1000$ and $F=10$. Since the main conclusions under NSHD are broadly consistent with those under AUROC and AUPRC, we omit the NSHD results for other settings for brevity.

\section{cMLP ablation experiments}
\label{app_ablation}
To better understand the source of cMLP's robustness under model-assumption violations, we conduct two additional ablations. These variants keep the optimization pipeline and hyperparameter selection protocol consistent with the original cMLP experiment, while modifying one model component at a time.

\begin{itemize}[leftmargin=*]
    \item \textbf{cMLP-linear.} This variant preserves the target-wise lagged modeling framework, network width, and training pipeline of the original cMLP, but replaces the ReLU activations between hidden layers with identity mappings. The resulting model therefore behaves as a linear mapping. This ablation evaluates the contribution of nonlinear expressive power.

    \item \textbf{cMLP (no sparsity).} This variant preserves the original nonlinear cMLP architecture and training pipeline, but sets the hierarchical sparsity regularization coefficient to zero, removing the explicit sparsity constraint on the learned causal graph. This ablation evaluates the role of sparsity regularization in causal edge selection.
\end{itemize}

\begin{table}[!htb]
\centering
\caption{Combined ablation summary for the 10-node, $T=1000$ setting. For each method and each scenario, we first select the best-performing result within that scenario, then aggregate the five seeds from vanilla and eight assumption-violation scenarios (45 values in total) to report mean and standard deviation.}
\label{tab:combined_ablation_summary}
\begin{tabular}{@{}lccc@{}}
\toprule
\textbf{Method} & \textbf{Setting} & \textbf{AUROC} & \textbf{AUPRC} \\ \midrule
\multirow{2}{*}{VAR} & Linear & 85.22$\pm$15.26 & 61.43$\pm$29.16 \\
 & Nonlinear & 74.07$\pm$11.43 & 54.99$\pm$12.67 \\
\cmidrule(lr){2-4}
\multirow{2}{*}{LGC} & Linear & 86.82$\pm$15.75 & 68.34$\pm$29.55 \\
 & Nonlinear & 76.76$\pm$10.55 & 59.44$\pm$13.22 \\
\cmidrule(lr){2-4}
\multirow{2}{*}{VARLiNGAM} & Linear & 50.86$\pm$6.63 & 12.67$\pm$3.30 \\
 & Nonlinear & 67.91$\pm$6.94 & 51.72$\pm$9.86 \\
\cmidrule(lr){2-4}
\multirow{2}{*}{PCMCI} & Linear & 88.38$\pm$13.35 & 56.72$\pm$26.87 \\
 & Nonlinear & 74.59$\pm$8.39 & 54.51$\pm$8.94 \\
\cmidrule(lr){2-4}
\multirow{2}{*}{DYNOTEARS} & Linear & 89.21$\pm$14.22 & 72.57$\pm$29.62 \\
 & Nonlinear & 72.52$\pm$14.26 & 56.75$\pm$18.39 \\
\cmidrule(lr){2-4}
\multirow{2}{*}{NTS-NOTEARS} & Linear & 70.44$\pm$18.67 & 41.16$\pm$31.26 \\
 & Nonlinear & 63.31$\pm$17.32 & 46.89$\pm$20.91 \\
\cmidrule(lr){2-4}
\multirow{2}{*}{TSCI} & Linear & 57.71$\pm$9.68 & 22.63$\pm$9.78 \\
 & Nonlinear & 72.38$\pm$13.84 & 61.13$\pm$16.20 \\
\cmidrule(lr){2-4}
\multirow{2}{*}{cMLP} & Linear & 89.52$\pm$17.41 & 79.22$\pm$29.95 \\
 & Nonlinear & 71.77$\pm$19.46 & 62.12$\pm$24.95 \\
\cmidrule(lr){2-4}
\multirow{2}{*}{cLSTM} & Linear & \textbf{95.61$\pm$12.04} & \textbf{89.89$\pm$23.03} \\
 & Nonlinear & 80.75$\pm$18.59 & 73.88$\pm$23.70 \\
\cmidrule(lr){2-4}
\multirow{2}{*}{CUTS} & Linear & 77.84$\pm$21.41 & 56.34$\pm$31.99 \\
 & Nonlinear & 75.57$\pm$18.81 & 67.45$\pm$22.91 \\
\cmidrule(lr){2-4}
\multirow{2}{*}{CUTS+} & Linear & 90.81$\pm$14.99 & 77.77$\pm$28.98 \\
 & Nonlinear & \textbf{86.13$\pm$17.72} & \textbf{81.63$\pm$22.67} \\
\cmidrule(lr){2-4}
\multirow{2}{*}{cMLP-linear} & Linear & 91.08$\pm$15.59 & 81.50$\pm$26.44 \\
 & Nonlinear & 58.34$\pm$7.15 & 44.46$\pm$8.95 \\
\cmidrule(lr){2-4}
\multirow{2}{*}{cMLP (no sparsity)} & Linear & 76.80$\pm$18.65 & 52.06$\pm$31.03 \\
 & Nonlinear & 67.11$\pm$18.38 & 56.71$\pm$22.41 \\
\bottomrule
\end{tabular}
\end{table}

\clearpage

\section{Table results under different levels of measurement error}
\label{app_mea}
Tables~\ref{tab:measurement_error_VAR_T1000}--\ref{tab:measurement_error_Lorenz_T1000_F40} report results under measurement error with $\alpha=1.2$ and $\alpha=10.0$. Larger values of $\alpha$ indicate stronger measurement error. The results indicate that, overall, method performance generally degrades as the severity of model assumption violations increases, while the best-performing methods across these settings are predominantly deep learning-based approaches. Considering that NTS-NOTEARS exhibits relatively slow runtime in certain settings and does not impact our main conclusions, we do not report its results in Appendices~\ref{app_mea}--\ref{app_confounder}.

\begin{table*}[!htbp]
\centering
\caption{Linear setting under measurement error with $\alpha = 1.2$ and $\alpha = 10.0$ ($T = 1000$).}
\label{tab:measurement_error_VAR_T1000}
\small
\setlength{\tabcolsep}{4pt}
\begin{tabular}{ccccccc}
\toprule
& \multicolumn{2}{c}{Vanilla} & \multicolumn{2}{c}{ME ($\alpha=1.2$)} & \multicolumn{2}{c}{ME ($\alpha=10.0$)} \\
\cmidrule(lr){2-3} \cmidrule(lr){4-5} \cmidrule(lr){6-7}
10 nodes & AUROC$\uparrow$ & AUPRC$\uparrow$ & AUROC$\uparrow$ & AUPRC$\uparrow$ & AUROC$\uparrow$ & AUPRC$\uparrow$ \\
\midrule
VAR & 99.7\footnotesize{$\pm$0.3} & 96.3\footnotesize{$\pm$4.4} & 85.2\footnotesize{$\pm$8.9} & 65.6\footnotesize{$\pm$18.8} & 60.9\footnotesize{$\pm$5.6} & 14.6\footnotesize{$\pm$2.3} \\
LGC & \textbf{100.0\footnotesize{$\pm$0.0}} & \textbf{100.0\footnotesize{$\pm$0.0}} & 84.5\footnotesize{$\pm$7.4} & 66.9\footnotesize{$\pm$17.4} & 55.6\footnotesize{$\pm$5.0} & 17.9\footnotesize{$\pm$6.8} \\
VARLiNGAM & 48.2\footnotesize{$\pm$5.5} & 11.8\footnotesize{$\pm$0.8} & 58.2\footnotesize{$\pm$5.1} & 13.7\footnotesize{$\pm$1.8} & 50.2\footnotesize{$\pm$2.4} & 12.8\footnotesize{$\pm$3.5} \\
PCMCI & 98.2\footnotesize{$\pm$0.9} & 79.1\footnotesize{$\pm$8.8} & 91.3\footnotesize{$\pm$6.0} & 68.5\footnotesize{$\pm$11.9} & 66.3\footnotesize{$\pm$7.5} & 19.8\footnotesize{$\pm$5.6} \\
DYNOTEARS & \textbf{100.0\footnotesize{$\pm$0.0}} & \textbf{100.0\footnotesize{$\pm$0.0}} & 76.8\footnotesize{$\pm$9.3} & 57.4\footnotesize{$\pm$17.1} & 64.5\footnotesize{$\pm$10.4} & 18.8\footnotesize{$\pm$6.2} \\
TSCI & 58.8\footnotesize{$\pm$6.7} & 27.0\footnotesize{$\pm$9.3} & 47.3\footnotesize{$\pm$6.1} & 11.7\footnotesize{$\pm$2.0} & 58.3\footnotesize{$\pm$4.8} & 17.3\footnotesize{$\pm$4.2} \\
cMLP & 99.9\footnotesize{$\pm$0.0} & 99.8\footnotesize{$\pm$0.3} & 84.1\footnotesize{$\pm$9.8} & 58.4\footnotesize{$\pm$16.6} & 53.5\footnotesize{$\pm$8.2} & 19.0\footnotesize{$\pm$7.2} \\
cLSTM & \textbf{100.0\footnotesize{$\pm$0.0}} & \textbf{100.0\footnotesize{$\pm$0.0}} & \textbf{99.6\footnotesize{$\pm$0.1}} & \textbf{97.9\footnotesize{$\pm$0.8}} & 79.7\footnotesize{$\pm$5.1} & 43.3\footnotesize{$\pm$12.1} \\
CUTS & 96.8\footnotesize{$\pm$1.3} & 80.8\footnotesize{$\pm$8.9} & 90.1\footnotesize{$\pm$7.1} & 64.6\footnotesize{$\pm$14.9} & 65.6\footnotesize{$\pm$4.1} & 25.5\footnotesize{$\pm$6.4} \\
CUTS+ & \textbf{100.0\footnotesize{$\pm$0.0}} & \textbf{100.0\footnotesize{$\pm$0.0}} & 99.4\footnotesize{$\pm$0.3} & 96.0\footnotesize{$\pm$2.9} & \textbf{82.0\footnotesize{$\pm$4.7}} & \textbf{45.9\footnotesize{$\pm$10.0}} \\
\bottomrule
\end{tabular}
\end{table*}

\begin{table*}[!htbp]
\centering
\caption{Nonlinear setting under measurement error with $\alpha = 1.2$ and $\alpha = 10.0$ ($T = 1000$, $F = 10$).}
\label{tab:measurement_error_Lorenz_T1000_F10}
\small
\setlength{\tabcolsep}{4pt}
\begin{tabular}{ccccccc}
\toprule
& \multicolumn{2}{c}{Vanilla} & \multicolumn{2}{c}{ME ($\alpha=1.2$)} & \multicolumn{2}{c}{ME ($\alpha=10.0$)} \\
\cmidrule(lr){2-3} \cmidrule(lr){4-5} \cmidrule(lr){6-7}
10 nodes & AUROC$\uparrow$ & AUPRC$\uparrow$ & AUROC$\uparrow$ & AUPRC$\uparrow$ & AUROC$\uparrow$ & AUPRC$\uparrow$ \\
\midrule
VAR & 83.3\footnotesize{$\pm$3.6} & 66.6\footnotesize{$\pm$5.6} & 73.8\footnotesize{$\pm$3.4} & 50.3\footnotesize{$\pm$3.8} & 55.3\footnotesize{$\pm$4.7} & 36.5\footnotesize{$\pm$2.8} \\
LGC & 83.8\footnotesize{$\pm$2.6} & 72.4\footnotesize{$\pm$4.5} & 73.5\footnotesize{$\pm$3.9} & 50.3\footnotesize{$\pm$4.4} & 55.1\footnotesize{$\pm$4.7} & 36.4\footnotesize{$\pm$2.8} \\
VARLiNGAM & 72.0\footnotesize{$\pm$1.9} & 60.3\footnotesize{$\pm$2.7} & 62.6\footnotesize{$\pm$5.8} & 41.8\footnotesize{$\pm$4.4} & 50.0\footnotesize{$\pm$1.0} & 33.8\footnotesize{$\pm$0.8} \\
PCMCI & 77.8\footnotesize{$\pm$4.9} & 55.9\footnotesize{$\pm$6.1} & 75.1\footnotesize{$\pm$5.6} & 56.2\footnotesize{$\pm$7.9} & 55.3\footnotesize{$\pm$4.4} & 36.6\footnotesize{$\pm$2.7} \\
DYNOTEARS & 86.1\footnotesize{$\pm$2.1} & 71.2\footnotesize{$\pm$3.4} & 61.1\footnotesize{$\pm$1.9} & 39.5\footnotesize{$\pm$1.3} & 51.6\footnotesize{$\pm$2.4} & 34.1\footnotesize{$\pm$1.2} \\
TSCI & 88.4\footnotesize{$\pm$1.7} & 80.5\footnotesize{$\pm$2.2} & 57.3\footnotesize{$\pm$6.0} & 45.0\footnotesize{$\pm$6.1} & 52.0\footnotesize{$\pm$4.8} & 38.2\footnotesize{$\pm$4.7} \\
cMLP & 99.9\footnotesize{$\pm$0.0} & 99.9\footnotesize{$\pm$0.0} & 57.7\footnotesize{$\pm$1.5} & 42.4\footnotesize{$\pm$1.5} & 51.9\footnotesize{$\pm$5.0} & 38.8\footnotesize{$\pm$4.3} \\
cLSTM & \textbf{100.0\footnotesize{$\pm$0.0}} & \textbf{99.9\footnotesize{$\pm$0.0}} & 77.5\footnotesize{$\pm$2.3} & 69.2\footnotesize{$\pm$2.7} & 52.2\footnotesize{$\pm$8.3} & 38.2\footnotesize{$\pm$7.4} \\
CUTS & 98.3\footnotesize{$\pm$1.1} & 97.5\footnotesize{$\pm$1.6} & 63.8\footnotesize{$\pm$2.6} & 55.0\footnotesize{$\pm$2.1} & 54.9\footnotesize{$\pm$3.4} & 38.4\footnotesize{$\pm$3.6} \\
CUTS+ & 99.9\footnotesize{$\pm$0.0} & 99.9\footnotesize{$\pm$0.1} & \textbf{96.7\footnotesize{$\pm$2.1}} & \textbf{94.6\footnotesize{$\pm$3.1}} & \textbf{59.1\footnotesize{$\pm$8.3}} & \textbf{50.0\footnotesize{$\pm$8.4}} \\
\bottomrule
\end{tabular}
\end{table*}

\begin{table*}[!htbp]
\centering
\caption{Nonlinear setting under measurement error with $\alpha = 1.2$ and $\alpha = 10.0$ ($T = 1000$, $F = 40$).}
\label{tab:measurement_error_Lorenz_T1000_F40}
\small
\setlength{\tabcolsep}{4pt}
\begin{tabular}{ccccccc}
\toprule
& \multicolumn{2}{c}{Vanilla} & \multicolumn{2}{c}{ME ($\alpha=1.2$)} & \multicolumn{2}{c}{ME ($\alpha=10.0$)} \\
\cmidrule(lr){2-3} \cmidrule(lr){4-5} \cmidrule(lr){6-7}
10 nodes & AUROC$\uparrow$ & AUPRC$\uparrow$ & AUROC$\uparrow$ & AUPRC$\uparrow$ & AUROC$\uparrow$ & AUPRC$\uparrow$ \\
\midrule
VAR & 73.0\footnotesize{$\pm$2.0} & 58.5\footnotesize{$\pm$3.1} & 61.9\footnotesize{$\pm$3.5} & 40.8\footnotesize{$\pm$2.7} & 50.3\footnotesize{$\pm$5.2} & 34.0\footnotesize{$\pm$2.4} \\
LGC & 72.8\footnotesize{$\pm$2.2} & 58.7\footnotesize{$\pm$4.2} & 62.3\footnotesize{$\pm$3.2} & 41.1\footnotesize{$\pm$2.5} & 50.3\footnotesize{$\pm$5.2} & 34.0\footnotesize{$\pm$2.4} \\
VARLiNGAM & 65.5\footnotesize{$\pm$2.7} & 45.0\footnotesize{$\pm$3.0} & 53.9\footnotesize{$\pm$3.5} & 36.6\footnotesize{$\pm$3.2} & 50.0\footnotesize{$\pm$1.3} & 33.8\footnotesize{$\pm$0.8} \\
PCMCI & 75.5\footnotesize{$\pm$1.5} & 59.8\footnotesize{$\pm$4.0} & 62.1\footnotesize{$\pm$5.9} & 42.7\footnotesize{$\pm$6.0} & 48.3\footnotesize{$\pm$5.1} & 33.7\footnotesize{$\pm$2.0} \\
DYNOTEARS & 69.6\footnotesize{$\pm$2.8} & 50.1\footnotesize{$\pm$3.5} & 54.0\footnotesize{$\pm$2.1} & 36.9\footnotesize{$\pm$2.1} & 50.3\footnotesize{$\pm$5.7} & 33.9\footnotesize{$\pm$2.2} \\
TSCI & 71.3\footnotesize{$\pm$4.6} & 56.1\footnotesize{$\pm$4.6} & 52.3\footnotesize{$\pm$7.1} & 35.1\footnotesize{$\pm$4.3} & 48.7\footnotesize{$\pm$5.0} & 35.8\footnotesize{$\pm$4.7} \\
cMLP & 87.8\footnotesize{$\pm$3.4} & 81.5\footnotesize{$\pm$5.1} & 52.6\footnotesize{$\pm$3.4} & 39.7\footnotesize{$\pm$3.3} & 50.5\footnotesize{$\pm$3.9} & 37.4\footnotesize{$\pm$2.6} \\
cLSTM & 93.1\footnotesize{$\pm$0.4} & 87.9\footnotesize{$\pm$0.9} & 60.7\footnotesize{$\pm$6.9} & 48.7\footnotesize{$\pm$6.9} & 49.0\footnotesize{$\pm$6.5} & 37.8\footnotesize{$\pm$4.6} \\
CUTS & 90.2\footnotesize{$\pm$2.1} & 76.0\footnotesize{$\pm$6.1} & 56.7\footnotesize{$\pm$5.0} & 44.1\footnotesize{$\pm$5.2} & \textbf{54.7\footnotesize{$\pm$4.8}} & \textbf{38.8\footnotesize{$\pm$4.0}} \\
CUTS+ & \textbf{99.8\footnotesize{$\pm$0.1}} & \textbf{99.7\footnotesize{$\pm$0.2}} & \textbf{78.8\footnotesize{$\pm$4.8}} & \textbf{67.9\footnotesize{$\pm$6.6}} & 51.9\footnotesize{$\pm$6.9} & 38.1\footnotesize{$\pm$6.7} \\
\bottomrule
\end{tabular}
\end{table*}

\section{Table results under different levels of nonstationarity}
\label{app_non}

Tables~\ref{tab:nonstationarity_var_T500}--\ref{tab:nonstationarity_lorenz_T1000_F40} present results under different levels of nonstationarity. We consider both moderate nonstationarity (linear: $m = 1$, $\nu = 1$; nonlinear with $F = 10$: $m = 2.5$, $\nu = 2.0$; nonlinear with $F = 40$: $m = 3.5$, $\nu = 2.0$) and stronger nonstationarity (linear: $m = 1.8$, $\nu = 1.5$; nonlinear with $F = 10$: $m = 4.2$, $\nu = 0.3$; nonlinear with $F = 40$: $m = 5.2$, $\nu = 0.3$). The results indicate that, overall, method performance generally degrades as the degree of nonstationarity increases, while the best-performing methods across these settings are predominantly deep learning-based approaches.

\begin{table*}[!htbp]
\centering
\caption{Linear setting under different levels of nonstationarity ($T = 500$).}
\label{tab:nonstationarity_var_T500}
\small
\setlength{\tabcolsep}{4pt}
\begin{tabular}{ccccccc}
\toprule
& \multicolumn{2}{c}{Vanilla} & \multicolumn{2}{c}{Nonstationarity} & \multicolumn{2}{c}{Stronger nonstationarity} \\
\cmidrule(lr){2-3} \cmidrule(lr){4-5} \cmidrule(lr){6-7}
10 nodes & AUROC$\uparrow$ & AUPRC$\uparrow$ & AUROC$\uparrow$ & AUPRC$\uparrow$ & AUROC$\uparrow$ & AUPRC$\uparrow$ \\
\midrule
VAR & 95.2\footnotesize{$\pm$1.0} & 57.3\footnotesize{$\pm$5.0} & 78.8\footnotesize{$\pm$8.2} & 25.4\footnotesize{$\pm$8.7} & 56.7\footnotesize{$\pm$8.7} & 22.0\footnotesize{$\pm$14.3} \\
LGC & 97.6\footnotesize{$\pm$1.1} & 73.6\footnotesize{$\pm$9.1} & 90.4\footnotesize{$\pm$9.9} & 59.3\footnotesize{$\pm$31.2} & 80.0\footnotesize{$\pm$10.0} & 27.4\footnotesize{$\pm$9.7} \\
VARLiNGAM & 53.9\footnotesize{$\pm$6.1} & 12.7\footnotesize{$\pm$2.0} & 50.0\footnotesize{$\pm$9.3} & 12.2\footnotesize{$\pm$1.9} & 48.5\footnotesize{$\pm$8.7} & 11.6\footnotesize{$\pm$1.7} \\
PCMCI & 97.8\footnotesize{$\pm$0.8} & 75.3\footnotesize{$\pm$6.8} & 83.5\footnotesize{$\pm$8.1} & 35.4\footnotesize{$\pm$16.7} & 77.1\footnotesize{$\pm$9.4} & 26.4\footnotesize{$\pm$12.6} \\
DYNOTEARS & \textbf{100.0\footnotesize{$\pm$0.0}} & \textbf{100.0\footnotesize{$\pm$0.0}} & 92.2\footnotesize{$\pm$4.8} & 66.3\footnotesize{$\pm$22.2} & 78.9\footnotesize{$\pm$7.9} & 29.9\footnotesize{$\pm$11.5} \\
TSCI & 64.5\footnotesize{$\pm$12.1} & 23.5\footnotesize{$\pm$10.4} & 52.4\footnotesize{$\pm$10.9} & 19.6\footnotesize{$\pm$8.2} & 55.5\footnotesize{$\pm$4.4} & 20.5\footnotesize{$\pm$4.3} \\
cMLP & 99.9\footnotesize{$\pm$0.0} & 99.6\footnotesize{$\pm$0.4} & 93.7\footnotesize{$\pm$4.0} & 72.8\footnotesize{$\pm$14.1} & 89.3\footnotesize{$\pm$5.6} & 55.4\footnotesize{$\pm$14.5} \\
cLSTM & \textbf{100.0\footnotesize{$\pm$0.0}} & \textbf{100.0\footnotesize{$\pm$0.0}} & 93.9\footnotesize{$\pm$6.3} & 80.7\footnotesize{$\pm$22.3} & 82.9\footnotesize{$\pm$7.8} & 52.2\footnotesize{$\pm$23.8} \\
CUTS & 92.8\footnotesize{$\pm$8.5} & 76.0\footnotesize{$\pm$12.2} & 95.2\footnotesize{$\pm$4.9} & 81.3\footnotesize{$\pm$15.6} & 96.5\footnotesize{$\pm$3.0} & \textbf{85.1\footnotesize{$\pm$13.3}} \\
CUTS+ & 99.6\footnotesize{$\pm$0.3} & 97.3\footnotesize{$\pm$2.7} & \textbf{99.0\footnotesize{$\pm$0.4}} & \textbf{93.1\footnotesize{$\pm$3.8}} & \textbf{97.2\footnotesize{$\pm$1.6}} & 84.4\footnotesize{$\pm$7.2} \\
\bottomrule
\end{tabular}
\end{table*}

\begin{table*}[!htbp]
\centering
\caption{Linear setting under different levels of nonstationarity ($T = 1000$).}
\label{tab:nonstationarity_var_T1000}
\small
\setlength{\tabcolsep}{4pt}
\begin{tabular}{ccccccc}
\toprule
& \multicolumn{2}{c}{Vanilla} & \multicolumn{2}{c}{Nonstationarity} & \multicolumn{2}{c}{Stronger nonstationarity} \\
\cmidrule(lr){2-3} \cmidrule(lr){4-5} \cmidrule(lr){6-7}
10 nodes & AUROC$\uparrow$ & AUPRC$\uparrow$ & AUROC$\uparrow$ & AUPRC$\uparrow$ & AUROC$\uparrow$ & AUPRC$\uparrow$ \\
\midrule
VAR & 99.7\footnotesize{$\pm$0.3} & 96.3\footnotesize{$\pm$4.4} & 92.7\footnotesize{$\pm$2.6} & 48.4\footnotesize{$\pm$10.9} & 88.7\footnotesize{$\pm$5.4} & 40.2\footnotesize{$\pm$14.5} \\
LGC & \textbf{100.0\footnotesize{$\pm$0.0}} & \textbf{100.0\footnotesize{$\pm$0.0}} & 97.7\footnotesize{$\pm$2.2} & 78.6\footnotesize{$\pm$19.1} & 94.3\footnotesize{$\pm$3.0} & 56.5\footnotesize{$\pm$15.1} \\
VARLiNGAM & 48.2\footnotesize{$\pm$5.5} & 11.8\footnotesize{$\pm$0.8} & 47.6\footnotesize{$\pm$5.7} & 11.6\footnotesize{$\pm$0.9} & 46.7\footnotesize{$\pm$7.1} & 11.3\footnotesize{$\pm$0.9} \\
PCMCI & 98.2\footnotesize{$\pm$0.9} & 79.1\footnotesize{$\pm$8.8} & 85.1\footnotesize{$\pm$5.9} & 32.2\footnotesize{$\pm$9.5} & 78.4\footnotesize{$\pm$7.8} & 24.4\footnotesize{$\pm$6.9} \\
DYNOTEARS & \textbf{100.0\footnotesize{$\pm$0.0}} & \textbf{100.0\footnotesize{$\pm$0.0}} & 94.8\footnotesize{$\pm$5.8} & 80.4\footnotesize{$\pm$16.9} & 92.0\footnotesize{$\pm$6.4} & 52.0\footnotesize{$\pm$19.6} \\
TSCI & 58.8\footnotesize{$\pm$6.7} & 27.0\footnotesize{$\pm$9.3} & 68.5\footnotesize{$\pm$11.3} & 29.4\footnotesize{$\pm$11.7} & 70.3\footnotesize{$\pm$10.2} & 32.5\footnotesize{$\pm$6.2} \\
cMLP & 99.9\footnotesize{$\pm$0.0} & 99.8\footnotesize{$\pm$0.3} & 81.8\footnotesize{$\pm$22.1} & 67.2\footnotesize{$\pm$38.9} & 73.6\footnotesize{$\pm$15.3} & 38.3\footnotesize{$\pm$29.6} \\
cLSTM & \textbf{100.0\footnotesize{$\pm$0.0}} & \textbf{100.0\footnotesize{$\pm$0.0}} & 99.4\footnotesize{$\pm$0.5} & 95.8\footnotesize{$\pm$5.0} & 77.1\footnotesize{$\pm$12.8} & 48.5\footnotesize{$\pm$29.9} \\
CUTS & 96.8\footnotesize{$\pm$1.3} & 80.8\footnotesize{$\pm$8.9} & 98.6\footnotesize{$\pm$1.9} & 93.8\footnotesize{$\pm$8.9} & 99.4\footnotesize{$\pm$0.5} & 95.5\footnotesize{$\pm$4.8} \\
CUTS+ & \textbf{100.0\footnotesize{$\pm$0.0}} & \textbf{100.0\footnotesize{$\pm$0.0}} & \textbf{99.9\footnotesize{$\pm$0.1}} & \textbf{99.4\footnotesize{$\pm$0.6}} & \textbf{99.7\footnotesize{$\pm$0.3}} & \textbf{98.2\footnotesize{$\pm$2.0}} \\
\bottomrule
\end{tabular}
\end{table*}

\begin{table*}[!htbp]
\centering
\caption{Nonlinear setting under different levels of nonstationarity ($T = 500$, $F = 10$).}
\label{tab:nonstationarity_lorenz_T500_F10}
\small
\setlength{\tabcolsep}{4pt}
\begin{tabular}{ccccccc}
\toprule
& \multicolumn{2}{c}{Vanilla} & \multicolumn{2}{c}{Nonstationarity} & \multicolumn{2}{c}{Stronger nonstationarity} \\
\cmidrule(lr){2-3} \cmidrule(lr){4-5} \cmidrule(lr){6-7}
10 nodes & AUROC$\uparrow$ & AUPRC$\uparrow$ & AUROC$\uparrow$ & AUPRC$\uparrow$ & AUROC$\uparrow$ & AUPRC$\uparrow$ \\
\midrule
VAR & 77.0\footnotesize{$\pm$3.1} & 55.5\footnotesize{$\pm$4.3} & 66.8\footnotesize{$\pm$9.2} & 46.8\footnotesize{$\pm$8.0} & 61.5\footnotesize{$\pm$5.6} & 42.6\footnotesize{$\pm$5.1} \\
LGC & 76.9\footnotesize{$\pm$2.4} & 62.9\footnotesize{$\pm$3.8} & 68.5\footnotesize{$\pm$11.5} & 52.7\footnotesize{$\pm$15.0} & 61.3\footnotesize{$\pm$5.6} & 42.7\footnotesize{$\pm$5.1} \\
VARLiNGAM & 72.1\footnotesize{$\pm$6.4} & 61.1\footnotesize{$\pm$9.4} & 60.5\footnotesize{$\pm$11.5} & 44.9\footnotesize{$\pm$13.1} & 54.1\footnotesize{$\pm$4.6} & 36.4\footnotesize{$\pm$3.2} \\
PCMCI & 71.3\footnotesize{$\pm$5.2} & 50.4\footnotesize{$\pm$7.0} & 65.0\footnotesize{$\pm$8.1} & 46.1\footnotesize{$\pm$7.3} & 66.3\footnotesize{$\pm$6.8} & 46.3\footnotesize{$\pm$6.4} \\
DYNOTEARS & 75.8\footnotesize{$\pm$6.9} & 65.9\footnotesize{$\pm$10.5} & 64.6\footnotesize{$\pm$10.6} & 47.7\footnotesize{$\pm$13.1} & 56.1\footnotesize{$\pm$3.1} & 37.0\footnotesize{$\pm$2.1} \\
TSCI & 86.3\footnotesize{$\pm$2.7} & 77.3\footnotesize{$\pm$3.3} & 71.0\footnotesize{$\pm$11.2} & 56.6\footnotesize{$\pm$13.2} & 56.7\footnotesize{$\pm$7.4} & 41.5\footnotesize{$\pm$8.1} \\
cMLP & 99.7\footnotesize{$\pm$0.1} & 99.5\footnotesize{$\pm$0.2} & 72.1\footnotesize{$\pm$22.4} & 64.9\footnotesize{$\pm$28.0} & 54.4\footnotesize{$\pm$5.5} & 39.8\footnotesize{$\pm$1.8} \\
cLSTM & \textbf{100.0\footnotesize{$\pm$0.0}} & \textbf{99.9\footnotesize{$\pm$0.0}} & 74.1\footnotesize{$\pm$21.5} & 67.5\footnotesize{$\pm$27.2} & 63.3\footnotesize{$\pm$10.2} & 52.8\footnotesize{$\pm$10.6} \\
CUTS & 93.6\footnotesize{$\pm$1.7} & 91.6\footnotesize{$\pm$2.3} & 66.8\footnotesize{$\pm$20.6} & 58.8\footnotesize{$\pm$24.2} & 57.0\footnotesize{$\pm$9.5} & 44.4\footnotesize{$\pm$7.5} \\
CUTS+ & 99.7\footnotesize{$\pm$0.2} & 99.4\footnotesize{$\pm$0.5} & \textbf{80.1\footnotesize{$\pm$19.1}} & \textbf{74.1\footnotesize{$\pm$22.3}} & \textbf{75.0\footnotesize{$\pm$14.5}} & \textbf{65.3\footnotesize{$\pm$17.0}} \\
\bottomrule
\end{tabular}
\end{table*}

\begin{table*}[!htbp]
\centering
\caption{Nonlinear setting under different levels of nonstationarity ($T = 1000$, $F = 10$).}
\label{tab:nonstationarity_lorenz_T1000_F10}
\small
\setlength{\tabcolsep}{4pt}
\begin{tabular}{ccccccc}
\toprule
& \multicolumn{2}{c}{Vanilla} & \multicolumn{2}{c}{Nonstationarity} & \multicolumn{2}{c}{Stronger nonstationarity} \\
\cmidrule(lr){2-3} \cmidrule(lr){4-5} \cmidrule(lr){6-7}
10 nodes & AUROC$\uparrow$ & AUPRC$\uparrow$ & AUROC$\uparrow$ & AUPRC$\uparrow$ & AUROC$\uparrow$ & AUPRC$\uparrow$ \\
\midrule
VAR & 83.3\footnotesize{$\pm$3.6} & 66.6\footnotesize{$\pm$5.6} & 65.1\footnotesize{$\pm$14.9} & 48.3\footnotesize{$\pm$14.6} & 67.8\footnotesize{$\pm$5.8} & 45.2\footnotesize{$\pm$5.0} \\
LGC & 83.8\footnotesize{$\pm$2.6} & 72.4\footnotesize{$\pm$4.5} & 71.1\footnotesize{$\pm$14.4} & 50.7\footnotesize{$\pm$12.8} & 68.3\footnotesize{$\pm$6.2} & 45.8\footnotesize{$\pm$5.5} \\
VARLiNGAM & 72.0\footnotesize{$\pm$1.9} & 60.3\footnotesize{$\pm$2.7} & 63.0\footnotesize{$\pm$12.6} & 45.1\footnotesize{$\pm$10.8} & 57.9\footnotesize{$\pm$5.3} & 39.0\footnotesize{$\pm$3.8} \\
PCMCI & 77.8\footnotesize{$\pm$4.9} & 55.9\footnotesize{$\pm$6.1} & 70.3\footnotesize{$\pm$13.2} & 52.6\footnotesize{$\pm$14.0} & 69.3\footnotesize{$\pm$9.2} & 50.0\footnotesize{$\pm$10.2} \\
DYNOTEARS & 86.1\footnotesize{$\pm$2.1} & 71.2\footnotesize{$\pm$3.4} & 67.3\footnotesize{$\pm$16.9} & 49.2\footnotesize{$\pm$17.6} & 56.1\footnotesize{$\pm$5.6} & 39.4\footnotesize{$\pm$6.1} \\
TSCI & 88.4\footnotesize{$\pm$1.7} & 80.5\footnotesize{$\pm$2.2} & 74.7\footnotesize{$\pm$11.5} & 60.2\footnotesize{$\pm$15.6} & 60.9\footnotesize{$\pm$7.2} & 45.9\footnotesize{$\pm$8.4} \\
cMLP & 99.9\footnotesize{$\pm$0.0} & 99.9\footnotesize{$\pm$0.0} & 73.8\footnotesize{$\pm$22.8} & 65.4\footnotesize{$\pm$29.6} & 54.7\footnotesize{$\pm$7.3} & 41.2\footnotesize{$\pm$8.4} \\
cLSTM & \textbf{100.0\footnotesize{$\pm$0.0}} & \textbf{99.9\footnotesize{$\pm$0.0}} & 75.4\footnotesize{$\pm$22.4} & 67.2\footnotesize{$\pm$28.9} & 60.8\footnotesize{$\pm$15.7} & 49.7\footnotesize{$\pm$18.4} \\
CUTS & 98.3\footnotesize{$\pm$1.1} & 97.5\footnotesize{$\pm$1.6} & 70.7\footnotesize{$\pm$22.9} & 62.8\footnotesize{$\pm$27.7} & 62.5\footnotesize{$\pm$10.6} & 50.7\footnotesize{$\pm$6.8} \\
CUTS+ & 99.9\footnotesize{$\pm$0.0} & 99.9\footnotesize{$\pm$0.1} & \textbf{83.4\footnotesize{$\pm$19.3}} & \textbf{78.0\footnotesize{$\pm$24.9}} & \textbf{86.3\footnotesize{$\pm$9.6}} & \textbf{83.8\footnotesize{$\pm$10.0}} \\
\bottomrule
\end{tabular}
\end{table*}

\begin{table*}[!htbp]
\centering
\caption{Nonlinear setting under different levels of nonstationarity ($T = 500$, $F = 40$).}
\label{tab:nonstationarity_lorenz_T500_F40}
\small
\setlength{\tabcolsep}{4pt}
\begin{tabular}{ccccccc}
\toprule
& \multicolumn{2}{c}{Vanilla} & \multicolumn{2}{c}{Nonstationarity} & \multicolumn{2}{c}{Stronger nonstationarity} \\
\cmidrule(lr){2-3} \cmidrule(lr){4-5} \cmidrule(lr){6-7}
10 nodes & AUROC$\uparrow$ & AUPRC$\uparrow$ & AUROC$\uparrow$ & AUPRC$\uparrow$ & AUROC$\uparrow$ & AUPRC$\uparrow$ \\
\midrule
VAR & 68.8\footnotesize{$\pm$4.1} & 48.3\footnotesize{$\pm$4.2} & 60.6\footnotesize{$\pm$9.1} & 42.3\footnotesize{$\pm$7.8} & 56.1\footnotesize{$\pm$4.6} & 37.6\footnotesize{$\pm$3.4} \\
LGC & 68.5\footnotesize{$\pm$3.7} & 48.0\footnotesize{$\pm$3.8} & 60.6\footnotesize{$\pm$9.1} & 42.3\footnotesize{$\pm$7.8} & 55.8\footnotesize{$\pm$4.3} & 37.3\footnotesize{$\pm$3.2} \\
VARLiNGAM & 61.8\footnotesize{$\pm$3.8} & 42.0\footnotesize{$\pm$3.4} & 56.0\footnotesize{$\pm$4.3} & 37.8\footnotesize{$\pm$3.0} & 51.5\footnotesize{$\pm$2.0} & 35.2\footnotesize{$\pm$2.5} \\
PCMCI & 66.1\footnotesize{$\pm$5.5} & 47.9\footnotesize{$\pm$7.6} & 60.1\footnotesize{$\pm$8.9} & 43.1\footnotesize{$\pm$10.4} & 57.3\footnotesize{$\pm$6.7} & 38.4\footnotesize{$\pm$5.1} \\
DYNOTEARS & 66.0\footnotesize{$\pm$2.5} & 43.6\footnotesize{$\pm$2.0} & 58.6\footnotesize{$\pm$7.6} & 39.1\footnotesize{$\pm$4.6} & 55.1\footnotesize{$\pm$4.0} & 36.5\footnotesize{$\pm$2.5} \\
TSCI & 60.4\footnotesize{$\pm$3.2} & 46.8\footnotesize{$\pm$5.8} & 55.6\footnotesize{$\pm$8.3} & 40.4\footnotesize{$\pm$8.0} & 49.7\footnotesize{$\pm$3.5} & 36.3\footnotesize{$\pm$3.9} \\
cMLP & 83.0\footnotesize{$\pm$3.6} & 72.5\footnotesize{$\pm$4.6} & 69.1\footnotesize{$\pm$21.8} & 61.5\footnotesize{$\pm$26.9} & 57.9\footnotesize{$\pm$11.9} & 46.7\footnotesize{$\pm$14.8} \\
cLSTM & 94.2\footnotesize{$\pm$0.4} & 89.3\footnotesize{$\pm$0.3} & 70.9\footnotesize{$\pm$20.1} & 63.3\footnotesize{$\pm$21.7} & 57.2\footnotesize{$\pm$8.1} & 45.5\footnotesize{$\pm$10.6} \\
CUTS & 85.8\footnotesize{$\pm$3.7} & 69.3\footnotesize{$\pm$6.9} & 69.4\footnotesize{$\pm$15.3} & 56.4\footnotesize{$\pm$15.2} & 55.7\footnotesize{$\pm$7.2} & 45.0\footnotesize{$\pm$9.5} \\
CUTS+ & \textbf{98.8\footnotesize{$\pm$1.0}} & \textbf{97.9\footnotesize{$\pm$1.8}} & \textbf{77.3\footnotesize{$\pm$19.6}} & \textbf{71.5\footnotesize{$\pm$23.8}} & \textbf{63.5\footnotesize{$\pm$9.5}} & \textbf{53.2\footnotesize{$\pm$6.4}} \\
\bottomrule
\end{tabular}
\end{table*}

\begin{table*}[!htbp]
\centering
\caption{Nonlinear setting under different levels of nonstationarity ($T = 1000$, $F = 40$).}
\label{tab:nonstationarity_lorenz_T1000_F40}
\small
\setlength{\tabcolsep}{4pt}
\begin{tabular}{ccccccc}
\toprule
& \multicolumn{2}{c}{Vanilla} & \multicolumn{2}{c}{Nonstationarity} & \multicolumn{2}{c}{Stronger nonstationarity} \\
\cmidrule(lr){2-3} \cmidrule(lr){4-5} \cmidrule(lr){6-7}
10 nodes & AUROC$\uparrow$ & AUPRC$\uparrow$ & AUROC$\uparrow$ & AUPRC$\uparrow$ & AUROC$\uparrow$ & AUPRC$\uparrow$ \\
\midrule
VAR & 73.0\footnotesize{$\pm$2.0} & 58.5\footnotesize{$\pm$3.1} & 60.5\footnotesize{$\pm$9.7} & 44.7\footnotesize{$\pm$11.1} & 59.5\footnotesize{$\pm$7.8} & 39.1\footnotesize{$\pm$5.2} \\
LGC & 72.8\footnotesize{$\pm$2.2} & 58.7\footnotesize{$\pm$4.2} & 60.5\footnotesize{$\pm$9.7} & 44.7\footnotesize{$\pm$11.1} & 59.8\footnotesize{$\pm$7.9} & 39.3\footnotesize{$\pm$5.5} \\
VARLiNGAM & 65.5\footnotesize{$\pm$2.7} & 45.0\footnotesize{$\pm$3.0} & 54.8\footnotesize{$\pm$7.6} & 37.1\footnotesize{$\pm$5.1} & 51.8\footnotesize{$\pm$2.9} & 34.9\footnotesize{$\pm$1.8} \\
PCMCI & 75.5\footnotesize{$\pm$1.5} & 59.8\footnotesize{$\pm$4.0} & 62.5\footnotesize{$\pm$10.3} & 46.1\footnotesize{$\pm$12.4} & 58.8\footnotesize{$\pm$6.1} & 39.0\footnotesize{$\pm$4.7} \\
DYNOTEARS & 69.6\footnotesize{$\pm$2.8} & 50.1\footnotesize{$\pm$3.5} & 59.0\footnotesize{$\pm$9.7} & 41.2\footnotesize{$\pm$7.8} & 53.6\footnotesize{$\pm$1.9} & 36.4\footnotesize{$\pm$1.5} \\
TSCI & 71.3\footnotesize{$\pm$4.6} & 56.1\footnotesize{$\pm$4.6} & 60.5\footnotesize{$\pm$7.7} & 48.6\footnotesize{$\pm$7.6} & 54.0\footnotesize{$\pm$3.3} & 39.7\footnotesize{$\pm$3.5} \\
cMLP & 87.8\footnotesize{$\pm$3.4} & 81.5\footnotesize{$\pm$5.1} & 74.5\footnotesize{$\pm$20.8} & 66.3\footnotesize{$\pm$27.7} & 60.4\footnotesize{$\pm$11.5} & 44.7\footnotesize{$\pm$11.8} \\
cLSTM & 93.1\footnotesize{$\pm$0.4} & 87.9\footnotesize{$\pm$0.9} & 72.0\footnotesize{$\pm$20.9} & 63.9\footnotesize{$\pm$24.6} & 60.7\footnotesize{$\pm$9.0} & 48.5\footnotesize{$\pm$11.1} \\
CUTS & 90.2\footnotesize{$\pm$2.1} & 76.0\footnotesize{$\pm$6.1} & 70.0\footnotesize{$\pm$19.9} & 56.1\footnotesize{$\pm$17.8} & 57.7\footnotesize{$\pm$4.6} & 43.5\footnotesize{$\pm$4.5} \\
CUTS+ & \textbf{99.8\footnotesize{$\pm$0.1}} & \textbf{99.7\footnotesize{$\pm$0.2}} & \textbf{80.1\footnotesize{$\pm$20.0}} & \textbf{75.2\footnotesize{$\pm$24.9}} & \textbf{67.3\footnotesize{$\pm$13.7}} & \textbf{57.2\footnotesize{$\pm$16.5}} \\
\bottomrule
\end{tabular}
\end{table*}

\section{Table results under time-varying coefficient nonstationarity}
\label{app_non_coeff}

In the main text, we consider nonstationarity induced by time-varying noise variance. Building on this setting, we further investigate a nonstationary scenario in which the coefficients of the linear vanilla model also vary over time. Table~\ref{tab:tvcoef_VAR_T1000} reports the results under coefficient nonstationarity with $\sigma_{\text{TV}} = 0.3$ and $\sigma_{\text{TV}} = 1.0$. The parameter $\sigma_{\text{TV}}$ controls the strength of coefficient nonstationarity, with larger values corresponding to stronger nonstationarity. The results indicate that, overall, method performance generally degrades as the degree of nonstationarity increases, while the best-performing methods across these settings are predominantly deep learning-based approaches.

\begin{table*}[!htbp]
\centering
\caption{Linear setting under nonstationary time-varying coefficients ($T = 1000$).}
\label{tab:tvcoef_VAR_T1000}
\small
\setlength{\tabcolsep}{4pt}
\begin{tabular}{ccccccc}
\toprule
& \multicolumn{2}{c}{Vanilla} & \multicolumn{2}{c}{TV Coef ($\sigma_{\text{TV}}=0.3$)} & \multicolumn{2}{c}{TV Coef ($\sigma_{\text{TV}}=1.0$)} \\
\cmidrule(lr){2-3} \cmidrule(lr){4-5} \cmidrule(lr){6-7}
10 nodes & AUROC$\uparrow$ & AUPRC$\uparrow$ & AUROC$\uparrow$ & AUPRC$\uparrow$ & AUROC$\uparrow$ & AUPRC$\uparrow$ \\
\midrule
VAR & 99.7\footnotesize{$\pm$0.3} & 96.3\footnotesize{$\pm$4.4} & 89.5\footnotesize{$\pm$5.2} & 47.5\footnotesize{$\pm$17.2} & 82.5\footnotesize{$\pm$8.0} & 29.7\footnotesize{$\pm$11.0} \\
LGC & \textbf{100.0\footnotesize{$\pm$0.0}} & \textbf{100.0\footnotesize{$\pm$0.0}} & 88.8\footnotesize{$\pm$5.8} & 58.3\footnotesize{$\pm$15.2} & 79.5\footnotesize{$\pm$12.0} & 31.8\footnotesize{$\pm$15.6} \\
VARLiNGAM & 48.2\footnotesize{$\pm$5.5} & 11.8\footnotesize{$\pm$0.8} & 51.6\footnotesize{$\pm$9.7} & 13.2\footnotesize{$\pm$2.7} & 64.1\footnotesize{$\pm$6.4} & 17.2\footnotesize{$\pm$5.2} \\
PCMCI & 98.2\footnotesize{$\pm$0.9} & 79.1\footnotesize{$\pm$8.8} & 84.5\footnotesize{$\pm$6.6} & 31.3\footnotesize{$\pm$9.0} & 77.7\footnotesize{$\pm$14.2} & 28.3\footnotesize{$\pm$12.9} \\
DYNOTEARS & \textbf{100.0\footnotesize{$\pm$0.0}} & \textbf{100.0\footnotesize{$\pm$0.0}} & 88.5\footnotesize{$\pm$5.3} & 46.2\footnotesize{$\pm$13.6} & 73.2\footnotesize{$\pm$12.1} & 33.8\footnotesize{$\pm$24.6} \\
TSCI & 58.8\footnotesize{$\pm$6.7} & 27.0\footnotesize{$\pm$9.3} & 61.3\footnotesize{$\pm$14.6} & 28.2\footnotesize{$\pm$14.1} & 63.4\footnotesize{$\pm$7.5} & 28.5\footnotesize{$\pm$7.9} \\
cMLP & 99.9\footnotesize{$\pm$0.0} & 99.8\footnotesize{$\pm$0.3} & 83.8\footnotesize{$\pm$2.8} & 53.7\footnotesize{$\pm$6.4} & \textbf{84.6\footnotesize{$\pm$3.2}} & \textbf{54.2\footnotesize{$\pm$9.0}} \\
cLSTM & \textbf{100.0\footnotesize{$\pm$0.0}} & \textbf{100.0\footnotesize{$\pm$0.0}} & 74.0\footnotesize{$\pm$9.4} & 38.6\footnotesize{$\pm$20.8} & 63.9\footnotesize{$\pm$15.3} & 33.8\footnotesize{$\pm$22.2} \\
CUTS & 96.8\footnotesize{$\pm$1.3} & 80.8\footnotesize{$\pm$8.9} & \textbf{90.4\footnotesize{$\pm$7.3}} & \textbf{84.2\footnotesize{$\pm$10.8}} & 77.7\footnotesize{$\pm$14.5} & 49.3\footnotesize{$\pm$29.0} \\
CUTS+ & \textbf{100.0\footnotesize{$\pm$0.0}} & \textbf{100.0\footnotesize{$\pm$0.0}} & 89.0\footnotesize{$\pm$6.2} & 70.7\footnotesize{$\pm$12.7} & 74.2\footnotesize{$\pm$17.4} & 50.2\footnotesize{$\pm$23.1} \\
\bottomrule
\end{tabular}
\end{table*}

\section{Table results under different levels of confounding}
\label{app_confounder}

Table~\ref{tab:confounder_Lorenz_T1000_F10} reports the results under latent confounding with $\zeta = 0.5$ and $\zeta = 0.9$. Larger values of $\zeta$ correspond to stronger latent confounding. The results indicate that, overall, method performance generally degrades as the severity of model assumption violations increases, while the best-performing methods across these settings are predominantly deep learning-based approaches.

\begin{table*}[!htbp]
\centering
\caption{Nonlinear setting under different levels of confounding ($T = 1000$, $F = 10$).}
\label{tab:confounder_Lorenz_T1000_F10}
\small
\setlength{\tabcolsep}{4pt}
\begin{tabular}{ccccccc}
\toprule
& \multicolumn{2}{c}{Vanilla} & \multicolumn{2}{c}{Confounders ($\zeta=0.5$)} & \multicolumn{2}{c}{Confounders ($\zeta=0.9$)} \\
\cmidrule(lr){2-3} \cmidrule(lr){4-5} \cmidrule(lr){6-7}
10 nodes & AUROC$\uparrow$ & AUPRC$\uparrow$ & AUROC$\uparrow$ & AUPRC$\uparrow$ & AUROC$\uparrow$ & AUPRC$\uparrow$ \\
\midrule
VAR & 83.3\footnotesize{$\pm$3.6} & 66.6\footnotesize{$\pm$5.6} & 59.8\footnotesize{$\pm$9.5} & 39.7\footnotesize{$\pm$5.9} & 54.0\footnotesize{$\pm$3.4} & 38.4\footnotesize{$\pm$4.2} \\
LGC & 83.8\footnotesize{$\pm$2.6} & 72.4\footnotesize{$\pm$4.5} & 60.0\footnotesize{$\pm$9.6} & 39.9\footnotesize{$\pm$6.0} & 54.0\footnotesize{$\pm$3.4} & 38.4\footnotesize{$\pm$4.2} \\
VARLiNGAM & 72.0\footnotesize{$\pm$1.9} & 60.3\footnotesize{$\pm$2.7} & 61.9\footnotesize{$\pm$2.7} & 42.4\footnotesize{$\pm$2.7} & 57.1\footnotesize{$\pm$1.9} & 38.3\footnotesize{$\pm$1.7} \\
PCMCI & 77.8\footnotesize{$\pm$4.9} & 55.9\footnotesize{$\pm$6.1} & 61.3\footnotesize{$\pm$7.1} & 41.5\footnotesize{$\pm$6.0} & 55.8\footnotesize{$\pm$5.8} & 38.2\footnotesize{$\pm$4.4} \\
DYNOTEARS & 86.1\footnotesize{$\pm$2.1} & 71.2\footnotesize{$\pm$3.4} & 61.8\footnotesize{$\pm$2.9} & 42.7\footnotesize{$\pm$3.8} & 57.3\footnotesize{$\pm$3.3} & 38.7\footnotesize{$\pm$3.1} \\
TSCI & 88.4\footnotesize{$\pm$1.7} & 80.5\footnotesize{$\pm$2.2} & 61.1\footnotesize{$\pm$6.2} & 45.3\footnotesize{$\pm$3.6} & 57.0\footnotesize{$\pm$6.6} & 43.5\footnotesize{$\pm$8.9} \\
cMLP & 99.9\footnotesize{$\pm$0.0} & 99.9\footnotesize{$\pm$0.0} & 66.5\footnotesize{$\pm$7.6} & 52.0\footnotesize{$\pm$10.3} & 59.3\footnotesize{$\pm$3.8} & 45.8\footnotesize{$\pm$5.3} \\
cLSTM & \textbf{100.0\footnotesize{$\pm$0.0}} & \textbf{99.9\footnotesize{$\pm$0.0}} & 91.7\footnotesize{$\pm$1.9} & 86.8\footnotesize{$\pm$1.4} & 85.4\footnotesize{$\pm$6.1} & 77.7\footnotesize{$\pm$8.5} \\
CUTS & 98.3\footnotesize{$\pm$1.1} & 97.5\footnotesize{$\pm$1.6} & 78.5\footnotesize{$\pm$7.6} & 57.8\footnotesize{$\pm$10.1} & 60.9\footnotesize{$\pm$7.0} & 38.9\footnotesize{$\pm$5.3} \\
CUTS+ & 99.9\footnotesize{$\pm$0.0} & 99.9\footnotesize{$\pm$0.1} & \textbf{96.7\footnotesize{$\pm$3.7}} & \textbf{93.9\footnotesize{$\pm$7.4}} & \textbf{91.6\footnotesize{$\pm$3.3}} & \textbf{87.3\footnotesize{$\pm$6.4}} \\
\bottomrule
\end{tabular}
\end{table*}

\clearpage

\section{Table results under different imputation strategies}
\label{app_imputation}

In the main text, we adopt zero-order hold (ZOH) interpolation to impute missing entries before applying causal discovery, as described in Section~\ref{sec:misspecified}. 
The ``imputation + causal discovery'' pipeline serves as an operational proxy for evaluating how methods without native missing-data support perform in the presence of missing data.
To demonstrate that the choice of imputation strategy does not substantially affect the main conclusions, we additionally evaluate benchmark methods under two alternative imputation approaches: Gaussian process (GP) interpolation and linear interpolation. 
Tables~\ref{tab:imputation_VAR_T1000} and~\ref{tab:imputation_Lorenz_T1000_F10} report results under MCAR missingness with $\gamma = 0.4$ for the linear and nonlinear settings, respectively. The column ``ImputeZOH'' corresponds to ZOH imputation (used in the main text), while ``ImputeGP'' and ``ImputeLinear'' denote GP and linear interpolation, respectively. The results show that deep learning-based methods achieve better overall performance across the three imputation strategies, confirming that the main conclusions are robust to the choice of imputation method.

\begin{table*}[!htbp]
	\centering
	\caption{Linear setting, 10-node case with $T = 1000$ under MCAR missingness ($\gamma=0.4$) and different imputation strategies.}
	\label{tab:imputation_VAR_T1000}
	\small
	\setlength{\tabcolsep}{4pt}
	\begin{tabular}{ccccccc}
		\toprule
		& \multicolumn{2}{c}{ImputeZOH} & \multicolumn{2}{c}{ImputeGP} & \multicolumn{2}{c}{ImputeLinear} \\
		\cmidrule(lr){2-3} \cmidrule(lr){4-5} \cmidrule(lr){6-7}
		10 nodes & AUROC$\uparrow$ & AUPRC$\uparrow$ & AUROC$\uparrow$ & AUPRC$\uparrow$ & AUROC$\uparrow$ & AUPRC$\uparrow$ \\
		\midrule
		VAR & 81.8\footnotesize{$\pm$10.5} & 58.3\footnotesize{$\pm$21.7} & 93.6\footnotesize{$\pm$1.7} & 91.6\footnotesize{$\pm$0.8} & 81.8\footnotesize{$\pm$10.5} & 58.3\footnotesize{$\pm$21.7} \\
		LGC & 76.0\footnotesize{$\pm$5.8} & 57.3\footnotesize{$\pm$10.3} & 98.6\footnotesize{$\pm$0.3} & 93.8\footnotesize{$\pm$0.7} & 76.0\footnotesize{$\pm$5.8} & 57.3\footnotesize{$\pm$10.3} \\
		VARLiNGAM & 53.1\footnotesize{$\pm$9.0} & 13.0\footnotesize{$\pm$3.9} & 57.5\footnotesize{$\pm$4.3} & 13.3\footnotesize{$\pm$5.2} & 53.1\footnotesize{$\pm$9.0} & 13.0\footnotesize{$\pm$3.9} \\
		PCMCI & 93.3\footnotesize{$\pm$4.5} & 66.0\footnotesize{$\pm$14.7} & 99.3\footnotesize{$\pm$0.3} & 90.9\footnotesize{$\pm$2.8} & 93.3\footnotesize{$\pm$4.5} & 66.0\footnotesize{$\pm$14.7} \\
		DYNOTEARS & 99.4\footnotesize{$\pm$0.4} & 93.0\footnotesize{$\pm$6.3} & 96.7\footnotesize{$\pm$2.1} & 93.9\footnotesize{$\pm$3.6} & 99.4\footnotesize{$\pm$0.4} & 93.0\footnotesize{$\pm$6.3} \\
		NTS-NOTEARS & 54.0\footnotesize{$\pm$7.9} & 18.2\footnotesize{$\pm$14.2} & 50.0\footnotesize{$\pm$0.0} & 11.1\footnotesize{$\pm$0.0} & 54.0\footnotesize{$\pm$7.9} & 18.2\footnotesize{$\pm$14.2} \\
		TSCI & 59.6\footnotesize{$\pm$4.9} & 20.8\footnotesize{$\pm$3.8} & 59.3\footnotesize{$\pm$3.7} & 21.6\footnotesize{$\pm$2.6} & 59.6\footnotesize{$\pm$4.9} & 20.8\footnotesize{$\pm$3.8} \\
		cMLP & 90.8\footnotesize{$\pm$3.4} & 75.9\footnotesize{$\pm$5.5} & 93.2\footnotesize{$\pm$3.9} & 85.0\footnotesize{$\pm$5.8} & 90.8\footnotesize{$\pm$3.4} & 75.9\footnotesize{$\pm$5.5} \\
		cLSTM & 99.8\footnotesize{$\pm$0.2} & 98.6\footnotesize{$\pm$1.9} & 91.9\footnotesize{$\pm$4.6} & 56.8\footnotesize{$\pm$3.2} & 99.8\footnotesize{$\pm$0.2} & 98.6\footnotesize{$\pm$1.9} \\
		CUTS & 76.7\footnotesize{$\pm$11.6} & 49.4\footnotesize{$\pm$15.2} & 88.3\footnotesize{$\pm$6.5} & 61.7\footnotesize{$\pm$5.7} & 78.0\footnotesize{$\pm$12.0} & 50.1\footnotesize{$\pm$13.7} \\
		CUTS+ & \textbf{99.9\footnotesize{$\pm$0.0}} & \textbf{99.4\footnotesize{$\pm$0.4}} & \textbf{100.0\footnotesize{$\pm$0.0}} & \textbf{100.0\footnotesize{$\pm$0.0}} & \textbf{99.9\footnotesize{$\pm$0.0}} & \textbf{99.4\footnotesize{$\pm$0.4}} \\
		\bottomrule
	\end{tabular}
\end{table*}

\begin{table*}[!htbp]
	\centering
	\caption{Nonlinear setting, 10-node case with $T = 1000$ and $F = 10$ under MCAR missingness ($\gamma=0.4$) and different imputation strategies.}
	\label{tab:imputation_Lorenz_T1000_F10}
	\small
	\setlength{\tabcolsep}{4pt}
	\begin{tabular}{ccccccc}
		\toprule
		& \multicolumn{2}{c}{ImputeZOH} & \multicolumn{2}{c}{ImputeGP} & \multicolumn{2}{c}{ImputeLinear} \\
		\cmidrule(lr){2-3} \cmidrule(lr){4-5} \cmidrule(lr){6-7}
		10 nodes & AUROC$\uparrow$ & AUPRC$\uparrow$ & AUROC$\uparrow$ & AUPRC$\uparrow$ & AUROC$\uparrow$ & AUPRC$\uparrow$ \\
		\midrule
		VAR & 74.8\footnotesize{$\pm$2.3} & 50.7\footnotesize{$\pm$2.2} & 69.1\footnotesize{$\pm$3.7} & 45.4\footnotesize{$\pm$2.6} & 74.8\footnotesize{$\pm$2.3} & 50.7\footnotesize{$\pm$2.2} \\
		LGC & 81.3\footnotesize{$\pm$2.6} & 60.1\footnotesize{$\pm$3.4} & 73.3\footnotesize{$\pm$2.8} & 49.5\footnotesize{$\pm$3.1} & 81.3\footnotesize{$\pm$2.6} & 60.1\footnotesize{$\pm$3.4} \\
		VARLiNGAM & 67.6\footnotesize{$\pm$3.5} & 44.7\footnotesize{$\pm$2.8} & 67.5\footnotesize{$\pm$3.1} & 44.3\footnotesize{$\pm$2.5} & 67.6\footnotesize{$\pm$3.5} & 44.7\footnotesize{$\pm$2.8} \\
		PCMCI & 78.1\footnotesize{$\pm$3.3} & 59.9\footnotesize{$\pm$5.7} & 82.5\footnotesize{$\pm$2.2} & 62.2\footnotesize{$\pm$3.4} & 78.1\footnotesize{$\pm$3.3} & 59.9\footnotesize{$\pm$5.7} \\
		DYNOTEARS & 69.5\footnotesize{$\pm$3.7} & 47.4\footnotesize{$\pm$3.9} & 57.5\footnotesize{$\pm$1.6} & 39.1\footnotesize{$\pm$1.7} & 69.5\footnotesize{$\pm$3.7} & 47.4\footnotesize{$\pm$3.9} \\
		NTS-NOTEARS & 50.0\footnotesize{$\pm$0.0} & 33.3\footnotesize{$\pm$0.0} & 50.0\footnotesize{$\pm$0.0} & 33.3\footnotesize{$\pm$0.0} & 50.0\footnotesize{$\pm$0.0} & 33.3\footnotesize{$\pm$0.0} \\
		TSCI & 70.7\footnotesize{$\pm$4.1} & 56.4\footnotesize{$\pm$4.6} & 73.3\footnotesize{$\pm$2.6} & 64.4\footnotesize{$\pm$3.2} & 70.7\footnotesize{$\pm$4.1} & 56.4\footnotesize{$\pm$4.6} \\
		cMLP & 82.2\footnotesize{$\pm$2.4} & 70.7\footnotesize{$\pm$5.3} & 60.7\footnotesize{$\pm$3.6} & 49.5\footnotesize{$\pm$2.3} & 82.2\footnotesize{$\pm$2.4} & 70.7\footnotesize{$\pm$5.3} \\
		cLSTM & 96.4\footnotesize{$\pm$0.7} & 94.9\footnotesize{$\pm$1.1} & 98.7\footnotesize{$\pm$0.2} & 97.6\footnotesize{$\pm$0.5} & 96.4\footnotesize{$\pm$0.7} & 94.9\footnotesize{$\pm$1.1} \\
		CUTS & 89.7\footnotesize{$\pm$3.3} & 86.3\footnotesize{$\pm$5.4} & 92.5\footnotesize{$\pm$1.9} & 88.3\footnotesize{$\pm$3.6} & 90.5\footnotesize{$\pm$3.1} & 87.5\footnotesize{$\pm$4.9} \\
		CUTS+ & \textbf{99.5\footnotesize{$\pm$0.4}} & \textbf{99.2\footnotesize{$\pm$0.6}} & \textbf{100.0\footnotesize{$\pm$0.0}} & \textbf{100.0\footnotesize{$\pm$0.0}} & \textbf{99.5\footnotesize{$\pm$0.4}} & \textbf{99.2\footnotesize{$\pm$0.6}} \\
		\bottomrule
	\end{tabular}
\end{table*}

\section{Table results for non-Gaussian noise}
\label{app_non_gaussian}

We consider the linear vanilla model with exponential noise. The results in Table~\ref{tab:linear-10-1000-non-gaussian-1} and Table~\ref{tab:linear-10-1000-non-gaussian-2} indicate that, overall, the methods achieving the best performance in most scenarios are deep learning-based approaches. We also find that NTS-NOTEARS relies on standardized preprocessing in practical applications: it performs poorly in the vanilla setting but exhibits strong performance once standardization is applied. Our results indicate that Gaussian noise does not affect the main conclusions of the paper.

\begin{table*}[!htbp]
	\centering
	\caption{Linear setting under exponential noise, 10-node case with $T = 1000$ (Part I).}
	\setlength{\tabcolsep}{8.3pt}
	\begin{tabular}{ccccccccc}
		\toprule
		& \multicolumn{2}{c}{Vanilla} & \multicolumn{2}{c}{Mixed data} & \multicolumn{2}{c}{Trend and seasonality} & \multicolumn{2}{c}{Min--max normalization} \\
		\cmidrule(lr){2-3} \cmidrule(lr){4-5} \cmidrule(lr){6-7} \cmidrule(lr){8-9}
		10 nodes & AUROC$\uparrow$ & AUPRC$\uparrow$ & AUROC$\uparrow$ & AUPRC$\uparrow$ & AUROC$\uparrow$ & AUPRC$\uparrow$ & AUROC$\uparrow$ & AUPRC$\uparrow$ \\
		\midrule
		VAR & 99.8\footnotesize{$\pm$0.2} & 98.1\footnotesize{$\pm$3.6} & 74.8\footnotesize{$\pm$8.3} & 29.7\footnotesize{$\pm$8.7} & 59.0\footnotesize{$\pm$4.8} & \textbf{27.1\footnotesize{$\pm$8.7}} & 99.4\footnotesize{$\pm$0.4} & 93.0\footnotesize{$\pm$6.3} \\
		LGC & \textbf{100.0\footnotesize{$\pm$0.0}} & \textbf{100.0\footnotesize{$\pm$0.0}} & 72.2\footnotesize{$\pm$8.2} & 37.6\footnotesize{$\pm$13.8} & 51.0\footnotesize{$\pm$2.0} & 12.8\footnotesize{$\pm$3.5} & \textbf{100.0\footnotesize{$\pm$0.0}} & \textbf{100.0\footnotesize{$\pm$0.0}} \\
		VARLiNGAM & 48.2\footnotesize{$\pm$5.5} & 11.8\footnotesize{$\pm$0.8} & 48.1\footnotesize{$\pm$7.5} & 11.7\footnotesize{$\pm$0.8} & 51.5\footnotesize{$\pm$2.6} & 13.6\footnotesize{$\pm$3.5} & 48.2\footnotesize{$\pm$5.5} & 11.8\footnotesize{$\pm$0.8} \\
		PCMCI & 98.5\footnotesize{$\pm$0.3} & 80.7\footnotesize{$\pm$3.1} & 88.1\footnotesize{$\pm$0.8} & 61.3\footnotesize{$\pm$7.7} & 58.6\footnotesize{$\pm$1.0} & 13.1\footnotesize{$\pm$0.2} & 98.5\footnotesize{$\pm$0.3} & 80.7\footnotesize{$\pm$3.1} \\
		DYNOTEARS & \textbf{100.0\footnotesize{$\pm$0.0}} & \textbf{100.0\footnotesize{$\pm$0.0}} & 74.6\footnotesize{$\pm$7.3} & 42.6\footnotesize{$\pm$11.7} & 58.8\footnotesize{$\pm$6.3} & 14.8\footnotesize{$\pm$3.0} & \textbf{100.0\footnotesize{$\pm$0.0}} & \textbf{100.0\footnotesize{$\pm$0.0}} \\
		NTS-NOTEARS & 55.9\footnotesize{$\pm$12.0} & 21.7\footnotesize{$\pm$21.3} & 56.0\footnotesize{$\pm$3.7} & 21.7\footnotesize{$\pm$6.6} & 57.4\footnotesize{$\pm$6.5} & 13.1\footnotesize{$\pm$1.8} & 53.8\footnotesize{$\pm$7.7} & 16.6\footnotesize{$\pm$11.0} \\
		TSCI & 77.7\footnotesize{$\pm$8.1} & 38.0\footnotesize{$\pm$13.1} & 53.1\footnotesize{$\pm$9.0} & 21.3\footnotesize{$\pm$8.1} & 55.5\footnotesize{$\pm$11.3} & 22.0\footnotesize{$\pm$11.6} & 77.7\footnotesize{$\pm$8.1} & 38.0\footnotesize{$\pm$13.1} \\
		cMLP & \textbf{100.0\footnotesize{$\pm$0.0}} & \textbf{100.0\footnotesize{$\pm$0.0}} & 76.4\footnotesize{$\pm$11.8} & 54.8\footnotesize{$\pm$16.1} & 50.7\footnotesize{$\pm$3.5} & 18.2\footnotesize{$\pm$3.4} & \textbf{100.0\footnotesize{$\pm$0.0}} & \textbf{100.0\footnotesize{$\pm$0.0}} \\
		cLSTM & \textbf{100.0\footnotesize{$\pm$0.0}} & \textbf{100.0\footnotesize{$\pm$0.0}} & \textbf{93.8\footnotesize{$\pm$4.5}} & \textbf{76.5\footnotesize{$\pm$16.2}} & \textbf{64.0\footnotesize{$\pm$11.5}} & 24.4\footnotesize{$\pm$9.8} & \textbf{100.0\footnotesize{$\pm$0.0}} & \textbf{100.0\footnotesize{$\pm$0.0}} \\
		CUTS & 95.3\footnotesize{$\pm$3.2} & 80.1\footnotesize{$\pm$11.9} & 63.8\footnotesize{$\pm$3.7} & 27.5\footnotesize{$\pm$6.9} & 49.5\footnotesize{$\pm$11.6} & 20.8\footnotesize{$\pm$11.6} & 59.3\footnotesize{$\pm$7.4} & 19.6\footnotesize{$\pm$9.0} \\
		CUTS+ & \textbf{100.0\footnotesize{$\pm$0.0}} & \textbf{100.0\footnotesize{$\pm$0.0}} & 84.3\footnotesize{$\pm$5.7} & 54.8\footnotesize{$\pm$5.3} & 55.2\footnotesize{$\pm$4.9} & 17.0\footnotesize{$\pm$4.7} & 92.2\footnotesize{$\pm$0.8} & 68.8\footnotesize{$\pm$7.3} \\
		\bottomrule
	\end{tabular}
    \label{tab:linear-10-1000-non-gaussian-1}

\end{table*}

\begin{table*}[!htbp]
	\centering
	\caption{Linear setting under exponential noise, 10-node case with $T = 1000$ (Part II).}
	\setlength{\tabcolsep}{4pt}
	\begin{tabular}{ccccccccccc}
		\toprule
		& \multicolumn{2}{c}{Latent confounders} & \multicolumn{2}{c}{Measurement error} & \multicolumn{2}{c}{Standardized} & \multicolumn{2}{c}{Missing} & \multicolumn{2}{c}{Nonstationary} \\
		\cmidrule(lr){2-3} \cmidrule(lr){4-5} \cmidrule(lr){6-7} \cmidrule(lr){8-9} \cmidrule(lr){10-11}
		10 nodes & AUROC$\uparrow$ & AUPRC$\uparrow$ & AUROC$\uparrow$ & AUPRC$\uparrow$ & AUROC$\uparrow$ & AUPRC$\uparrow$ & AUROC$\uparrow$ & AUPRC$\uparrow$ & AUROC$\uparrow$ & AUPRC$\uparrow$ \\
		\midrule
		VAR & 86.2\footnotesize{$\pm$2.5} & 31.7\footnotesize{$\pm$4.2} & 88.6\footnotesize{$\pm$5.5} & 66.2\footnotesize{$\pm$12.4} & 99.8\footnotesize{$\pm$0.2} & 98.1\footnotesize{$\pm$3.6} & 80.2\footnotesize{$\pm$4.8} & 57.0\footnotesize{$\pm$9.2} & 94.1\footnotesize{$\pm$3.2} & 55.3\footnotesize{$\pm$14.4} \\
		LGC & 91.8\footnotesize{$\pm$2.4} & 44.6\footnotesize{$\pm$7.2} & 88.8\footnotesize{$\pm$5.4} & 68.9\footnotesize{$\pm$12.0} & \textbf{100.0\footnotesize{$\pm$0.0}} & \textbf{100.0\footnotesize{$\pm$0.0}} & 79.6\footnotesize{$\pm$5.3} & 59.2\footnotesize{$\pm$8.8} & 98.0\footnotesize{$\pm$1.7} & 78.9\footnotesize{$\pm$15.4} \\
		VARLiNGAM & 48.0\footnotesize{$\pm$6.5} & 12.0\footnotesize{$\pm$1.9} & 58.7\footnotesize{$\pm$9.8} & 15.7\footnotesize{$\pm$6.7} & 48.2\footnotesize{$\pm$5.5} & 11.8\footnotesize{$\pm$0.8} & 50.6\footnotesize{$\pm$7.0} & 13.1\footnotesize{$\pm$2.5} & 51.3\footnotesize{$\pm$7.5} & 12.9\footnotesize{$\pm$1.8} \\
		PCMCI & 75.3\footnotesize{$\pm$3.5} & 20.5\footnotesize{$\pm$2.3} & 92.0\footnotesize{$\pm$4.3} & 66.8\footnotesize{$\pm$8.3} & 98.5\footnotesize{$\pm$0.3} & 80.7\footnotesize{$\pm$3.1} & 94.8\footnotesize{$\pm$2.9} & 72.3\footnotesize{$\pm$10.0} & 87.7\footnotesize{$\pm$4.9} & 36.5\footnotesize{$\pm$10.7} \\
		DYNOTEARS & 91.1\footnotesize{$\pm$1.9} & 43.2\footnotesize{$\pm$2.8} & 83.7\footnotesize{$\pm$9.0} & 68.0\footnotesize{$\pm$15.2} & \textbf{100.0\footnotesize{$\pm$0.0}} & \textbf{100.0\footnotesize{$\pm$0.0}} & 99.7\footnotesize{$\pm$0.3} & 96.3\footnotesize{$\pm$4.4} & 97.5\footnotesize{$\pm$2.3} & 89.4\footnotesize{$\pm$6.4} \\
		NTS-NOTEARS & 82.1\footnotesize{$\pm$11.8} & 39.8\footnotesize{$\pm$15.3} & 54.8\footnotesize{$\pm$9.7} & 14.5\footnotesize{$\pm$6.9} & 99.8\footnotesize{$\pm$0.2} & 98.1\footnotesize{$\pm$3.6} & 59.8\footnotesize{$\pm$14.9} & 27.1\footnotesize{$\pm$23.4} & 81.8\footnotesize{$\pm$18.2} & 56.4\footnotesize{$\pm$35.1} \\
		TSCI & 53.0\footnotesize{$\pm$7.7} & 17.4\footnotesize{$\pm$6.3} & 54.1\footnotesize{$\pm$6.2} & 18.2\footnotesize{$\pm$8.9} & 77.7\footnotesize{$\pm$8.1} & 38.0\footnotesize{$\pm$13.1} & 63.6\footnotesize{$\pm$7.6} & 22.8\footnotesize{$\pm$5.0} & 71.7\footnotesize{$\pm$4.0} & 30.2\footnotesize{$\pm$7.8} \\
		cMLP & \textbf{100.0\footnotesize{$\pm$0.0}} & \textbf{100.0\footnotesize{$\pm$0.0}} & 85.2\footnotesize{$\pm$6.4} & 55.6\footnotesize{$\pm$15.0} & \textbf{100.0\footnotesize{$\pm$0.0}} & \textbf{100.0\footnotesize{$\pm$0.0}} & 94.6\footnotesize{$\pm$3.2} & 78.8\footnotesize{$\pm$10.7} & 96.5\footnotesize{$\pm$4.4} & 84.2\footnotesize{$\pm$16.4} \\
		cLSTM & 99.9\footnotesize{$\pm$0.0} & 99.8\footnotesize{$\pm$0.3} & \textbf{99.9\footnotesize{$\pm$0.1}} & \textbf{99.4\footnotesize{$\pm$0.6}} & \textbf{100.0\footnotesize{$\pm$0.0}} & \textbf{100.0\footnotesize{$\pm$0.0}} & \textbf{99.9\footnotesize{$\pm$0.1}} & \textbf{99.4\footnotesize{$\pm$0.6}} & 98.3\footnotesize{$\pm$2.4} & 92.8\footnotesize{$\pm$10.3} \\
		CUTS & 84.2\footnotesize{$\pm$5.7} & 52.1\footnotesize{$\pm$16.8} & 90.4\footnotesize{$\pm$7.0} & 68.6\footnotesize{$\pm$14.3} & \textbf{100.0\footnotesize{$\pm$0.0}} & \textbf{100.0\footnotesize{$\pm$0.0}} & 76.7\footnotesize{$\pm$11.6} & 49.4\footnotesize{$\pm$15.2} & 98.6\footnotesize{$\pm$1.3} & 92.5\footnotesize{$\pm$6.9} \\
		CUTS+ & 91.2\footnotesize{$\pm$2.8} & 62.5\footnotesize{$\pm$11.2} & 99.6\footnotesize{$\pm$0.3} & 97.3\footnotesize{$\pm$3.0} & 99.9\footnotesize{$\pm$0.1} & 99.4\footnotesize{$\pm$0.6} & \textbf{99.9\footnotesize{$\pm$0.0}} & 99.4\footnotesize{$\pm$0.4} & \textbf{99.6\footnotesize{$\pm$0.5}} & \textbf{97.1\footnotesize{$\pm$3.8}} \\
		\bottomrule
	\end{tabular}
    \label{tab:linear-10-1000-non-gaussian-2}

\end{table*}

\section{Table results under different levels of trend and seasonality}
\label{app_stronger_trend_seasonality}

Table~\ref{tab:stronger_trend_seasonality} reports results under stronger trend and seasonality in the nonlinear Lorenz--96 setting with $D=10$, $T=1000$, and $F=10$.
We increase $\rho$ from $0.01$ to $0.02$ and $\eta$ from $0.5$ to $12$.
Larger values of $\rho$ and $\eta$ correspond to stronger trend and seasonal components, respectively. The results indicate that, under the weaker perturbation setting, simple methods such as VAR and LGC exhibit little change in performance, whereas deep learning-based methods experience substantial performance degradation.
Under the stronger perturbation setting, these simple methods also exhibit performance degradation. Understanding why deep learning-based methods are vulnerable to trend and seasonality even under the weaker setting is an important direction for future research.

\begin{table*}[!htbp]
    \centering
    \caption{
    AUROC/AUPRC results under vanilla, original, and stronger trend
    and seasonality settings for Lorenz--96 with $D=10$, $T=1000$,
    and $F=10$.
    }
    \label{tab:stronger_trend_seasonality}
    \small
    \setlength{\tabcolsep}{4pt}
    \resizebox{\textwidth}{!}{
    \begin{tabular}{lcccccc}
        \toprule
        & \multicolumn{2}{c}{Vanilla}
        & \multicolumn{2}{c}{Original: $\rho=0.01$, $\eta=0.5$}
        & \multicolumn{2}{c}{Stronger: $\rho=0.02$, $\eta=12$} \\
        \cmidrule(lr){2-3}
        \cmidrule(lr){4-5}
        \cmidrule(lr){6-7}
        Method & AUROC & AUPRC & AUROC & AUPRC & AUROC & AUPRC \\
        \midrule
        VAR
        & 83.3{\footnotesize$\pm$3.6}
        & 66.6{\footnotesize$\pm$5.6}
        & 82.5{\footnotesize$\pm$2.4}
        & 65.7{\footnotesize$\pm$3.8}
        & 69.5{\footnotesize$\pm$3.7}
        & 58.9{\footnotesize$\pm$5.6} \\
        LGC
        & 83.8{\footnotesize$\pm$2.6}
        & 72.4{\footnotesize$\pm$4.5}
        & 81.8{\footnotesize$\pm$2.3}
        & 69.3{\footnotesize$\pm$3.8}
        & 67.8{\footnotesize$\pm$4.4}
        & 56.7{\footnotesize$\pm$6.6} \\
        VARLiNGAM
        & 72.0{\footnotesize$\pm$1.9}
        & 60.3{\footnotesize$\pm$2.7}
        & 71.6{\footnotesize$\pm$2.6}
        & 58.7{\footnotesize$\pm$5.2}
        & 66.0{\footnotesize$\pm$2.6}
        & 53.9{\footnotesize$\pm$4.0} \\
        PCMCI
        & 77.8{\footnotesize$\pm$4.9}
        & 55.9{\footnotesize$\pm$6.1}
        & 79.5{\footnotesize$\pm$3.5}
        & 58.2{\footnotesize$\pm$5.4}
        & 56.3{\footnotesize$\pm$1.7}
        & 36.4{\footnotesize$\pm$0.9} \\
        DYNOTEARS
        & 86.1{\footnotesize$\pm$2.1}
        & 71.2{\footnotesize$\pm$3.4}
        & \textbf{87.8{\footnotesize$\pm$1.7}}
        & \textbf{73.6{\footnotesize$\pm$1.6}}
        & \textbf{75.5{\footnotesize$\pm$3.5}}
        & \textbf{60.0{\footnotesize$\pm$5.1}} \\
        NTS-NOTEARS
        & 82.9{\footnotesize$\pm$1.4}
        & 59.5{\footnotesize$\pm$2.0}
        & 60.6{\footnotesize$\pm$4.0}
        & 39.0{\footnotesize$\pm$2.6}
        & 59.3{\footnotesize$\pm$2.4}
        & 38.1{\footnotesize$\pm$1.4} \\
        TSCI
        & 88.4{\footnotesize$\pm$1.7}
        & 80.5{\footnotesize$\pm$2.2}
        & 55.9{\footnotesize$\pm$6.2}
        & 45.5{\footnotesize$\pm$9.1}
        & 59.7{\footnotesize$\pm$4.1}
        & 40.4{\footnotesize$\pm$4.5} \\
        cMLP
        & 99.9{\footnotesize$\pm$0.0}
        & 99.9{\footnotesize$\pm$0.0}
        & 51.2{\footnotesize$\pm$1.0}
        & 46.1{\footnotesize$\pm$1.5}
        & 51.3{\footnotesize$\pm$1.0}
        & 46.1{\footnotesize$\pm$1.4} \\
        cLSTM
        & \textbf{100.0{\footnotesize$\pm$0.0}}
        & \textbf{100.0{\footnotesize$\pm$0.0}}
        & 76.1{\footnotesize$\pm$2.7}
        & 60.5{\footnotesize$\pm$5.3}
        & 58.0{\footnotesize$\pm$0.9}
        & 46.1{\footnotesize$\pm$1.4} \\
        CUTS
        & 98.3{\footnotesize$\pm$1.1}
        & 97.5{\footnotesize$\pm$1.6}
        & 46.3{\footnotesize$\pm$2.2}
        & 41.9{\footnotesize$\pm$5.5}
        & 59.5{\footnotesize$\pm$1.5}
        & 46.0{\footnotesize$\pm$3.8} \\
        CUTS+
        & 99.9{\footnotesize$\pm$0.0}
        & 99.9{\footnotesize$\pm$0.1}
        & 49.2{\footnotesize$\pm$1.0}
        & 35.2{\footnotesize$\pm$2.4}
        & 54.6{\footnotesize$\pm$4.7}
        & 41.2{\footnotesize$\pm$4.1} \\
        \bottomrule
    \end{tabular}
    }
\end{table*}

\section{Table results under different time steps}
\label{app_time_steps}

To assess the effect of discrete-time sampling on the experimental results, we further conduct experiments with time steps of $\Delta t = 0.05$ and $\Delta t = 0.01$ in the Lorenz-96 model. The results for $\Delta t = 0.1$ are reported in the main text. As $\Delta t$ decreases, the resulting statistical dependencies better approximate the continuous-time topology. The results in Table~\ref{tab:time_steps} show that the
performance metrics generally decline as $\Delta t$ decreases. Nevertheless, deep learning-based methods achieve better overall performance. Therefore, sampling the data at discrete time steps does not alter the main conclusions. The CausalDynamics benchmark~\cite{herdeanu2025causaldynamics}
also generates data from continuous-time ordinary differential
equations using discrete-time sampling, which is a commonly adopted approximation in this field.

\begin{table*}[!htbp]
    \centering
    \caption{AUROC/AUPRC results under different time steps.}
    \label{tab:time_steps}
    \small
    \begin{tabular}{lcc}
        \toprule
        Method
        & $\Delta t = 0.05$
        & $\Delta t = 0.01$ \\
        \midrule
        VAR
        & 62.33/42.12
        & 57.17/37.26 \\

        LGC
        & 64.33/42.56
        & 53.33/35.72 \\

        VARLiNGAM
        & 69.83/50.24
        & 60.17/43.13 \\

        PCMCI
        & 68.83/49.47
        & 61.50/40.68 \\

        DYNOTEARS
        & 75.50/52.56
        & 71.00/53.86 \\

        NTS-NOTEARS
        & 69.17/51.81
        & 54.50/40.90 \\

        TSCI
        & 83.78/74.81
        & 62.66/52.45 \\

        cMLP
        & 99.90/99.81
        & \textbf{82.10}/\textbf{72.16} \\

        cLSTM
        & \textbf{99.98}/\textbf{99.96}
        & 67.24/48.41 \\

        CUTS
        & 90.43/86.98
        & 55.51/42.37 \\

        CUTS+
        & 91.52/82.13
        & 63.03/45.96 \\
        \bottomrule
    \end{tabular}
\end{table*}

\section{Table results under different conditional independence tests}
\label{app_ci_tests}

We further evaluate PCMCI and PCMCI+ using the nonlinear
conditional independence test GPDC on Lorenz-96 datasets. We report only the best result
among the deep learning-based methods in each scenario,
denoted by DL. The results in Table~\ref{tab:ci_tests} show that
deep learning-based methods achieve better overall performance.
The average runtime per run is 3.43 seconds for PCMCI-Linear,
5172.7 seconds for PCMCI-GPDC, and 6401.7 seconds
for PCMCI+-GPDC, versus 178.1--240.8 seconds
for the deep learning-based methods.
Despite its substantially greater computational cost,
PCMCI-GPDC performs similarly overall to PCMCI-Linear,
and PCMCI+-GPDC does not consistently improve upon PCMCI-GPDC.
Given the large number of benchmark datasets,
our use of the considerably faster PCMCI-Linear
in the main experiments is therefore a reasonable design choice.

\begin{table*}[!htbp]
    \centering
    \caption{AUROC/AUPRC results under different conditional
    independence tests on Lorenz-96 datasets
    ($D = 10$, $T = 1000$, $F = 10$).}
    \label{tab:ci_tests}
    \small
    \setlength{\tabcolsep}{3pt}
    \begin{adjustbox}{max width=\textwidth}
    \begin{tabular}{lccccccccc}
        \toprule
        Method
        & Vanilla
        & \makecell{Mixed\\data}
        & \makecell{Trend and\\seasonality}
        & \makecell{Min--max\\normalization}
        & \makecell{Latent\\confounders}
        & \makecell{Measurement\\error}
        & Standardized
        & Missing
        & Nonstationary \\
        \midrule
        PCMCI-Linear
        & 77.8/55.9
        & \textbf{73.3}/54.0
        & \textbf{79.5}/58.2
        & 77.8/55.9
        & 61.3/41.5
        & 75.1/56.2
        & 77.8/55.9
        & 78.1/59.9
        & 70.3/52.6 \\

        PCMCI-GPDC
        & 78.3/54.0
        & 68.3/47.7
        & 74.2/51.9
        & \textbf{78.3}/54.0
        & 70.0/56.4
        & 64.2/47.9
        & 78.3/54.0
        & 72.5/52.3
        & 73.3/58.2 \\

        PCMCI+-GPDC
        & 65.8/49.9
        & 67.5/54.7
        & 65.8/52.5
        & 65.8/49.9
        & 65.8/52.5
        & 73.3/58.2
        & 65.8/49.9
        & 62.5/45.8
        & 80.0/64.4 \\

        DL
        & \textbf{100.0}/\textbf{99.9}
        & 72.3/\textbf{66.1}
        & 76.1/\textbf{60.5}
        & 77.1/\textbf{67.5}
        & \textbf{96.7}/\textbf{93.9}
        & \textbf{96.7}/\textbf{94.6}
        & \textbf{100.0}/\textbf{99.9}
        & \textbf{99.5}/\textbf{99.2}
        & \textbf{83.4}/\textbf{78.0} \\
        \bottomrule
    \end{tabular}
    \end{adjustbox}
\end{table*}

\section{Table results on real-world datasets}
\label{app_real_datasets}

We further evaluate the eleven benchmark methods on eight real-world datasets:
an fMRI brain-network dataset~\cite{smith2011network},
the Arctic Sea Ice dataset~\cite{huang2021benchmarking},
the Finance dataset~\cite{kleinberg2011logic},
and five DREAM3 gene regulatory network datasets~\cite{prill2010towards}.
In particular, DREAM3 extends our evaluation to 100 variables, providing further evidence of method performance in high-dimensional settings. The fMRI, Arctic Sea Ice, and Finance datasets contain 10, 12, and 25 nodes, with time-series lengths of 1,200, 468, and 4,000, respectively. Each DREAM3 dataset contains 100 nodes and 966 time points. The results in Table~\ref{tab:real_datasets} show that deep learning-based methods achieve better overall performance across these 8 real-world datasets. This finding is particularly important because it provides empirical evidence supporting the effectiveness of these methods in complex real-world settings.

\begin{table*}[!htbp]
    \centering
    \caption{AUROC/AUPRC results on real-world datasets.}
    \label{tab:real_datasets}
    \small
    \setlength{\tabcolsep}{3pt}
    \begin{adjustbox}{max width=\textwidth}
    \begin{tabular}{lcccccccc}
        \toprule
        Method
        & fMRI
        & \makecell{Arctic\\Sea Ice}
        & Finance
        & DREAM3-E1
        & DREAM3-E2
        & DREAM3-Y1
        & DREAM3-Y2
        & DREAM3-Y3 \\
        \midrule
        VAR
        & 68.2/30.5 & 52.8/37.9 & 97.6/57.2
        & 54.1/1.3 & 55.2/1.3 & 52.5/1.7 & 51.0/4.0 & 50.0/5.5 \\

        LGC
        & 78.0/30.0 & \textbf{56.9}/40.4 & 96.7/68.7
        & 51.7/1.6 & 51.8/1.5 & 52.4/2.3 & 53.2/4.3 & 53.3/5.9 \\

        PCMCI
        & 78.4/33.8 & 54.4/40.1 & 96.9/56.1
        & 58.1/2.5 & 57.0/2.4 & 56.3/2.7 & 52.9/4.5 & 51.7/5.9 \\

        VARLiNGAM
        & 60.7/17.1 & 55.2/39.3 & 97.9/67.1
        & 55.7/1.4 & 51.0/1.2 & 51.0/1.7 & 51.1/4.0 & 50.4/5.6 \\

        DYNOTEARS
        & 71.4/31.7 & 56.8/40.8 & 95.2/32.3
        & 51.5/2.6 & 54.2/2.2 & 51.8/5.2 & 50.8/5.2 & 51.4/6.1 \\

        NTS-NOTEARS
        & 79.9/34.1 & 50.8/36.9 & 91.3/49.5
        & 50.0/1.2 & 50.0/1.2 & 50.0/1.6 & 50.0/3.9 & 50.0/5.5 \\

        TSCI
        & 78.7/42.7 & 49.7/37.2 & 47.0/3.2
        & \textbf{64.3}/2.8 & 59.6/2.5 & 51.6/2.0 & 52.2/4.2 & 46.6/5.1 \\

        cMLP
        & 86.2/54.7 & 54.9/40.9 & 98.5/87.8
        & 58.9/\textbf{5.8} & 55.8/\textbf{4.9}
        & \textbf{58.5}/\textbf{8.2} & 53.0/\textbf{6.9}
        & \textbf{54.5}/\textbf{7.9} \\

        cLSTM
        & 59.9/21.8 & 56.6/\textbf{44.8}
        & \textbf{98.7}/\textbf{92.6}
        & 58.4/4.6 & 56.5/3.6 & 58.4/7.5 & 52.2/6.5 & 53.0/7.3 \\

        CUTS
        & \textbf{86.8}/\textbf{58.3} & 54.9/41.9 & 91.3/84.0
        & 56.2/2.5 & \textbf{60.1}/2.1 & 52.6/2.6 & 53.3/4.7 & 50.6/6.0 \\

        CUTS+
        & 82.4/48.0 & 49.2/39.1 & \textbf{98.7}/92.4
        & 58.6/3.3 & 42.6/3.0 & 57.4/3.0 & \textbf{53.4}/6.8 & 52.6/5.8 \\
        \bottomrule
    \end{tabular}
    \end{adjustbox}
\end{table*}

\section{Table results under unified threshold-search and learning-rate settings}
\label{app_unified_protocol}

\paragraph{Unified threshold-search protocol.}
To assess the effect of threshold selection on comparisons across methods, we apply the same automatic threshold-search protocol to all 11 methods and report NSHD, true positive rate (TPR), and false discovery rate (FDR). Each cell in Tables~\ref{tab:unified_var} and~\ref{tab:unified_lorenz} reports NSHD/TPR/FDR, with TPR and FDR expressed as percentages. For FDR, an em dash indicates that all runs included in the corresponding entry yield empty predicted graphs. Under the original learning-rate settings, the results show that, in most scenarios, the methods exhibiting superior overall performance are deep learning-based approaches. Under this unified automatic threshold-search protocol, the sensitivity of NTS-NOTEARS to standardization is observed only in the evaluated linear setting. In practical applications, however, the ground-truth causal graph is typically unavailable, preventing automatic threshold selection by minimizing NSHD against the ground truth. Our observations of the sensitivity of NTS-NOTEARS to standardization in both linear and nonlinear settings under manually specified thresholds therefore remain practically relevant.

\paragraph{Learning-rate settings.}
Our original learning-rate search ranges for cMLP and cLSTM are informed by the experimental configurations of Khanna and Tan~\cite{khanna2019economy}, who use higher learning rates for VAR data and lower learning rates for Lorenz data in Tables~7 and~8 of their paper. Following this practice, we use $\mathcal{H}=\{0.01,0.1\}$ for linear VAR data and $\mathcal{L}=\{5\times10^{-4},0.001\}$ for nonlinear Lorenz-96 data to better characterize the performance of these methods under suitable optimization settings. However, the underlying data-generating process may be unknown in practice. We therefore additionally evaluate two shared learning-rate settings, using either $\mathcal{H}$ or $\mathcal{L}$ for both data types. The results in Tables~\ref{tab:unified_var} and~\ref{tab:unified_lorenz} show that deep learning-based methods achieve the best or tied-best NSHD in 15/18, 9/18, and 13/18 scenarios under the shared higher range $\mathcal{H}$, the shared lower range $\mathcal{L}$, and the original split ranges, respectively. 
The original split ranges use $\mathcal{H}$ for VAR and $\mathcal{L}$ for Lorenz-96. Taken together, these results support our conclusion under the original learning-rate settings that, in most scenarios, the methods exhibiting superior overall performance are deep learning-based approaches, while also showing that their relative performance depends on the learning-rate settings.
Given the experimental settings used by Khanna and Tan~\cite{khanna2019economy}, we do not recommend interpreting results obtained under other settings, such as the uniform lower range ($\mathcal{L}=\{5\times10^{-4},0.001\}$), as a reasonable characterization of the performance of deep learning-based methods.

\begin{table*}[htbp]
\centering
\caption{NSHD/TPR/FDR results on VAR data under the unified threshold-search protocol ($p=10$, $T=1000$). $\mathcal{H}$ and $\mathcal{L}$ denote the higher and lower learning-rate search ranges, respectively. TPR and FDR are reported in percentages. Bold indicates the lowest NSHD among all displayed rows in each column.}
\label{tab:unified_var}
\scriptsize
\setlength{\tabcolsep}{2pt}
\renewcommand{\arraystretch}{1.15}
\begin{tabular}{@{}l*{9}{c}@{}}
\toprule
Method & Vanilla & Mixed & Trend & Min--max & Confounders & Measurement & Standardized & Missing & Nonstationary \\
\midrule
VAR
& \textbf{0.00}/100.0/0.0
& 0.76/24.0/0.0
& \textbf{0.30}/76.0/7.9
& 0.02/100.0/1.8
& 0.24/84.0/7.8
& 0.22/93.9/13.4
& \textbf{0.00}/100.0/0.0
& 0.32/76.0/7.8
& 0.13/90.0/3.6 \\

LGC
& \textbf{0.00}/100.0/0.0
& 0.42/65.9/10.7
& 0.94/6.0/0.0
& \textbf{0.00}/100.0/0.0
& 0.32/78.0/10.4
& 0.22/96.0/15.2
& \textbf{0.00}/100.0/0.0
& 0.18/92.0/9.0
& 0.13/100.0/10.3 \\

VARLiNGAM
& \textbf{0.00}/100.0/0.0
& 0.54/74.0/26.7
& 0.48/64.0/14.0
& \textbf{0.00}/100.0/0.0
& 0.93/10.0/20.8
& 0.20/98.0/14.8
& \textbf{0.00}/100.0/0.0
& 0.08/98.0/5.6
& \textbf{0.04}/100.0/3.3 \\

PCMCI
& 0.24/100.0/18.4
& 0.34/98.0/24.5
& 1.00/0.0/\textemdash
& 0.24/100.0/18.4
& 1.00/0.0/\textemdash
& 0.32/96.0/21.9
& 0.24/100.0/18.4
& 0.46/94.0/27.6
& 1.00/20.0/50.0 \\

DYNOTEARS
& \textbf{0.00}/100.0/0.0
& 0.38/68.0/6.0
& 0.96/8.0/16.6
& \textbf{0.00}/100.0/0.0
& 0.70/36.0/9.0
& 0.13/96.0/9.0
& \textbf{0.00}/100.0/0.0
& \textbf{0.04}/98.0/1.8
& \textbf{0.04}/100.0/3.3 \\

NTS-NOTEARS
& 0.96/4.0/0.0
& 0.44/68.0/14.2
& 0.98/2.0/0.0
& 0.64/36.0/0.0
& 0.78/41.9/16.8
& 0.93/6.0/0.0
& \textbf{0.00}/100.0/0.0
& 0.92/8.0/0.0
& 0.26/80.0/5.6 \\

TSCI
& 0.87/13.9/8.3
& 0.98/6.0/40.0
& 1.00/10.0/50.0
& 0.87/13.9/8.3
& 0.98/2.0/0.0
& 1.00/0.0/\textemdash
& 0.87/13.9/8.3
& 0.96/4.0/0.0
& 0.93/8.0/12.5 \\

cMLP ($\mathcal{H}$)
& 0.02/98.0/0.0
& \textbf{0.10}/98.0/6.9
& 0.96/4.0/0.0
& 0.02/100.0/1.8
& \textbf{0.02}/98.0/0.0
& 0.68/42.0/16.3
& \textbf{0.00}/100.0/0.0
& 0.46/62.0/8.8
& 0.34/92.0/21.2 \\

cMLP ($\mathcal{L}$)
& 1.00/0.0/\textemdash
& 1.00/0.0/\textemdash
& 0.90/10.0/0.0
& 1.00/0.0/\textemdash
& 0.80/20.0/0.0
& 1.00/0.0/\textemdash
& 0.30/70.0/0.0
& 1.00/0.0/\textemdash
& 0.90/10.0/0.0 \\

cLSTM ($\mathcal{H}$)
& \textbf{0.00}/100.0/0.0
& 0.20/82.0/2.5
& 0.94/8.0/12.5
& \textbf{0.00}/100.0/0.0
& 0.04/98.0/1.8
& \textbf{0.12}/96.0/6.9
& \textbf{0.00}/100.0/0.0
& 0.06/98.0/3.8
& 0.10/96.0/5.3 \\

cLSTM ($\mathcal{L}$)
& 1.00/0.0/\textemdash
& 0.80/20.0/0.0
& 0.90/10.0/0.0
& 1.00/0.0/\textemdash
& 0.70/40.0/20.0
& 1.00/0.0/\textemdash
& \textbf{0.00}/100.0/0.0
& 1.00/0.0/\textemdash
& 1.00/0.0/\textemdash \\

CUTS
& 0.41/84.0/22.7
& 0.94/12.0/25.0
& 0.96/4.0/0.0
& 0.98/2.0/0.0
& 0.68/38.0/9.0
& 0.60/55.9/20.0
& \textbf{0.00}/100.0/0.0
& 0.72/36.0/17.5
& 0.16/86.0/1.8 \\

CUTS+
& \textbf{0.00}/100.0/0.0
& 0.74/42.0/23.2
& 0.98/4.0/25.0
& 0.50/74.0/23.7
& 0.61/53.9/23.4
& 0.15/92.0/7.1
& 0.02/100.0/1.8
& 0.06/98.0/3.6
& \textbf{0.04}/98.0/1.8 \\
\bottomrule
\end{tabular}
\end{table*}

\begin{table*}[htbp]
\centering
\caption{NSHD/TPR/FDR results on Lorenz-96 data under the unified threshold-search protocol ($p=10$, $T=1000$, $F=10$). $\mathcal{H}$ and $\mathcal{L}$ denote the higher and lower learning-rate search ranges, respectively. TPR and FDR are reported in percentages. Bold indicates the lowest NSHD among all displayed rows in each column.}
\label{tab:unified_lorenz}
\scriptsize
\setlength{\tabcolsep}{2pt}
\renewcommand{\arraystretch}{1.15}
\begin{tabular}{@{}l*{9}{c}@{}}
\toprule
Method & Vanilla & Mixed & Trend & Min--max & Confounders & Measurement & Standardized & Missing & Nonstationary \\
\midrule
VAR
& 0.34/76.0/10.4
& 0.90/34.6/31.6
& 0.38/72.0/11.1
& 0.35/78.6/12.9
& 0.80/40.0/34.8
& 0.60/59.3/23.7
& 0.34/73.3/7.6
& 0.66/65.9/30.7
& 0.62/56.0/30.0 \\

LGC
& 0.34/74.0/10.4
& 0.69/50.6/27.0
& 0.38/68.6/8.4
& 0.30/78.0/8.8
& 0.80/40.0/34.8
& 0.59/60.0/23.4
& 0.34/74.0/7.8
& 0.55/82.6/29.7
& 0.62/56.6/30.3 \\

VARLiNGAM
& 0.31/85.3/15.9
& 0.61/67.3/26.5
& 0.44/68.6/14.5
& 0.31/83.3/14.3
& 0.84/26.6/27.2
& 0.50/66.6/19.9
& 0.32/82.0/13.8
& 0.40/74.0/15.5
& 0.57/54.0/13.4 \\

PCMCI
& 0.71/84.0/39.1
& 0.65/71.3/32.7
& 0.55/84.0/32.0
& 0.71/84.0/39.1
& 0.90/36.6/43.0
& 0.60/68.0/29.0
& 0.71/84.0/39.1
& 0.48/72.0/21.3
& 0.70/55.9/30.9 \\

DYNOTEARS
& 0.36/80.6/17.0
& 0.84/24.6/21.4
& 0.34/87.3/19.6
& 0.66/36.6/6.0
& 0.78/48.6/34.1
& 0.86/37.3/35.9
& 0.08/96.6/4.4
& 0.72/41.3/14.8
& 0.69/36.0/14.0 \\

NTS-NOTEARS
& \textbf{0.00}/100.0/0.0
& 0.59/44.6/7.9
& \textbf{0.04}/98.0/1.9
& 1.00/0.0/\textemdash
& 0.71/52.0/29.8
& 0.86/30.0/30.5
& \textbf{0.00}/100.0/0.0
& 0.39/70.0/11.2
& 0.59/49.3/23.8 \\

TSCI
& 0.52/76.6/26.5
& 0.80/41.3/30.7
& 0.90/13.3/17.6
& 0.52/76.6/26.5
& 0.90/22.6/28.9
& 0.88/20.0/23.1
& 0.52/76.6/26.5
& 0.82/34.6/30.7
& 0.73/48.6/31.4 \\

cMLP ($\mathcal{L}$)
& 0.01/100.0/1.2
& 0.96/19.3/40.3
& 0.87/18.0/16.6
& 0.98/16.6/43.0
& 0.86/36.0/37.2
& 0.95/20.0/36.3
& 0.02/99.3/1.2
& 0.59/57.9/22.6
& 0.55/47.3/13.4 \\

cMLP ($\mathcal{H}$)
& \textbf{0.00}/100.0/0.0
& 0.50/53.3/5.8
& 0.93/16.6/37.5
& 0.10/93.3/3.4
& 0.93/8.0/12.5
& 0.72/36.0/17.5
& \textbf{0.00}/100.0/0.0
& 0.56/60.0/21.7
& 0.96/10.0/40.0 \\

cLSTM ($\mathcal{L}$)
& \textbf{0.00}/100.0/0.0
& 0.86/21.3/22.9
& 0.79/45.3/28.3
& 0.96/8.0/38.3
& 0.42/82.6/21.6
& 0.65/54.6/22.9
& 0.05/98.6/3.8
& 0.20/85.3/5.3
& 0.51/52.6/21.3 \\

cLSTM ($\mathcal{H}$)
& \textbf{0.00}/100.0/0.0
& \textbf{0.20}/93.3/12.5
& 0.86/33.3/37.5
& \textbf{0.00}/100.0/0.0
& 0.36/76.6/14.8
& 0.63/60.0/28.0
& \textbf{0.00}/100.0/0.0
& 0.26/93.3/17.6
& 0.63/50.0/21.0 \\

CUTS
& 0.16/92.0/7.8
& 0.85/20.0/21.6
& 0.96/5.3/20.8
& 0.86/32.0/35.6
& 0.76/65.3/39.1
& 0.82/29.3/25.0
& \textbf{0.00}/100.0/0.0
& 0.46/73.3/19.8
& 0.64/44.6/18.7 \\

CUTS+
& 0.01/100.0/1.2
& 0.68/45.3/22.5
& 0.97/3.3/16.6
& 0.66/49.3/24.4
& \textbf{0.22}/89.3/8.8
& \textbf{0.22}/89.3/11.7
& \textbf{0.00}/100.0/0.0
& \textbf{0.06}/98.0/4.4
& \textbf{0.40}/64.6/13.0 \\
\bottomrule
\end{tabular}
\end{table*}

\section{Table results with dataset-specific optimal hyperparameters}
\label{app_table_results_best}
We determine the hyperparameters of each method as dataset-specific optimal values. Additional experimental results under both linear and nonlinear settings, spanning different numbers of nodes, time series lengths, and external forcing intensities, are reported in the Appendix (Tables~\ref{tab:linear-10-500-1}--\ref{tab:nonlinear-15-1000-f40-2}).


\section{Table results with hyperparameters selected by average performance}
\label{app_table_results_avg_scen}
We select, for each causal discovery method, a single hyperparameter configuration that maximizes average performance across all scenarios. Additional experimental results under both linear and nonlinear settings, spanning different numbers of nodes, time series lengths, and external forcing intensities, are reported in the Appendix (Tables~\ref{tab:linear-10-500-avg-scen-1}--\ref{tab:nonlinear-15-1000-f40-avg-scen-2}).

\section{Table results aggregated over all hyperparameters}
\label{app_table_results_avg_hyper}
For each scenario, we compute the mean and standard deviation of the evaluation metrics for each method across different hyperparameter settings. Additional experimental results under both linear and nonlinear settings, spanning different numbers of nodes, time series lengths, and external forcing intensities, are reported in the Appendix (Tables~\ref{tab:linear-10-500-avg-hyper-1}--\ref{tab:nonlinear-15-1000-f40-avg-hyper-2}).

\section{Figure results with dataset-specific optimal hyperparameters}
\label{app_fig_results_best}
We determine the hyperparameters of each method as dataset-specific optimal values. Additional figure results under both linear and nonlinear settings, spanning different numbers of nodes, time series lengths, and external forcing intensities, are reported in the Appendix (Figures~\ref{fig:experiments_10_500_f10}--~\ref{fig:experiments_15_500_1000_f40}).

\section{Figure results with hyperparameters selected by average performance}\label{app_fig_results_avg_scen}
\begin{figure*}[t]
     \centering
     \begin{subfigure}[b]{0.49\textwidth}
         \centering
        \includegraphics[width=\textwidth]{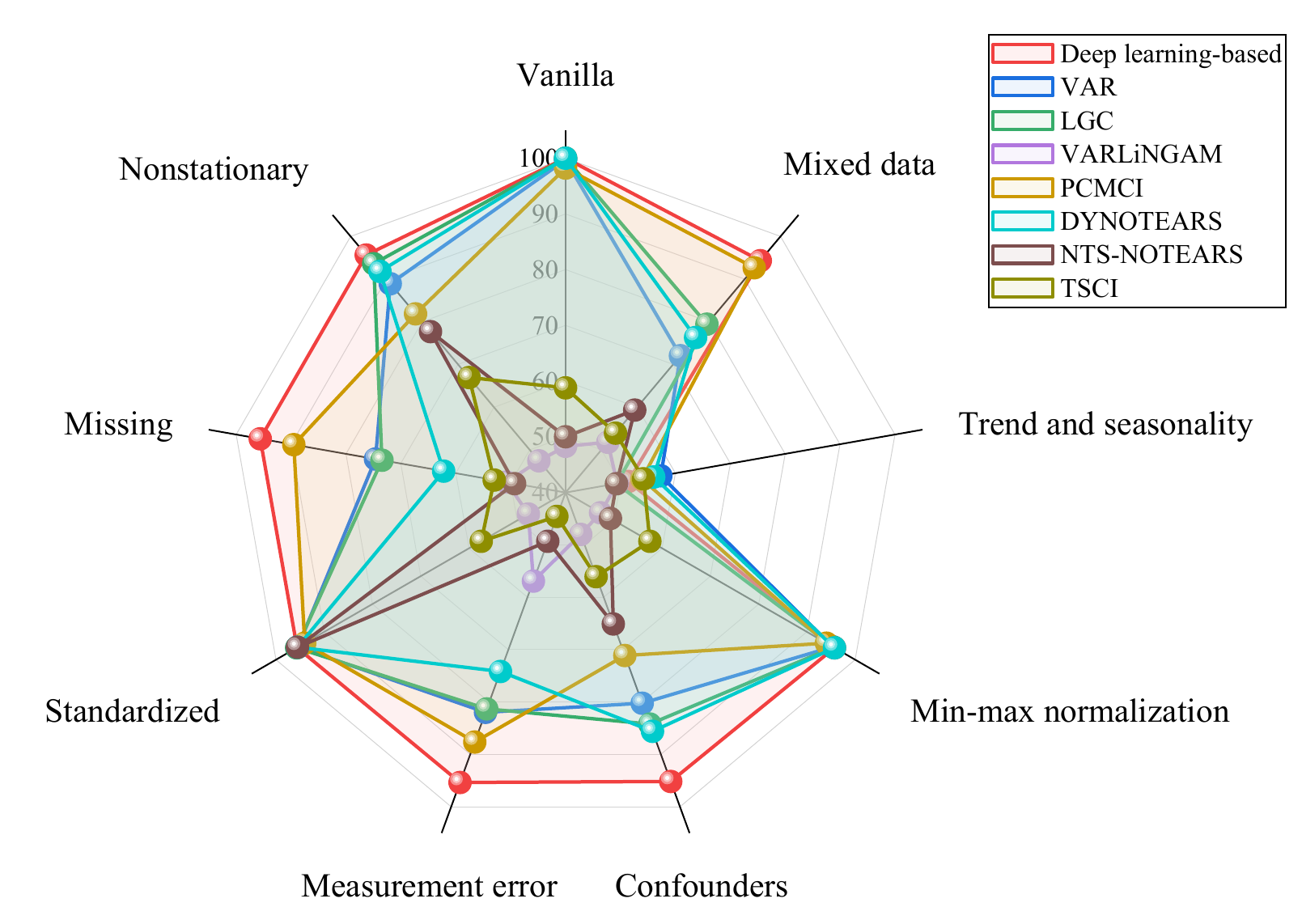}
         \caption{AUROC for linear 10-node case with $T = 1000$. Hyperparameters are selected to maximize average performance across all scenarios.}
         \label{fig:linear_10_1000_auroc_avg_scen}
     \end{subfigure}%
     \hfill
     \begin{subfigure}[b]{0.49\textwidth}
         \centering
         \includegraphics[width=\textwidth]{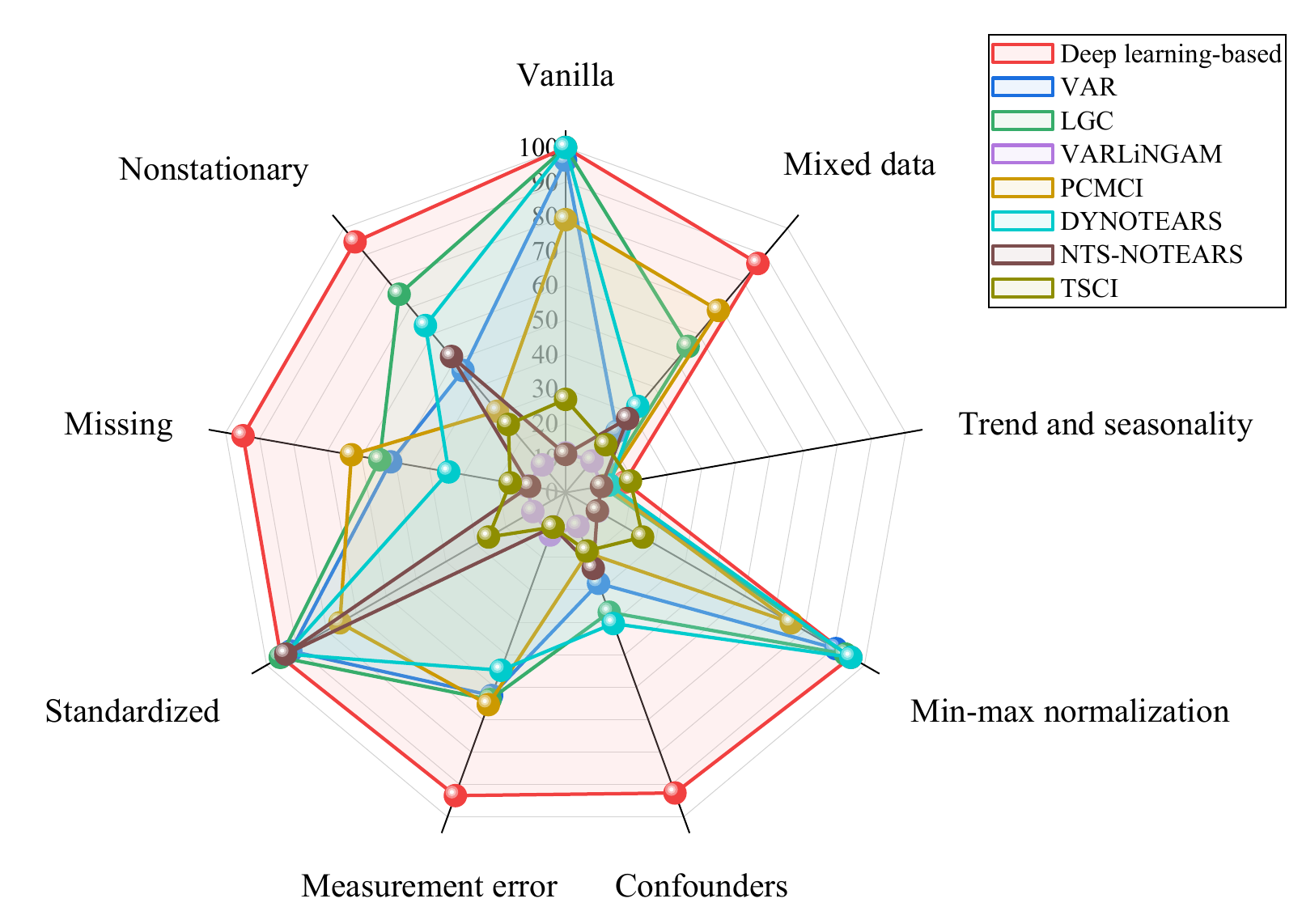}
         \caption{AUPRC for linear 10-node case with $T = 1000$. Hyperparameters are selected to maximize average performance across all scenarios.}
         \label{fig:linear_10_1000_auprc_avg_scen}
     \end{subfigure}

     \medskip  

     \begin{subfigure}[b]{0.49\textwidth}
         \centering
        \includegraphics[width=\textwidth]{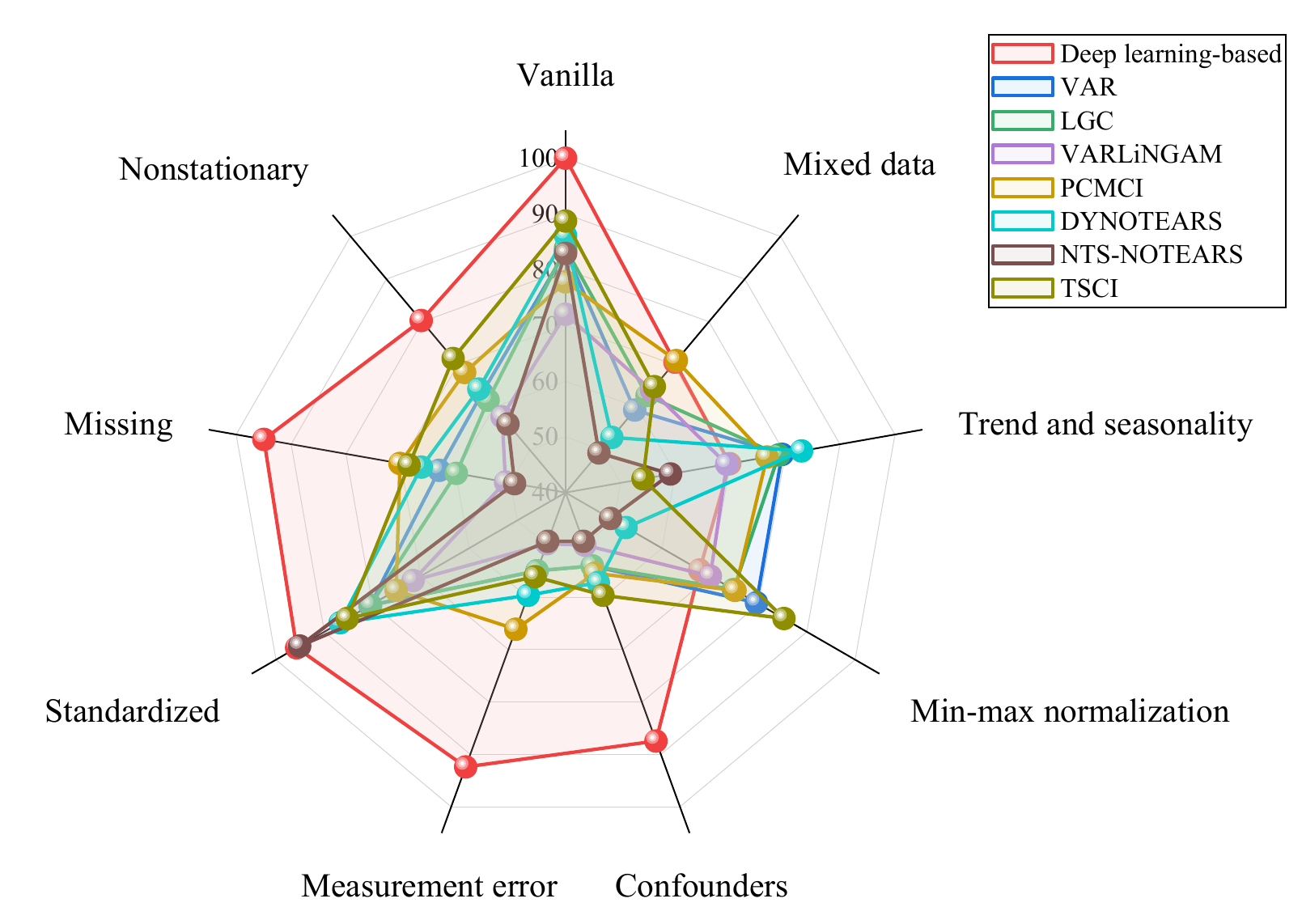}
         \caption{AUROC for nonlinear 10-node case with $T = 1000$ and $F=10$. Hyperparameters are selected to maximize average performance across all scenarios.}
         \label{fig:nonlinear_10_1000_f10_auroc_avg_scen}
     \end{subfigure}%
     \hfill
     \begin{subfigure}[b]{0.49\textwidth}
         \centering
         \includegraphics[width=\textwidth]{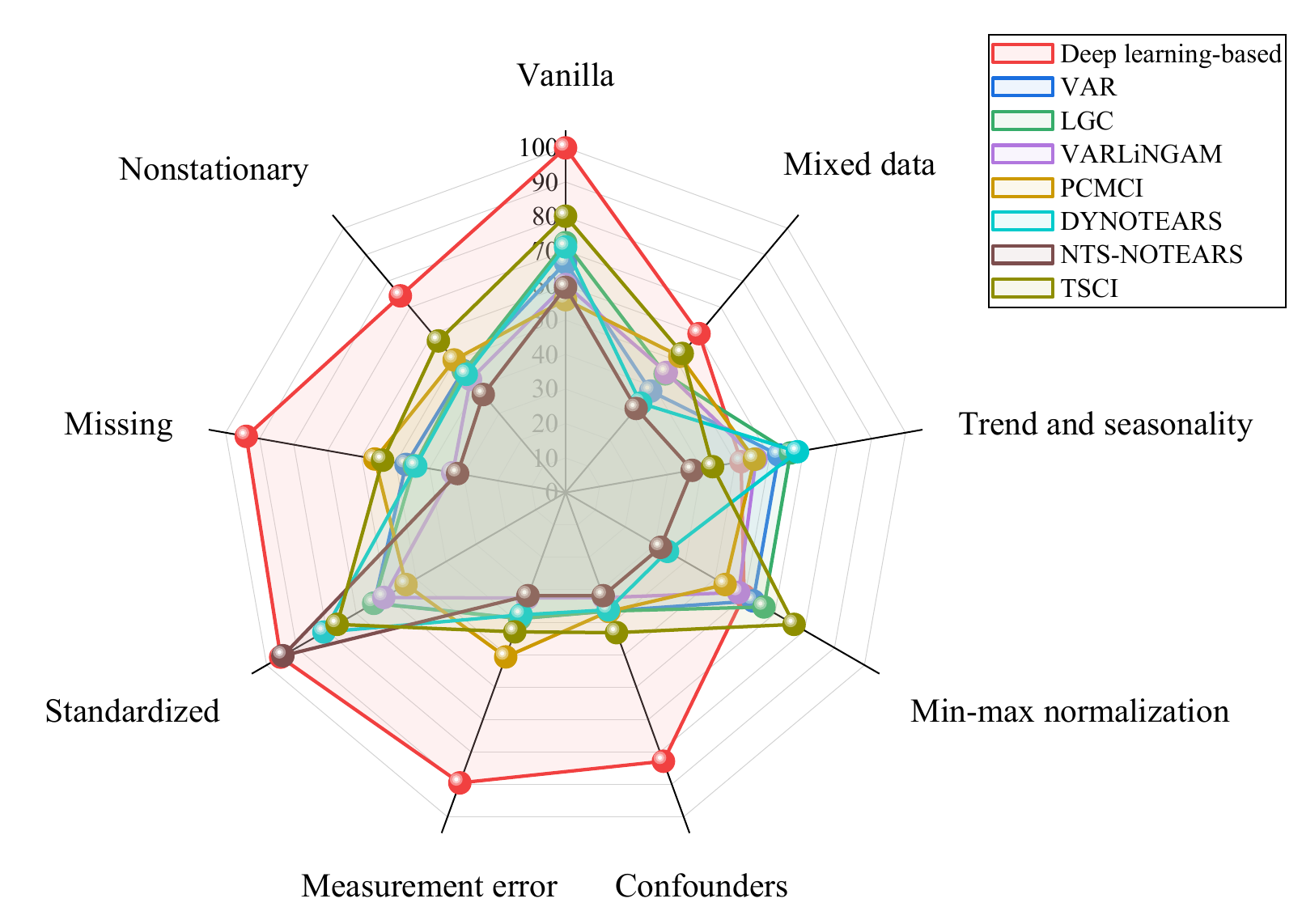}
         \caption{AUPRC for nonlinear 10-node case with $T = 1000$ and $F=10$. Hyperparameters are selected to maximize average performance across all scenarios.}
         \label{fig:nonlinear_10_1000_f10_auprc_avg_scen}
     \end{subfigure}

\Description{Four radar charts comparing causal discovery methods on 10-node networks with T=1000 across 9 scenarios. Superior performance is predominantly achieved by deep learning-based approaches. Hyperparameters that achieve the best average performance across all scenarios are used.}
\caption{Experimental results under the linear and nonlinear settings across the vanilla scenario and eight assumption violation scenarios. AUROC and AUPRC (the higher the better) are evaluated over 5 trials for the 10-node case with $T = 1000$. For the deep learning-based methods, we present only the optimal results. Hyperparameters are selected to maximize average performance across all scenarios.}
\label{fig:experiments_10_1000_f10_avg_scen}
\end{figure*}

\begin{figure*}[t]
     \centering
     \begin{subfigure}[b]{0.49\textwidth}
         \centering
        \includegraphics[width=\textwidth]{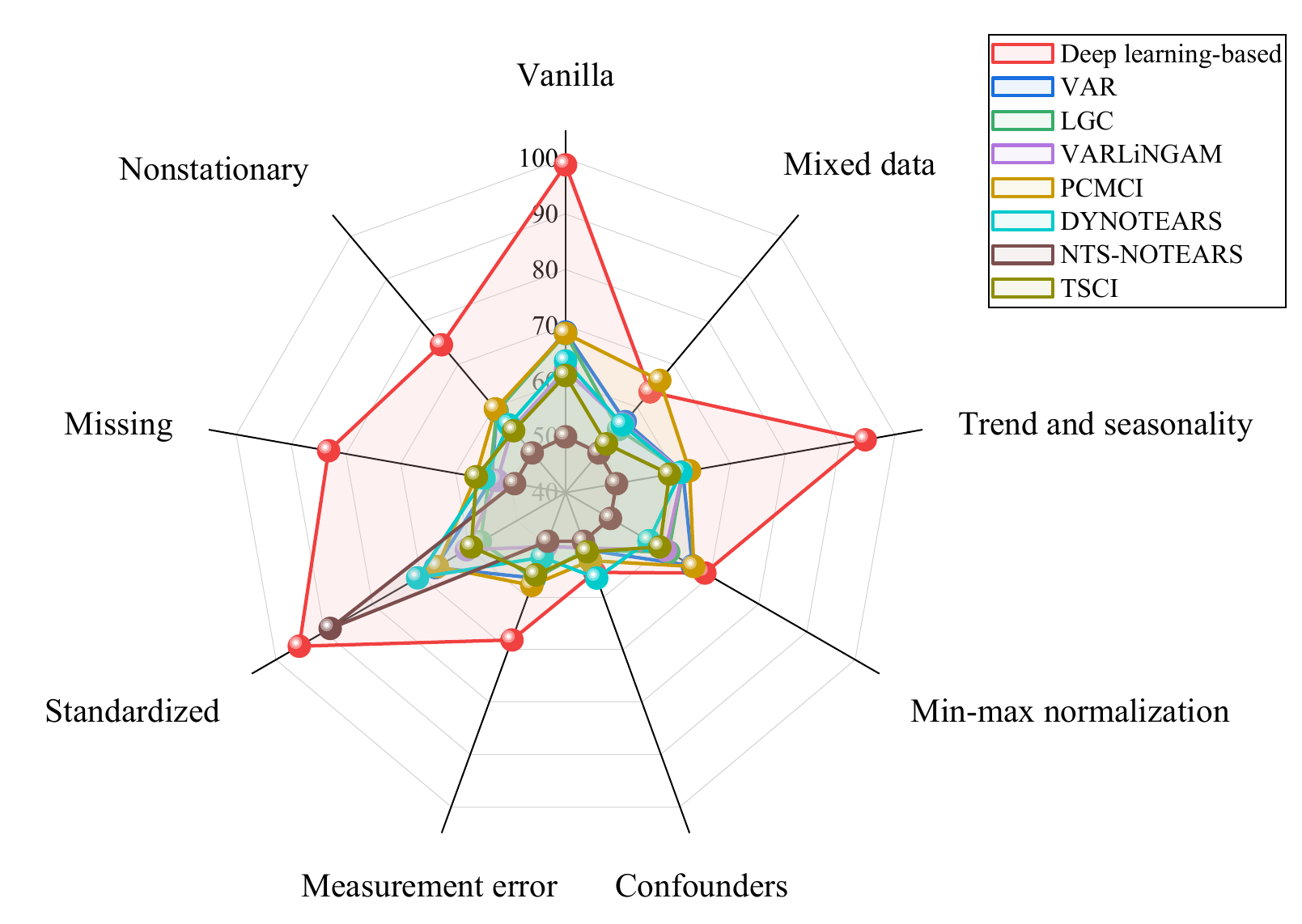}
         \caption{AUROC for nonlinear 10-node case with $T = 500$ and $F=40$. Hyperparameters are selected to maximize average performance across all scenarios.}
         \label{fig:nonlinear_10_500_f40_auroc_avg_scen}
     \end{subfigure}%
     \hfill
     \begin{subfigure}[b]{0.49\textwidth}
         \centering
         \includegraphics[width=\textwidth]{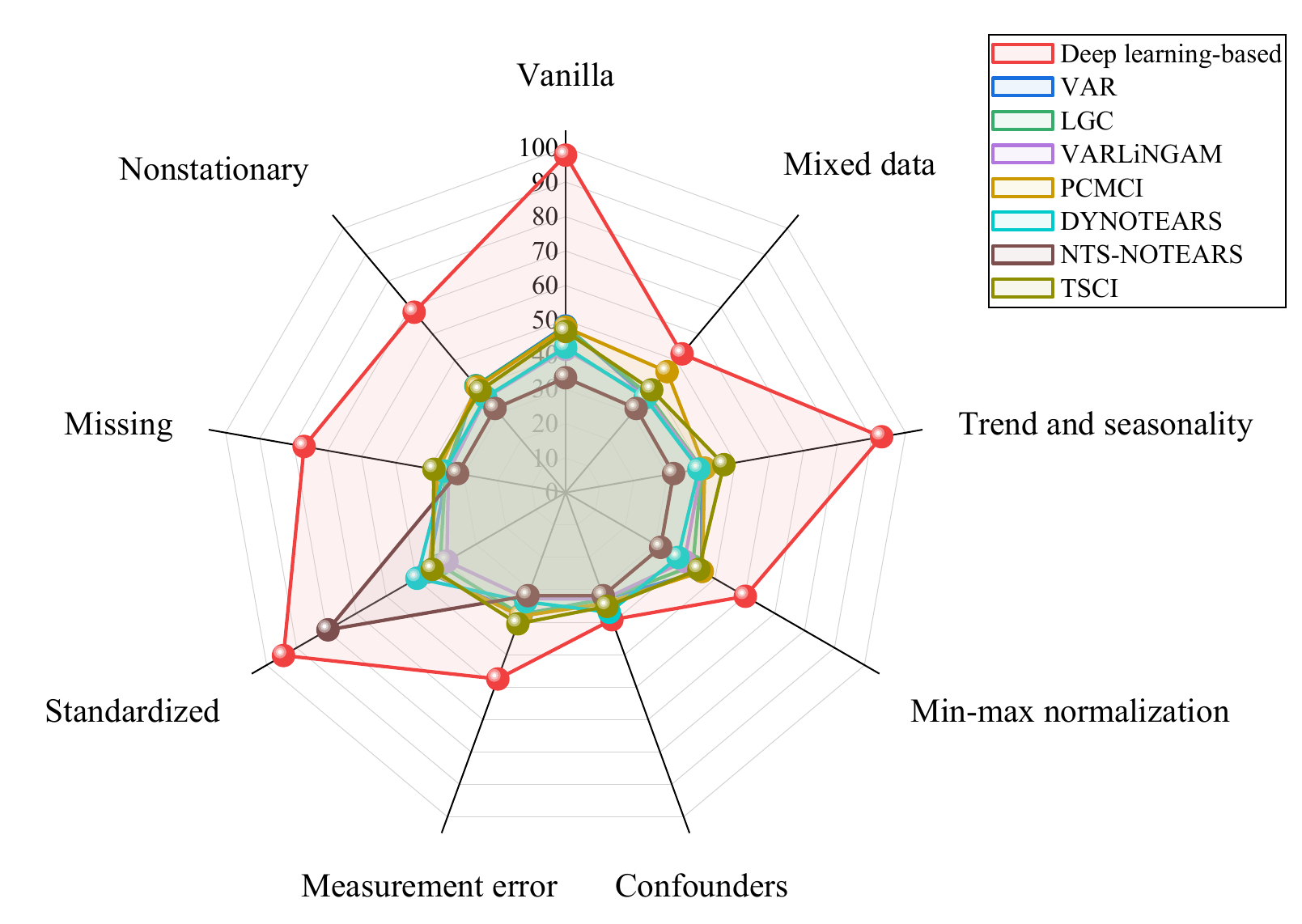}
         \caption{AUPRC for nonlinear 10-node case with $T = 500$ and $F=40$. Hyperparameters are selected to maximize average performance across all scenarios.}
         \label{fig:nonlinear_10_500_f40_auprc_avg_scen}
     \end{subfigure}

     \medskip  

     \begin{subfigure}[b]{0.49\textwidth}
         \centering
        \includegraphics[width=\textwidth]{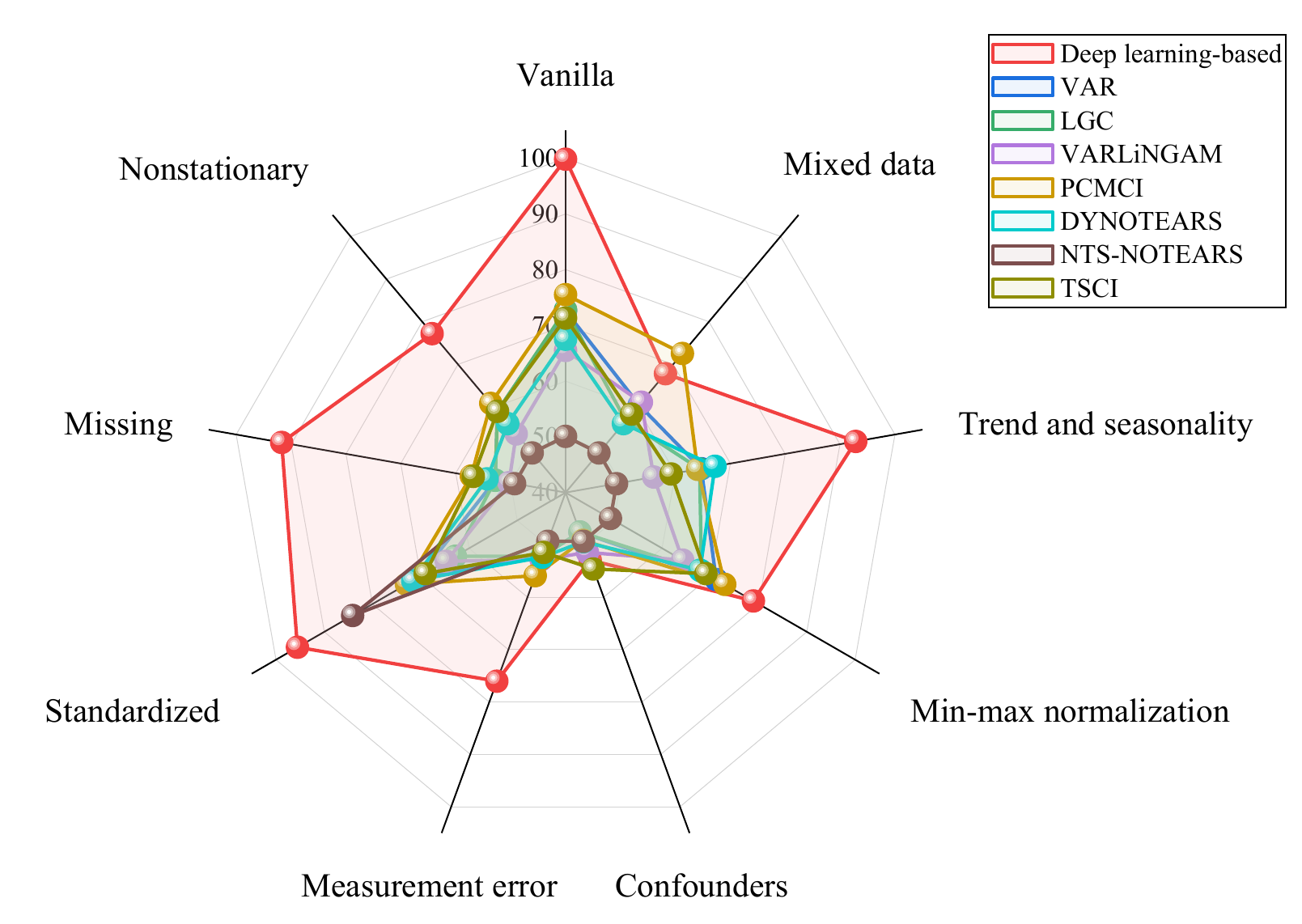}
         \caption{AUROC for nonlinear 10-node case with $T = 1000$ and $F=40$. Hyperparameters are selected to maximize average performance across all scenarios.}
         \label{fig:nonlinear_10_1000_f40_auroc_avg_scen}
     \end{subfigure}%
     \hfill
     \begin{subfigure}[b]{0.49\textwidth}
         \centering
         \includegraphics[width=\textwidth]{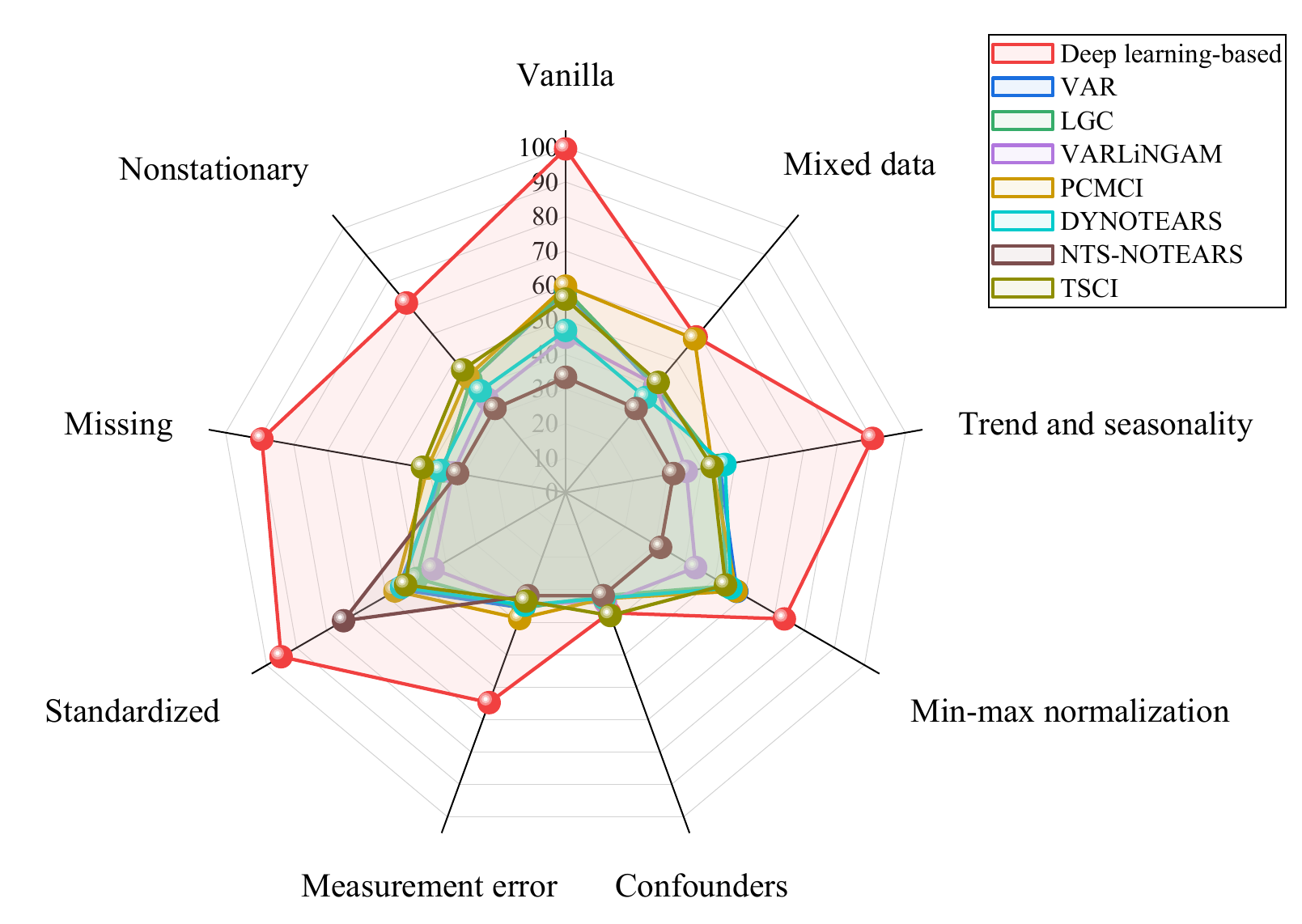}
         \caption{AUPRC for nonlinear 10-node case with $T = 1000$ and $F=40$. Hyperparameters are selected to maximize average performance across all scenarios.}
         \label{fig:nonlinear_10_1000_f40_auprc_avg_scen}
     \end{subfigure}

\Description{Four radar charts comparing causal discovery methods on 10-node networks with F=40 across 9 scenarios. Superior performance is predominantly achieved by deep learning-based approaches. Hyperparameters that achieve the best average performance across all scenarios are used.}
\caption{Experimental results under the nonlinear settings across the vanilla scenario and eight assumption violation scenarios. AUROC and AUPRC (the higher the better) are evaluated over 5 trials for the 10-node case with $F = 40$. For the deep learning-based methods, we present only the optimal results. Hyperparameters are selected to maximize average performance across all scenarios.}
\label{fig:experiments_10_500_1000_f40_avg_scen}
\end{figure*}

\begin{figure*}[t]
     \centering
     \begin{subfigure}[b]{0.49\textwidth}
         \centering
        \includegraphics[width=\textwidth]{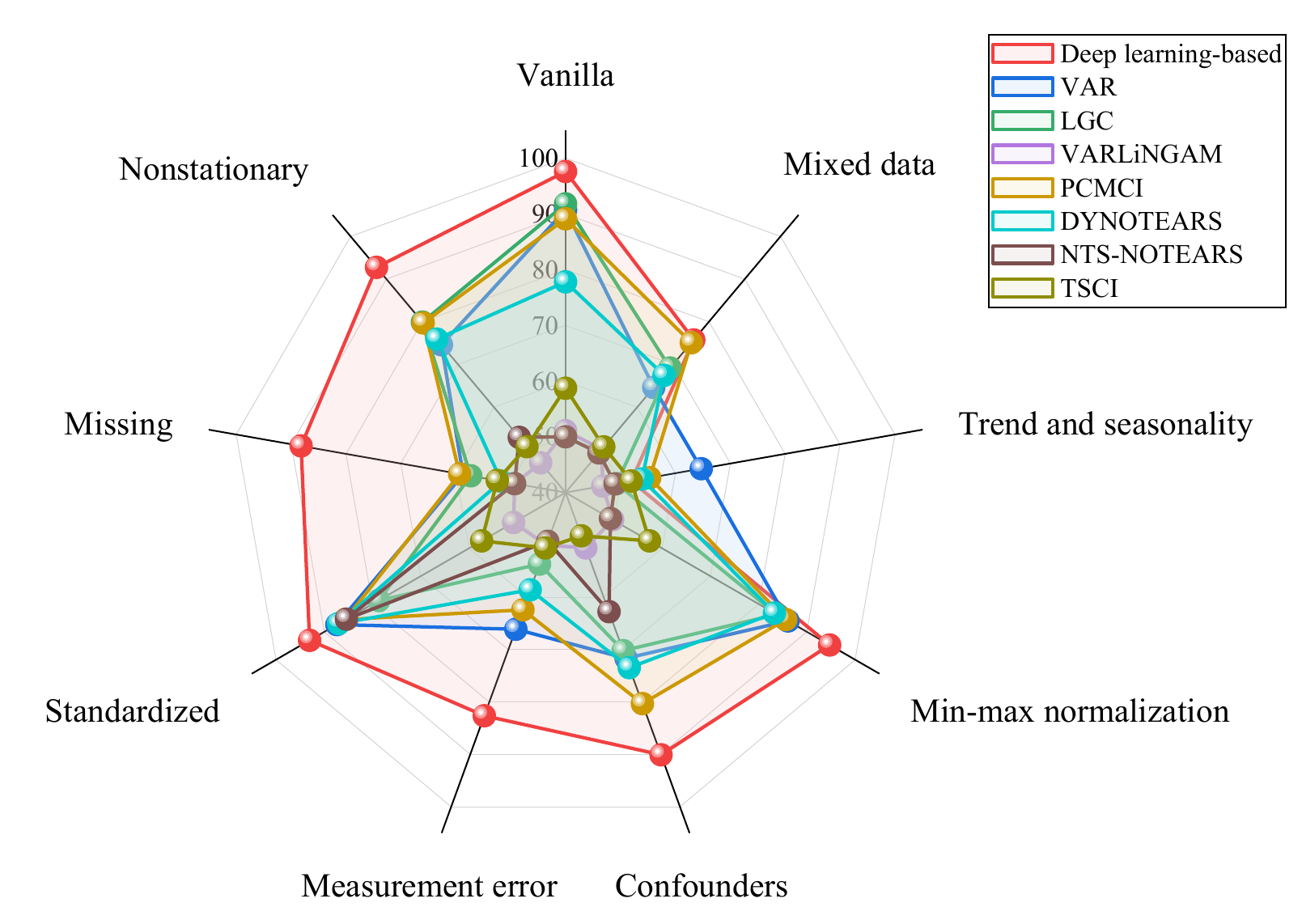}
         \caption{AUROC for linear 15-node case with $T = 500$. Hyperparameters are selected to maximize average performance across all scenarios.}
         \label{fig:linear_15_500_auroc_avg_scen}
     \end{subfigure}%
     \hfill
     \begin{subfigure}[b]{0.49\textwidth}
         \centering
         \includegraphics[width=\textwidth]{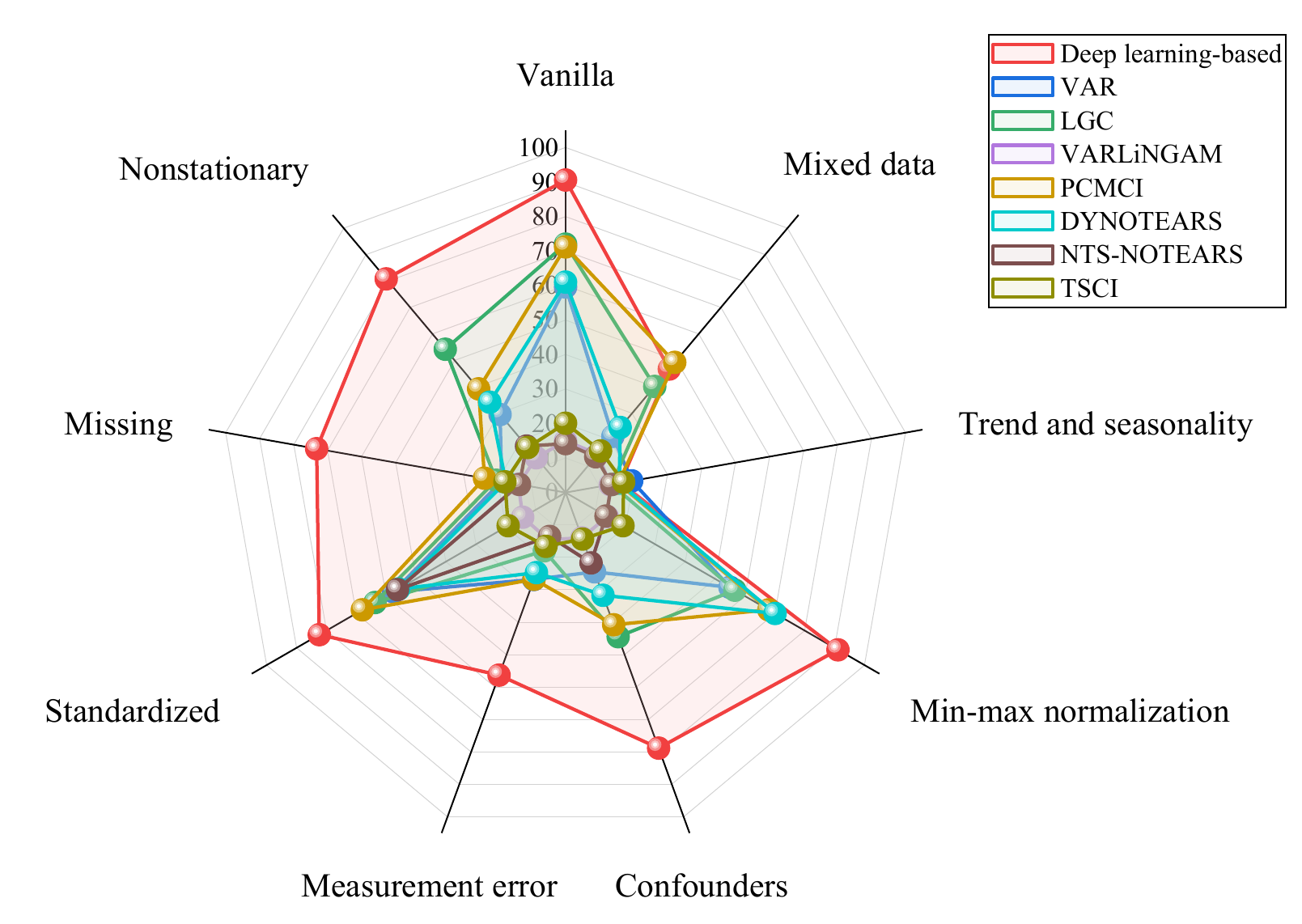}
         \caption{AUPRC for linear 15-node case with $T = 500$. Hyperparameters are selected to maximize average performance across all scenarios.}
         \label{fig:linear_15_500_auprc_avg_scen}
     \end{subfigure}

     \medskip  

     \begin{subfigure}[b]{0.49\textwidth}
         \centering
        \includegraphics[width=\textwidth]{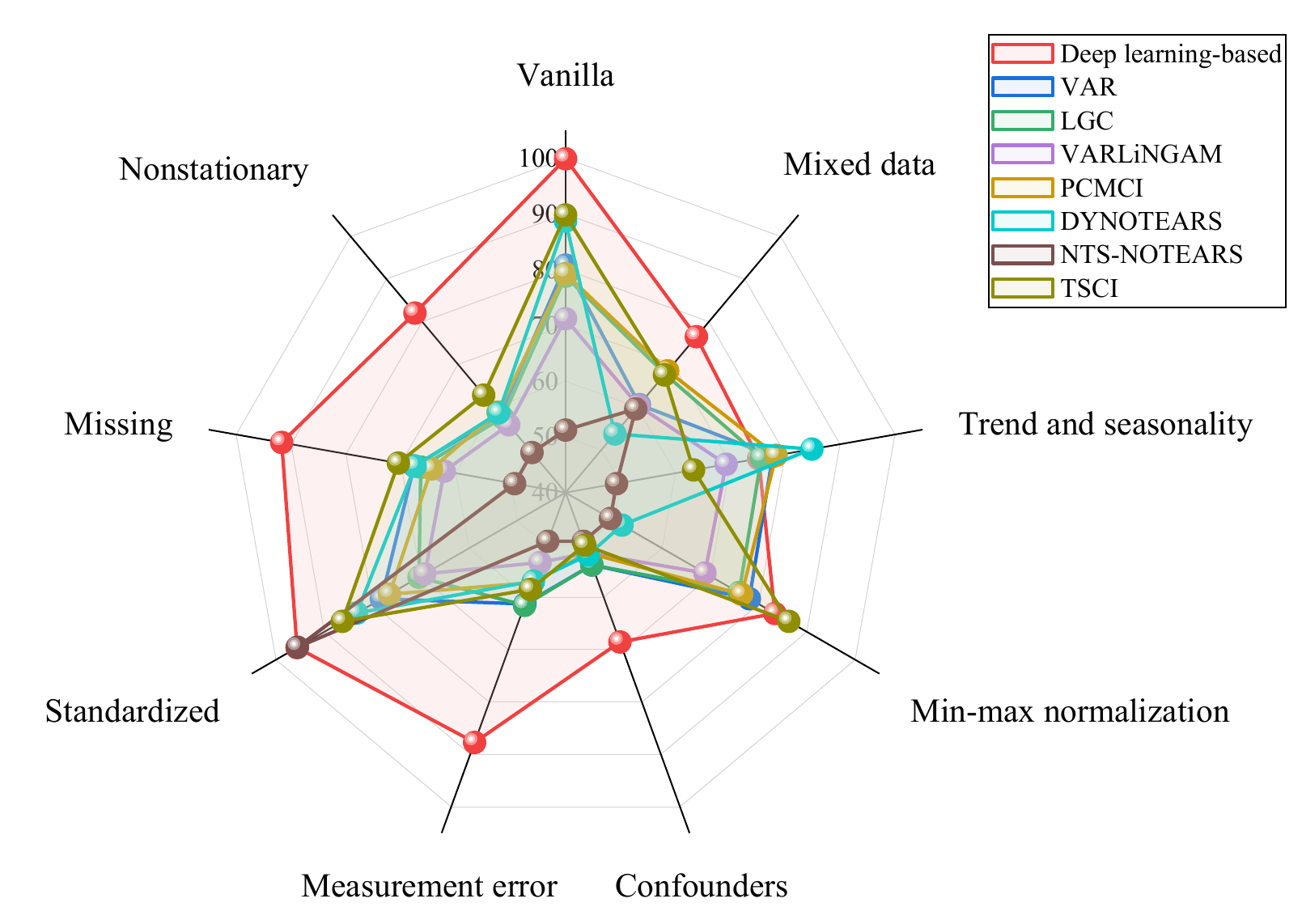}
         \caption{AUROC for nonlinear 15-node case with $T = 500$ and $F=10$. Hyperparameters are selected to maximize average performance across all scenarios.}
         \label{fig:nonlinear_15_500_f10_auroc_avg_scen}
     \end{subfigure}%
     \hfill
     \begin{subfigure}[b]{0.49\textwidth}
         \centering
         \includegraphics[width=\textwidth]{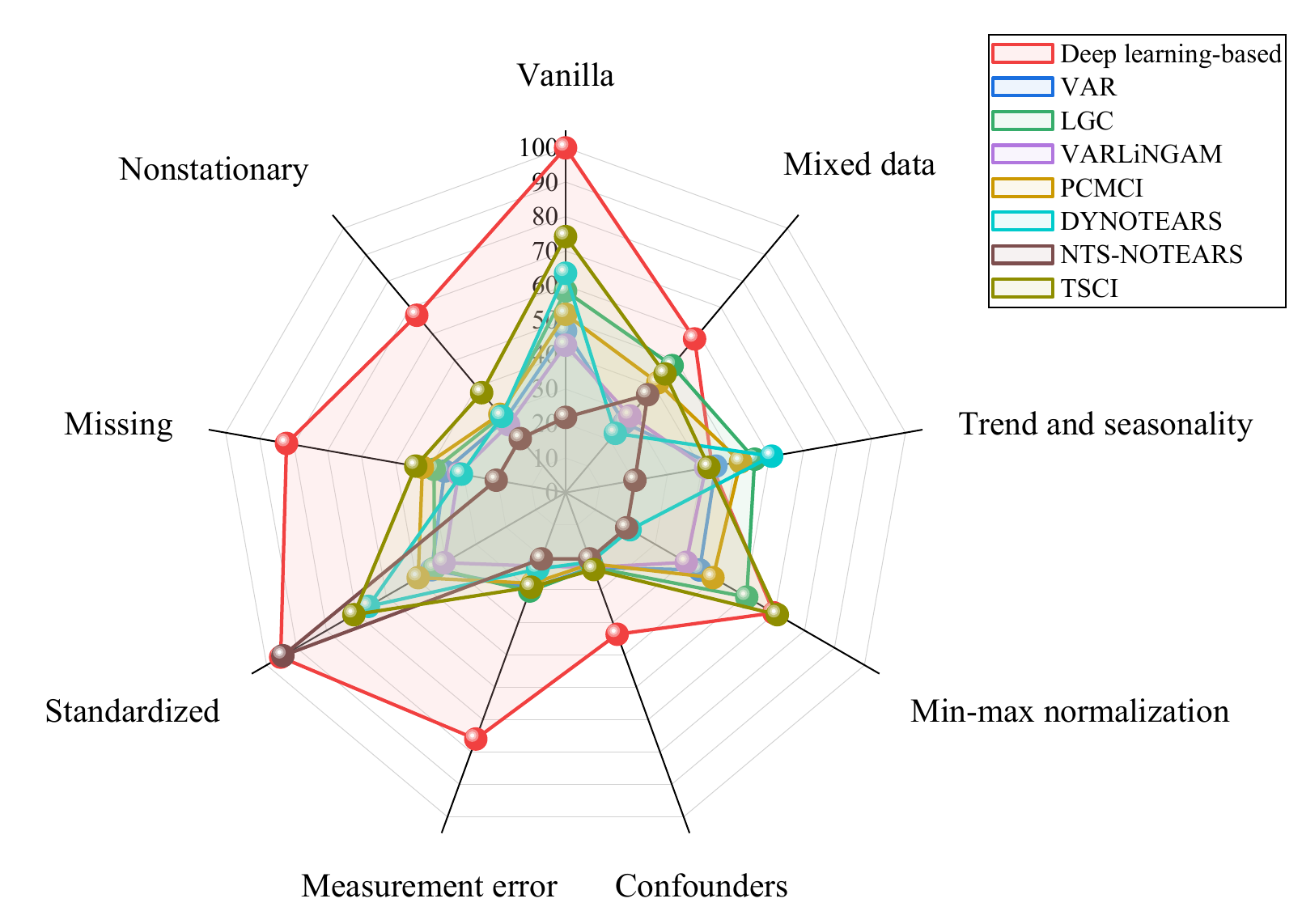}
         \caption{AUPRC for nonlinear 15-node case with $T = 500$ and $F=10$. Hyperparameters are selected to maximize average performance across all scenarios.}
         \label{fig:nonlinear_15_500_f10_auprc_avg_scen}
     \end{subfigure}

\Description{Four radar charts comparing causal discovery methods on 15-node networks with T=500 across 9 scenarios. Superior performance is predominantly achieved by deep learning-based approaches. Hyperparameters that achieve the best average performance across all scenarios are used.}
\caption{Experimental results under the linear and nonlinear settings across the vanilla scenario and eight assumption violation scenarios. AUROC and AUPRC (the higher the better) are evaluated over 5 trials for the 15-node case with $T = 500$. For the deep learning-based methods, we present only the optimal results. Hyperparameters are selected to maximize average performance across all scenarios.}
\label{fig:experiments_15_500_f10_avg_scen}
\end{figure*}

\begin{figure*}[t]
     \centering
     \begin{subfigure}[b]{0.49\textwidth}
         \centering
        \includegraphics[width=\textwidth]{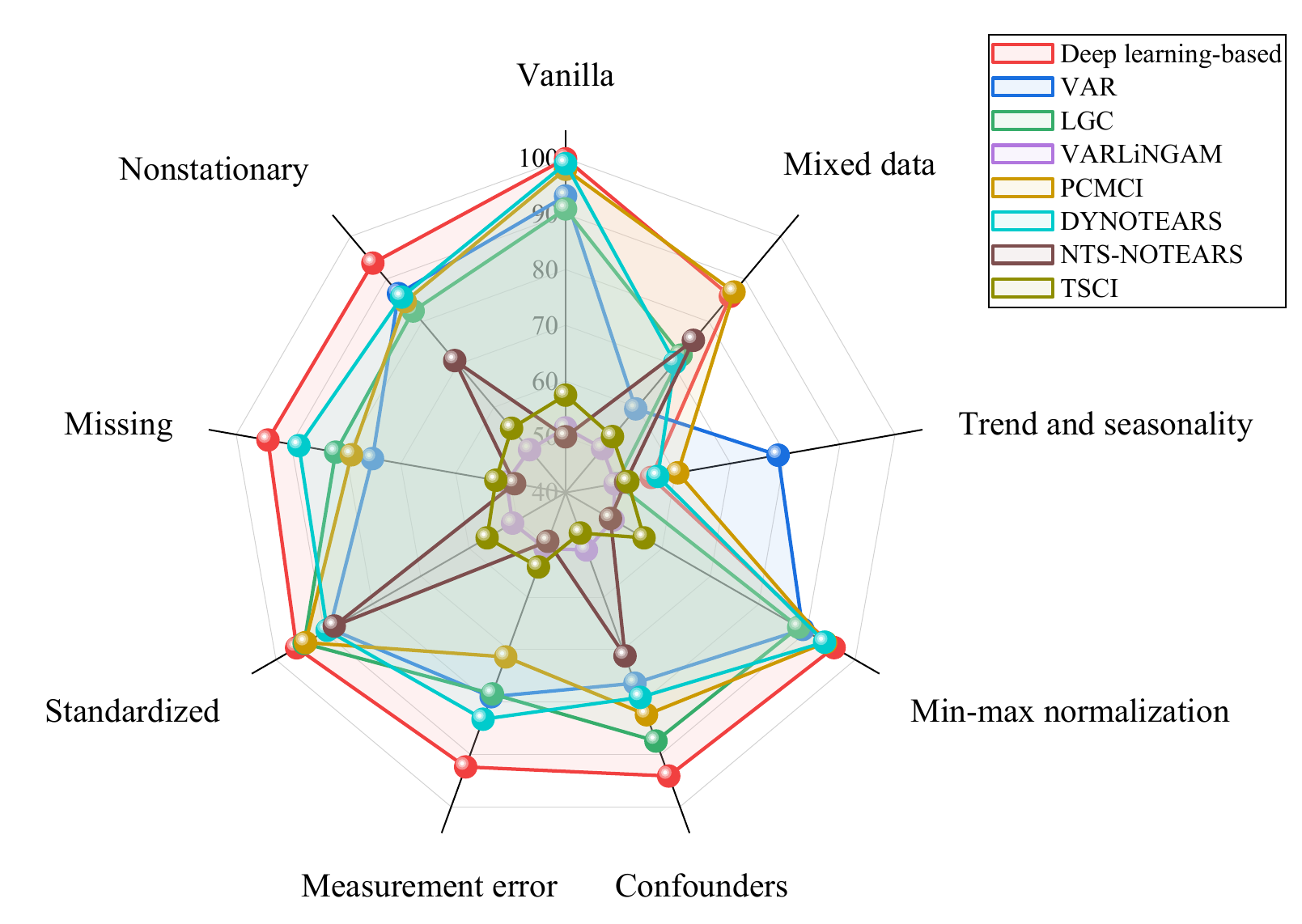}
         \caption{AUROC for linear 15-node case with $T = 1000$. Hyperparameters are selected to maximize average performance across all scenarios.}
         \label{fig:linear_15_1000_auroc_avg_scen}
     \end{subfigure}%
     \hfill
     \begin{subfigure}[b]{0.49\textwidth}
         \centering
         \includegraphics[width=\textwidth]{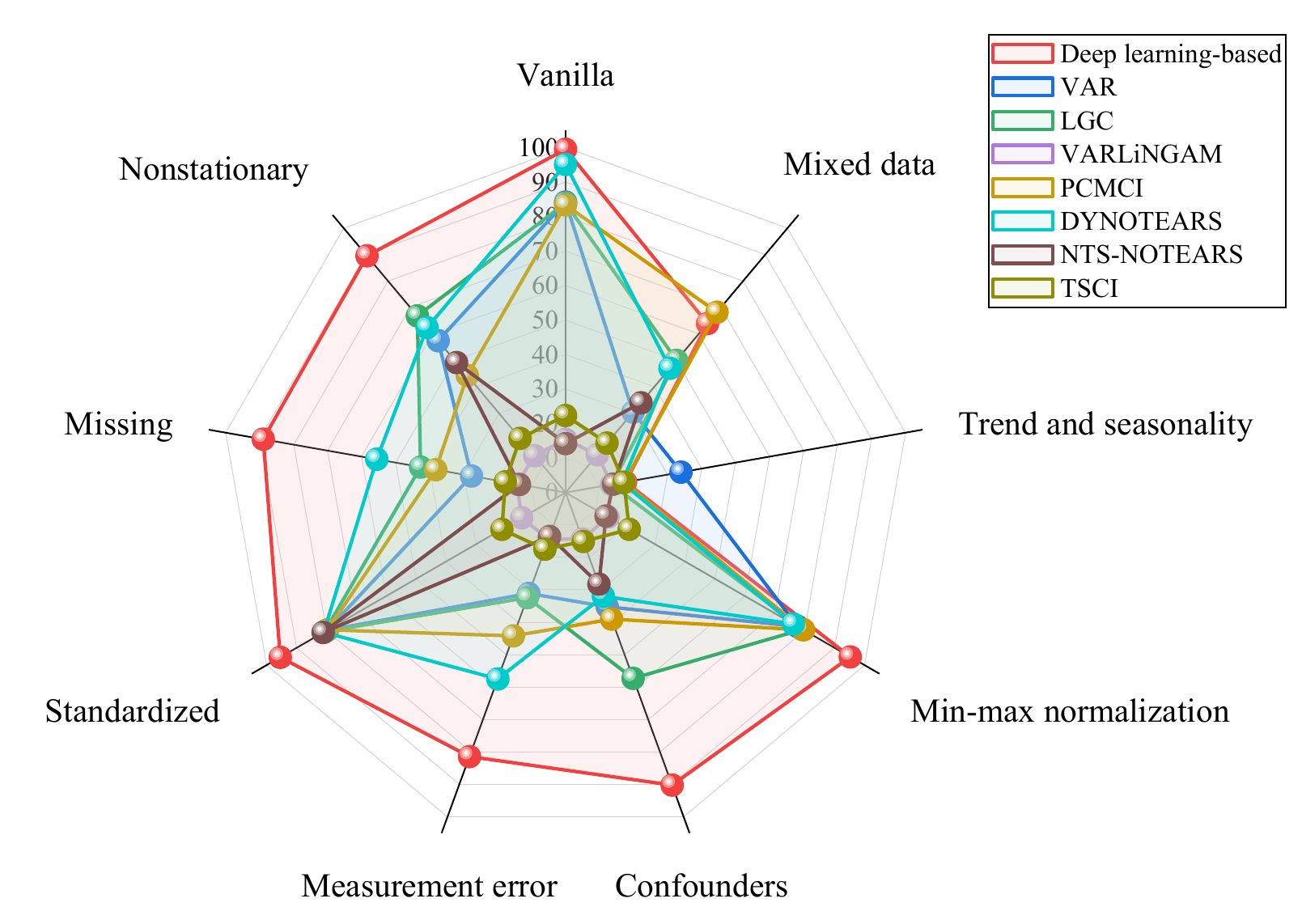}
         \caption{AUPRC for linear 15-node case with $T = 1000$. Hyperparameters are selected to maximize average performance across all scenarios.}
         \label{fig:linear_15_1000_auprc_avg_scen}
     \end{subfigure}

     \medskip  

     \begin{subfigure}[b]{0.49\textwidth}
         \centering
        \includegraphics[width=\textwidth]{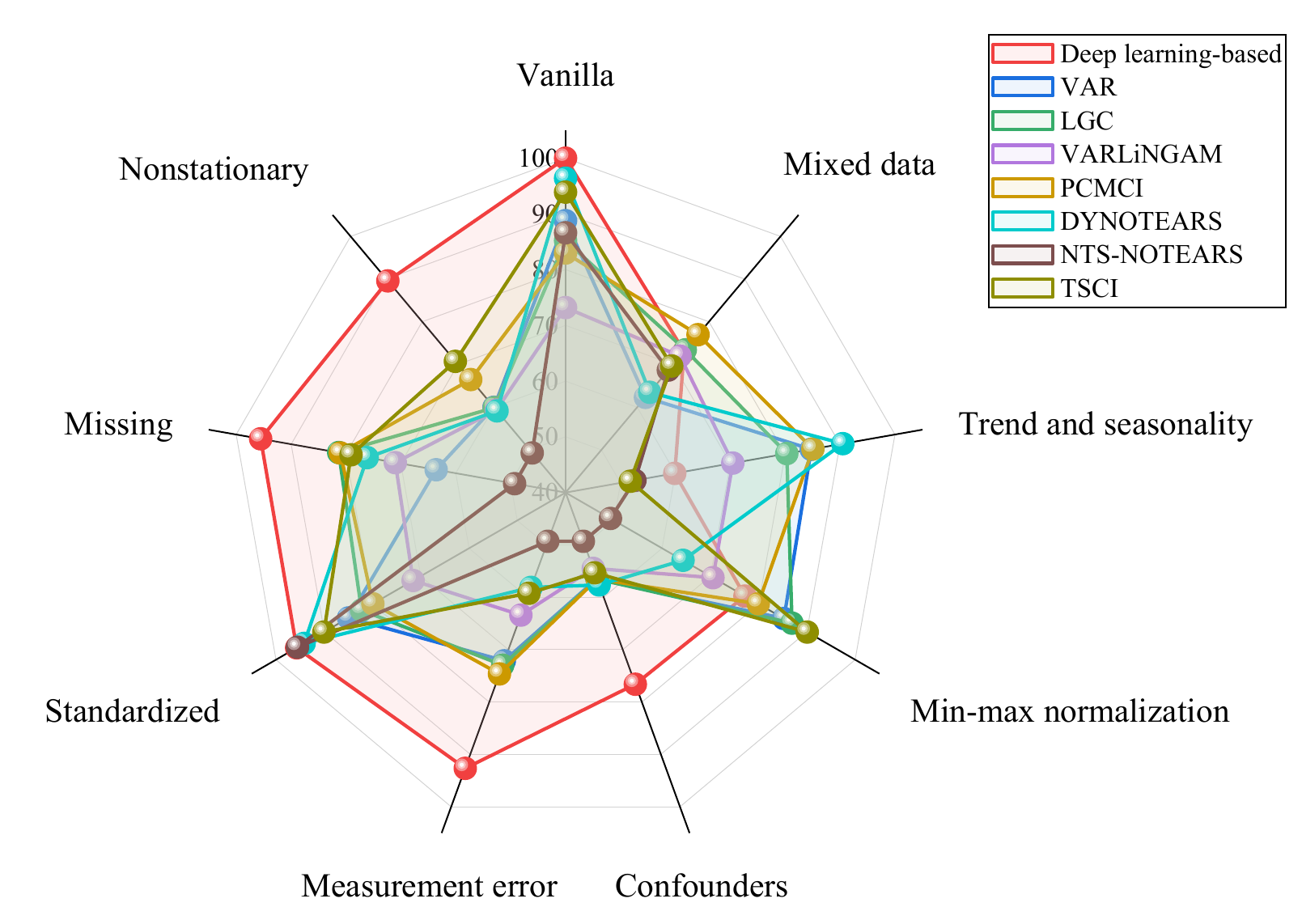}
         \caption{AUROC for nonlinear 15-node case with $T = 1000$ and $F=10$. Hyperparameters are selected to maximize average performance across all scenarios.}
         \label{fig:nonlinear_15_1000_f10_auroc_avg_scen}
     \end{subfigure}%
     \hfill
     \begin{subfigure}[b]{0.49\textwidth}
         \centering
         \includegraphics[width=\textwidth]{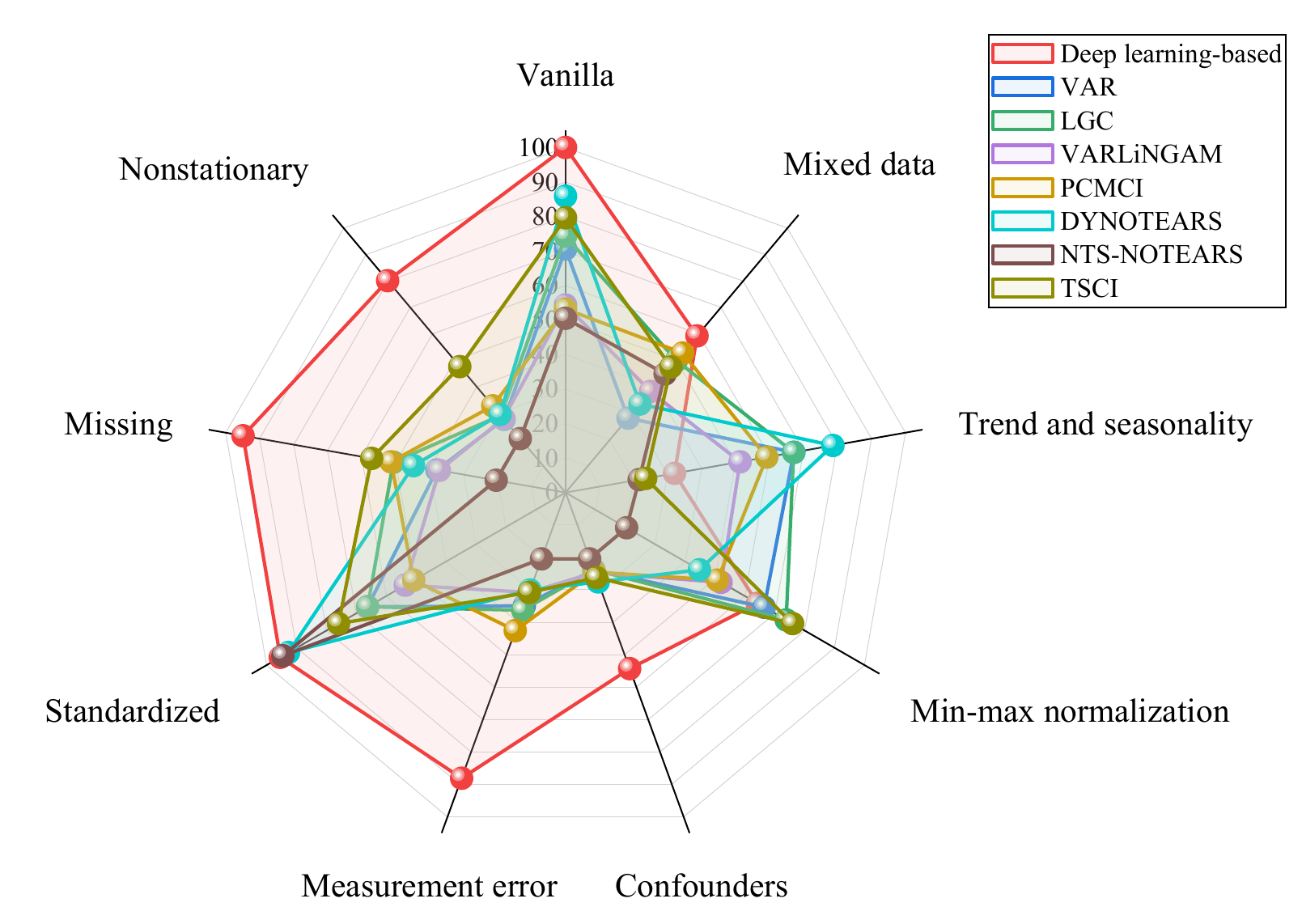}
         \caption{AUPRC for nonlinear 15-node case with $T = 1000$ and $F=10$. Hyperparameters are selected to maximize average performance across all scenarios.}
         \label{fig:nonlinear_15_1000_f10_auprc_avg_scen}
     \end{subfigure}

\Description{Four radar charts comparing causal discovery methods on 15-node networks with T=1000 across 9 scenarios. Superior performance is predominantly achieved by deep learning-based approaches. Hyperparameters that achieve the best average performance across all scenarios are used.}
\caption{Experimental results under the linear and nonlinear settings across the vanilla scenario and eight assumption violation scenarios. AUROC and AUPRC (the higher the better) are evaluated over 5 trials for the 15-node case with $T = 1000$. For the deep learning-based methods, we present only the optimal results. Hyperparameters are selected to maximize average performance across all scenarios.}
\label{fig:experiments_15_1000_f10_avg_scen}
\end{figure*}

\begin{figure*}[t]
     \centering
     \begin{subfigure}[b]{0.49\textwidth}
         \centering
        \includegraphics[width=\textwidth]{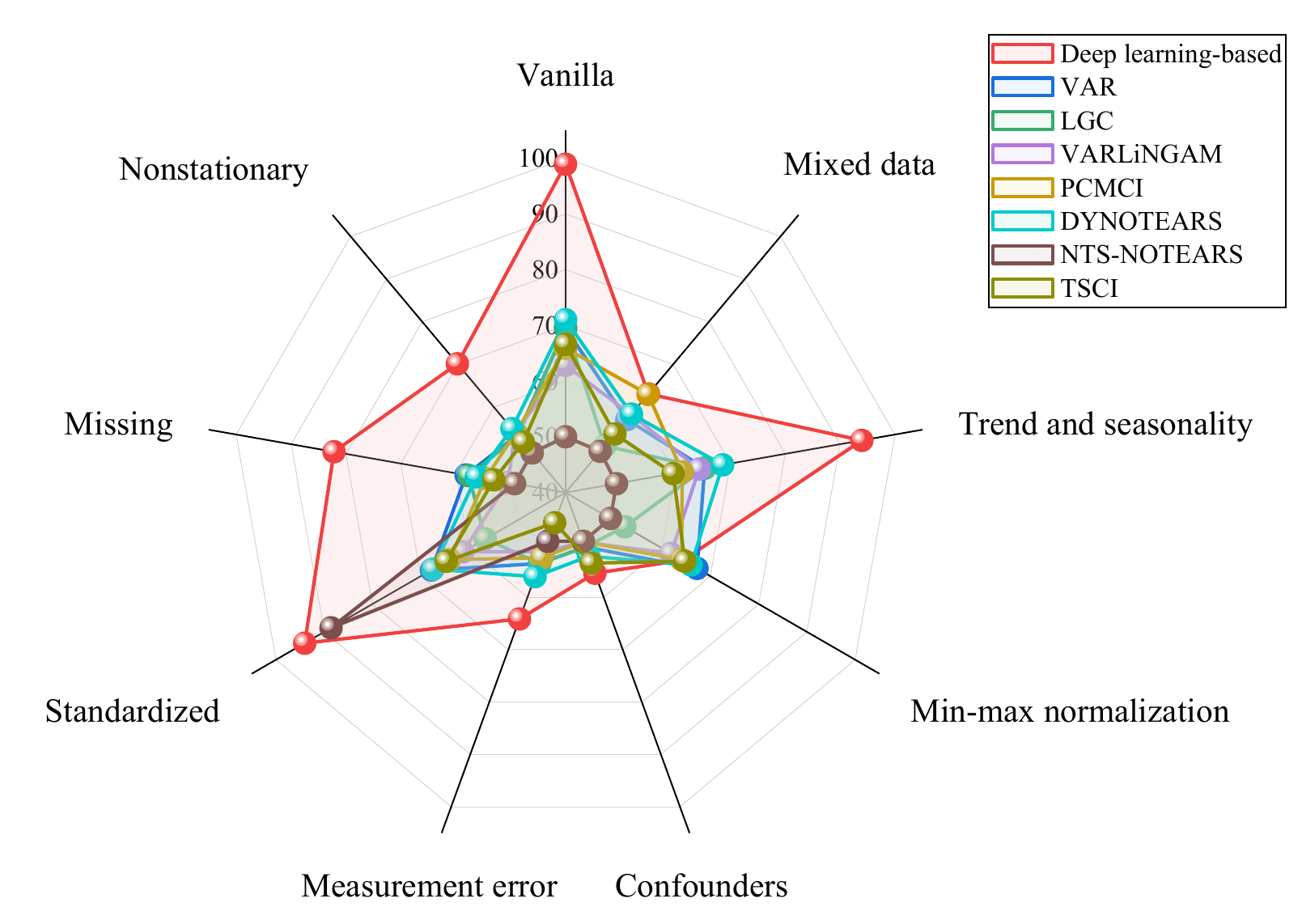}
         \caption{AUROC for nonlinear 15-node case with $T = 500$ and $F=40$. Hyperparameters are selected to maximize average performance across all scenarios.}
         \label{fig:nonlinear_15_500_f40_auroc_avg_scen}
     \end{subfigure}%
     \hfill
     \begin{subfigure}[b]{0.49\textwidth}
         \centering
         \includegraphics[width=\textwidth]{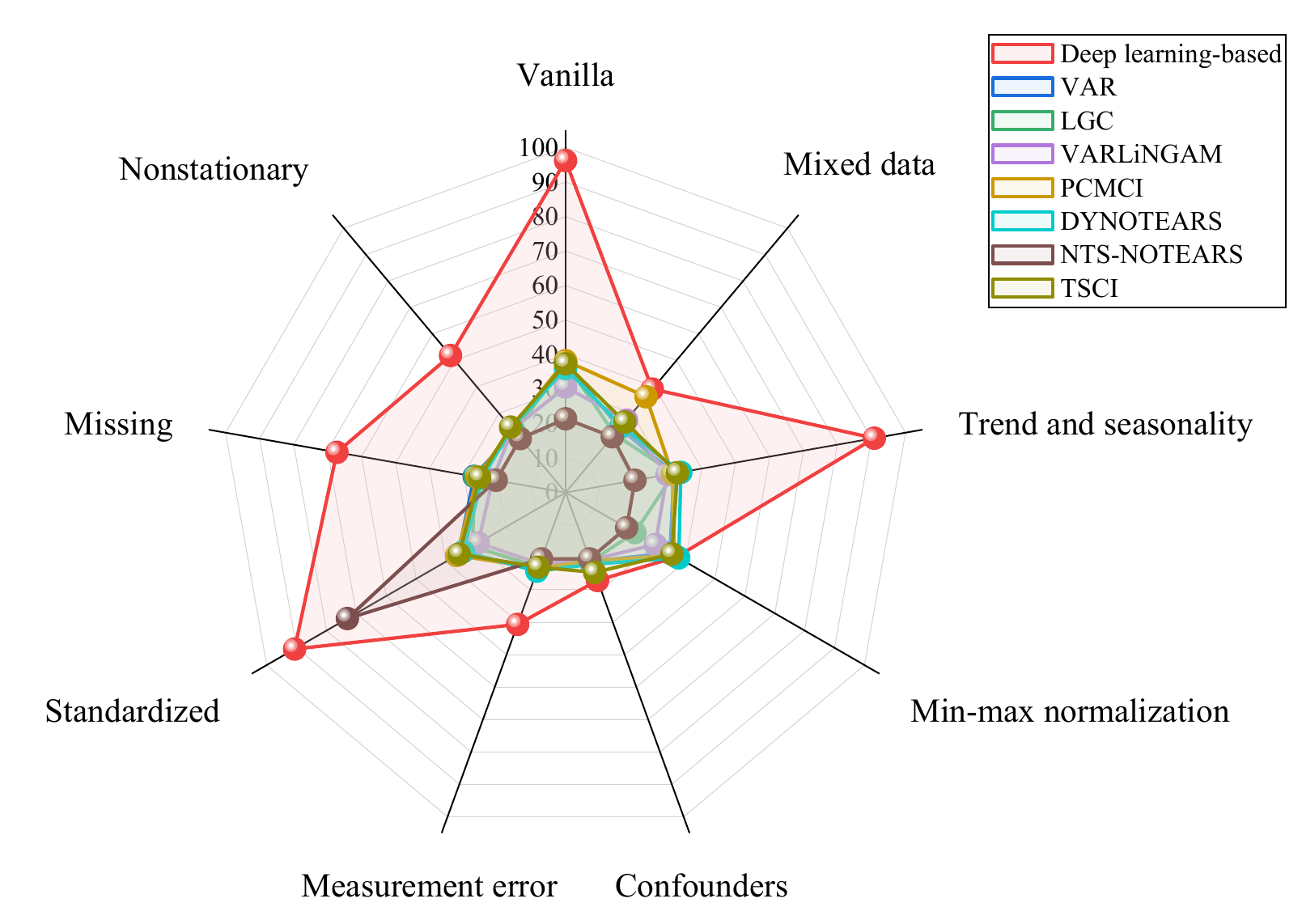}
         \caption{AUPRC for nonlinear 15-node case with $T = 500$ and $F=40$. Hyperparameters are selected to maximize average performance across all scenarios.}
         \label{fig:nonlinear_15_500_f40_auprc_avg_scen}
     \end{subfigure}

     \medskip  

     \begin{subfigure}[b]{0.49\textwidth}
         \centering
        \includegraphics[width=\textwidth]{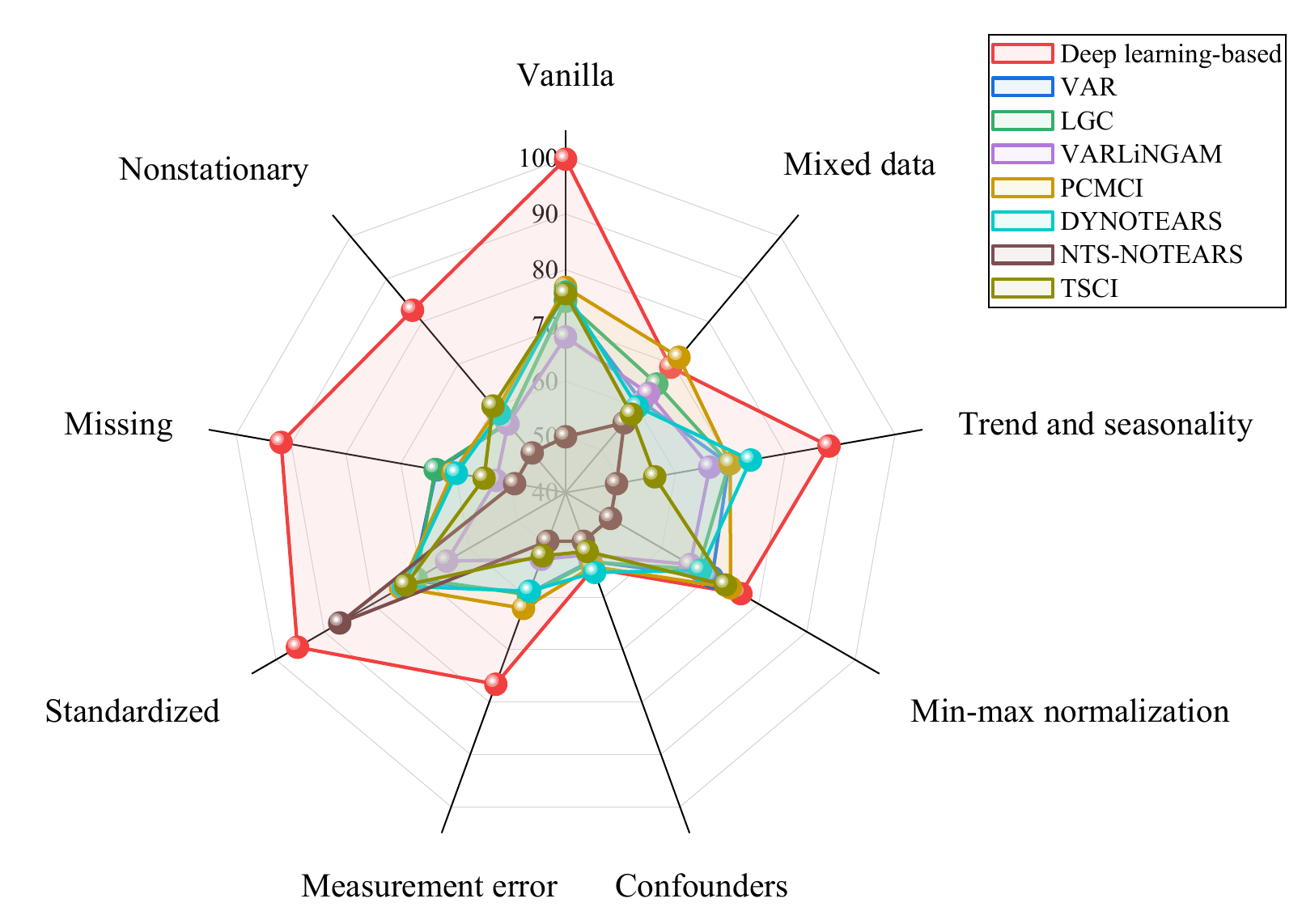}
         \caption{AUROC for nonlinear 15-node case with $T = 1000$ and $F=40$. Hyperparameters are selected to maximize average performance across all scenarios.}
         \label{fig:nonlinear_15_1000_f40_auroc_avg_scen}
     \end{subfigure}%
     \hfill
     \begin{subfigure}[b]{0.49\textwidth}
         \centering
         \includegraphics[width=\textwidth]{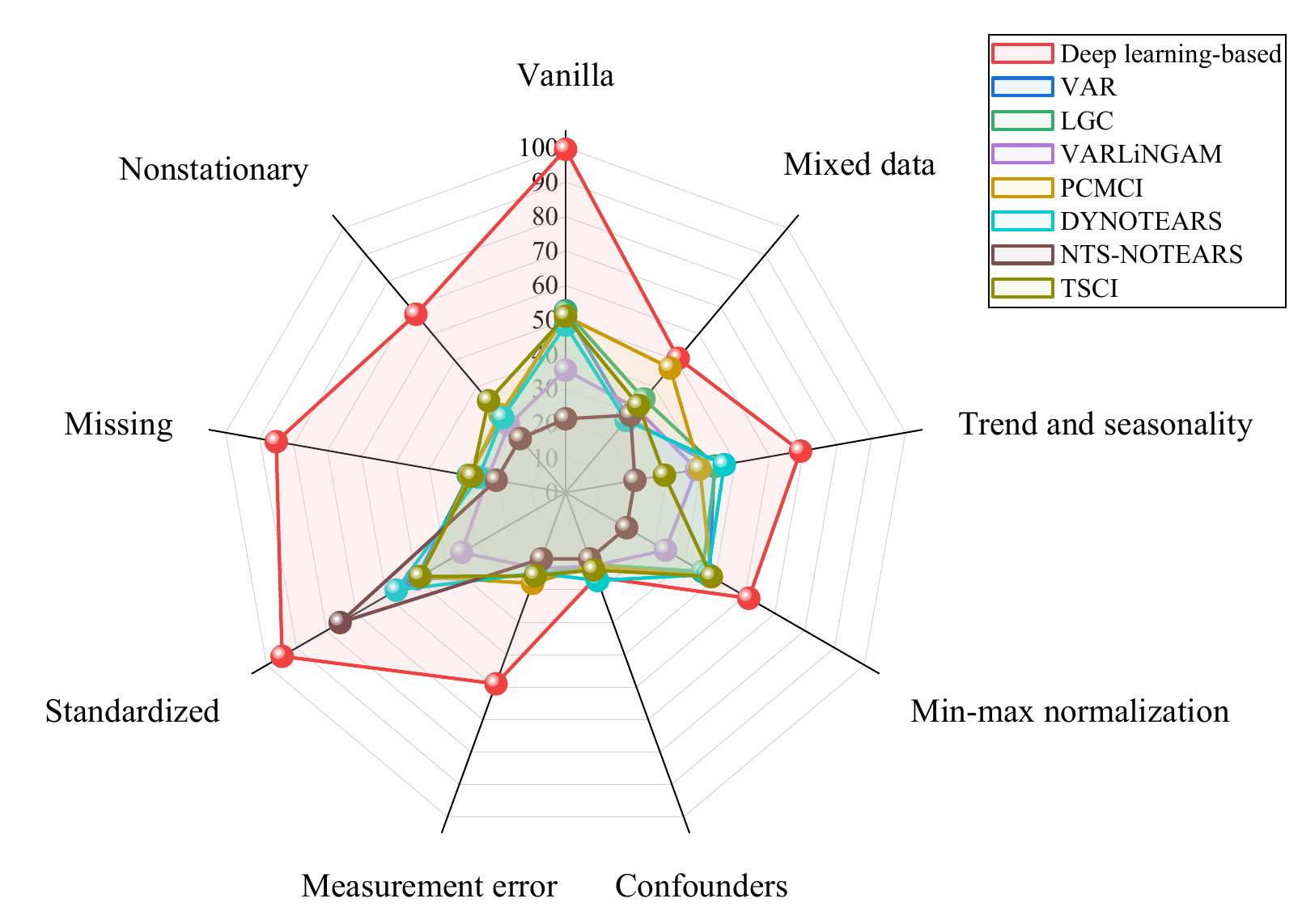}
         \caption{AUPRC for nonlinear 15-node case with $T = 1000$ and $F=40$. Hyperparameters are selected to maximize average performance across all scenarios.}
         \label{fig:nonlinear_15_1000_f40_auprc_avg_scen}
     \end{subfigure}

\Description{Four radar charts comparing causal discovery methods on 15-node networks with F=40 across 9 scenarios. Superior performance is predominantly achieved by deep learning-based approaches. Hyperparameters that achieve the best average performance across all scenarios are used.}
\caption{Experimental results under the nonlinear settings across the vanilla scenario and eight assumption violation scenarios. AUROC and AUPRC (the higher the better) are evaluated over 5 trials for the 15-node case with $F = 40$. For the deep learning-based methods, we present only the optimal results. Hyperparameters are selected to maximize average performance across all scenarios.}
\label{fig:experiments_15_500_1000_f40_avg_scen}
\end{figure*}

We select, for each causal discovery method, a single hyperparameter configuration that maximizes average performance across all scenarios. Additional figure results under both linear and nonlinear settings, spanning different numbers of nodes, time series lengths, and external forcing intensities, are reported in the Appendix (Figures~\ref{fig:experiments_10_500_f10_avg_scen}--\ref{fig:experiments_15_500_1000_f40_avg_scen}).

\section{Figure results aggregated over all hyperparameters}
\label{app_fig_results_avg_hyper}
\begin{figure*}[t]
     \centering
     \begin{subfigure}[b]{0.49\textwidth}
         \centering
        \includegraphics[width=\textwidth]{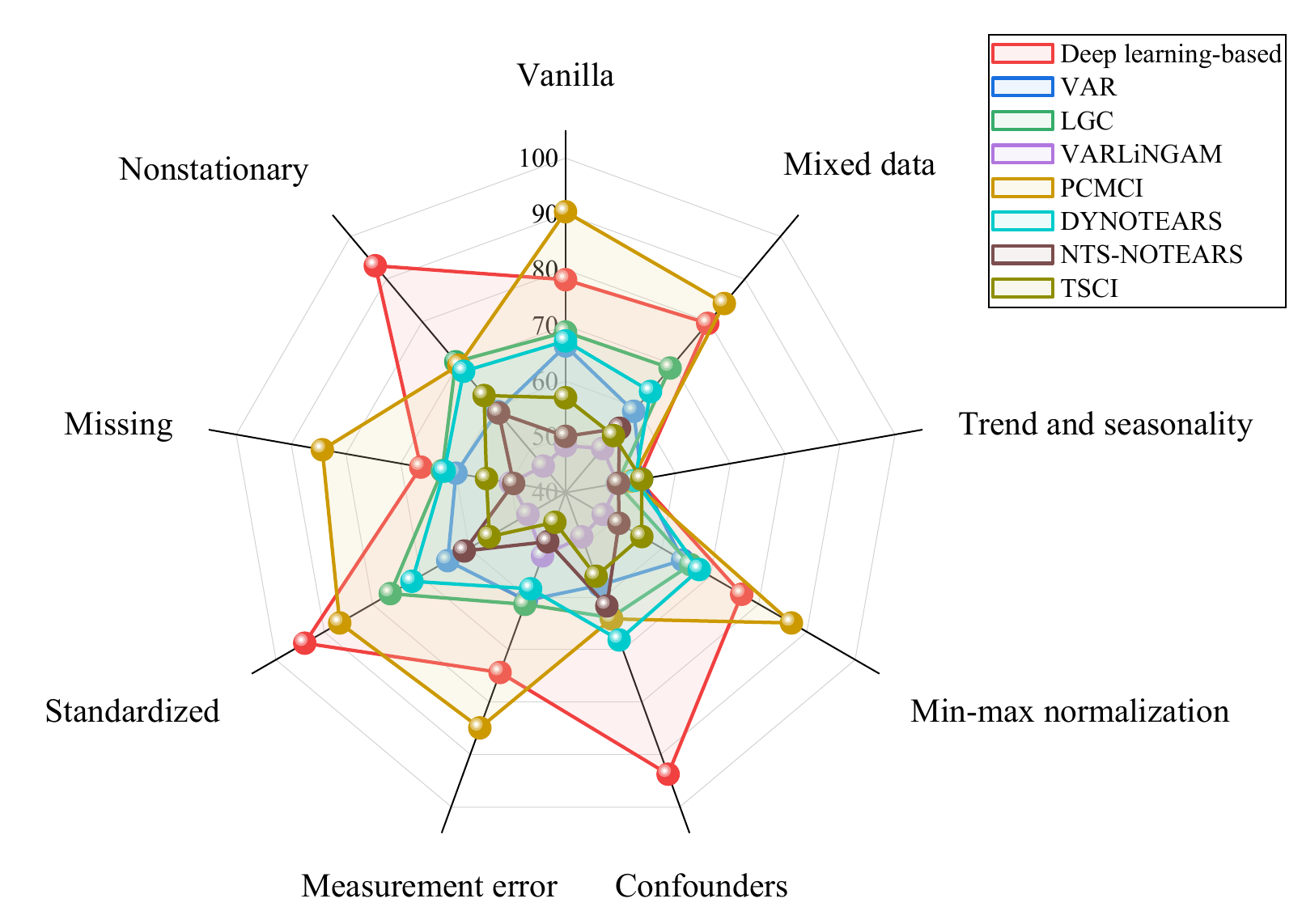}
         \caption{AUROC for linear 10-node case with $T = 1000$. Results aggregated over all hyperparameters.}
         \label{fig:linear_10_1000_auroc_avg_hyper}
     \end{subfigure}%
     \hfill
     \begin{subfigure}[b]{0.49\textwidth}
         \centering
         \includegraphics[width=\textwidth]{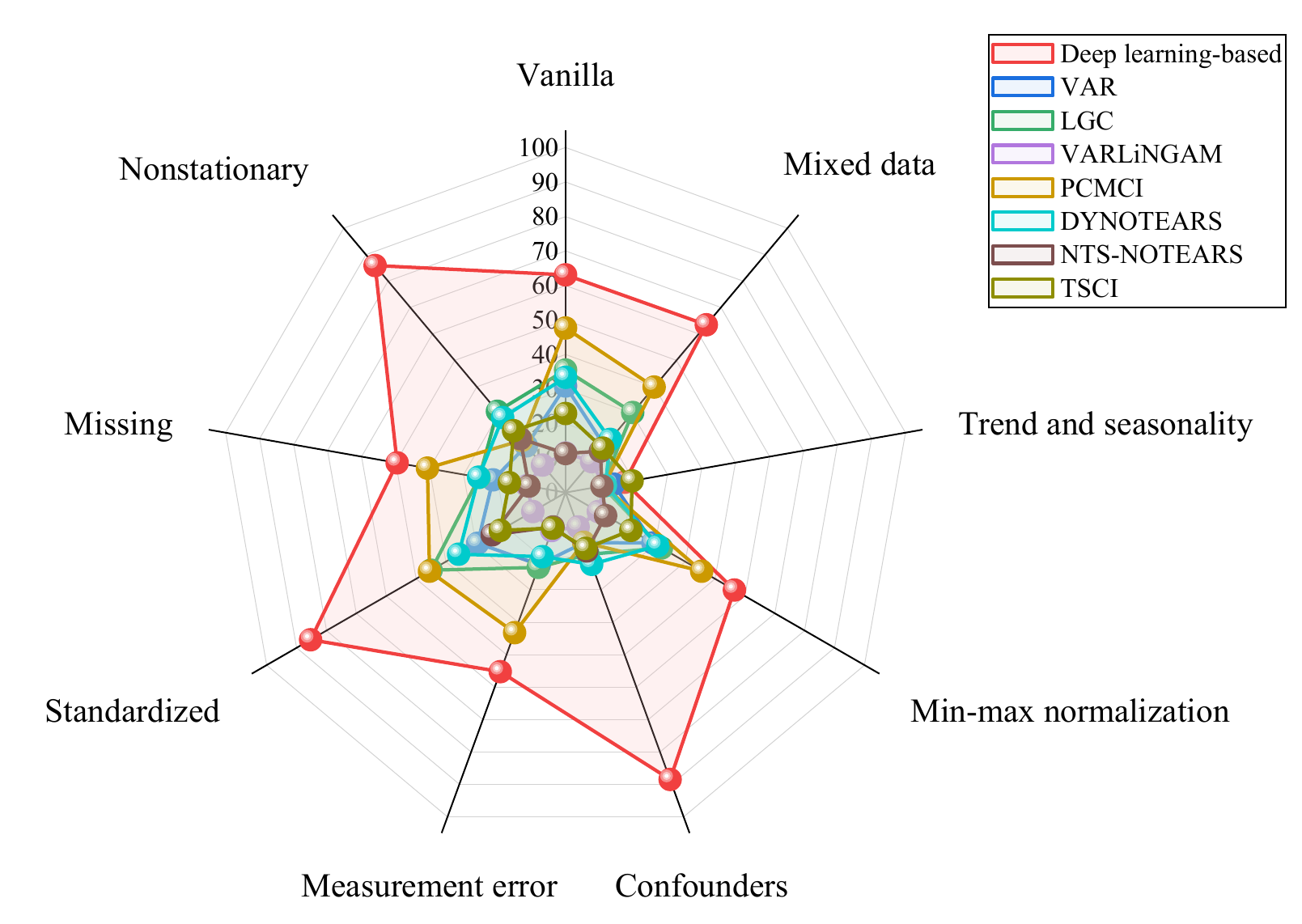}
         \caption{AUPRC for linear 10-node case with $T = 1000$. Results aggregated over all hyperparameters.}
         \label{fig:linear_10_1000_auprc_avg_hyper}
     \end{subfigure}

     \medskip  

     \begin{subfigure}[b]{0.49\textwidth}
         \centering
        \includegraphics[width=\textwidth]{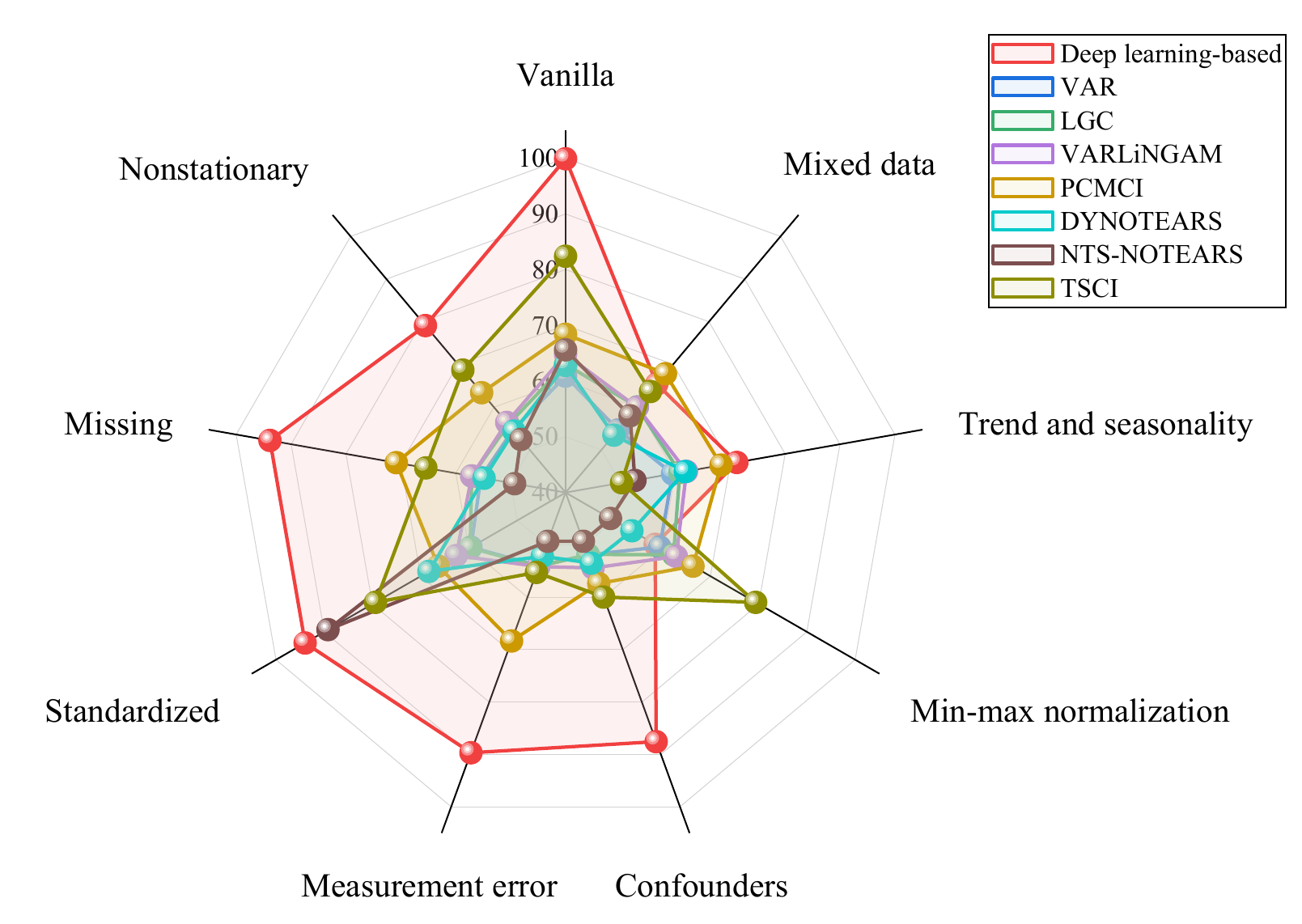}
         \caption{AUROC for nonlinear 10-node case with $T = 1000$ and $F=10$. Results aggregated over all hyperparameters.}
         \label{fig:nonlinear_10_1000_f10_auroc_avg_hyper}
     \end{subfigure}%
     \hfill
     \begin{subfigure}[b]{0.49\textwidth}
         \centering
         \includegraphics[width=\textwidth]{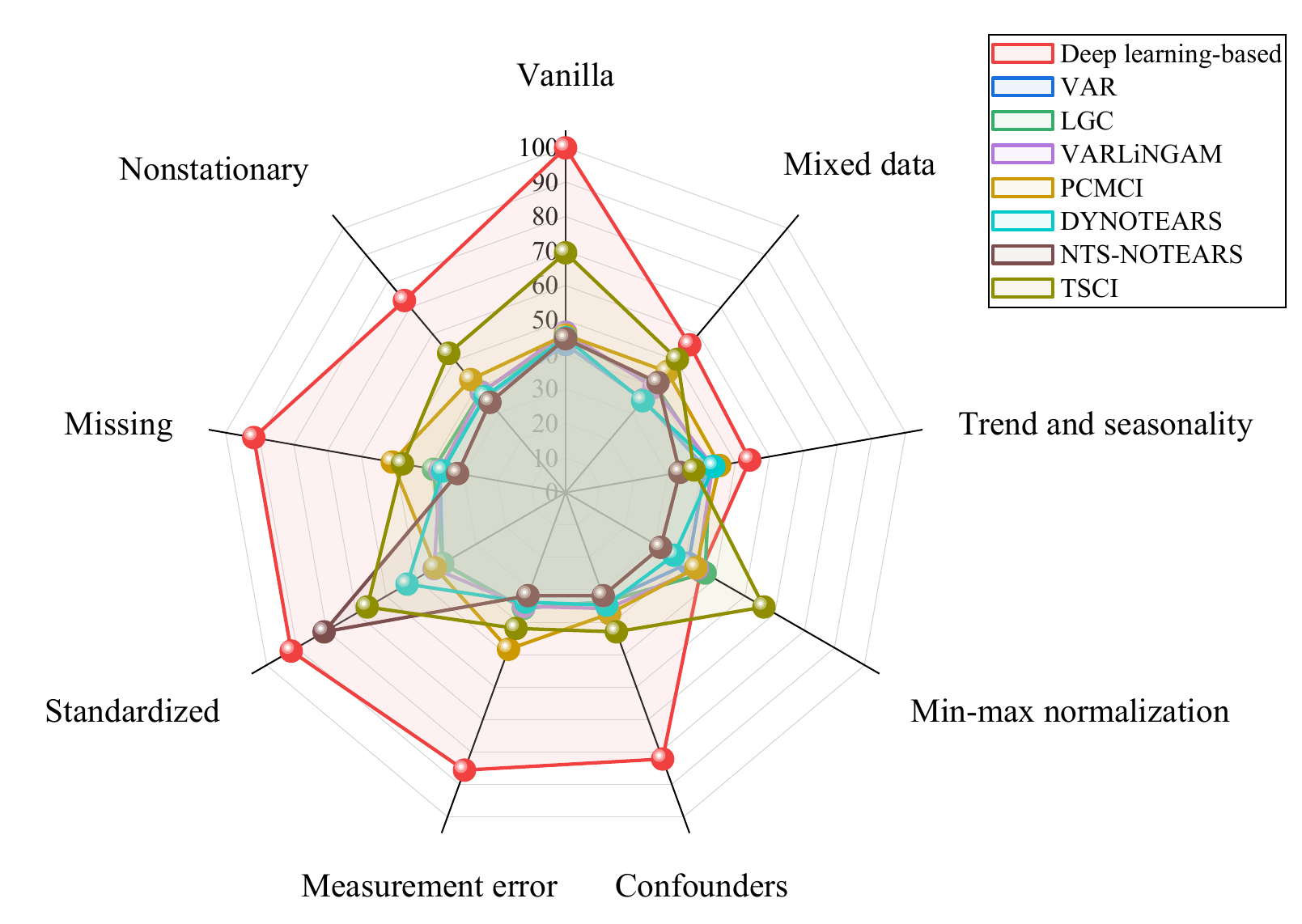}
         \caption{AUPRC for nonlinear 10-node case with $T = 1000$ and $F=10$. Results aggregated over all hyperparameters.}
         \label{fig:nonlinear_10_1000_f10_auprc_avg_hyper}
     \end{subfigure}

\Description{Four radar charts comparing causal discovery methods on 10-node networks with T=1000 across 9 scenarios. Superior performance is predominantly achieved by deep learning-based approaches. Results obtained by aggregating over all hyperparameter configurations.}
\caption{Experimental results under the linear and nonlinear settings across the vanilla scenario and eight assumption violation scenarios. AUROC and AUPRC (the higher the better) are evaluated over 5 trials for the 10-node case with $T = 1000$. For the deep learning-based methods, we present only the optimal results. Results aggregated over all hyperparameters.}
\label{fig:experiments_10_1000_f10_avg_hyper}
\end{figure*}

\begin{figure*}[t]
     \centering
     \begin{subfigure}[b]{0.49\textwidth}
         \centering
        \includegraphics[width=\textwidth]{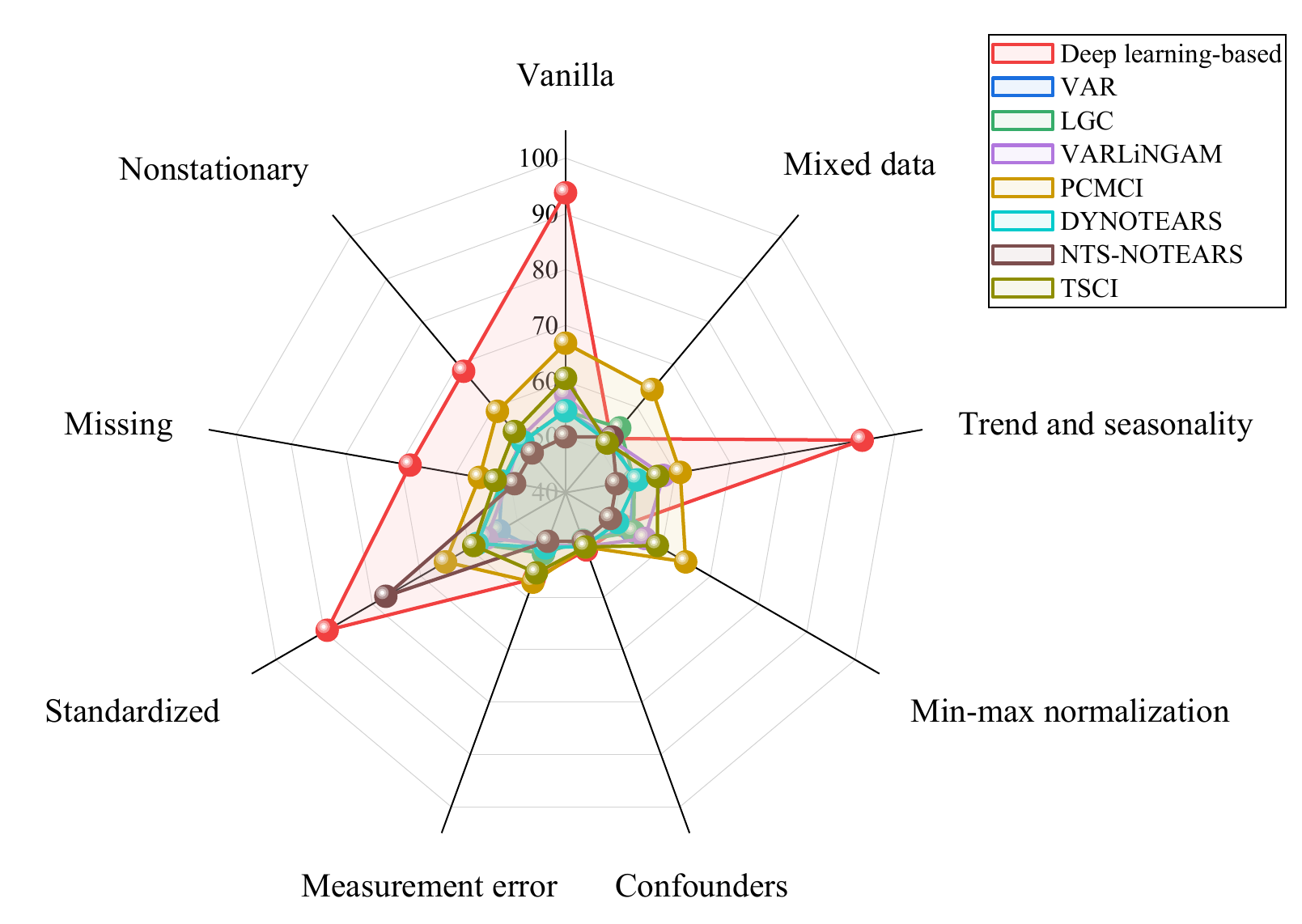}
         \caption{AUROC for nonlinear 10-node case with $T = 500$ and $F=40$. Results aggregated over all hyperparameters.}
         \label{fig:nonlinear_10_500_f40_auroc_avg_hyper}
     \end{subfigure}%
     \hfill
     \begin{subfigure}[b]{0.49\textwidth}
         \centering
         \includegraphics[width=\textwidth]{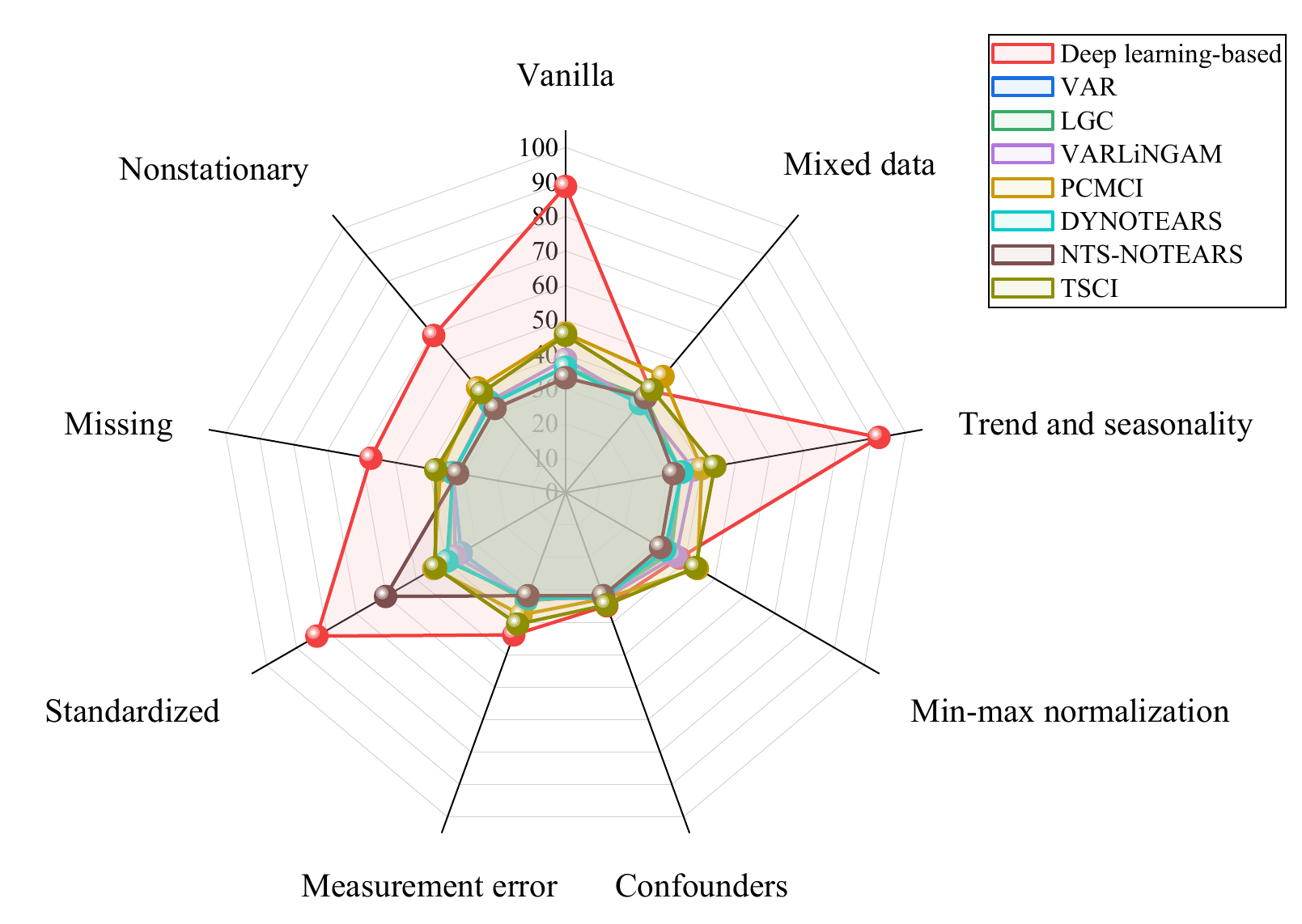}
         \caption{AUPRC for nonlinear 10-node case with $T = 500$ and $F=40$. Results aggregated over all hyperparameters.}
         \label{fig:nonlinear_10_500_f40_auprc_avg_hyper}
     \end{subfigure}

     \medskip  

     \begin{subfigure}[b]{0.49\textwidth}
         \centering
        \includegraphics[width=\textwidth]{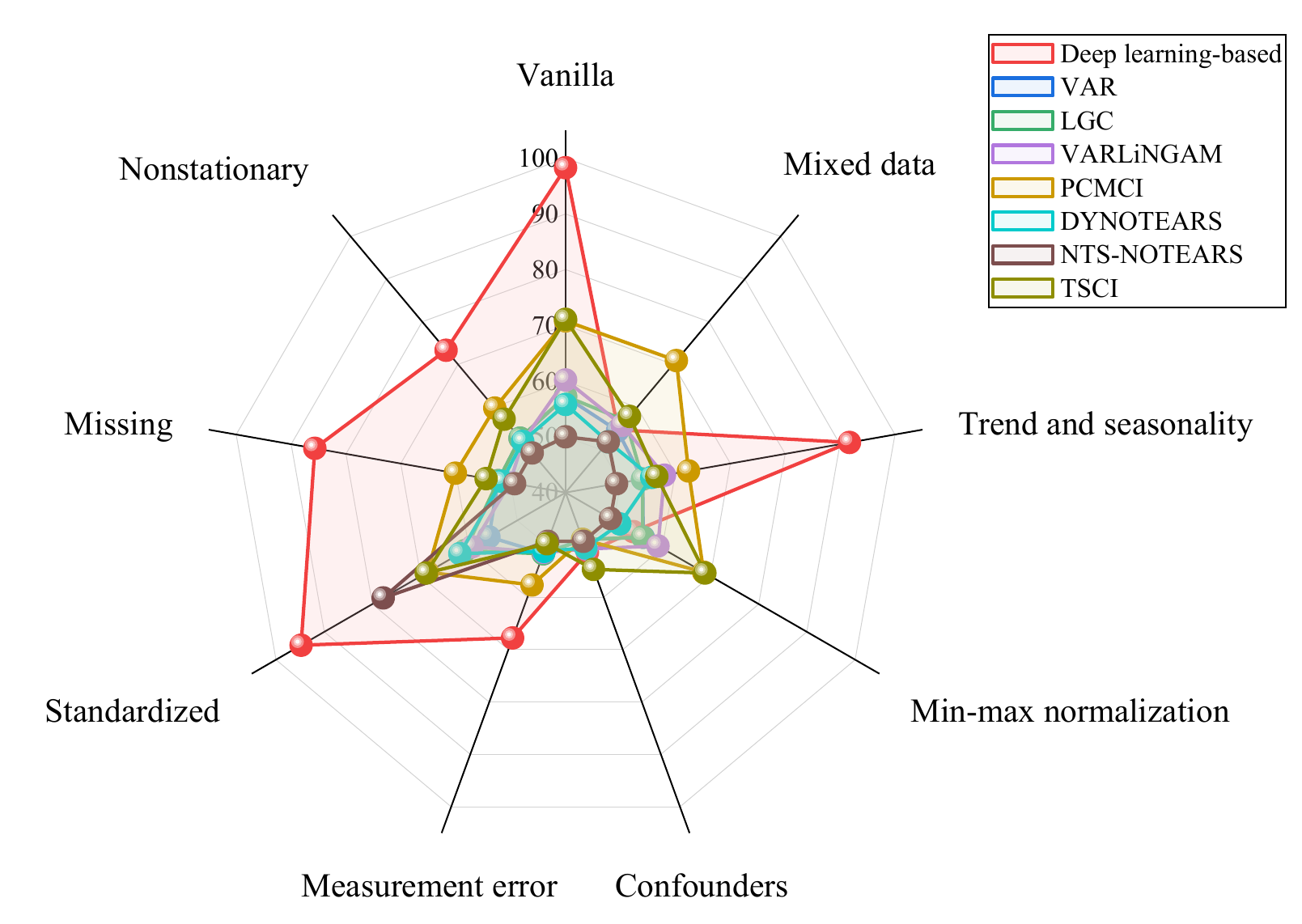}
         \caption{AUROC for nonlinear 10-node case with $T = 1000$ and $F=40$. Results aggregated over all hyperparameters.}
         \label{fig:nonlinear_10_1000_f40_auroc_avg_hyper}
     \end{subfigure}%
     \hfill
     \begin{subfigure}[b]{0.49\textwidth}
         \centering
         \includegraphics[width=\textwidth]{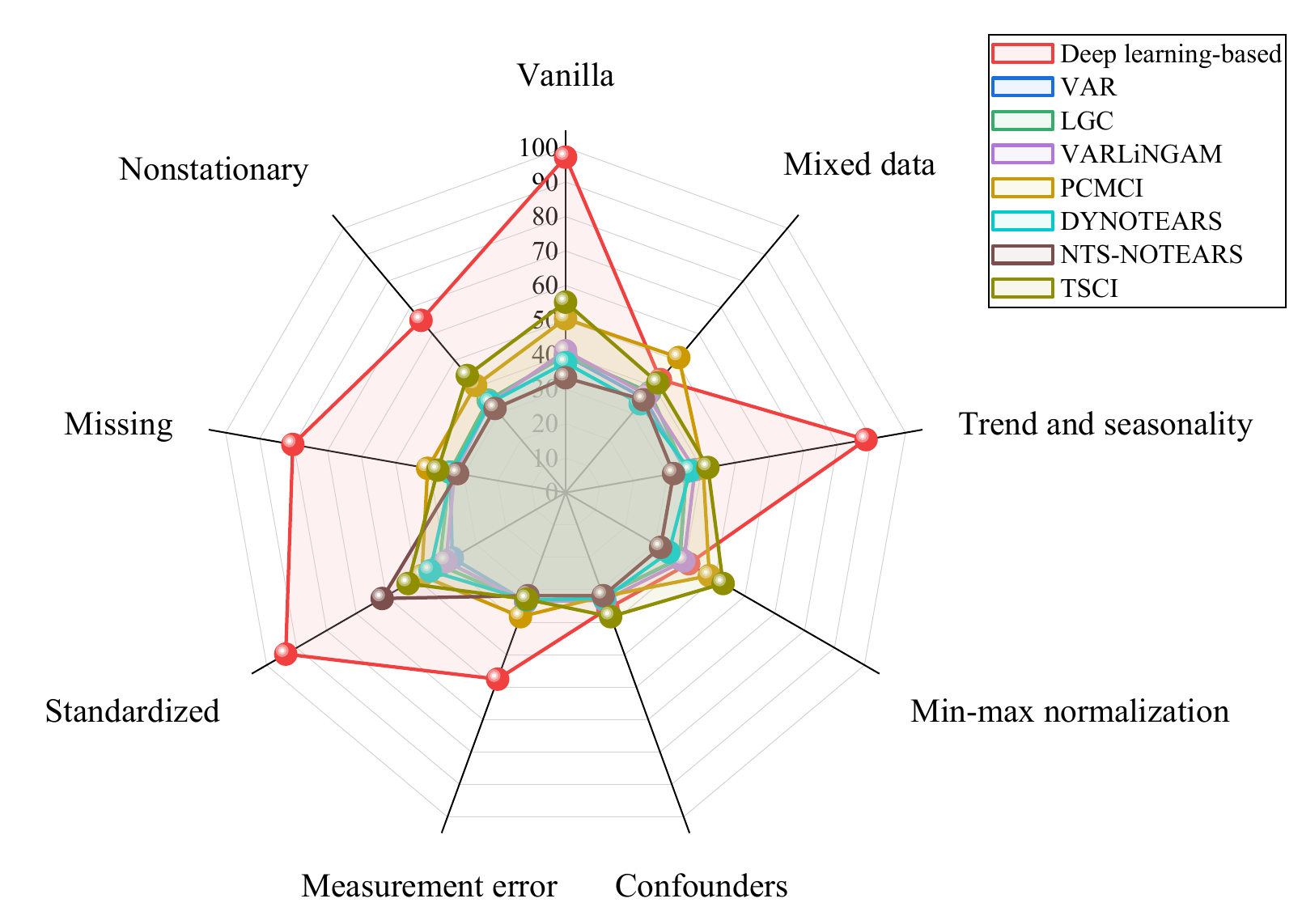}
         \caption{AUPRC for nonlinear 10-node case with $T = 1000$ and $F=40$. Results aggregated over all hyperparameters.}
         \label{fig:nonlinear_10_1000_f40_auprc_avg_hyper}
     \end{subfigure}
     
\Description{Four radar charts comparing causal discovery methods on 10-node networks with F=40 across 9 scenarios. Superior performance is predominantly achieved by deep learning-based approaches. Results obtained by aggregating over all hyperparameter configurations.}   
\caption{Experimental results under the nonlinear settings across the vanilla scenario and eight assumption violation scenarios. AUROC and AUPRC (the higher the better) are evaluated over 5 trials for the 10-node case with $F = 40$. For the deep learning-based methods, we present only the optimal results. Results aggregated over all hyperparameters.}
\label{fig:experiments_10_500_1000_f40_avg_hyper}
\end{figure*}

\begin{figure*}[t]
     \centering
     \begin{subfigure}[b]{0.49\textwidth}
         \centering
        \includegraphics[width=\textwidth]{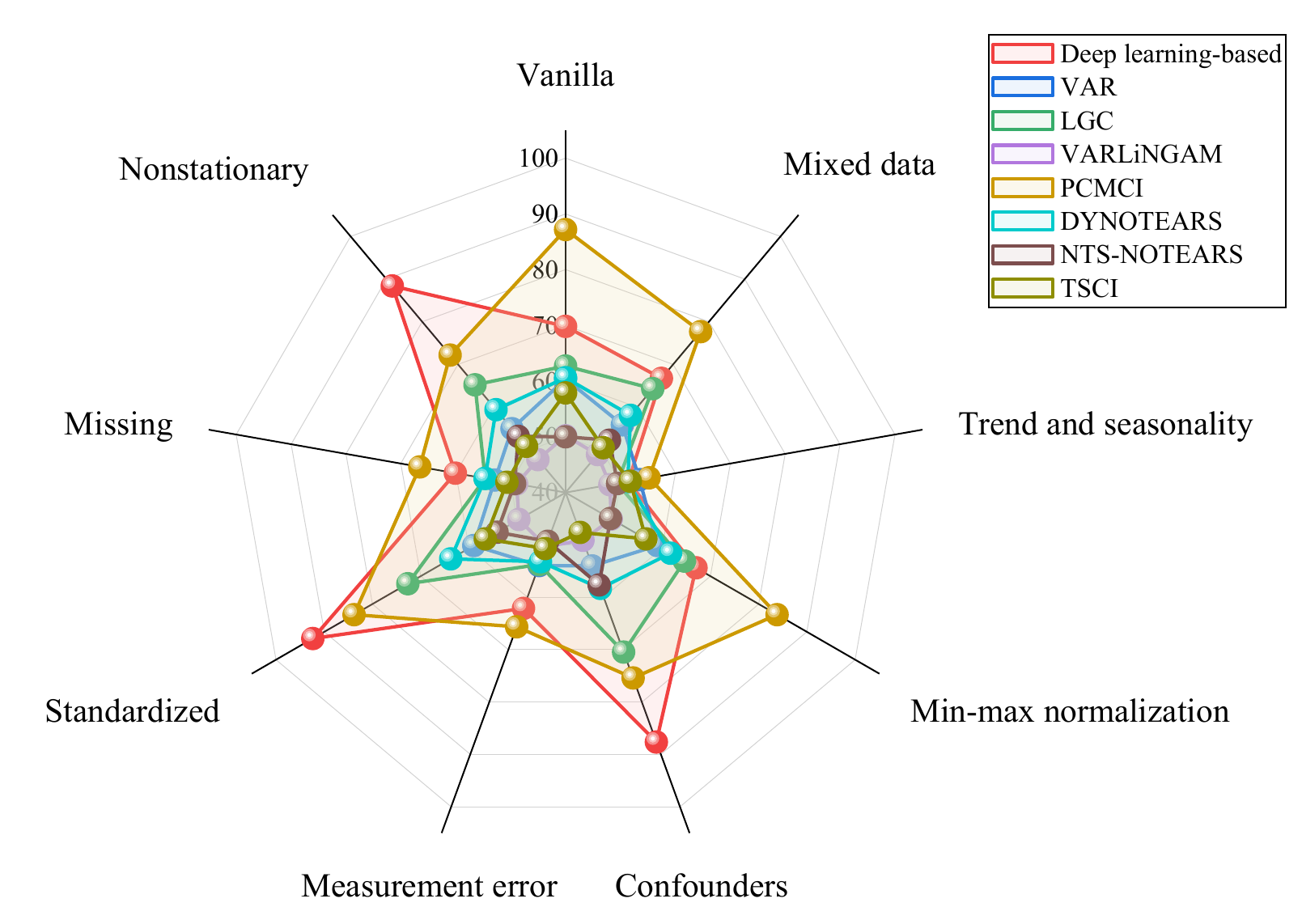}
         \caption{AUROC for linear 15-node case with $T = 500$. Results aggregated over all hyperparameters.}
         \label{fig:linear_15_500_auroc_avg_hyper}
     \end{subfigure}%
     \hfill
     \begin{subfigure}[b]{0.49\textwidth}
         \centering
         \includegraphics[width=\textwidth]{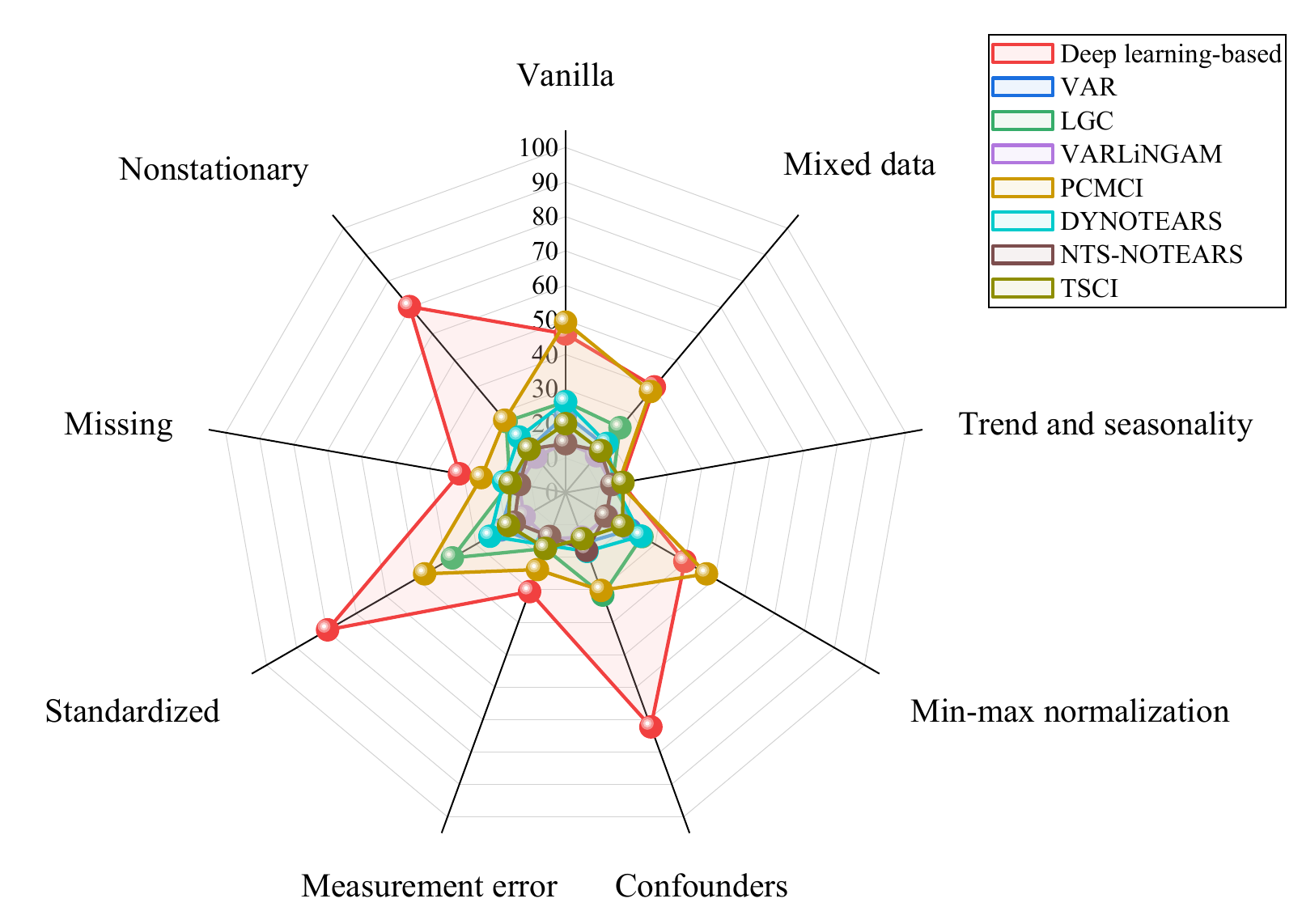}
         \caption{AUPRC for linear 15-node case with $T = 500$. Results aggregated over all hyperparameters.}
         \label{fig:linear_15_500_auprc_avg_hyper}
     \end{subfigure}

     \medskip  

     \begin{subfigure}[b]{0.49\textwidth}
         \centering
        \includegraphics[width=\textwidth]{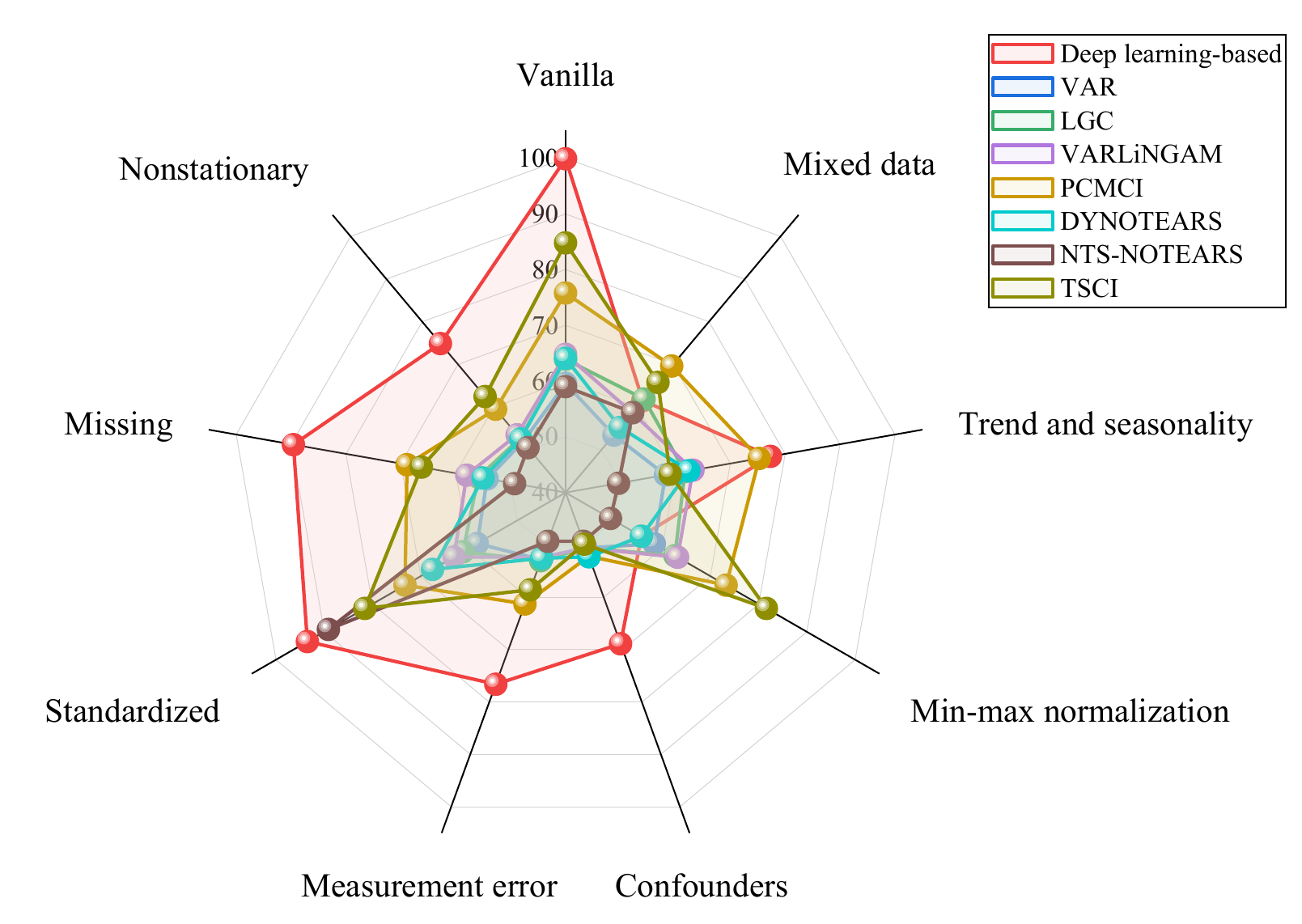}
         \caption{AUROC for nonlinear 15-node case with $T = 500$ and $F=10$. Results aggregated over all hyperparameters.}
         \label{fig:nonlinear_15_500_f10_auroc_avg_hyper}
     \end{subfigure}%
     \hfill
     \begin{subfigure}[b]{0.49\textwidth}
         \centering
         \includegraphics[width=\textwidth]{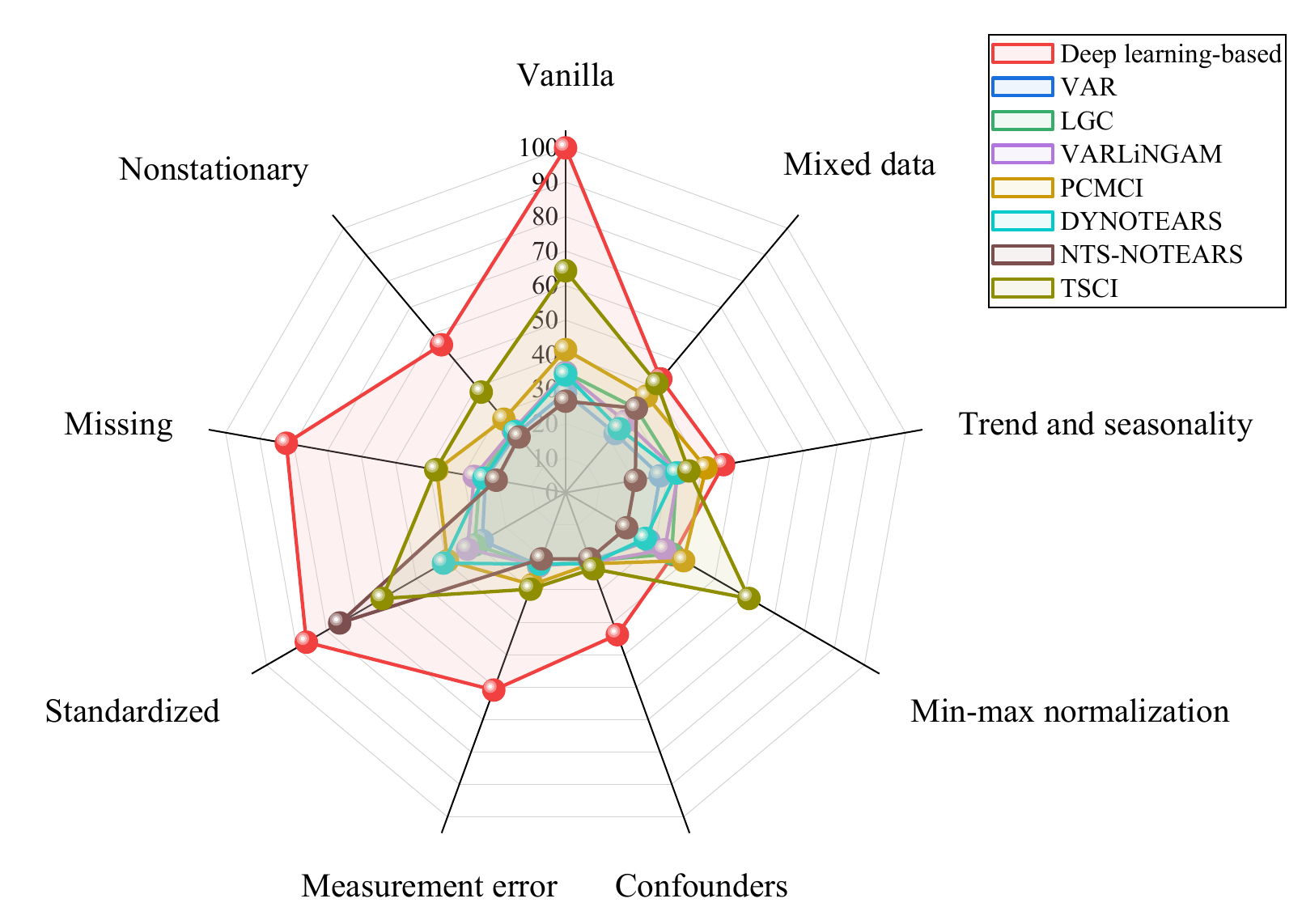}
         \caption{AUPRC for nonlinear 15-node case with $T = 500$ and $F=10$. Results aggregated over all hyperparameters.}
         \label{fig:nonlinear_15_500_f10_auprc_avg_hyper}
     \end{subfigure}
\Description{Four radar charts comparing causal discovery methods on 15-node networks with T=500 across 9 scenarios. Superior performance is predominantly achieved by deep learning-based approaches. Results obtained by aggregating over all hyperparameter configurations.}
\caption{Experimental results under the linear and nonlinear settings across the vanilla scenario and eight assumption violation scenarios. AUROC and AUPRC (the higher the better) are evaluated over 5 trials for the 15-node case with $T = 500$. For the deep learning-based methods, we present only the optimal results. Results aggregated over all hyperparameters.}
\label{fig:experiments_15_500_f10_avg_hyper}
\end{figure*}

\begin{figure*}[t]
     \centering
     \begin{subfigure}[b]{0.49\textwidth}
         \centering
        \includegraphics[width=\textwidth]{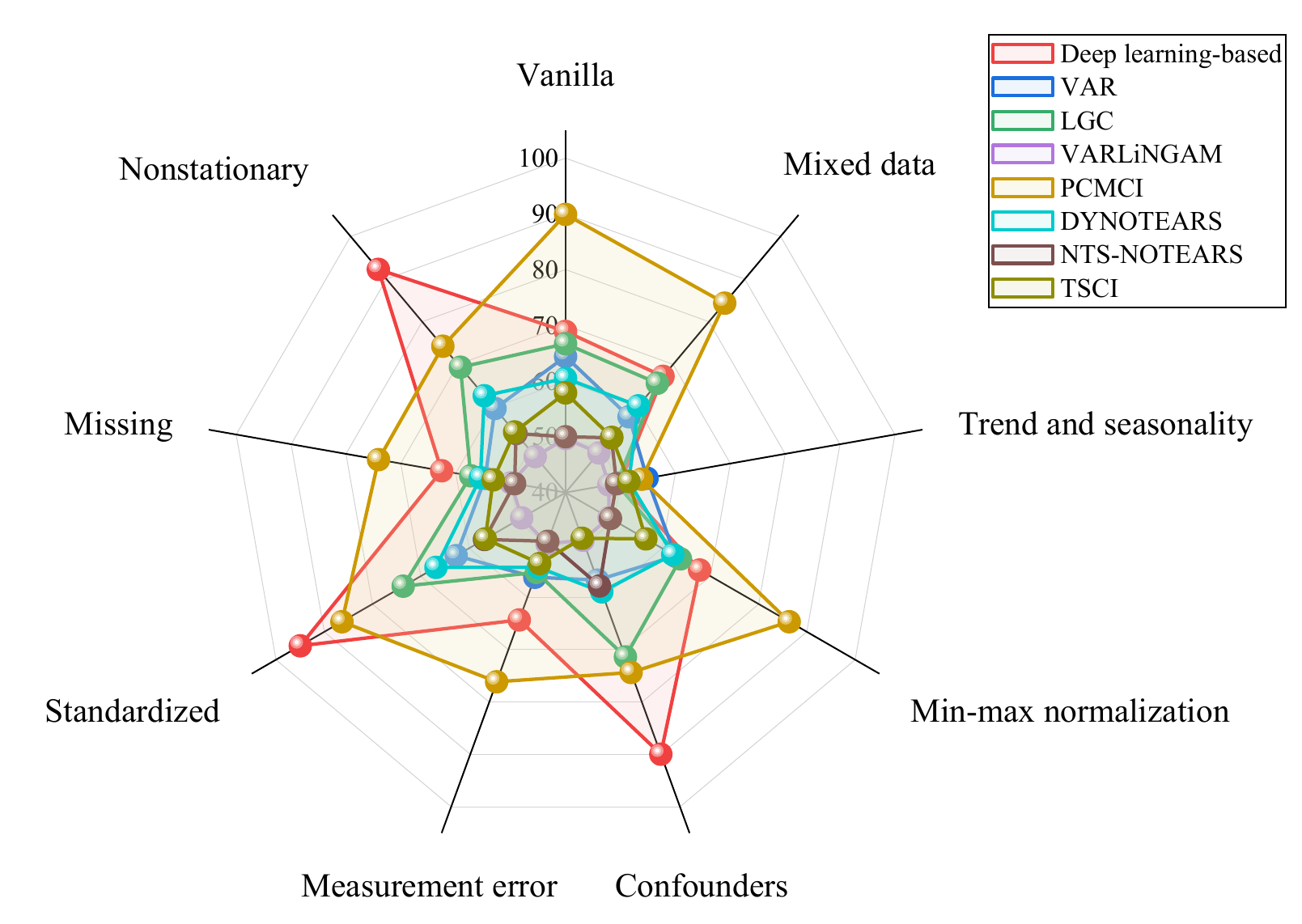}
         \caption{AUROC for linear 15-node case with $T = 1000$. Results aggregated over all hyperparameters.}
         \label{fig:linear_15_1000_auroc_avg_hyper}
     \end{subfigure}%
     \hfill
     \begin{subfigure}[b]{0.49\textwidth}
         \centering
         \includegraphics[width=\textwidth]{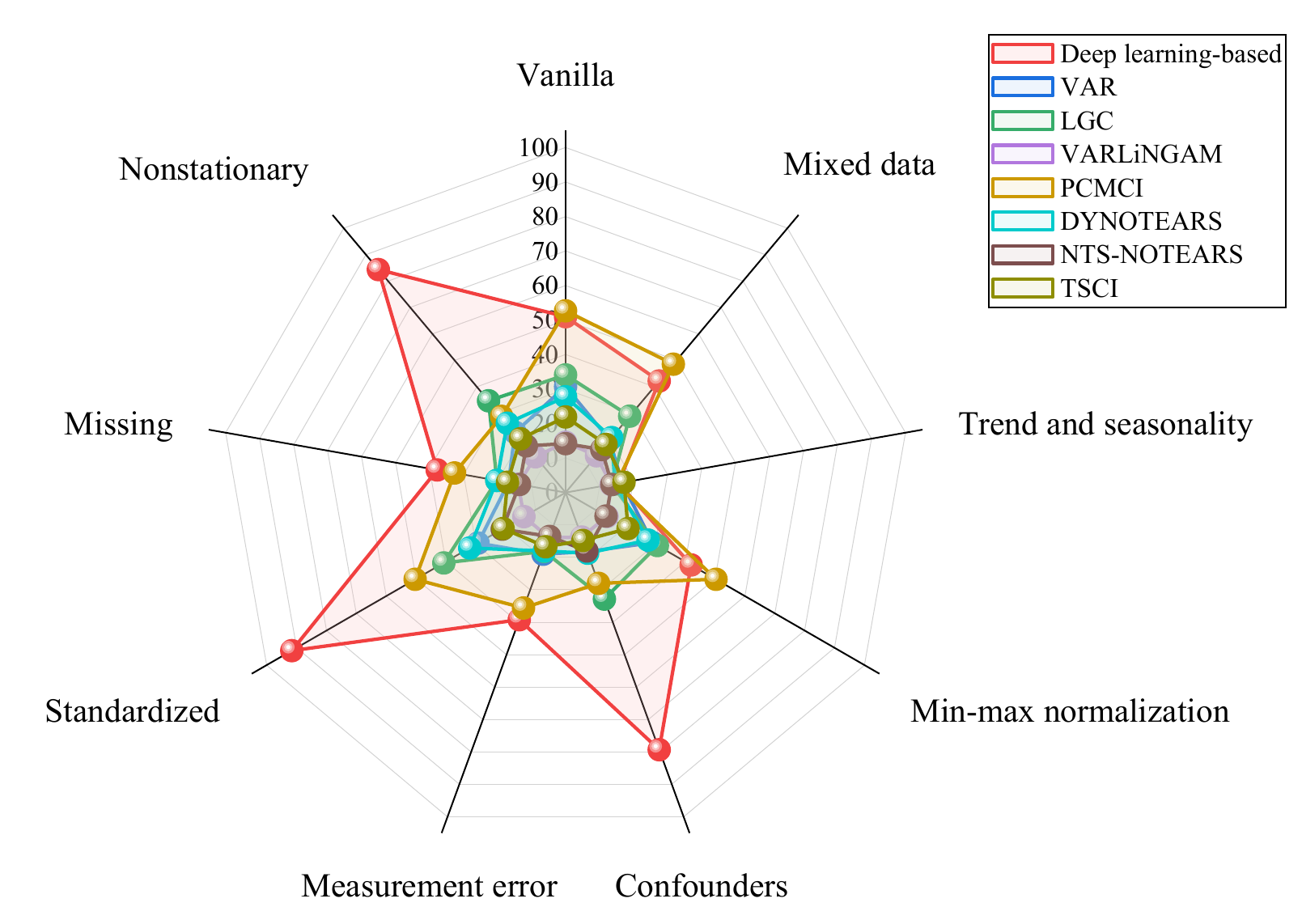}
         \caption{AUPRC for linear 15-node case with $T = 1000$. Results aggregated over all hyperparameters.}
         \label{fig:linear_15_1000_auprc_avg_hyper}
     \end{subfigure}

     \medskip  

     \begin{subfigure}[b]{0.49\textwidth}
         \centering
        \includegraphics[width=\textwidth]{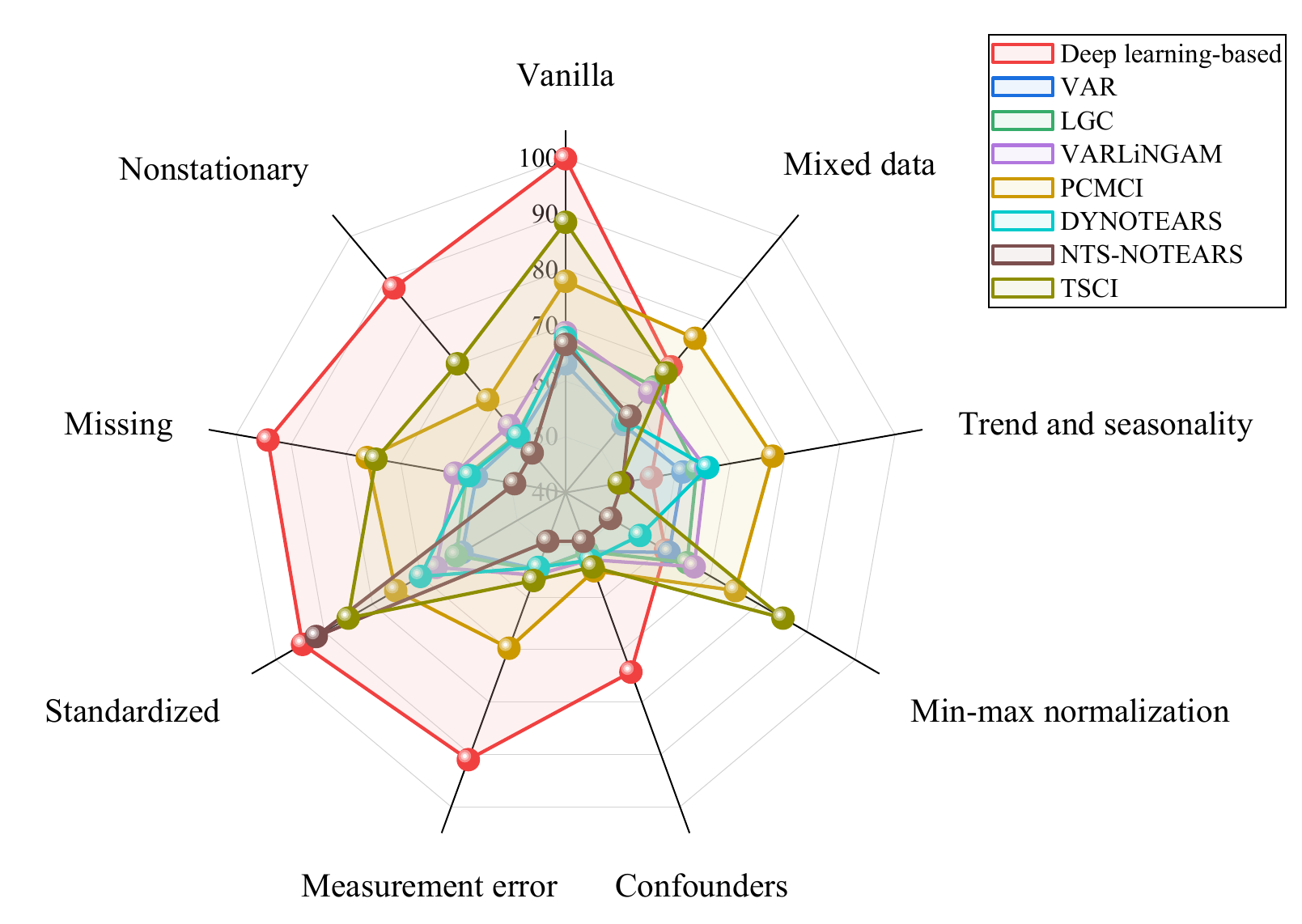}
         \caption{AUROC for nonlinear 15-node case with $T = 1000$ and $F=10$. Results aggregated over all hyperparameters.}
         \label{fig:nonlinear_15_1000_f10_auroc_avg_hyper}
     \end{subfigure}%
     \hfill
     \begin{subfigure}[b]{0.49\textwidth}
         \centering
         \includegraphics[width=\textwidth]{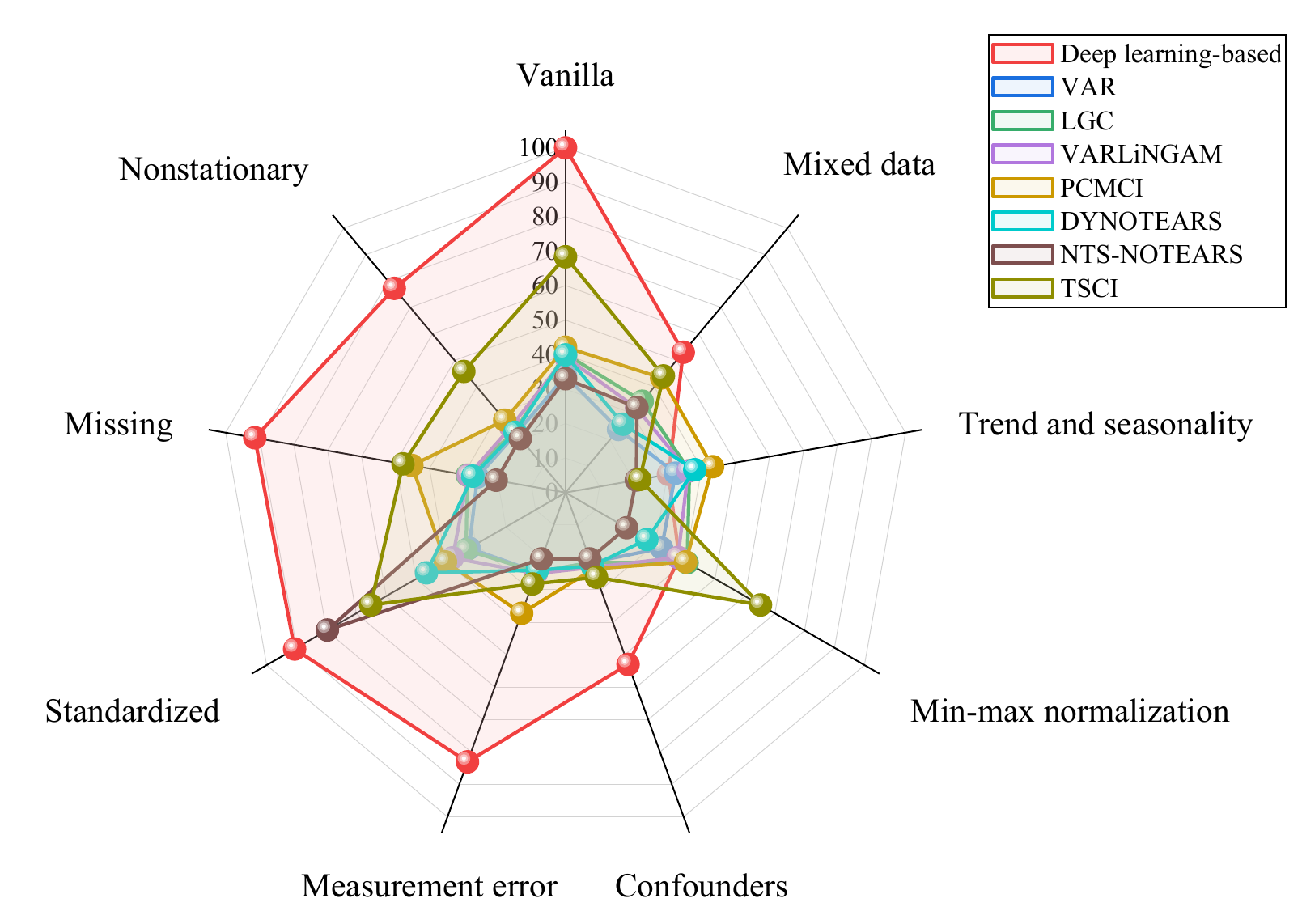}
         \caption{AUPRC for nonlinear 15-node case with $T = 1000$ and $F=10$. Results aggregated over all hyperparameters.}
         \label{fig:nonlinear_15_1000_f10_auprc_avg_hyper}
     \end{subfigure}

\Description{Four radar charts comparing causal discovery methods on 15-node networks with T=1000 across 9 scenarios. Superior performance is predominantly achieved by deep learning-based approaches. Results obtained by aggregating over all hyperparameter configurations.}
\caption{Experimental results under the linear and nonlinear settings across the vanilla scenario and eight assumption violation scenarios. AUROC and AUPRC (the higher the better) are evaluated over 5 trials for the 15-node case with $T = 1000$. For the deep learning-based methods, we present only the optimal results. Results aggregated over all hyperparameters.}
\label{fig:experiments_15_1000_f10_avg_hyper}
\end{figure*}

\begin{figure*}[t]
     \centering
     \begin{subfigure}[b]{0.49\textwidth}
         \centering
        \includegraphics[width=\textwidth]{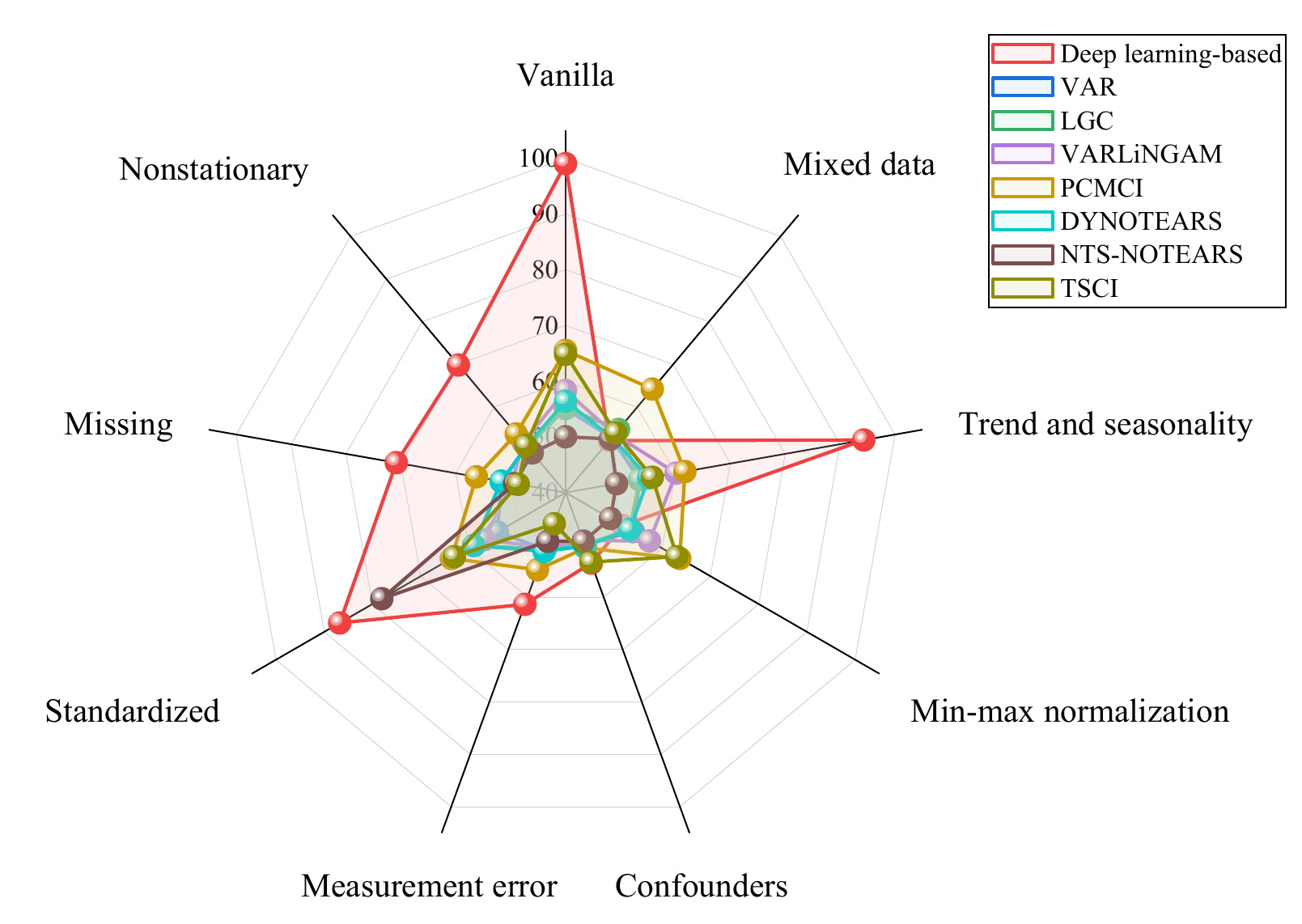}
         \caption{AUROC for nonlinear 15-node case with $T = 500$ and $F=40$. Results aggregated over all hyperparameters.}
         \label{fig:nonlinear_15_500_f40_auroc_avg_hyper}
     \end{subfigure}%
     \hfill
     \begin{subfigure}[b]{0.49\textwidth}
         \centering
         \includegraphics[width=\textwidth]{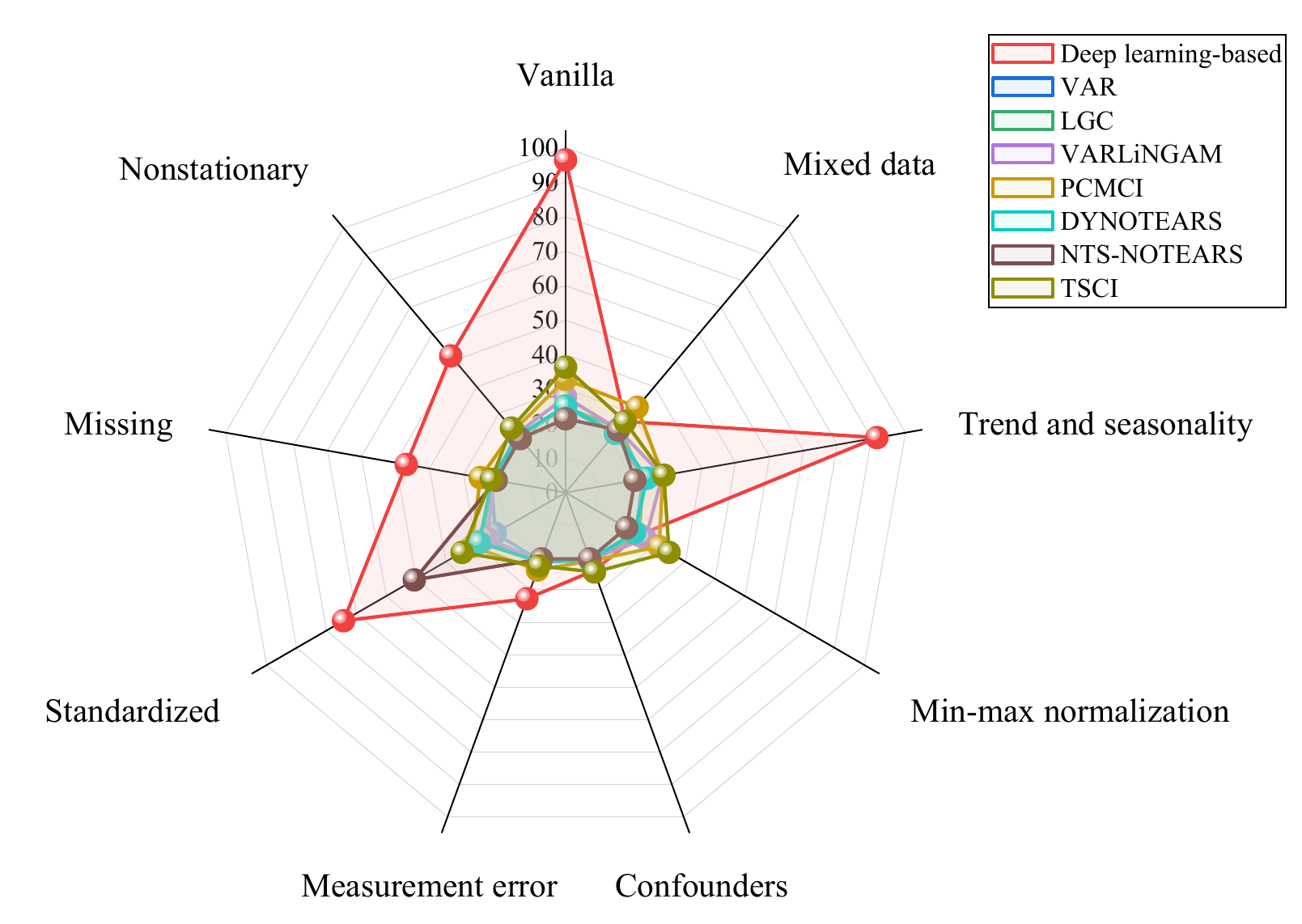}
         \caption{AUPRC for nonlinear 15-node case with $T = 500$ and $F=40$. Results aggregated over all hyperparameters.}
         \label{fig:nonlinear_15_500_f40_auprc_avg_hyper}
     \end{subfigure}

     \medskip  

     \begin{subfigure}[b]{0.49\textwidth}
         \centering
        \includegraphics[width=\textwidth]{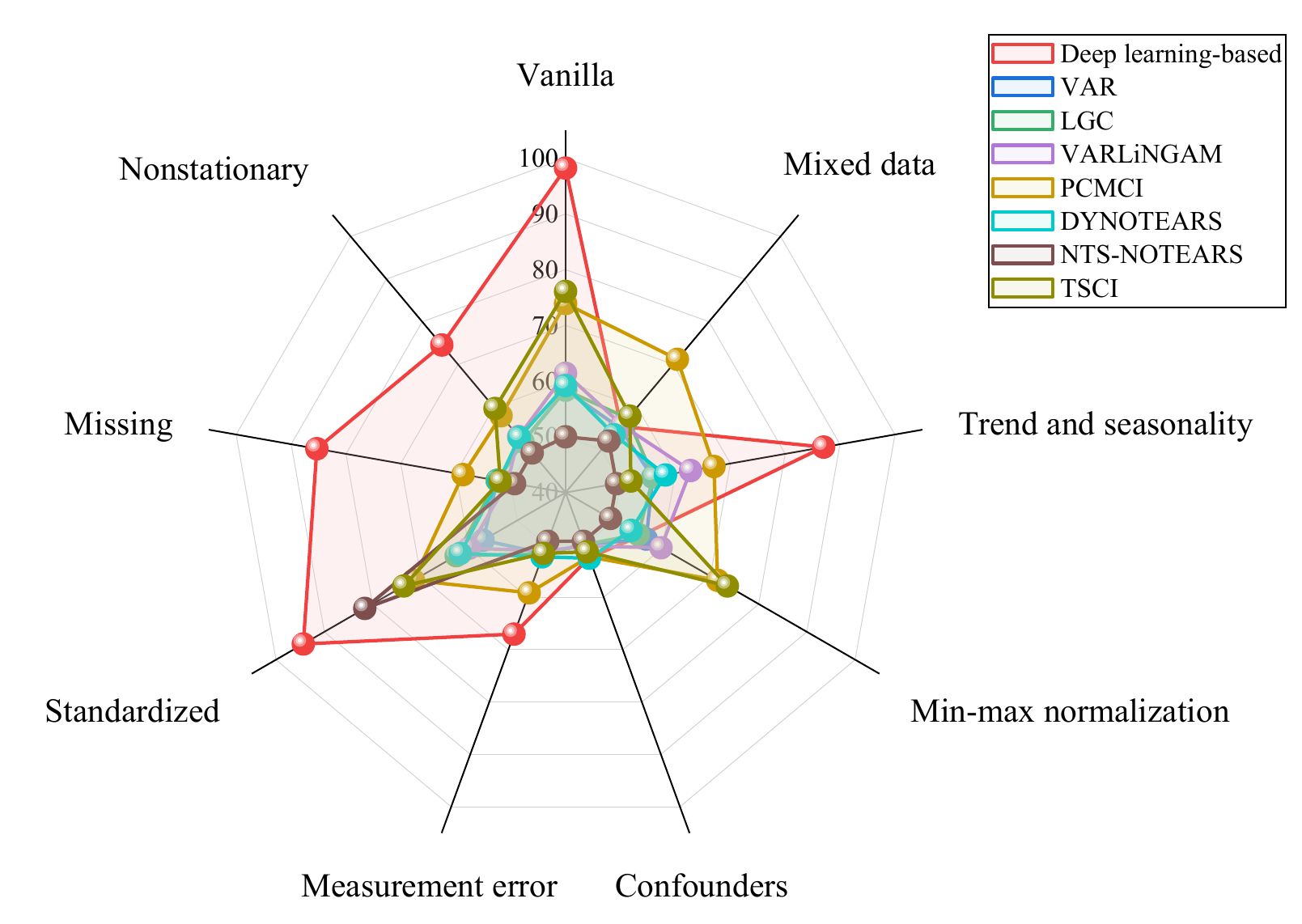}
         \caption{AUROC for nonlinear 15-node case with $T = 1000$ and $F=40$. Results aggregated over all hyperparameters.}
         \label{fig:nonlinear_15_1000_f40_auroc_avg_hyper}
     \end{subfigure}%
     \hfill
     \begin{subfigure}[b]{0.49\textwidth}
         \centering
         \includegraphics[width=\textwidth]{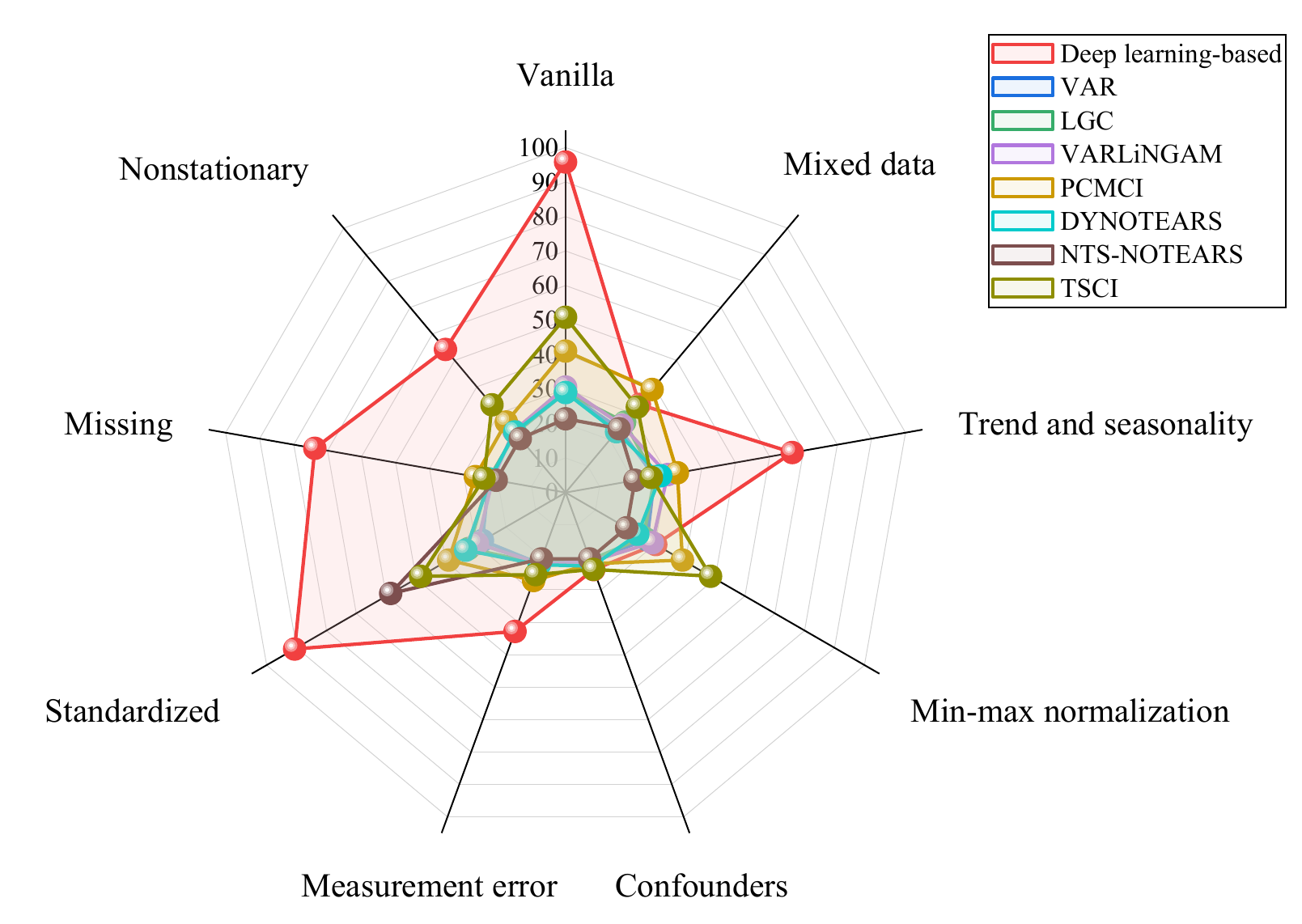}
         \caption{AUPRC for nonlinear 15-node case with $T = 1000$ and $F=40$. Results aggregated over all hyperparameters.}
         \label{fig:nonlinear_15_1000_f40_auprc_avg_hyper}
     \end{subfigure}
     
\Description{Four radar charts comparing causal discovery methods on 15-node networks with F=40 across 9 scenarios. Superior performance is predominantly achieved by deep learning-based approaches. Results obtained by aggregating over all hyperparameter configurations.}   
\caption{Experimental results under the nonlinear settings across the vanilla scenario and eight assumption violation scenarios. AUROC and AUPRC (the higher the better) are evaluated over 5 trials for the 15-node case with $F = 40$. For the deep learning-based methods, we present only the optimal results. Results aggregated over all hyperparameters.}
\label{fig:experiments_15_500_1000_f40_avg_hyper}
\end{figure*}

For each scenario, we compute the mean and standard deviation of the evaluation metrics for each method across different hyperparameter settings. Additional figure results under both linear and nonlinear settings, spanning different numbers of nodes, time series lengths, and external forcing intensities, are reported in the Appendix (Figures~\ref{fig:experiments_10_500_f10_avg_hyper}--\ref{fig:experiments_15_500_1000_f40_avg_hyper}).

\begin{table*}[!htbp]
	\centering
	\caption{Linear setting, 10-node case with $T = 500$ (Part I).}
	\setlength{\tabcolsep}{8.3pt}
\begin{tabular}{ccccccccc}
		\toprule
		& \multicolumn{2}{c}{Vanilla} & \multicolumn{2}{c}{Mixed data} & \multicolumn{2}{c}{Trend and seasonality} & \multicolumn{2}{c}{Min--max normalization} \\
		\cmidrule(lr){2-3} \cmidrule(lr){4-5} \cmidrule(lr){6-7} \cmidrule(lr){8-9}
		10 nodes & AUROC$\uparrow$ & AUPRC$\uparrow$ & AUROC$\uparrow$ & AUPRC$\uparrow$ & AUROC$\uparrow$ & AUPRC$\uparrow$ & AUROC$\uparrow$ & AUPRC$\uparrow$ \\
		\midrule
		VAR & 95.2\footnotesize{$\pm$1.0} & 57.3\footnotesize{$\pm$5.0} & 59.8\footnotesize{$\pm$4.1} & 22.7\footnotesize{$\pm$9.7} & 59.5\footnotesize{$\pm$3.3} & \textbf{24.5\footnotesize{$\pm$8.0}} & 95.2\footnotesize{$\pm$0.7} & 57.0\footnotesize{$\pm$3.8} \\
		LGC & 97.6\footnotesize{$\pm$1.1} & 73.6\footnotesize{$\pm$9.1} & 82.1\footnotesize{$\pm$8.1} & 51.0\footnotesize{$\pm$10.3} & 50.5\footnotesize{$\pm$3.5} & 12.6\footnotesize{$\pm$1.9} & 97.1\footnotesize{$\pm$1.0} & 69.3\footnotesize{$\pm$7.4} \\
		VARLiNGAM & 53.9\footnotesize{$\pm$6.1} & 12.7\footnotesize{$\pm$2.0} & 51.3\footnotesize{$\pm$4.1} & 12.9\footnotesize{$\pm$3.0} & 50.5\footnotesize{$\pm$2.9} & 12.3\footnotesize{$\pm$1.5} & 51.2\footnotesize{$\pm$8.5} & 12.9\footnotesize{$\pm$1.9} \\
		PCMCI & 97.8\footnotesize{$\pm$0.8} & 75.3\footnotesize{$\pm$6.8} & \textbf{91.1\footnotesize{$\pm$4.0}} & 67.3\footnotesize{$\pm$12.0} & 63.2\footnotesize{$\pm$2.3} & 14.6\footnotesize{$\pm$0.8} & 97.8\footnotesize{$\pm$0.8} & 75.3\footnotesize{$\pm$6.8} \\
		DYNOTEARS & \textbf{100.0\footnotesize{$\pm$0.0}} & \textbf{100.0\footnotesize{$\pm$0.0}} & 78.4\footnotesize{$\pm$7.2} & 48.0\footnotesize{$\pm$14.2} & 60.1\footnotesize{$\pm$8.8} & 15.2\footnotesize{$\pm$3.9} & \textbf{100.0\footnotesize{$\pm$0.0}} & \textbf{100.0\footnotesize{$\pm$0.0}} \\
\bottomrule
	\end{tabular}
    \label{tab:linear-10-500-1}

\end{table*}

\begin{table*}[!htbp]
	\centering
	\caption{Linear setting, 10-node case with $T = 500$ (Part II).}
	\setlength{\tabcolsep}{4pt}
	\begin{tabular}{ccccccccccc}
		\toprule
		& \multicolumn{2}{c}{Latent confounders} & \multicolumn{2}{c}{Measurement error} & \multicolumn{2}{c}{Standardized} & \multicolumn{2}{c}{Missing} & \multicolumn{2}{c}{Nonstationary} \\
		\cmidrule(lr){2-3} \cmidrule(lr){4-5} \cmidrule(lr){6-7} \cmidrule(lr){8-9} \cmidrule(lr){10-11}
		10 nodes & AUROC$\uparrow$ & AUPRC$\uparrow$ & AUROC$\uparrow$ & AUPRC$\uparrow$ & AUROC$\uparrow$ & AUPRC$\uparrow$ & AUROC$\uparrow$ & AUPRC$\uparrow$ & AUROC$\uparrow$ & AUPRC$\uparrow$ \\
		\midrule
		VAR & 55.0\footnotesize{$\pm$3.1} & 20.0\footnotesize{$\pm$5.6} & 90.7\footnotesize{$\pm$2.3} & 46.3\footnotesize{$\pm$3.5} & 94.6\footnotesize{$\pm$1.1} & 54.3\footnotesize{$\pm$5.5} & 86.0\footnotesize{$\pm$6.2} & 42.9\footnotesize{$\pm$12.8} & 78.8\footnotesize{$\pm$8.2} & 25.4\footnotesize{$\pm$8.7} \\
		LGC & 91.8\footnotesize{$\pm$4.4} & 59.2\footnotesize{$\pm$18.2} & 85.8\footnotesize{$\pm$5.5} & 47.5\footnotesize{$\pm$9.9} & 99.8\footnotesize{$\pm$0.2} & 98.1\footnotesize{$\pm$3.6} & 85.7\footnotesize{$\pm$6.5} & 55.1\footnotesize{$\pm$14.3} & 90.4\footnotesize{$\pm$9.9} & 59.3\footnotesize{$\pm$31.2} \\
		VARLiNGAM & 55.0\footnotesize{$\pm$9.1} & 15.4\footnotesize{$\pm$6.5} & 55.6\footnotesize{$\pm$8.5} & 13.7\footnotesize{$\pm$2.6} & 53.9\footnotesize{$\pm$6.1} & 12.7\footnotesize{$\pm$2.0} & 54.2\footnotesize{$\pm$11.5} & 14.1\footnotesize{$\pm$2.8} & 50.0\footnotesize{$\pm$9.3} & 12.2\footnotesize{$\pm$1.9} \\
		PCMCI & 83.1\footnotesize{$\pm$4.5} & 28.2\footnotesize{$\pm$5.2} & 85.6\footnotesize{$\pm$8.1} & 54.8\footnotesize{$\pm$15.3} & 97.8\footnotesize{$\pm$0.8} & 75.3\footnotesize{$\pm$6.8} & 80.7\footnotesize{$\pm$8.3} & 47.6\footnotesize{$\pm$14.4} & 83.5\footnotesize{$\pm$8.1} & 35.4\footnotesize{$\pm$16.7} \\
\bottomrule
	\end{tabular}
    \label{tab:linear-10-500-2}

\end{table*}

\begin{table*}[!htbp]
	\centering
	\caption{Linear setting, 10-node case with $T = 1000$ (Part I).}
	\setlength{\tabcolsep}{8.3pt}
	\begin{tabular}{ccccccccc}
		\toprule
		& \multicolumn{2}{c}{Vanilla} & \multicolumn{2}{c}{Mixed data} & \multicolumn{2}{c}{Trend and seasonality} & \multicolumn{2}{c}{Min--max normalization} \\
		\cmidrule(lr){2-3} \cmidrule(lr){4-5} \cmidrule(lr){6-7} \cmidrule(lr){8-9}
		10 nodes & AUROC$\uparrow$ & AUPRC$\uparrow$ & AUROC$\uparrow$ & AUPRC$\uparrow$ & AUROC$\uparrow$ & AUPRC$\uparrow$ & AUROC$\uparrow$ & AUPRC$\uparrow$ \\
		\midrule
		VAR & 99.7$\pm$0.3 & 96.3$\pm$4.4 & 62.0$\pm$2.4 & 32.4$\pm$4.3 & 60.0$\pm$3.1 & \textbf{28.8$\pm$5.6} & 99.6$\pm$0.5 & 94.8$\pm$6.7 \\
		LGC & \textbf{100.0$\pm$0.0} & \textbf{100.0$\pm$0.0} & 81.2$\pm$5.5 & 58.4$\pm$6.5 & 51.0$\pm$2.0 & 12.8$\pm$3.5 & 99.8$\pm$0.2 & 98.1$\pm$3.6 \\
		VARLiNGAM & 48.2$\pm$5.5 & 11.8$\pm$0.8 & 50.7$\pm$6.1 & 12.9$\pm$3.8 & 52.1$\pm$4.5 & 15.4$\pm$6.8 & 48.2$\pm$5.5 & 11.8$\pm$0.8 \\
		PCMCI & 98.2$\pm$0.9 & 79.1$\pm$8.8 & 96.7$\pm$1.9 & 72.2$\pm$5.8 & 57.6$\pm$2.0 & 12.8$\pm$0.5 & 98.2$\pm$0.9 & 79.1$\pm$8.8 \\
		DYNOTEARS & \textbf{100.0$\pm$0.0} & \textbf{100.0$\pm$0.0} & 80.5$\pm$5.0 & 59.3$\pm$10.4 & 60.0$\pm$7.9 & 15.6$\pm$4.0 & \textbf{100.0$\pm$0.0} & \textbf{100.0$\pm$0.0} \\
		NTS-NOTEARS & 52.0$\pm$3.9 & 14.6$\pm$7.1 & 75.0$\pm$3.1 & 55.5$\pm$5.6 & 56.2$\pm$5.1 & 12.6$\pm$1.3 & 79.7$\pm$12.8 & 60.9$\pm$21.3 \\
\bottomrule
	\end{tabular}
	\label{tab:linear-10-1000-1}

\end{table*}

\begin{table*}[!htbp]
	\centering
	\caption{Linear setting, 10-node case with $T = 1000$ (Part II).}
	\setlength{\tabcolsep}{4pt}
	\begin{tabular}{ccccccccccc}
		\toprule
		& \multicolumn{2}{c}{Latent confounders} & \multicolumn{2}{c}{Measurement error} & \multicolumn{2}{c}{Standardized} & \multicolumn{2}{c}{Missing} & \multicolumn{2}{c}{Nonstationary} \\
		\cmidrule(lr){2-3} \cmidrule(lr){4-5} \cmidrule(lr){6-7} \cmidrule(lr){8-9} \cmidrule(lr){10-11}
		10 nodes & AUROC$\uparrow$ & AUPRC$\uparrow$ & AUROC$\uparrow$ & AUPRC$\uparrow$ & AUROC$\uparrow$ & AUPRC$\uparrow$ & AUROC$\uparrow$ & AUPRC$\uparrow$ & AUROC$\uparrow$ & AUPRC$\uparrow$ \\
		\midrule
		VAR & 86.0\footnotesize{$\pm$2.7} & 31.4\footnotesize{$\pm$4.0} & 85.2\footnotesize{$\pm$8.9} & 65.6\footnotesize{$\pm$18.8} & 99.7\footnotesize{$\pm$0.3} & 96.3\footnotesize{$\pm$4.4} & 81.8\footnotesize{$\pm$10.5} & 58.3\footnotesize{$\pm$21.7} & 92.7\footnotesize{$\pm$2.6} & 48.4\footnotesize{$\pm$10.9} \\
		LGC & 90.9\footnotesize{$\pm$3.3} & 42.5\footnotesize{$\pm$7.3} & 84.5\footnotesize{$\pm$7.4} & 66.9\footnotesize{$\pm$17.4} & \textbf{100.0\footnotesize{$\pm$0.0}} & \textbf{100.0\footnotesize{$\pm$0.0}} & 76.0\footnotesize{$\pm$5.8} & 57.3\footnotesize{$\pm$10.3} & 97.7\footnotesize{$\pm$2.2} & 78.6\footnotesize{$\pm$19.1} \\
		VARLiNGAM & 51.1\footnotesize{$\pm$3.3} & 11.7\footnotesize{$\pm$0.7} & 58.2\footnotesize{$\pm$5.1} & 13.7\footnotesize{$\pm$1.8} & 48.2\footnotesize{$\pm$5.5} & 11.8\footnotesize{$\pm$0.8} & 53.1\footnotesize{$\pm$9.0} & 13.0\footnotesize{$\pm$3.9} & 47.6\footnotesize{$\pm$5.7} & 11.6\footnotesize{$\pm$0.9} \\
		PCMCI & 76.3\footnotesize{$\pm$2.2} & 21.0\footnotesize{$\pm$1.4} & 91.3\footnotesize{$\pm$6.0} & 68.5\footnotesize{$\pm$11.9} & 98.2\footnotesize{$\pm$0.9} & 79.1\footnotesize{$\pm$8.8} & 93.3\footnotesize{$\pm$4.5} & 66.0\footnotesize{$\pm$14.7} & 85.1\footnotesize{$\pm$5.9} & 32.2\footnotesize{$\pm$9.5} \\
\bottomrule
	\end{tabular}
	\label{tab:linear-10-1000-2}

\end{table*}

\begin{table*}[!htbp]
	\centering
    \caption{Linear NSHD results for the 10-node case with $T = 1000$ (Part I).}
	\setlength{\tabcolsep}{8.3pt}
	\begin{tabular}{ccccc}
		\toprule
		& \multicolumn{1}{c}{Vanilla} & \multicolumn{1}{c}{Mixed data} & \multicolumn{1}{c}{Trend and seasonality} & \multicolumn{1}{c}{Min--max normalization} \\
		\cmidrule(lr){2-2} \cmidrule(lr){3-3} \cmidrule(lr){4-4} \cmidrule(lr){5-5}
		10 nodes & NSHD$\downarrow$ & NSHD$\downarrow$ & NSHD$\downarrow$ & NSHD$\downarrow$ \\
		\midrule
		VAR & \textbf{0.00\footnotesize{$\pm$0.00}} & 0.76\footnotesize{$\pm$0.04} & \textbf{0.80\footnotesize{$\pm$0.06}} & 0.02\footnotesize{$\pm$0.04} \\
		LGC & \textbf{0.00\footnotesize{$\pm$0.00}} & 0.42\footnotesize{$\pm$0.07} & 0.94\footnotesize{$\pm$0.07} & \textbf{0.00\footnotesize{$\pm$0.00}} \\
		VARLiNGAM & 1.00\footnotesize{$\pm$0.00} & 1.00\footnotesize{$\pm$0.00} & 0.92\footnotesize{$\pm$0.11} & 1.00\footnotesize{$\pm$0.00} \\
		PCMCI & 0.24\footnotesize{$\pm$0.13} & 0.34\footnotesize{$\pm$0.04} & 1.00\footnotesize{$\pm$0.00} & 0.24\footnotesize{$\pm$0.13} \\
		DYNOTEARS & \textbf{0.00\footnotesize{$\pm$0.00}} & 0.38\footnotesize{$\pm$0.11} & 0.96\footnotesize{$\pm$0.04} & \textbf{0.00\footnotesize{$\pm$0.00}} \\
		NTS-NOTEARS & 0.96\footnotesize{$\pm$0.07} & 0.44\footnotesize{$\pm$0.08} & 0.98\footnotesize{$\pm$0.03} & 0.64\footnotesize{$\pm$0.30} \\
		TSCI & 0.87\footnotesize{$\pm$0.07} & 0.98\footnotesize{$\pm$0.03} & 1.00\footnotesize{$\pm$0.00} & 0.87\footnotesize{$\pm$0.07} \\
		cMLP & 0.02\footnotesize{$\pm$0.04} & \textbf{0.10\footnotesize{$\pm$0.06}} & 0.96\footnotesize{$\pm$0.04} & 0.02\footnotesize{$\pm$0.04} \\
		cLSTM & \textbf{0.00\footnotesize{$\pm$0.00}} & 0.20\footnotesize{$\pm$0.14} & 0.94\footnotesize{$\pm$0.07} & \textbf{0.00\footnotesize{$\pm$0.00}} \\
		CUTS & 0.41\footnotesize{$\pm$0.17} & 0.94\footnotesize{$\pm$0.07} & 0.96\footnotesize{$\pm$0.04} & 0.98\footnotesize{$\pm$0.03} \\
		CUTS+ & 0.84\footnotesize{$\pm$0.20} & 0.78\footnotesize{$\pm$0.16} & 0.96\footnotesize{$\pm$0.04} & 0.92\footnotesize{$\pm$0.07} \\
		\bottomrule
	\end{tabular}
	\label{tab:nshd-linear-10-1000-1}
\end{table*}

\begin{table*}[!htbp]
	\centering
    \caption{Linear NSHD results for the 10-node case with $T = 1000$ (Part II).}
	\setlength{\tabcolsep}{4pt}
	\begin{tabular}{cccccc}
		\toprule
		& \multicolumn{1}{c}{Latent confounders} & \multicolumn{1}{c}{Measurement error} & \multicolumn{1}{c}{Standardized} & \multicolumn{1}{c}{Missing} & \multicolumn{1}{c}{Nonstationary} \\
		\cmidrule(lr){2-2} \cmidrule(lr){3-3} \cmidrule(lr){4-4} \cmidrule(lr){5-5} \cmidrule(lr){6-6}
		10 nodes & NSHD$\downarrow$ & NSHD$\downarrow$ & NSHD$\downarrow$ & NSHD$\downarrow$ & NSHD$\downarrow$ \\
		\midrule
		VAR & 0.24\footnotesize{$\pm$0.13} & 0.22\footnotesize{$\pm$0.11} & \textbf{0.00\footnotesize{$\pm$0.00}} & 0.32\footnotesize{$\pm$0.14} & 0.13\footnotesize{$\pm$0.18} \\
		LGC & 0.32\footnotesize{$\pm$0.11} & 0.22\footnotesize{$\pm$0.11} & \textbf{0.00\footnotesize{$\pm$0.00}} & 0.18\footnotesize{$\pm$0.16} & 0.13\footnotesize{$\pm$0.17} \\
		VARLiNGAM & 0.98\footnotesize{$\pm$0.03} & 0.98\footnotesize{$\pm$0.03} & 1.00\footnotesize{$\pm$0.00} & 1.00\footnotesize{$\pm$0.00} & 0.98\footnotesize{$\pm$0.03} \\
		PCMCI & 1.00\footnotesize{$\pm$0.00} & 0.32\footnotesize{$\pm$0.11} & 0.24\footnotesize{$\pm$0.13} & 0.46\footnotesize{$\pm$0.26} & 1.00\footnotesize{$\pm$0.00} \\
		DYNOTEARS & 0.70\footnotesize{$\pm$0.08} & 0.13\footnotesize{$\pm$0.07} & \textbf{0.00\footnotesize{$\pm$0.00}} & \textbf{0.04\footnotesize{$\pm$0.04}} & \textbf{0.04\footnotesize{$\pm$0.08}} \\
		NTS-NOTEARS & 0.78\footnotesize{$\pm$0.07} & 0.93\footnotesize{$\pm$0.07} & \textbf{0.00\footnotesize{$\pm$0.00}} & 0.92\footnotesize{$\pm$0.16} & 0.26\footnotesize{$\pm$0.32} \\
\bottomrule
	\end{tabular}
	\label{tab:nshd-linear-10-1000-2}
\end{table*}

\begin{table*}[!htbp]
	\centering
	\caption{Linear setting, 15-node case with $T = 500$ (Part I).}
	\setlength{\tabcolsep}{8.3pt}
	\begin{tabular}{ccccccccc}
		\toprule
		& \multicolumn{2}{c}{Vanilla} & \multicolumn{2}{c}{Mixed data} & \multicolumn{2}{c}{Trend and seasonality} & \multicolumn{2}{c}{Min--max normalization} \\
		\cmidrule(lr){2-3} \cmidrule(lr){4-5} \cmidrule(lr){6-7} \cmidrule(lr){8-9}
		15 nodes & AUROC$\uparrow$ & AUPRC$\uparrow$ & AUROC$\uparrow$ & AUPRC$\uparrow$ & AUROC$\uparrow$ & AUPRC$\uparrow$ & AUROC$\uparrow$ & AUPRC$\uparrow$ \\
		\midrule
		VAR & 90.7\footnotesize{$\pm$3.2} & 59.6\footnotesize{$\pm$8.1} & 62.5\footnotesize{$\pm$1.9} & 29.0\footnotesize{$\pm$4.6} & \textbf{72.3\footnotesize{$\pm$4.3}} & \textbf{23.9\footnotesize{$\pm$2.7}} & 89.6\footnotesize{$\pm$4.6} & 57.6\footnotesize{$\pm$9.6} \\
		LGC & 91.8\footnotesize{$\pm$2.5} & 72.0\footnotesize{$\pm$4.9} & 71.4\footnotesize{$\pm$5.2} & 42.1\footnotesize{$\pm$7.2} & 50.8\footnotesize{$\pm$1.0} & 15.0\footnotesize{$\pm$1.0} & 87.1\footnotesize{$\pm$5.9} & 59.3\footnotesize{$\pm$8.7} \\
		VARLiNGAM & 51.1\footnotesize{$\pm$4.6} & 15.0\footnotesize{$\pm$1.3} & 50.9\footnotesize{$\pm$5.6} & 15.2\footnotesize{$\pm$2.4} & 50.7\footnotesize{$\pm$1.7} & 15.0\footnotesize{$\pm$1.0} & 51.1\footnotesize{$\pm$3.6} & 14.7\footnotesize{$\pm$1.1} \\
		PCMCI & 89.1\footnotesize{$\pm$4.7} & 71.2\footnotesize{$\pm$9.7} & 77.8\footnotesize{$\pm$5.1} & 51.6\footnotesize{$\pm$10.6} & 64.5\footnotesize{$\pm$6.0} & 19.9\footnotesize{$\pm$3.0} & 89.1\footnotesize{$\pm$4.7} & 71.2\footnotesize{$\pm$9.7} \\
		DYNOTEARS & 96.5\footnotesize{$\pm$0.7} & 82.0\footnotesize{$\pm$3.8} & 70.6\footnotesize{$\pm$4.0} & 37.4\footnotesize{$\pm$5.2} & 55.0\footnotesize{$\pm$3.2} & 16.6\footnotesize{$\pm$1.8} & 86.9\footnotesize{$\pm$4.4} & 74.0\footnotesize{$\pm$9.4} \\
		NTS-NOTEARS & 50.0\footnotesize{$\pm$0.0} & 14.2\footnotesize{$\pm$0.0} & 63.2\footnotesize{$\pm$1.2} & 33.9\footnotesize{$\pm$3.0} & 52.6\footnotesize{$\pm$3.0} & 15.0\footnotesize{$\pm$0.8} & 50.0\footnotesize{$\pm$0.0} & 14.2\footnotesize{$\pm$0.0} \\
\bottomrule
	\end{tabular}
    \label{tab:linear-15-500-1}

\end{table*}

\begin{table*}[!htbp]
	\centering
	\caption{Linear setting, 15-node case with $T = 500$ (Part II).}
	\setlength{\tabcolsep}{4pt}
	\begin{tabular}{ccccccccccc}
		\toprule
		& \multicolumn{2}{c}{Latent confounders} & \multicolumn{2}{c}{Measurement error} & \multicolumn{2}{c}{Standardized} & \multicolumn{2}{c}{Missing} & \multicolumn{2}{c}{Nonstationary} \\
		\cmidrule(lr){2-3} \cmidrule(lr){4-5} \cmidrule(lr){6-7} \cmidrule(lr){8-9} \cmidrule(lr){10-11}
		15 nodes & AUROC$\uparrow$ & AUPRC$\uparrow$ & AUROC$\uparrow$ & AUPRC$\uparrow$ & AUROC$\uparrow$ & AUPRC$\uparrow$ & AUROC$\uparrow$ & AUPRC$\uparrow$ & AUROC$\uparrow$ & AUPRC$\uparrow$ \\
		\midrule
		VAR & 74.1\footnotesize{$\pm$4.0} & 25.6\footnotesize{$\pm$2.6} & 68.1\footnotesize{$\pm$3.8} & 28.1\footnotesize{$\pm$3.9} & 91.0\footnotesize{$\pm$4.3} & 60.3\footnotesize{$\pm$10.7} & 60.1\footnotesize{$\pm$1.4} & 19.7\footnotesize{$\pm$1.2} & 77.3\footnotesize{$\pm$8.1} & 30.9\footnotesize{$\pm$8.1} \\
		LGC & 74.8\footnotesize{$\pm$5.7} & 50.5\footnotesize{$\pm$10.2} & 62.3\footnotesize{$\pm$4.1} & 26.2\footnotesize{$\pm$4.4} & 95.2\footnotesize{$\pm$2.2} & 70.1\footnotesize{$\pm$8.6} & 71.4\footnotesize{$\pm$2.2} & 23.6\footnotesize{$\pm$1.6} & 83.0\footnotesize{$\pm$7.9} & 56.8\footnotesize{$\pm$19.4} \\
		VARLiNGAM & 53.0\footnotesize{$\pm$1.9} & 15.7\footnotesize{$\pm$1.3} & 51.3\footnotesize{$\pm$3.3} & 15.9\footnotesize{$\pm$2.6} & 51.6\footnotesize{$\pm$3.7} & 15.0\footnotesize{$\pm$1.2} & 50.6\footnotesize{$\pm$1.0} & 14.7\footnotesize{$\pm$0.4} & 49.5\footnotesize{$\pm$4.3} & 14.4\footnotesize{$\pm$0.6} \\
		PCMCI & 83.4\footnotesize{$\pm$3.5} & 42.7\footnotesize{$\pm$4.6} & 64.1\footnotesize{$\pm$3.0} & 28.0\footnotesize{$\pm$3.8} & 89.1\footnotesize{$\pm$4.7} & 71.2\footnotesize{$\pm$9.7} & 72.8\footnotesize{$\pm$1.7} & 29.8\footnotesize{$\pm$1.8} & 82.8\footnotesize{$\pm$5.5} & 41.1\footnotesize{$\pm$9.8} \\
\bottomrule
	\end{tabular}
    \label{tab:linear-15-500-2}

\end{table*}

\begin{table*}[!htbp]
	\centering
	\caption{Linear setting, 15-node case with $T = 1000$ (Part I).}
	\setlength{\tabcolsep}{8.3pt}
	\begin{tabular}{ccccccccc}
		\toprule
		& \multicolumn{2}{c}{Vanilla} & \multicolumn{2}{c}{Mixed data} & \multicolumn{2}{c}{Trend and seasonality} & \multicolumn{2}{c}{Min--max normalization} \\
		\cmidrule(lr){2-3} \cmidrule(lr){4-5} \cmidrule(lr){6-7} \cmidrule(lr){8-9}
		15 nodes & AUROC$\uparrow$ & AUPRC$\uparrow$ & AUROC$\uparrow$ & AUPRC$\uparrow$ & AUROC$\uparrow$ & AUPRC$\uparrow$ & AUROC$\uparrow$ & AUPRC$\uparrow$ \\
		\midrule
		VAR & 93.2\footnotesize{$\pm$2.6} & 84.2\footnotesize{$\pm$6.4} & 61.1\footnotesize{$\pm$2.0} & 31.9\footnotesize{$\pm$4.1} & \textbf{81.7\footnotesize{$\pm$2.2}} & \textbf{35.6\footnotesize{$\pm$2.9}} & 92.9\footnotesize{$\pm$2.1} & 81.3\footnotesize{$\pm$5.8} \\
		LGC & 90.9\footnotesize{$\pm$2.7} & 83.9\footnotesize{$\pm$5.0} & 74.7\footnotesize{$\pm$5.3} & 52.4\footnotesize{$\pm$12.2} & 50.0\footnotesize{$\pm$0.0} & 14.2\footnotesize{$\pm$0.0} & 92.0\footnotesize{$\pm$2.3} & 83.1\footnotesize{$\pm$4.9} \\
		VARLiNGAM & 51.6\footnotesize{$\pm$5.9} & 15.4\footnotesize{$\pm$2.0} & 51.1\footnotesize{$\pm$3.8} & 15.0\footnotesize{$\pm$1.5} & 49.7\footnotesize{$\pm$0.2} & 14.2\footnotesize{$\pm$0.0} & 50.6\footnotesize{$\pm$5.0} & 14.9\footnotesize{$\pm$1.4} \\
		PCMCI & 98.0\footnotesize{$\pm$0.8} & 83.4\footnotesize{$\pm$4.3} & \textbf{90.6\footnotesize{$\pm$3.3}} & \textbf{71.5\footnotesize{$\pm$11.3}} & 62.0\footnotesize{$\pm$2.2} & 18.1\footnotesize{$\pm$0.8} & 98.0\footnotesize{$\pm$0.8} & 83.4\footnotesize{$\pm$4.3} \\
		DYNOTEARS & 99.0\footnotesize{$\pm$0.8} & 95.1\footnotesize{$\pm$3.1} & 72.8\footnotesize{$\pm$3.9} & 49.3\footnotesize{$\pm$8.2} & 58.1\footnotesize{$\pm$4.6} & 17.4\footnotesize{$\pm$1.8} & 97.8\footnotesize{$\pm$0.3} & 79.8\footnotesize{$\pm$2.9} \\
		NTS-NOTEARS & 50.0\footnotesize{$\pm$0.0} & 14.2\footnotesize{$\pm$0.0} & 78.4\footnotesize{$\pm$4.5} & 35.6\footnotesize{$\pm$4.3} & 51.8\footnotesize{$\pm$3.6} & 14.8\footnotesize{$\pm$1.0} & 50.0\footnotesize{$\pm$0.0} & 14.2\footnotesize{$\pm$0.0} \\
\bottomrule
	\end{tabular}
    \label{tab:linear-15-1000-1}

\end{table*}

\begin{table*}[!htbp]
	\centering
	\caption{Linear setting, 15-node case with $T = 1000$ (Part II).}
	\setlength{\tabcolsep}{4pt}
	\begin{tabular}{ccccccccccc}
		\toprule
		& \multicolumn{2}{c}{Latent confounders} & \multicolumn{2}{c}{Measurement error} & \multicolumn{2}{c}{Standardized} & \multicolumn{2}{c}{Missing} & \multicolumn{2}{c}{Nonstationary} \\
		\cmidrule(lr){2-3} \cmidrule(lr){4-5} \cmidrule(lr){6-7} \cmidrule(lr){8-9} \cmidrule(lr){10-11}
		15 nodes & AUROC$\uparrow$ & AUPRC$\uparrow$ & AUROC$\uparrow$ & AUPRC$\uparrow$ & AUROC$\uparrow$ & AUPRC$\uparrow$ & AUROC$\uparrow$ & AUPRC$\uparrow$ & AUROC$\uparrow$ & AUPRC$\uparrow$ \\
		\midrule
		VAR & 79.2\footnotesize{$\pm$7.2} & 36.9\footnotesize{$\pm$9.1} & 82.0\footnotesize{$\pm$2.0} & 32.6\footnotesize{$\pm$2.5} & 92.6\footnotesize{$\pm$1.9} & 83.6\footnotesize{$\pm$3.8} & 77.9\footnotesize{$\pm$3.4} & 29.0\footnotesize{$\pm$3.6} & 90.2\footnotesize{$\pm$5.0} & 60.2\footnotesize{$\pm$13.6} \\
		LGC & 91.1\footnotesize{$\pm$2.9} & 60.0\footnotesize{$\pm$6.8} & 81.4\footnotesize{$\pm$2.0} & 34.0\footnotesize{$\pm$2.8} & 98.3\footnotesize{$\pm$0.8} & 84.5\footnotesize{$\pm$8.2} & 85.1\footnotesize{$\pm$2.1} & 44.7\footnotesize{$\pm$5.0} & 85.8\footnotesize{$\pm$1.9} & 70.0\footnotesize{$\pm$5.3} \\
		VARLiNGAM & 51.8\footnotesize{$\pm$5.7} & 15.0\footnotesize{$\pm$1.7} & 51.5\footnotesize{$\pm$3.8} & 15.1\footnotesize{$\pm$1.0} & 51.8\footnotesize{$\pm$5.8} & 15.4\footnotesize{$\pm$1.9} & 51.6\footnotesize{$\pm$3.7} & 15.0\footnotesize{$\pm$1.3} & 50.8\footnotesize{$\pm$3.2} & 14.6\footnotesize{$\pm$0.9} \\
		PCMCI & 85.7\footnotesize{$\pm$3.7} & 40.9\footnotesize{$\pm$5.8} & 73.8\footnotesize{$\pm$1.4} & 46.3\footnotesize{$\pm$3.4} & 98.0\footnotesize{$\pm$0.8} & 83.4\footnotesize{$\pm$4.3} & 82.0\footnotesize{$\pm$3.4} & 40.0\footnotesize{$\pm$6.1} & 88.2\footnotesize{$\pm$5.5} & 46.4\footnotesize{$\pm$9.6} \\
\bottomrule
	\end{tabular}
    \label{tab:linear-15-1000-2}

\end{table*}

\begin{table*}[!htbp]
	\centering
	\caption{Nonlinear setting, 10-node case with $T = 500$ and $F = 10$ (Part I).}
	\setlength{\tabcolsep}{8.3pt}
	\begin{tabular}{ccccccccc}
		\toprule
		& \multicolumn{2}{c}{Vanilla} & \multicolumn{2}{c}{Mixed data} & \multicolumn{2}{c}{Trend and seasonality} & \multicolumn{2}{c}{Min--max normalization} \\
		\cmidrule(lr){2-3} \cmidrule(lr){4-5} \cmidrule(lr){6-7} \cmidrule(lr){8-9}
		10 nodes & AUROC$\uparrow$ & AUPRC$\uparrow$ & AUROC$\uparrow$ & AUPRC$\uparrow$ & AUROC$\uparrow$ & AUPRC$\uparrow$ & AUROC$\uparrow$ & AUPRC$\uparrow$ \\
		\midrule
		VAR & 77.0\footnotesize{$\pm$3.1} & 55.5\footnotesize{$\pm$4.3} & 59.6\footnotesize{$\pm$5.6} & 39.4\footnotesize{$\pm$4.1} & 67.3\footnotesize{$\pm$5.5} & 56.4\footnotesize{$\pm$7.3} & 65.9\footnotesize{$\pm$5.6} & 54.6\footnotesize{$\pm$7.5} \\
		LGC & 76.9\footnotesize{$\pm$2.4} & 62.9\footnotesize{$\pm$3.8} & 67.8\footnotesize{$\pm$3.9} & 54.1\footnotesize{$\pm$4.8} & 76.3\footnotesize{$\pm$2.5} & 61.5\footnotesize{$\pm$5.1} & 76.0\footnotesize{$\pm$3.0} & 65.2\footnotesize{$\pm$5.9} \\
		VARLiNGAM & 72.1\footnotesize{$\pm$6.4} & 61.1\footnotesize{$\pm$9.4} & 66.1\footnotesize{$\pm$2.7} & 46.3\footnotesize{$\pm$2.7} & 71.5\footnotesize{$\pm$3.9} & 60.2\footnotesize{$\pm$6.2} & 74.3\footnotesize{$\pm$6.4} & 65.0\footnotesize{$\pm$9.3} \\
		PCMCI & 71.3\footnotesize{$\pm$5.2} & 50.4\footnotesize{$\pm$7.0} & 73.8\footnotesize{$\pm$0.8} & 52.7\footnotesize{$\pm$1.4} & 72.8\footnotesize{$\pm$5.9} & 52.2\footnotesize{$\pm$7.8} & 71.3\footnotesize{$\pm$5.2} & 50.4\footnotesize{$\pm$7.0} \\
		DYNOTEARS & 75.8\footnotesize{$\pm$6.9} & 65.9\footnotesize{$\pm$10.5} & 59.6\footnotesize{$\pm$2.2} & 42.8\footnotesize{$\pm$3.5} & 75.6\footnotesize{$\pm$6.4} & 66.1\footnotesize{$\pm$9.9} & 71.0\footnotesize{$\pm$3.1} & 53.0\footnotesize{$\pm$4.1} \\
		NTS-NOTEARS & 68.8\footnotesize{$\pm$3.2} & 44.6\footnotesize{$\pm$2.6} & 69.8\footnotesize{$\pm$0.6} & 52.7\footnotesize{$\pm$3.1} & 51.3\footnotesize{$\pm$0.4} & 33.9\footnotesize{$\pm$0.1} & 50.0\footnotesize{$\pm$0.0} & 33.3\footnotesize{$\pm$0.0} \\
\bottomrule
	\end{tabular}
    \label{tab:nonlinear-10-500-f10-1}

\end{table*}

\begin{table*}[!htbp]
	\centering
	\caption{Nonlinear setting, 10-node case with $T = 500$ and $F = 10$ (Part II).}
	\setlength{\tabcolsep}{4pt}
	\begin{tabular}{ccccccccccc}
		\toprule
		& \multicolumn{2}{c}{Latent confounders} & \multicolumn{2}{c}{Measurement error} & \multicolumn{2}{c}{Standardized} & \multicolumn{2}{c}{Missing} & \multicolumn{2}{c}{Nonstationary} \\
		\cmidrule(lr){2-3} \cmidrule(lr){4-5} \cmidrule(lr){6-7} \cmidrule(lr){8-9} \cmidrule(lr){10-11}
		10 nodes & AUROC$\uparrow$ & AUPRC$\uparrow$ & AUROC$\uparrow$ & AUPRC$\uparrow$ & AUROC$\uparrow$ & AUPRC$\uparrow$ & AUROC$\uparrow$ & AUPRC$\uparrow$ & AUROC$\uparrow$ & AUPRC$\uparrow$ \\
		\midrule
		VAR & 54.8\footnotesize{$\pm$4.0} & 36.6\footnotesize{$\pm$3.2} & 62.8\footnotesize{$\pm$6.5} & 44.0\footnotesize{$\pm$7.4} & 65.9\footnotesize{$\pm$5.8} & 54.6\footnotesize{$\pm$7.7} & 64.1\footnotesize{$\pm$5.2} & 44.5\footnotesize{$\pm$6.0} & 66.8\footnotesize{$\pm$9.2} & 46.8\footnotesize{$\pm$8.0} \\
		LGC & 55.1\footnotesize{$\pm$4.1} & 36.9\footnotesize{$\pm$3.6} & 63.1\footnotesize{$\pm$6.3} & 44.4\footnotesize{$\pm$7.2} & 74.6\footnotesize{$\pm$2.1} & 56.2\footnotesize{$\pm$2.4} & 64.1\footnotesize{$\pm$4.8} & 46.3\footnotesize{$\pm$6.0} & 68.5\footnotesize{$\pm$11.5} & 52.7\footnotesize{$\pm$15.0} \\
		VARLiNGAM & 57.0\footnotesize{$\pm$5.0} & 38.4\footnotesize{$\pm$4.6} & 55.8\footnotesize{$\pm$3.9} & 37.3\footnotesize{$\pm$3.1} & 73.6\footnotesize{$\pm$7.9} & 63.5\footnotesize{$\pm$11.4} & 62.6\footnotesize{$\pm$4.8} & 42.7\footnotesize{$\pm$4.4} & 60.5\footnotesize{$\pm$11.5} & 44.9\footnotesize{$\pm$13.1} \\
		PCMCI & 57.0\footnotesize{$\pm$5.3} & 38.5\footnotesize{$\pm$4.7} & 65.6\footnotesize{$\pm$5.1} & 45.6\footnotesize{$\pm$5.7} & 71.3\footnotesize{$\pm$5.2} & 50.4\footnotesize{$\pm$7.0} & 63.5\footnotesize{$\pm$3.2} & 47.7\footnotesize{$\pm$4.6} & 65.0\footnotesize{$\pm$8.1} & 46.1\footnotesize{$\pm$7.3} \\
\bottomrule
	\end{tabular}
    \label{tab:nonlinear-10-500-f10-2}

\end{table*}

\begin{table*}[!htbp]
	\centering
	\caption{Nonlinear setting, 10-node case with $T = 500$ and $F = 40$ (Part I).}
	\setlength{\tabcolsep}{8.3pt}
	\begin{tabular}{ccccccccc}
		\toprule
		& \multicolumn{2}{c}{Vanilla} & \multicolumn{2}{c}{Mixed data} & \multicolumn{2}{c}{Trend and seasonality} & \multicolumn{2}{c}{Min--max normalization} \\
		\cmidrule(lr){2-3} \cmidrule(lr){4-5} \cmidrule(lr){6-7} \cmidrule(lr){8-9}
		10 nodes & AUROC$\uparrow$ & AUPRC$\uparrow$ & AUROC$\uparrow$ & AUPRC$\uparrow$ & AUROC$\uparrow$ & AUPRC$\uparrow$ & AUROC$\uparrow$ & AUPRC$\uparrow$ \\
		\midrule
		VAR & 68.8\footnotesize{$\pm$4.1} & 48.3\footnotesize{$\pm$4.2} & 54.8\footnotesize{$\pm$2.6} & 38.3\footnotesize{$\pm$2.8} & 63.6\footnotesize{$\pm$5.5} & 42.2\footnotesize{$\pm$4.4} & 68.3\footnotesize{$\pm$2.9} & 47.8\footnotesize{$\pm$2.9} \\
		LGC & 68.5\footnotesize{$\pm$3.7} & 48.0\footnotesize{$\pm$3.8} & 61.9\footnotesize{$\pm$2.7} & 41.6\footnotesize{$\pm$2.9} & 63.8\footnotesize{$\pm$5.8} & 42.5\footnotesize{$\pm$5.0} & 63.0\footnotesize{$\pm$4.6} & 44.6\footnotesize{$\pm$5.0} \\
		VARLiNGAM & 61.8\footnotesize{$\pm$3.8} & 42.0\footnotesize{$\pm$3.4} & 57.3\footnotesize{$\pm$2.6} & 38.1\footnotesize{$\pm$1.5} & 63.5\footnotesize{$\pm$0.9} & 42.6\footnotesize{$\pm$0.8} & 62.3\footnotesize{$\pm$3.3} & 41.7\footnotesize{$\pm$2.3} \\
		PCMCI & 66.1\footnotesize{$\pm$5.5} & 47.9\footnotesize{$\pm$7.6} & 65.3\footnotesize{$\pm$5.6} & 49.3\footnotesize{$\pm$7.6} & 64.3\footnotesize{$\pm$4.3} & 45.6\footnotesize{$\pm$4.9} & 66.1\footnotesize{$\pm$5.5} & 47.9\footnotesize{$\pm$7.6} \\
		DYNOTEARS & 66.0\footnotesize{$\pm$2.5} & 43.6\footnotesize{$\pm$2.0} & 56.3\footnotesize{$\pm$2.3} & 38.2\footnotesize{$\pm$2.4} & 64.4\footnotesize{$\pm$3.5} & 42.5\footnotesize{$\pm$2.9} & 64.0\footnotesize{$\pm$4.1} & 45.0\footnotesize{$\pm$3.9} \\
		NTS-NOTEARS & 50.0\footnotesize{$\pm$0.0} & 33.3\footnotesize{$\pm$0.0} & 60.1\footnotesize{$\pm$1.7} & 43.9\footnotesize{$\pm$2.5} & 50.0\footnotesize{$\pm$0.0} & 33.3\footnotesize{$\pm$0.0} & 50.0\footnotesize{$\pm$0.0} & 33.3\footnotesize{$\pm$0.0} \\
\bottomrule
	\end{tabular}
    \label{tab:nonlinear-10-500-f40-1}

\end{table*}

\begin{table*}[!htbp]
	\centering
	\caption{Nonlinear setting, 10-node case with $T = 500$ and $F = 40$ (Part II).}
	\setlength{\tabcolsep}{4pt}
	\begin{tabular}{ccccccccccc}
		\toprule
		& \multicolumn{2}{c}{Latent confounders} & \multicolumn{2}{c}{Measurement error} & \multicolumn{2}{c}{Standardized} & \multicolumn{2}{c}{Missing} & \multicolumn{2}{c}{Nonstationary} \\
		\cmidrule(lr){2-3} \cmidrule(lr){4-5} \cmidrule(lr){6-7} \cmidrule(lr){8-9} \cmidrule(lr){10-11}
		10 nodes & AUROC$\uparrow$ & AUPRC$\uparrow$ & AUROC$\uparrow$ & AUPRC$\uparrow$ & AUROC$\uparrow$ & AUPRC$\uparrow$ & AUROC$\uparrow$ & AUPRC$\uparrow$ & AUROC$\uparrow$ & AUPRC$\uparrow$ \\
		\midrule
		VAR & 51.8\footnotesize{$\pm$2.1} & 34.4\footnotesize{$\pm$1.2} & 57.6\footnotesize{$\pm$2.6} & 39.3\footnotesize{$\pm$2.5} & 69.0\footnotesize{$\pm$3.6} & 48.3\footnotesize{$\pm$3.3} & 59.0\footnotesize{$\pm$5.2} & 38.2\footnotesize{$\pm$3.1} & 60.6\footnotesize{$\pm$9.1} & 42.3\footnotesize{$\pm$7.8} \\
		LGC & 51.8\footnotesize{$\pm$2.1} & 34.4\footnotesize{$\pm$1.2} & 57.8\footnotesize{$\pm$2.8} & 39.5\footnotesize{$\pm$2.7} & 68.5\footnotesize{$\pm$4.0} & 49.0\footnotesize{$\pm$5.2} & 58.8\footnotesize{$\pm$4.8} & 38.1\footnotesize{$\pm$2.9} & 60.6\footnotesize{$\pm$9.1} & 42.3\footnotesize{$\pm$7.8} \\
		VARLiNGAM & 52.0\footnotesize{$\pm$2.0} & 35.0\footnotesize{$\pm$1.6} & 52.0\footnotesize{$\pm$2.8} & 35.0\footnotesize{$\pm$1.9} & 62.1\footnotesize{$\pm$1.3} & 41.6\footnotesize{$\pm$1.0} & 53.4\footnotesize{$\pm$2.7} & 36.1\footnotesize{$\pm$2.5} & 56.0\footnotesize{$\pm$4.3} & 37.8\footnotesize{$\pm$3.0} \\
		PCMCI & 54.0\footnotesize{$\pm$2.5} & 35.7\footnotesize{$\pm$1.6} & 60.5\footnotesize{$\pm$4.7} & 40.5\footnotesize{$\pm$4.0} & 66.1\footnotesize{$\pm$5.5} & 47.9\footnotesize{$\pm$7.6} & 57.6\footnotesize{$\pm$5.9} & 39.2\footnotesize{$\pm$4.8} & 60.1\footnotesize{$\pm$8.9} & 43.1\footnotesize{$\pm$10.4} \\
\bottomrule
	\end{tabular}
    \label{tab:nonlinear-10-500-f40-2}

\end{table*}

\begin{table*}[!htbp]
	\centering
	\caption{Nonlinear setting, 10-node case with $T = 1000$ and $F = 10$ (Part I).}
	\setlength{\tabcolsep}{8.3pt}
	\begin{tabular}{ccccccccc}
		\toprule
		& \multicolumn{2}{c}{Vanilla} & \multicolumn{2}{c}{Mixed data} & \multicolumn{2}{c}{Trend and seasonality} & \multicolumn{2}{c}{Min--max normalization} \\
		\cmidrule(lr){2-3} \cmidrule(lr){4-5} \cmidrule(lr){6-7} \cmidrule(lr){8-9}
		10 nodes & AUROC$\uparrow$ & AUPRC$\uparrow$ & AUROC$\uparrow$ & AUPRC$\uparrow$ & AUROC$\uparrow$ & AUPRC$\uparrow$ & AUROC$\uparrow$ & AUPRC$\uparrow$ \\
		\midrule
		VAR & 83.3\footnotesize{$\pm$3.6} & 66.6\footnotesize{$\pm$5.6} & 60.8\footnotesize{$\pm$4.2} & 40.2\footnotesize{$\pm$3.4} & 82.5\footnotesize{$\pm$2.4} & 65.7\footnotesize{$\pm$3.8} & 82.6\footnotesize{$\pm$2.8} & 65.8\footnotesize{$\pm$4.8} \\
		LGC & 83.8\footnotesize{$\pm$2.6} & 72.4\footnotesize{$\pm$4.5} & 69.0\footnotesize{$\pm$5.4} & 49.7\footnotesize{$\pm$8.4} & 81.8\footnotesize{$\pm$2.3} & 69.3\footnotesize{$\pm$3.8} & 86.6\footnotesize{$\pm$1.0} & 74.8\footnotesize{$\pm$2.7} \\
		VARLiNGAM & 72.0\footnotesize{$\pm$1.9} & 60.3\footnotesize{$\pm$2.7} & 65.8\footnotesize{$\pm$4.2} & 47.6\footnotesize{$\pm$4.7} & 71.6\footnotesize{$\pm$2.6} & 58.7\footnotesize{$\pm$5.2} & 72.3\footnotesize{$\pm$1.8} & 60.7\footnotesize{$\pm$2.6} \\
		PCMCI & 77.8\footnotesize{$\pm$4.9} & 55.9\footnotesize{$\pm$6.1} & \textbf{73.3\footnotesize{$\pm$3.6}} & 54.0\footnotesize{$\pm$4.4} & 79.5\footnotesize{$\pm$3.5} & 58.2\footnotesize{$\pm$5.4} & 77.8\footnotesize{$\pm$4.9} & 55.9\footnotesize{$\pm$6.1} \\
		DYNOTEARS & 86.1\footnotesize{$\pm$2.1} & 71.2\footnotesize{$\pm$3.4} & 57.3\footnotesize{$\pm$2.9} & 40.7\footnotesize{$\pm$4.0} & \textbf{87.8\footnotesize{$\pm$1.7}} & \textbf{73.6\footnotesize{$\pm$1.6}} & 66.1\footnotesize{$\pm$0.6} & 53.8\footnotesize{$\pm$2.2} \\
		NTS-NOTEARS & 82.9\footnotesize{$\pm$1.4} & 59.5\footnotesize{$\pm$2.0} & 69.0\footnotesize{$\pm$2.2} & 51.7\footnotesize{$\pm$3.8} & 60.6\footnotesize{$\pm$4.0} & 39.0\footnotesize{$\pm$2.6} & 50.0\footnotesize{$\pm$0.0} & 33.3\footnotesize{$\pm$0.0} \\
\bottomrule
	\end{tabular}
	\label{tab:nonlinear-10-1000-f10-1}

\end{table*}

\begin{table*}[!htbp]
	\centering
	\caption{Nonlinear setting, 10-node case with $T = 1000$ and $F = 10$ (Part II).}
	\setlength{\tabcolsep}{4pt}
	\begin{tabular}{ccccccccccc}
		\toprule
		& \multicolumn{2}{c}{Latent confounders} & \multicolumn{2}{c}{Measurement error} & \multicolumn{2}{c}{Standardized} & \multicolumn{2}{c}{Missing} & \multicolumn{2}{c}{Nonstationary} \\
		\cmidrule(lr){2-3} \cmidrule(lr){4-5} \cmidrule(lr){6-7} \cmidrule(lr){8-9} \cmidrule(lr){10-11}
		10 nodes & AUROC$\uparrow$ & AUPRC$\uparrow$ & AUROC$\uparrow$ & AUPRC$\uparrow$ & AUROC$\uparrow$ & AUPRC$\uparrow$ & AUROC$\uparrow$ & AUPRC$\uparrow$ & AUROC$\uparrow$ & AUPRC$\uparrow$ \\
		\midrule
		VAR & 59.8\footnotesize{$\pm$9.5} & 39.7\footnotesize{$\pm$5.9} & 73.8\footnotesize{$\pm$3.4} & 50.3\footnotesize{$\pm$3.8} & 83.6\footnotesize{$\pm$4.1} & 67.1\footnotesize{$\pm$6.5} & 74.8\footnotesize{$\pm$2.3} & 50.7\footnotesize{$\pm$2.2} & 65.1\footnotesize{$\pm$14.9} & 48.3\footnotesize{$\pm$14.6} \\
		LGC & 60.0\footnotesize{$\pm$9.6} & 39.9\footnotesize{$\pm$6.0} & 73.5\footnotesize{$\pm$3.9} & 50.3\footnotesize{$\pm$4.4} & 83.5\footnotesize{$\pm$3.2} & 67.3\footnotesize{$\pm$5.0} & 81.3\footnotesize{$\pm$2.6} & 60.1\footnotesize{$\pm$3.4} & 71.1\footnotesize{$\pm$14.4} & 50.7\footnotesize{$\pm$12.8} \\
		VARLiNGAM & 61.9\footnotesize{$\pm$2.7} & 42.4\footnotesize{$\pm$2.7} & 62.6\footnotesize{$\pm$5.8} & 41.8\footnotesize{$\pm$4.4} & 74.0\footnotesize{$\pm$2.9} & 63.7\footnotesize{$\pm$4.0} & 67.6\footnotesize{$\pm$3.5} & 44.7\footnotesize{$\pm$2.8} & 63.0\footnotesize{$\pm$12.6} & 45.1\footnotesize{$\pm$10.8} \\
		PCMCI & 61.3\footnotesize{$\pm$7.1} & 41.5\footnotesize{$\pm$6.0} & 75.1\footnotesize{$\pm$5.6} & 56.2\footnotesize{$\pm$7.9} & 77.8\footnotesize{$\pm$4.9} & 55.9\footnotesize{$\pm$6.1} & 78.1\footnotesize{$\pm$3.3} & 59.9\footnotesize{$\pm$5.7} & 70.3\footnotesize{$\pm$13.2} & 52.6\footnotesize{$\pm$14.0} \\
\bottomrule
	\end{tabular}
	\label{tab:nonlinear-10-1000-f10-2}

\end{table*}

\begin{table*}[!htbp]
	\centering
    \caption{Nonlinear NSHD results for the 10-node case with $T = 1000$ and $F = 10$ (Part I).}
	\setlength{\tabcolsep}{8.3pt}
	\begin{tabular}{ccccc}
		\toprule
		& \multicolumn{1}{c}{Vanilla} & \multicolumn{1}{c}{Mixed data} & \multicolumn{1}{c}{Trend and seasonality} & \multicolumn{1}{c}{Min--max normalization} \\
		\cmidrule(lr){2-2} \cmidrule(lr){3-3} \cmidrule(lr){4-4} \cmidrule(lr){5-5}
		10 nodes & NSHD$\downarrow$ & NSHD$\downarrow$ & NSHD$\downarrow$ & NSHD$\downarrow$ \\
		\midrule
		VAR & 0.34\footnotesize{$\pm$0.05} & 0.90\footnotesize{$\pm$0.04} & 0.38\footnotesize{$\pm$0.06} & 0.35\footnotesize{$\pm$0.04} \\
		LGC & 0.34\footnotesize{$\pm$0.05} & 0.69\footnotesize{$\pm$0.09} & 0.38\footnotesize{$\pm$0.08} & \textbf{0.30\footnotesize{$\pm$0.04}} \\
		VARLiNGAM & 0.52\footnotesize{$\pm$0.04} & 0.75\footnotesize{$\pm$0.06} & 0.54\footnotesize{$\pm$0.06} & 0.51\footnotesize{$\pm$0.06} \\
		PCMCI & 0.71\footnotesize{$\pm$0.14} & \textbf{0.65\footnotesize{$\pm$0.10}} & 0.55\footnotesize{$\pm$0.07} & 0.71\footnotesize{$\pm$0.14} \\
		DYNOTEARS & 0.36\footnotesize{$\pm$0.05} & 0.84\footnotesize{$\pm$0.04} & \textbf{0.34\footnotesize{$\pm$0.08}} & 0.66\footnotesize{$\pm$0.02} \\
		NTS-NOTEARS & 0.66\footnotesize{$\pm$0.06} & 0.71\footnotesize{$\pm$0.06} & 1.61\footnotesize{$\pm$0.12} & 1.00\footnotesize{$\pm$0.00} \\
		TSCI & 0.52\footnotesize{$\pm$0.06} & 0.80\footnotesize{$\pm$0.05} & 0.90\footnotesize{$\pm$0.09} & 0.52\footnotesize{$\pm$0.06} \\
		cMLP & 0.01\footnotesize{$\pm$0.01} & 0.96\footnotesize{$\pm$0.02} & 0.87\footnotesize{$\pm$0.02} & 0.98\footnotesize{$\pm$0.02} \\
		cLSTM & \textbf{0.00\footnotesize{$\pm$0.00}} & 0.86\footnotesize{$\pm$0.06} & 0.79\footnotesize{$\pm$0.05} & 0.96\footnotesize{$\pm$0.02} \\
		CUTS & 0.16\footnotesize{$\pm$0.08} & 0.85\footnotesize{$\pm$0.07} & 0.96\footnotesize{$\pm$0.04} & 0.86\footnotesize{$\pm$0.10} \\
		CUTS+ & 0.00\footnotesize{$\pm$0.01} & 0.68\footnotesize{$\pm$0.05} & 0.96\footnotesize{$\pm$0.02} & 0.83\footnotesize{$\pm$0.05} \\
		\bottomrule
	\end{tabular}
	\label{tab:nshd-nonlinear-10-1000-f10-1}
\end{table*}

\begin{table*}[!htbp]
	\centering
    \caption{Nonlinear NSHD results for the 10-node case with $T = 1000$ and $F = 10$ (Part II).}

	\setlength{\tabcolsep}{4pt}
	\begin{tabular}{cccccc}
		\toprule
		& \multicolumn{1}{c}{Latent confounders} & \multicolumn{1}{c}{Measurement error} & \multicolumn{1}{c}{Standardized} & \multicolumn{1}{c}{Missing} & \multicolumn{1}{c}{Nonstationary} \\
		\cmidrule(lr){2-2} \cmidrule(lr){3-3} \cmidrule(lr){4-4} \cmidrule(lr){5-5} \cmidrule(lr){6-6}
		10 nodes & NSHD$\downarrow$ & NSHD$\downarrow$ & NSHD$\downarrow$ & NSHD$\downarrow$ & NSHD$\downarrow$ \\
		\midrule
		VAR & 0.80\footnotesize{$\pm$0.13} & 0.60\footnotesize{$\pm$0.11} & 0.34\footnotesize{$\pm$0.04} & 0.66\footnotesize{$\pm$0.11} & 0.62\footnotesize{$\pm$0.28} \\
		LGC & 0.80\footnotesize{$\pm$0.13} & 0.59\footnotesize{$\pm$0.10} & 0.34\footnotesize{$\pm$0.05} & 0.55\footnotesize{$\pm$0.11} & 0.62\footnotesize{$\pm$0.28} \\
		VARLiNGAM & 0.86\footnotesize{$\pm$0.07} & 0.91\footnotesize{$\pm$0.03} & 0.48\footnotesize{$\pm$0.08} & 0.82\footnotesize{$\pm$0.07} & 0.75\footnotesize{$\pm$0.22} \\
		PCMCI & 0.90\footnotesize{$\pm$0.08} & 0.60\footnotesize{$\pm$0.09} & 0.71\footnotesize{$\pm$0.14} & 0.48\footnotesize{$\pm$0.04} & 0.70\footnotesize{$\pm$0.26} \\
		DYNOTEARS & 0.78\footnotesize{$\pm$0.13} & 0.86\footnotesize{$\pm$0.04} & 0.08\footnotesize{$\pm$0.04} & 0.72\footnotesize{$\pm$0.08} & 0.69\footnotesize{$\pm$0.33} \\
		NTS-NOTEARS & 2.00\footnotesize{$\pm$0.00} & 2.00\footnotesize{$\pm$0.00} & \textbf{0.00\footnotesize{$\pm$0.00}} & 2.00\footnotesize{$\pm$0.00} & 1.71\footnotesize{$\pm$0.50} \\
\bottomrule
	\end{tabular}
	\label{tab:nshd-nonlinear-10-1000-f10-2}
\end{table*}

\begin{table*}[!htbp]
	\centering
	\caption{Nonlinear setting, 10-node case with $T = 1000$ and $F = 40$ (Part I).}
	\setlength{\tabcolsep}{8.3pt}
	\begin{tabular}{ccccccccc}
		\toprule
		& \multicolumn{2}{c}{Vanilla} & \multicolumn{2}{c}{Mixed data} & \multicolumn{2}{c}{Trend and seasonality} & \multicolumn{2}{c}{Min--max normalization} \\
		\cmidrule(lr){2-3} \cmidrule(lr){4-5} \cmidrule(lr){6-7} \cmidrule(lr){8-9}
		10 nodes & AUROC$\uparrow$ & AUPRC$\uparrow$ & AUROC$\uparrow$ & AUPRC$\uparrow$ & AUROC$\uparrow$ & AUPRC$\uparrow$ & AUROC$\uparrow$ & AUPRC$\uparrow$ \\
		\midrule
		VAR & 73.0\footnotesize{$\pm$2.0} & 58.5\footnotesize{$\pm$3.1} & 62.1\footnotesize{$\pm$1.3} & 42.0\footnotesize{$\pm$1.3} & 68.5\footnotesize{$\pm$3.2} & 49.7\footnotesize{$\pm$4.3} & 73.6\footnotesize{$\pm$2.3} & 60.0\footnotesize{$\pm$5.0} \\
		LGC & 72.8\footnotesize{$\pm$2.2} & 58.7\footnotesize{$\pm$4.2} & 66.3\footnotesize{$\pm$3.2} & 47.2\footnotesize{$\pm$3.3} & 68.5\footnotesize{$\pm$2.4} & 49.9\footnotesize{$\pm$3.6} & 70.5\footnotesize{$\pm$4.2} & 57.1\footnotesize{$\pm$6.1} \\
		VARLiNGAM & 65.5\footnotesize{$\pm$2.7} & 45.0\footnotesize{$\pm$3.0} & 63.0\footnotesize{$\pm$2.9} & 43.0\footnotesize{$\pm$2.7} & 68.6\footnotesize{$\pm$2.1} & 50.5\footnotesize{$\pm$3.0} & 66.1\footnotesize{$\pm$2.1} & 45.6\footnotesize{$\pm$2.6} \\
		PCMCI & 75.5\footnotesize{$\pm$1.5} & 59.8\footnotesize{$\pm$4.0} & \textbf{75.1\footnotesize{$\pm$4.8}} & 60.9\footnotesize{$\pm$6.1} & 66.6\footnotesize{$\pm$1.8} & 45.9\footnotesize{$\pm$1.4} & 75.5\footnotesize{$\pm$1.5} & 59.8\footnotesize{$\pm$4.0} \\
		DYNOTEARS & 69.6\footnotesize{$\pm$2.8} & 50.1\footnotesize{$\pm$3.5} & 56.5\footnotesize{$\pm$2.4} & 39.1\footnotesize{$\pm$2.8} & 70.6\footnotesize{$\pm$2.4} & 50.7\footnotesize{$\pm$3.6} & 70.0\footnotesize{$\pm$4.6} & 58.0\footnotesize{$\pm$7.4} \\
		NTS-NOTEARS & 50.1\footnotesize{$\pm$0.3} & 33.4\footnotesize{$\pm$0.1} & 57.3\footnotesize{$\pm$0.9} & 42.1\footnotesize{$\pm$2.2} & 50.0\footnotesize{$\pm$0.0} & 33.3\footnotesize{$\pm$0.0} & 50.0\footnotesize{$\pm$0.0} & 33.3\footnotesize{$\pm$0.0} \\
\bottomrule
	\end{tabular}
    \label{tab:nonlinear-10-1000-f40-1}

\end{table*}

\begin{table*}[!htbp]
	\centering
	\caption{Nonlinear setting, 10-node case with $T = 1000$ and $F = 40$ (Part II).}
	\setlength{\tabcolsep}{4pt}
	\begin{tabular}{ccccccccccc}
		\toprule
		& \multicolumn{2}{c}{Latent confounders} & \multicolumn{2}{c}{Measurement error} & \multicolumn{2}{c}{Standardized} & \multicolumn{2}{c}{Missing} & \multicolumn{2}{c}{Nonstationary} \\
		\cmidrule(lr){2-3} \cmidrule(lr){4-5} \cmidrule(lr){6-7} \cmidrule(lr){8-9} \cmidrule(lr){10-11}
		10 nodes & AUROC$\uparrow$ & AUPRC$\uparrow$ & AUROC$\uparrow$ & AUPRC$\uparrow$ & AUROC$\uparrow$ & AUPRC$\uparrow$ & AUROC$\uparrow$ & AUPRC$\uparrow$ & AUROC$\uparrow$ & AUPRC$\uparrow$ \\
		\midrule
		VAR & 50.6\footnotesize{$\pm$5.1} & 34.0\footnotesize{$\pm$2.2} & 61.9\footnotesize{$\pm$3.5} & 40.8\footnotesize{$\pm$2.7} & 72.8\footnotesize{$\pm$2.2} & 58.7\footnotesize{$\pm$4.2} & 62.8\footnotesize{$\pm$4.0} & 41.3\footnotesize{$\pm$3.2} & 60.5\footnotesize{$\pm$9.7} & 44.7\footnotesize{$\pm$11.1} \\
		LGC & 50.6\footnotesize{$\pm$5.1} & 34.0\footnotesize{$\pm$2.2} & 62.3\footnotesize{$\pm$3.2} & 41.1\footnotesize{$\pm$2.5} & 75.6\footnotesize{$\pm$4.9} & 57.4\footnotesize{$\pm$7.2} & 63.6\footnotesize{$\pm$3.6} & 42.0\footnotesize{$\pm$2.9} & 60.5\footnotesize{$\pm$9.7} & 44.7\footnotesize{$\pm$11.1} \\
		VARLiNGAM & 53.3\footnotesize{$\pm$1.5} & 36.5\footnotesize{$\pm$2.1} & 53.9\footnotesize{$\pm$3.5} & 36.6\footnotesize{$\pm$3.2} & 66.6\footnotesize{$\pm$3.4} & 46.4\footnotesize{$\pm$4.1} & 52.6\footnotesize{$\pm$2.5} & 35.3\footnotesize{$\pm$1.6} & 54.8\footnotesize{$\pm$7.6} & 37.1\footnotesize{$\pm$5.1} \\
		PCMCI & 51.8\footnotesize{$\pm$2.7} & 34.4\footnotesize{$\pm$1.5} & 62.1\footnotesize{$\pm$5.9} & 42.7\footnotesize{$\pm$6.0} & 75.5\footnotesize{$\pm$1.5} & 59.8\footnotesize{$\pm$4.0} & 63.5\footnotesize{$\pm$5.4} & 43.9\footnotesize{$\pm$5.7} & 62.5\footnotesize{$\pm$10.3} & 46.1\footnotesize{$\pm$12.4} \\
\bottomrule
	\end{tabular}
    \label{tab:nonlinear-10-1000-f40-2}

\end{table*}

\begin{table*}[!htbp]
	\centering
	\caption{Nonlinear setting, 15-node case with $T = 500$ and $F = 10$ (Part I).}
	\setlength{\tabcolsep}{8.3pt}
	\begin{tabular}{ccccccccc}
		\toprule
		& \multicolumn{2}{c}{Vanilla} & \multicolumn{2}{c}{Mixed data} & \multicolumn{2}{c}{Trend and seasonality} & \multicolumn{2}{c}{Min--max normalization} \\
		\cmidrule(lr){2-3} \cmidrule(lr){4-5} \cmidrule(lr){6-7} \cmidrule(lr){8-9}
		15 nodes & AUROC$\uparrow$ & AUPRC$\uparrow$ & AUROC$\uparrow$ & AUPRC$\uparrow$ & AUROC$\uparrow$ & AUPRC$\uparrow$ & AUROC$\uparrow$ & AUPRC$\uparrow$ \\
		\midrule
		VAR & 80.8\footnotesize{$\pm$3.6} & 46.8\footnotesize{$\pm$4.8} & 62.2\footnotesize{$\pm$2.9} & 27.7\footnotesize{$\pm$1.8} & 80.6\footnotesize{$\pm$1.8} & 46.3\footnotesize{$\pm$2.5} & 81.0\footnotesize{$\pm$3.7} & 47.0\footnotesize{$\pm$4.8} \\
		LGC & 78.8\footnotesize{$\pm$4.5} & 58.4\footnotesize{$\pm$5.2} & 69.9\footnotesize{$\pm$2.7} & 50.3\footnotesize{$\pm$5.2} & 78.4\footnotesize{$\pm$4.2} & 58.3\footnotesize{$\pm$4.6} & 78.7\footnotesize{$\pm$4.8} & 63.5\footnotesize{$\pm$7.0} \\
		VARLiNGAM & 65.3\footnotesize{$\pm$4.4} & 42.8\footnotesize{$\pm$5.3} & 66.3\footnotesize{$\pm$2.1} & 32.3\footnotesize{$\pm$1.9} & 66.9\footnotesize{$\pm$2.5} & 46.3\footnotesize{$\pm$3.4} & 65.7\footnotesize{$\pm$3.8} & 44.1\footnotesize{$\pm$4.5} \\
		PCMCI & 79.3\footnotesize{$\pm$4.3} & 51.7\footnotesize{$\pm$8.0} & 70.7\footnotesize{$\pm$3.3} & 43.5\footnotesize{$\pm$5.2} & 81.3\footnotesize{$\pm$2.7} & 53.9\footnotesize{$\pm$5.7} & 79.3\footnotesize{$\pm$4.3} & 51.7\footnotesize{$\pm$8.0} \\
		DYNOTEARS & 89.5\footnotesize{$\pm$3.0} & 64.7\footnotesize{$\pm$5.6} & 60.8\footnotesize{$\pm$2.0} & 31.9\footnotesize{$\pm$3.1} & \textbf{89.1\footnotesize{$\pm$4.0}} & \textbf{64.5\footnotesize{$\pm$6.8}} & 74.1\footnotesize{$\pm$4.1} & 48.2\footnotesize{$\pm$6.6} \\
		NTS-NOTEARS & 75.8\footnotesize{$\pm$2.6} & 36.2\footnotesize{$\pm$2.3} & 69.7\footnotesize{$\pm$1.6} & 40.0\footnotesize{$\pm$3.6} & 51.3\footnotesize{$\pm$0.4} & 21.9\footnotesize{$\pm$0.1} & 50.0\footnotesize{$\pm$0.0} & 21.4\footnotesize{$\pm$0.0} \\
\bottomrule
	\end{tabular}
    \label{tab:nonlinear-15-500-f10-1}

\end{table*}

\begin{table*}[!htbp]
	\centering
	\caption{Nonlinear setting, 15-node case with $T = 500$ and $F = 10$ (Part II).}
	\setlength{\tabcolsep}{4pt}
	\begin{tabular}{ccccccccccc}
		\toprule
		& \multicolumn{2}{c}{Latent confounders} & \multicolumn{2}{c}{Measurement error} & \multicolumn{2}{c}{Standardized} & \multicolumn{2}{c}{Missing} & \multicolumn{2}{c}{Nonstationary} \\
		\cmidrule(lr){2-3} \cmidrule(lr){4-5} \cmidrule(lr){6-7} \cmidrule(lr){8-9} \cmidrule(lr){10-11}
		15 nodes & AUROC$\uparrow$ & AUPRC$\uparrow$ & AUROC$\uparrow$ & AUPRC$\uparrow$ & AUROC$\uparrow$ & AUPRC$\uparrow$ & AUROC$\uparrow$ & AUPRC$\uparrow$ & AUROC$\uparrow$ & AUPRC$\uparrow$ \\
		\midrule
		VAR & 55.7\footnotesize{$\pm$3.7} & 24.3\footnotesize{$\pm$2.1} & 62.9\footnotesize{$\pm$2.4} & 30.9\footnotesize{$\pm$2.6} & 81.1\footnotesize{$\pm$3.2} & 46.9\footnotesize{$\pm$4.2} & 69.7\footnotesize{$\pm$1.9} & 37.3\footnotesize{$\pm$2.7} & 59.6\footnotesize{$\pm$13.8} & 28.7\footnotesize{$\pm$11.7} \\
		LGC & 54.9\footnotesize{$\pm$3.3} & 24.2\footnotesize{$\pm$2.6} & 63.2\footnotesize{$\pm$2.9} & 31.8\footnotesize{$\pm$3.8} & 81.4\footnotesize{$\pm$3.1} & 47.0\footnotesize{$\pm$5.1} & 68.4\footnotesize{$\pm$2.1} & 40.5\footnotesize{$\pm$3.3} & 58.9\footnotesize{$\pm$12.7} & 30.3\footnotesize{$\pm$15.1} \\
		VARLiNGAM & 52.4\footnotesize{$\pm$2.8} & 24.2\footnotesize{$\pm$3.9} & 56.9\footnotesize{$\pm$1.4} & 25.4\footnotesize{$\pm$1.2} & 65.2\footnotesize{$\pm$3.2} & 42.8\footnotesize{$\pm$3.2} & 68.9\footnotesize{$\pm$4.1} & 35.0\footnotesize{$\pm$4.7} & 57.1\footnotesize{$\pm$7.3} & 26.9\footnotesize{$\pm$6.9} \\
		PCMCI & 54.6\footnotesize{$\pm$4.6} & 23.6\footnotesize{$\pm$2.4} & 65.6\footnotesize{$\pm$2.9} & 32.0\footnotesize{$\pm$3.1} & 79.3\footnotesize{$\pm$4.3} & 51.7\footnotesize{$\pm$8.0} & 75.0\footnotesize{$\pm$3.1} & 44.6\footnotesize{$\pm$4.6} & 60.3\footnotesize{$\pm$11.5} & 31.0\footnotesize{$\pm$12.9} \\
\bottomrule
	\end{tabular}
    \label{tab:nonlinear-15-500-f10-2}

\end{table*}

\begin{table*}[!htbp]
	\centering
	\caption{Nonlinear setting, 15-node case with $T = 500$ and $F = 40$ (Part I).}
	\setlength{\tabcolsep}{8.3pt}
	\begin{tabular}{ccccccccc}
		\toprule
		& \multicolumn{2}{c}{Vanilla} & \multicolumn{2}{c}{Mixed data} & \multicolumn{2}{c}{Trend and seasonality} & \multicolumn{2}{c}{Min--max normalization} \\
		\cmidrule(lr){2-3} \cmidrule(lr){4-5} \cmidrule(lr){6-7} \cmidrule(lr){8-9}
		15 nodes & AUROC$\uparrow$ & AUPRC$\uparrow$ & AUROC$\uparrow$ & AUPRC$\uparrow$ & AUROC$\uparrow$ & AUPRC$\uparrow$ & AUROC$\uparrow$ & AUPRC$\uparrow$ \\
		\midrule
		VAR & 69.5\footnotesize{$\pm$5.3} & 37.5\footnotesize{$\pm$6.4} & 54.6\footnotesize{$\pm$1.1} & 26.6\footnotesize{$\pm$2.1} & 67.3\footnotesize{$\pm$3.3} & 33.4\footnotesize{$\pm$3.1} & \textbf{69.3\footnotesize{$\pm$3.4}} & 36.6\footnotesize{$\pm$4.1} \\
		LGC & 69.7\footnotesize{$\pm$5.3} & 37.9\footnotesize{$\pm$6.4} & 62.7\footnotesize{$\pm$1.1} & 29.5\footnotesize{$\pm$1.6} & 67.3\footnotesize{$\pm$3.3} & 33.5\footnotesize{$\pm$3.3} & 59.4\footnotesize{$\pm$3.0} & 34.4\footnotesize{$\pm$4.4} \\
		VARLiNGAM & 62.8\footnotesize{$\pm$1.8} & 31.0\footnotesize{$\pm$1.6} & 59.7\footnotesize{$\pm$4.8} & 28.7\footnotesize{$\pm$4.8} & 64.4\footnotesize{$\pm$3.3} & 36.8\footnotesize{$\pm$5.2} & 63.6\footnotesize{$\pm$2.2} & 31.8\footnotesize{$\pm$2.3} \\
		PCMCI & 65.9\footnotesize{$\pm$3.2} & 38.2\footnotesize{$\pm$5.3} & 64.8\footnotesize{$\pm$3.0} & 38.0\footnotesize{$\pm$4.7} & 63.7\footnotesize{$\pm$3.5} & 33.1\footnotesize{$\pm$4.6} & 65.9\footnotesize{$\pm$3.2} & 38.2\footnotesize{$\pm$5.3} \\
		DYNOTEARS & 71.5\footnotesize{$\pm$3.4} & 36.7\footnotesize{$\pm$3.3} & 59.0\footnotesize{$\pm$0.9} & 26.7\footnotesize{$\pm$0.8} & 71.6\footnotesize{$\pm$3.2} & 36.4\footnotesize{$\pm$3.1} & 68.0\footnotesize{$\pm$2.7} & \textbf{39.7\footnotesize{$\pm$4.1}} \\
		NTS-NOTEARS & 50.0\footnotesize{$\pm$0.0} & 21.4\footnotesize{$\pm$0.0} & 58.2\footnotesize{$\pm$1.8} & 28.1\footnotesize{$\pm$1.6} & 50.0\footnotesize{$\pm$0.0} & 21.4\footnotesize{$\pm$0.0} & 50.0\footnotesize{$\pm$0.0} & 21.4\footnotesize{$\pm$0.0} \\
\bottomrule
	\end{tabular}
    \label{tab:nonlinear-15-500-f40-1}

\end{table*}

\begin{table*}[!htbp]
	\centering
	\caption{Nonlinear setting, 15-node case with $T = 500$ and $F = 40$ (Part II).}
	\setlength{\tabcolsep}{4pt}
	\begin{tabular}{ccccccccccc}
		\toprule
		& \multicolumn{2}{c}{Latent confounders} & \multicolumn{2}{c}{Measurement error} & \multicolumn{2}{c}{Standardized} & \multicolumn{2}{c}{Missing} & \multicolumn{2}{c}{Nonstationary} \\
		\cmidrule(lr){2-3} \cmidrule(lr){4-5} \cmidrule(lr){6-7} \cmidrule(lr){8-9} \cmidrule(lr){10-11}
		15 nodes & AUROC$\uparrow$ & AUPRC$\uparrow$ & AUROC$\uparrow$ & AUPRC$\uparrow$ & AUROC$\uparrow$ & AUPRC$\uparrow$ & AUROC$\uparrow$ & AUPRC$\uparrow$ & AUROC$\uparrow$ & AUPRC$\uparrow$ \\
		\midrule
		VAR & 53.2\footnotesize{$\pm$3.9} & 22.7\footnotesize{$\pm$1.3} & 54.6\footnotesize{$\pm$2.2} & 24.0\footnotesize{$\pm$1.6} & 69.9\footnotesize{$\pm$4.0} & 37.7\footnotesize{$\pm$4.9} & 59.5\footnotesize{$\pm$2.8} & 28.2\footnotesize{$\pm$2.9} & 55.5\footnotesize{$\pm$7.1} & 24.3\footnotesize{$\pm$3.5} \\
		LGC & 53.2\footnotesize{$\pm$3.9} & 22.7\footnotesize{$\pm$1.3} & 54.6\footnotesize{$\pm$2.2} & 24.0\footnotesize{$\pm$1.6} & 68.2\footnotesize{$\pm$1.8} & 38.1\footnotesize{$\pm$4.4} & 59.0\footnotesize{$\pm$2.5} & 27.9\footnotesize{$\pm$2.4} & 54.3\footnotesize{$\pm$6.4} & 24.0\footnotesize{$\pm$3.4} \\
		VARLiNGAM & 50.2\footnotesize{$\pm$1.3} & 21.8\footnotesize{$\pm$0.7} & 52.4\footnotesize{$\pm$2.7} & 23.3\footnotesize{$\pm$2.3} & 63.5\footnotesize{$\pm$2.4} & 31.9\footnotesize{$\pm$2.6} & 51.7\footnotesize{$\pm$1.7} & 23.1\footnotesize{$\pm$2.5} & 55.3\footnotesize{$\pm$5.5} & 25.1\footnotesize{$\pm$4.0} \\
		PCMCI & 52.9\footnotesize{$\pm$2.4} & 22.9\footnotesize{$\pm$1.5} & 57.5\footnotesize{$\pm$4.2} & 26.3\footnotesize{$\pm$3.8} & 65.9\footnotesize{$\pm$3.2} & 38.2\footnotesize{$\pm$5.3} & 55.7\footnotesize{$\pm$3.4} & 27.2\footnotesize{$\pm$4.3} & 55.0\footnotesize{$\pm$5.3} & 25.5\footnotesize{$\pm$4.4} \\
\bottomrule
	\end{tabular}
    \label{tab:nonlinear-15-500-f40-2}

\end{table*}

\begin{table*}[!htbp]
	\centering
	\caption{Nonlinear setting, 15-node case with $T = 1000$ and $F = 10$ (Part I).}
	\setlength{\tabcolsep}{8.3pt}
	\begin{tabular}{ccccccccc}
		\toprule
		& \multicolumn{2}{c}{Vanilla} & \multicolumn{2}{c}{Mixed data} & \multicolumn{2}{c}{Trend and seasonality} & \multicolumn{2}{c}{Min--max normalization} \\
		\cmidrule(lr){2-3} \cmidrule(lr){4-5} \cmidrule(lr){6-7} \cmidrule(lr){8-9}
		15 nodes & AUROC$\uparrow$ & AUPRC$\uparrow$ & AUROC$\uparrow$ & AUPRC$\uparrow$ & AUROC$\uparrow$ & AUPRC$\uparrow$ & AUROC$\uparrow$ & AUPRC$\uparrow$ \\
		\midrule
		VAR & 88.8\footnotesize{$\pm$1.6} & 70.6\footnotesize{$\pm$6.0} & 64.0\footnotesize{$\pm$2.0} & 29.5\footnotesize{$\pm$1.6} & 88.2\footnotesize{$\pm$1.5} & 70.4\footnotesize{$\pm$4.4} & 88.7\footnotesize{$\pm$1.4} & 69.6\footnotesize{$\pm$3.3} \\
		LGC & 85.2\footnotesize{$\pm$2.1} & 74.3\footnotesize{$\pm$3.8} & 76.0\footnotesize{$\pm$3.1} & 52.3\footnotesize{$\pm$5.3} & 83.4\footnotesize{$\pm$1.7} & 70.3\footnotesize{$\pm$3.1} & 90.5\footnotesize{$\pm$0.4} & 77.3\footnotesize{$\pm$2.0} \\
		VARLiNGAM & 73.2\footnotesize{$\pm$1.7} & 54.4\footnotesize{$\pm$4.0} & 74.4\footnotesize{$\pm$2.5} & 40.1\footnotesize{$\pm$3.1} & 72.8\footnotesize{$\pm$0.3} & 53.8\footnotesize{$\pm$2.5} & 72.9\footnotesize{$\pm$1.1} & 54.5\footnotesize{$\pm$3.5} \\
		PCMCI & 83.0\footnotesize{$\pm$2.3} & 53.2\footnotesize{$\pm$3.7} & \textbf{79.8\footnotesize{$\pm$2.7}} & 55.3\footnotesize{$\pm$3.5} & 88.5\footnotesize{$\pm$1.3} & 62.0\footnotesize{$\pm$2.4} & 83.0\footnotesize{$\pm$2.3} & 53.2\footnotesize{$\pm$3.7} \\
		DYNOTEARS & 96.4\footnotesize{$\pm$0.7} & 85.9\footnotesize{$\pm$2.7} & 65.3\footnotesize{$\pm$1.7} & 35.1\footnotesize{$\pm$1.8} & \textbf{94.4\footnotesize{$\pm$1.1}} & \textbf{82.4\footnotesize{$\pm$2.3}} & 66.2\footnotesize{$\pm$0.5} & 46.9\footnotesize{$\pm$0.8} \\
		NTS-NOTEARS & 86.6\footnotesize{$\pm$0.5} & 50.5\footnotesize{$\pm$1.0} & 70.9\footnotesize{$\pm$1.7} & 47.0\footnotesize{$\pm$5.2} & 53.6\footnotesize{$\pm$1.9} & 22.7\footnotesize{$\pm$0.7} & 50.0\footnotesize{$\pm$0.0} & 21.4\footnotesize{$\pm$0.0} \\
\bottomrule
	\end{tabular}
    \label{tab:nonlinear-15-1000-f10-1}

\end{table*}

\begin{table*}[!htbp]
	\centering
	\caption{Nonlinear setting, 15-node case with $T = 1000$ and $F = 10$ (Part II).}
	\setlength{\tabcolsep}{4pt}
	\begin{tabular}{ccccccccccc}
		\toprule
		& \multicolumn{2}{c}{Latent confounders} & \multicolumn{2}{c}{Measurement error} & \multicolumn{2}{c}{Standardized} & \multicolumn{2}{c}{Missing} & \multicolumn{2}{c}{Nonstationary} \\
		\cmidrule(lr){2-3} \cmidrule(lr){4-5} \cmidrule(lr){6-7} \cmidrule(lr){8-9} \cmidrule(lr){10-11}
		15 nodes & AUROC$\uparrow$ & AUPRC$\uparrow$ & AUROC$\uparrow$ & AUPRC$\uparrow$ & AUROC$\uparrow$ & AUPRC$\uparrow$ & AUROC$\uparrow$ & AUPRC$\uparrow$ & AUROC$\uparrow$ & AUPRC$\uparrow$ \\
		\midrule
		VAR & 57.7\footnotesize{$\pm$4.2} & 24.9\footnotesize{$\pm$2.1} & 74.6\footnotesize{$\pm$4.0} & 36.6\footnotesize{$\pm$3.6} & 88.4\footnotesize{$\pm$2.3} & 69.4\footnotesize{$\pm$6.1} & 65.4\footnotesize{$\pm$3.9} & 39.7\footnotesize{$\pm$6.2} & 61.5\footnotesize{$\pm$6.8} & 29.0\footnotesize{$\pm$5.3} \\
		LGC & 57.8\footnotesize{$\pm$4.0} & 24.9\footnotesize{$\pm$2.1} & 75.5\footnotesize{$\pm$4.5} & 38.0\footnotesize{$\pm$4.4} & 85.5\footnotesize{$\pm$2.1} & 69.4\footnotesize{$\pm$3.9} & 84.5\footnotesize{$\pm$3.7} & 53.1\footnotesize{$\pm$5.9} & 61.4\footnotesize{$\pm$7.0} & 30.6\footnotesize{$\pm$8.2} \\
		VARLiNGAM & 55.6\footnotesize{$\pm$2.2} & 25.8\footnotesize{$\pm$2.2} & 65.2\footnotesize{$\pm$2.9} & 32.3\footnotesize{$\pm$2.5} & 74.0\footnotesize{$\pm$1.8} & 56.2\footnotesize{$\pm$4.3} & 73.4\footnotesize{$\pm$3.9} & 38.9\footnotesize{$\pm$3.9} & 60.8\footnotesize{$\pm$6.5} & 29.3\footnotesize{$\pm$6.8} \\
		PCMCI & 57.8\footnotesize{$\pm$3.6} & 25.7\footnotesize{$\pm$2.2} & 77.3\footnotesize{$\pm$0.6} & 44.6\footnotesize{$\pm$1.3} & 83.0\footnotesize{$\pm$2.3} & 53.2\footnotesize{$\pm$3.7} & 84.2\footnotesize{$\pm$4.1} & 53.8\footnotesize{$\pm$6.0} & 68.5\footnotesize{$\pm$6.9} & 34.4\footnotesize{$\pm$6.6} \\
\bottomrule
	\end{tabular}%
    \label{tab:nonlinear-15-1000-f40-2}

\end{table*}

\begin{table*}[!htbp]
	\centering
	\caption{Linear setting, 10-node case with $T = 500$ (Part I). Hyperparameters are selected to maximize average performance across all scenarios.}
	\setlength{\tabcolsep}{8.3pt}
	%
    \label{tab:linear-10-500-avg-scen-1}

\end{table*}

\begin{table*}[!htbp]
	\centering
	\caption{Linear setting, 10-node case with $T = 500$ (Part II). Hyperparameters are selected to maximize average performance across all scenarios.}
	\setlength{\tabcolsep}{4pt}
	%
    \label{tab:linear-10-500-avg-scen-2}

\end{table*}

\begin{table*}[!htbp]
	\centering
	\caption{Linear setting, 10-node case with $T = 1000$ (Part I). Hyperparameters are selected to maximize average performance across all scenarios.}
	\setlength{\tabcolsep}{8.3pt}
	%
    \label{tab:linear-10-1000-avg-scen-1}

\end{table*}

\begin{table*}[!htbp]
	\centering
	\caption{Linear setting, 10-node case with $T = 1000$ (Part II). Hyperparameters are selected to maximize average performance across all scenarios.}
	\setlength{\tabcolsep}{4pt}
	%
    \label{tab:linear-10-1000-avg-scen-2}

\end{table*}

\begin{table*}[!htbp]
	\centering
	\caption{Linear setting, 15-node case with $T = 500$ (Part I). Hyperparameters are selected to maximize average performance across all scenarios.}
	\setlength{\tabcolsep}{8.3pt}
	%
    \label{tab:linear-15-500-avg-scen-1}

\end{table*}

\begin{table*}[!htbp]
	\centering
	\caption{Linear setting, 15-node case with $T = 500$ (Part II). Hyperparameters are selected to maximize average performance across all scenarios.}
	\setlength{\tabcolsep}{4pt}
	%
    \label{tab:linear-15-500-avg-scen-2}

\end{table*}

\begin{table*}[!htbp]
	\centering
	\caption{Linear setting, 15-node case with $T = 1000$ (Part I). Hyperparameters are selected to maximize average performance across all scenarios.}
	\setlength{\tabcolsep}{8.3pt}
	%
    \label{tab:linear-15-1000-avg-scen-2}

\end{table*}

\begin{table*}[!htbp]
	\centering
	\caption{Nonlinear setting, 10-node case with $T = 500$ and $F = 10$ (Part I). Hyperparameters are selected to maximize average performance across all scenarios.}
	\setlength{\tabcolsep}{8.3pt}
	%
    \label{tab:nonlinear-10-500-f10-avg-scen-1}

\end{table*}

\begin{table*}[!htbp]
	\centering
	\caption{Nonlinear setting, 10-node case with $T = 500$ and $F = 10$ (Part II). Hyperparameters are selected to maximize average performance across all scenarios.}
	\setlength{\tabcolsep}{4pt}
	%
    \label{tab:nonlinear-10-500-f10-avg-scen-2}

\end{table*}

\begin{table*}[!htbp]
	\centering
	\caption{Nonlinear setting, 10-node case with $T = 500$ and $F = 40$ (Part I). Hyperparameters are selected to maximize average performance across all scenarios.}
	\setlength{\tabcolsep}{8.3pt}
	%
    \label{tab:nonlinear-10-500-f40-avg-scen-1}

\end{table*}

\begin{table*}[!htbp]
	\centering
	\caption{Nonlinear setting, 10-node case with $T = 500$ and $F = 40$ (Part II). Hyperparameters are selected to maximize average performance across all scenarios.}
	\setlength{\tabcolsep}{4pt}
	%
    \label{tab:nonlinear-10-500-f40-avg-scen-2}

\end{table*}

\begin{table*}[!htbp]
	\centering
	\caption{Nonlinear setting, 10-node case with $T = 1000$ and $F = 10$ (Part I). Hyperparameters are selected to maximize average performance across all scenarios.}
	\setlength{\tabcolsep}{8.3pt}
	%
    \label{tab:nonlinear-10-1000-f10-avg-scen-1}

\end{table*}

\begin{table*}[!htbp]
	\centering
	\caption{Nonlinear setting, 10-node case with $T = 1000$ and $F = 10$ (Part II). Hyperparameters are selected to maximize average performance across all scenarios.}
	\setlength{\tabcolsep}{4pt}
	%
    \label{tab:nonlinear-10-1000-f10-avg-scen-2}

\end{table*}

\begin{table*}[!htbp]
	\centering
	\caption{Nonlinear setting, 10-node case with $T = 1000$ and $F = 40$ (Part I). Hyperparameters are selected to maximize average performance across all scenarios.}
	\setlength{\tabcolsep}{8.3pt}
	%
    \label{tab:nonlinear-15-1000-f40-avg-scen-2}

\end{table*}

\begin{table*}[!htbp]
	\centering
	\caption{Linear setting, 10-node case with $T = 500$ (Part I). Results aggregated over all hyperparameters.}
	\setlength{\tabcolsep}{8.3pt}
	%
    \label{tab:linear-10-500-avg-hyper-1}

\end{table*}

\begin{table*}[!htbp]
	\centering
	\caption{Linear setting, 10-node case with $T = 500$ (Part II). Results aggregated over all hyperparameters.}
	\setlength{\tabcolsep}{4pt}
	%
    \label{tab:linear-10-500-avg-hyper-2}

\end{table*}

\begin{table*}[!htbp]
	\centering
	\caption{Linear setting, 10-node case with $T = 1000$ (Part I). Results aggregated over all hyperparameters.}
	\setlength{\tabcolsep}{8.3pt}
	%
    \label{tab:linear-10-1000-avg-hyper-1}

\end{table*}

\begin{table*}[!htbp]
	\centering
	\caption{Linear setting, 10-node case with $T = 1000$ (Part II). Results aggregated over all hyperparameters.}
	\setlength{\tabcolsep}{4pt}
	%
    \label{tab:linear-10-1000-avg-hyper-2}

\end{table*}

\begin{table*}[!htbp]
	\centering
	\caption{Linear setting, 15-node case with $T = 500$ (Part I). Results aggregated over all hyperparameters.}
	\setlength{\tabcolsep}{8.3pt}
	%
    \label{tab:linear-15-500-avg-hyper-1}

\end{table*}

\begin{table*}[!htbp]
	\centering
	\caption{Linear setting, 15-node case with $T = 500$ (Part II). Results aggregated over all hyperparameters.}
	\setlength{\tabcolsep}{4pt}
	%
    \label{tab:linear-15-500-avg-hyper-2}

\end{table*}

\begin{table*}[!htbp]
	\centering
	\caption{Linear setting, 15-node case with $T = 1000$ (Part I). Results aggregated over all hyperparameters.}
	\setlength{\tabcolsep}{8.3pt}
	%
    \label{tab:linear-15-1000-avg-hyper-2}

\end{table*}

\begin{table*}[!htbp]
	\centering
	\caption{Nonlinear setting, 10-node case with $T = 500$ and $F = 10$ (Part I). Results aggregated over all hyperparameters.}
	\setlength{\tabcolsep}{8.3pt}
	%
    \label{tab:nonlinear-10-500-f10-avg-hyper-1}

\end{table*}

\begin{table*}[!htbp]
	\centering
	\caption{Nonlinear setting, 10-node case with $T = 500$ and $F = 10$ (Part II). Results aggregated over all hyperparameters.}
	\setlength{\tabcolsep}{4pt}
	%
    \label{tab:nonlinear-10-500-f10-avg-hyper-2}

\end{table*}

\begin{table*}[!htbp]
	\centering
	\caption{Nonlinear setting, 10-node case with $T = 500$ and $F = 40$ (Part I). Results aggregated over all hyperparameters.}
	\setlength{\tabcolsep}{8.3pt}
	%
    \label{tab:nonlinear-10-500-f40-avg-hyper-1}

\end{table*}

\begin{table*}[!htbp]
	\centering
	\caption{Nonlinear setting, 10-node case with $T = 500$ and $F = 40$ (Part II). Results aggregated over all hyperparameters.}
	\setlength{\tabcolsep}{4pt}
	%
    \label{tab:nonlinear-10-500-f40-avg-hyper-2}

\end{table*}

\begin{table*}[!htbp]
	\centering
	\caption{Nonlinear setting, 10-node case with $T = 1000$ and $F = 10$ (Part I). Results aggregated over all hyperparameters.}
	\setlength{\tabcolsep}{8.3pt}
	%
    \label{tab:nonlinear-10-1000-f10-avg-hyper-1}

\end{table*}

\begin{table*}[!htbp]
	\centering
	\caption{Nonlinear setting, 10-node case with $T = 1000$ and $F = 10$ (Part II). Results aggregated over all hyperparameters.}
	\setlength{\tabcolsep}{4pt}
	%
    \label{tab:nonlinear-10-1000-f10-avg-hyper-2}

\end{table*}

\begin{table*}[!htbp]
	\centering
	\caption{Nonlinear setting, 10-node case with $T = 1000$ and $F = 40$ (Part I). Results aggregated over all hyperparameters.}
	\setlength{\tabcolsep}{8.3pt}
	%
    \label{tab:nonlinear-15-1000-f40-avg-hyper-2}

\end{table*}

\begin{figure*}[t]
     \centering
     \begin{subfigure}[b]{0.49\textwidth}
         \centering
        \includegraphics[width=\textwidth]{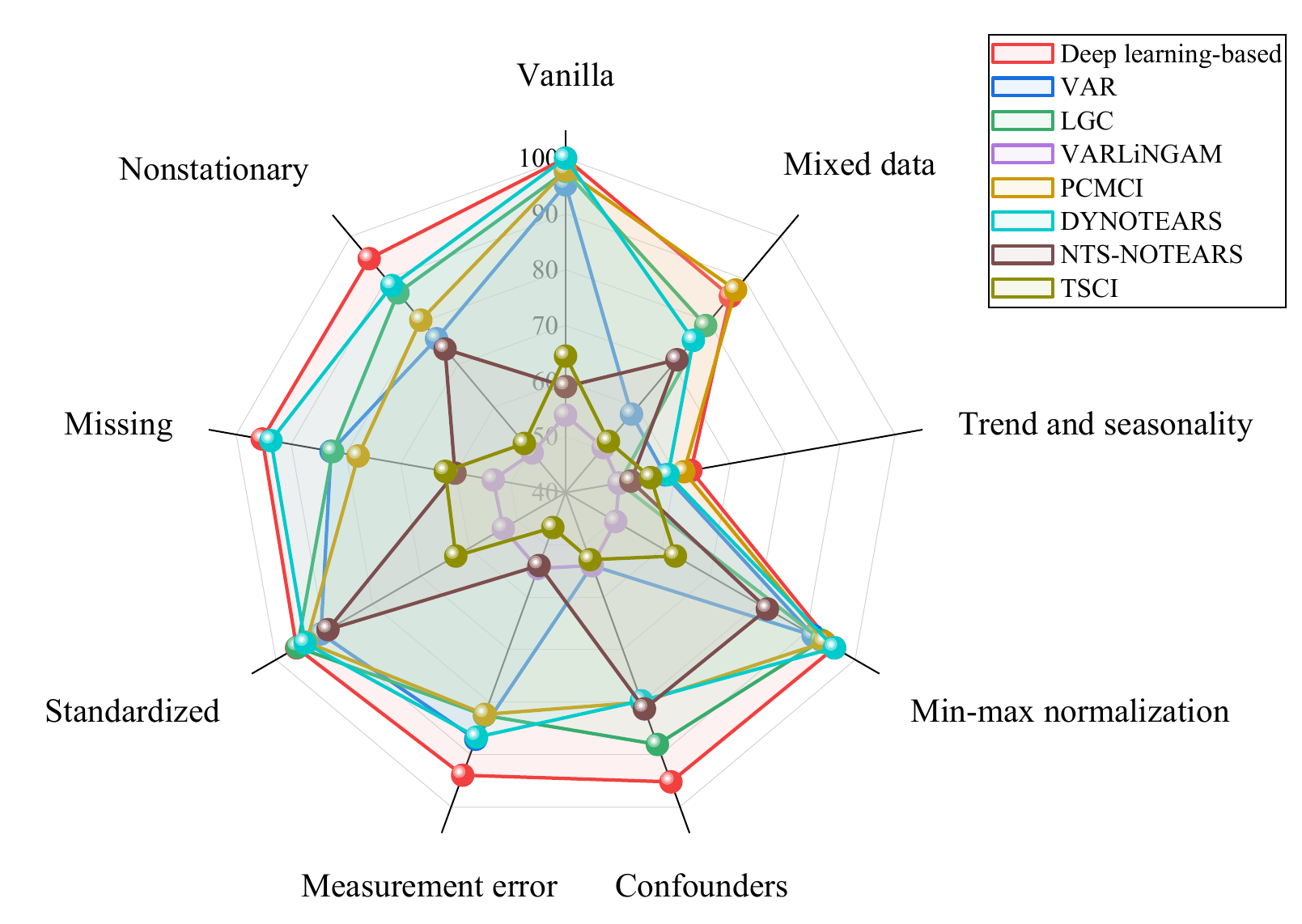}
         \caption{AUROC for linear 10-node case with $T = 500$.}
         \label{fig:linear_10_500_auroc}
     \end{subfigure}%
     \hfill  
     \begin{subfigure}[b]{0.49\textwidth}
         \centering
         \includegraphics[width=\textwidth]{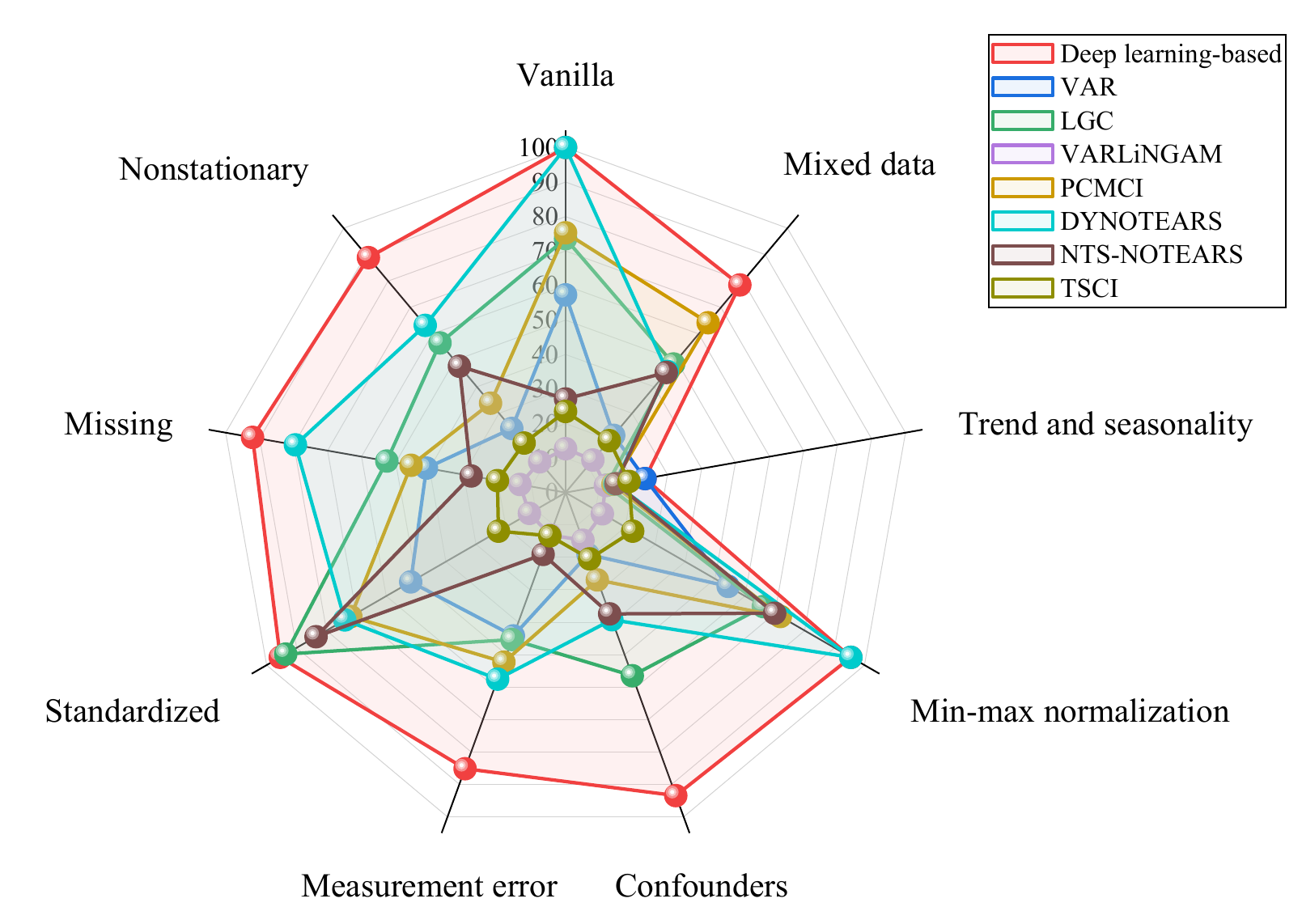}
         \caption{AUPRC for linear 10-node case with $T = 500$.}
         \label{fig:linear_10_500_auprc}
     \end{subfigure}

     \medskip  

     \begin{subfigure}[b]{0.49\textwidth}
         \centering
        \includegraphics[width=\textwidth]{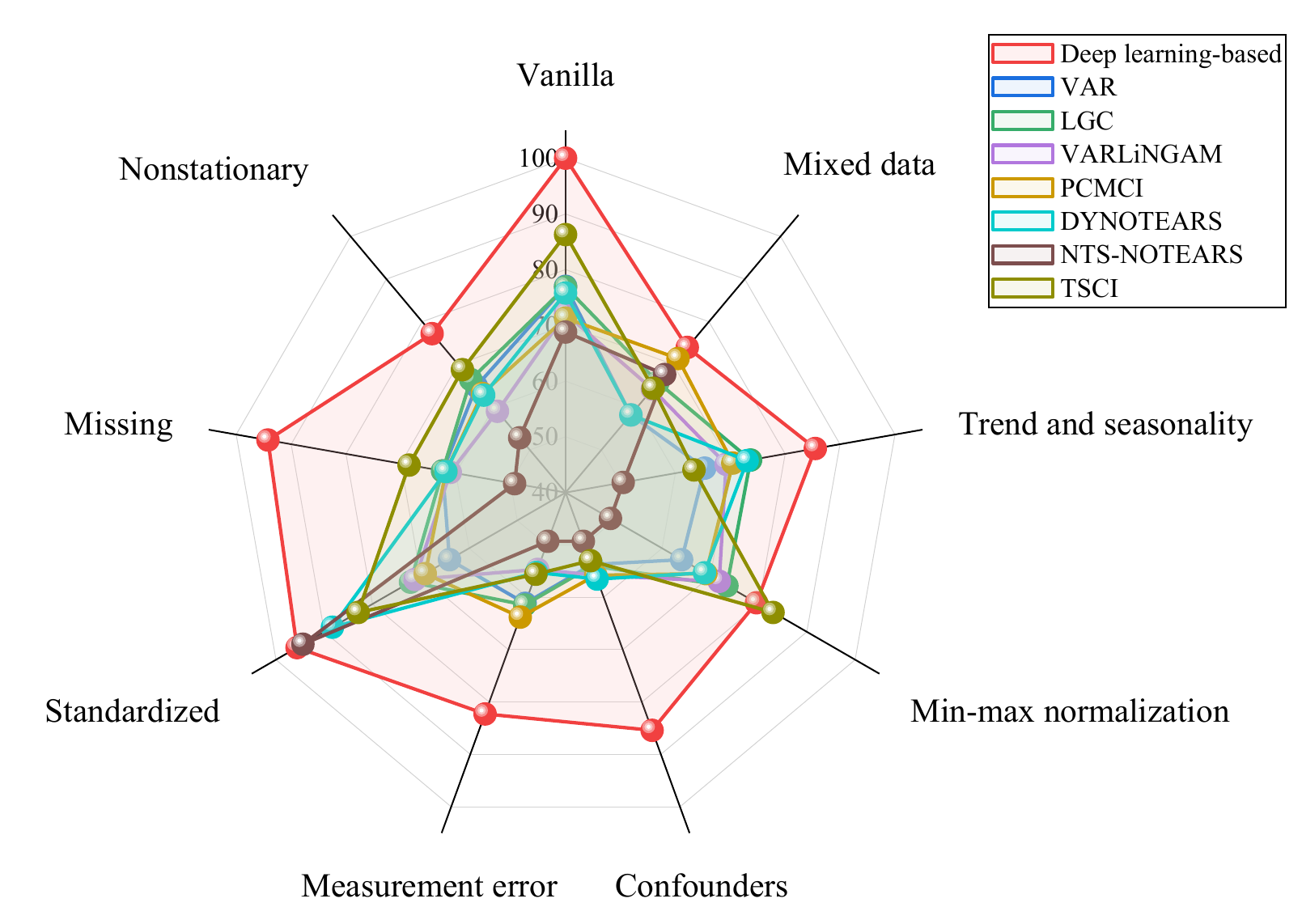}
        \caption{AUROC for nonlinear 10-node case with $T = 500$ and $F = 10$.}
         \label{fig:nonlinear_10_500_f10_auroc}
     \end{subfigure}%
     \hfill
     \begin{subfigure}[b]{0.49\textwidth}
         \centering
         \includegraphics[width=\textwidth]{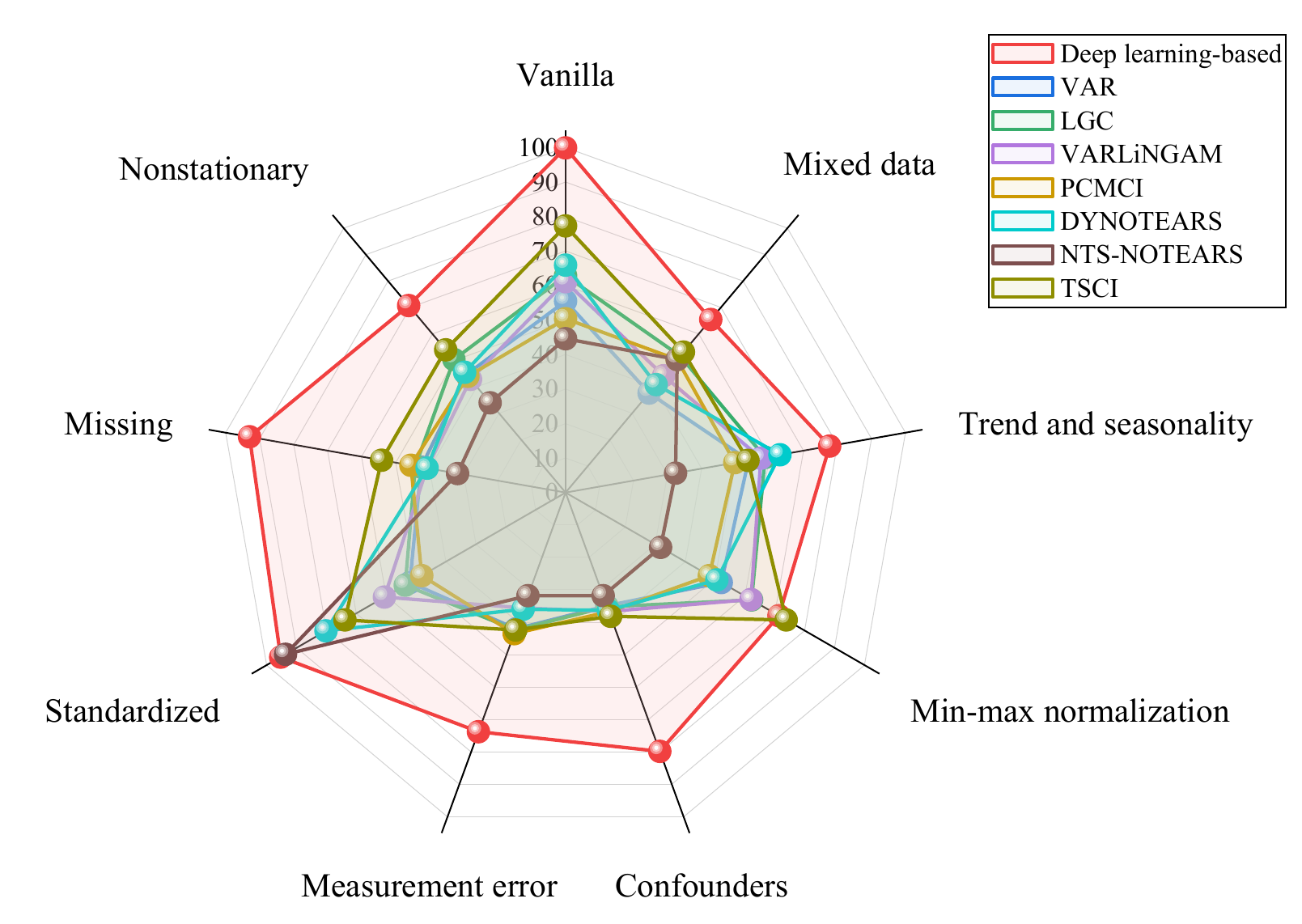}
         \caption{AUPRC for nonlinear 10-node case with $T = 500$ and $F = 10$.}
         \label{fig:nonlinear_10_500_f10_auprc}
     \end{subfigure}

\Description{Four radar charts comparing causal discovery methods on 10-node networks with T=500 across 9 scenarios. Superior performance is predominantly achieved by deep learning-based approaches.}
\caption{Experimental results under the linear and nonlinear settings across the vanilla scenario and eight assumption violation scenarios. AUROC and AUPRC (the higher the better) are evaluated over 5 trials for the 10-node case with $T = 500$. For the deep learning-based methods, we present only the optimal results.}
\label{fig:experiments_10_500_f10}
\end{figure*}

\clearpage
\begin{figure*}[t]
     \centering
     \begin{subfigure}[b]{0.49\textwidth}
         \centering
        \includegraphics[width=\textwidth]{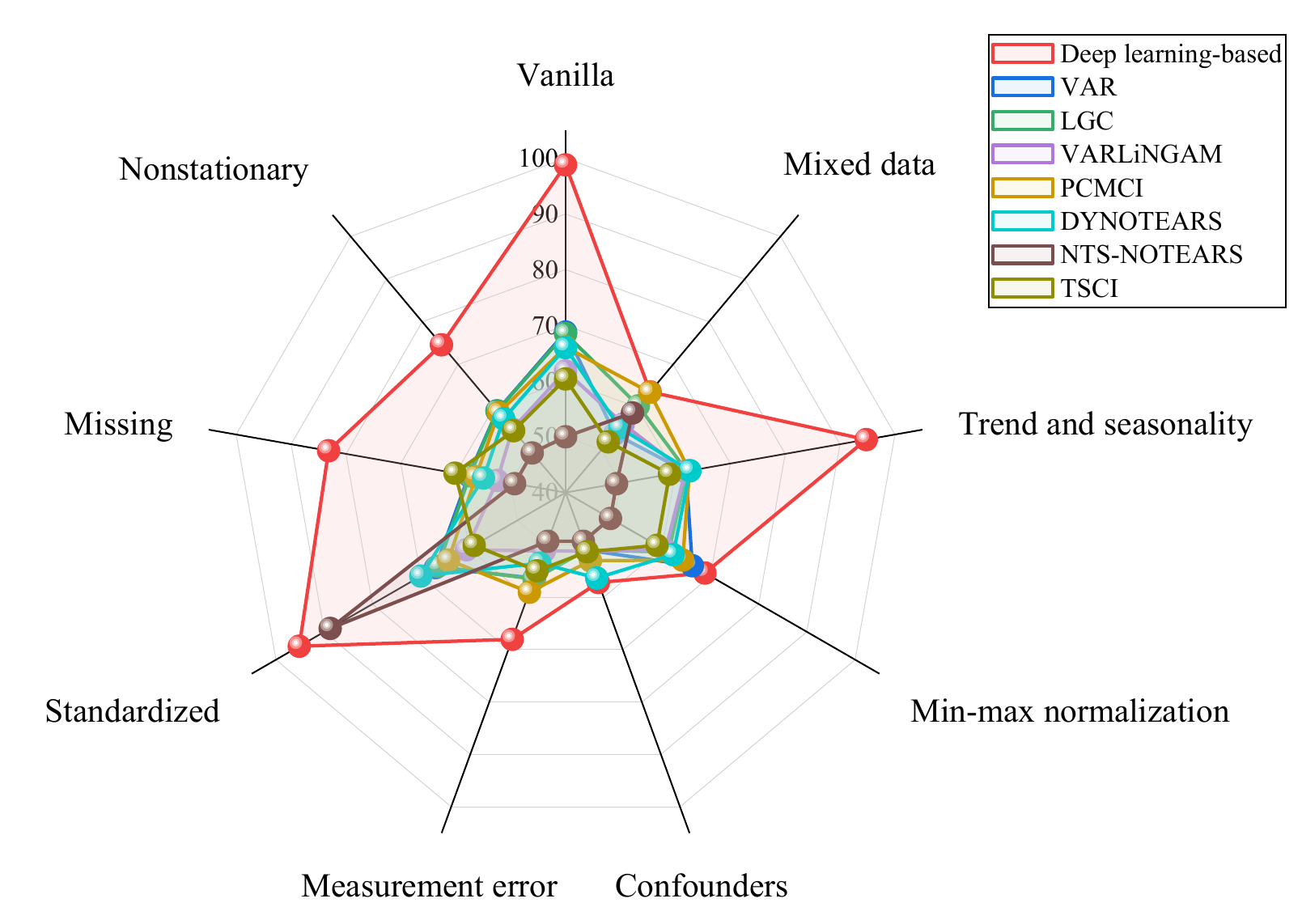}
         \caption{AUROC for nonlinear 10-node case with $T = 500$ and $F = 40$.}
         \label{fig:nonlinear_10_500_f40_auroc}
     \end{subfigure}%
     \hfill  
     \begin{subfigure}[b]{0.49\textwidth}
         \centering
         \includegraphics[width=\textwidth]{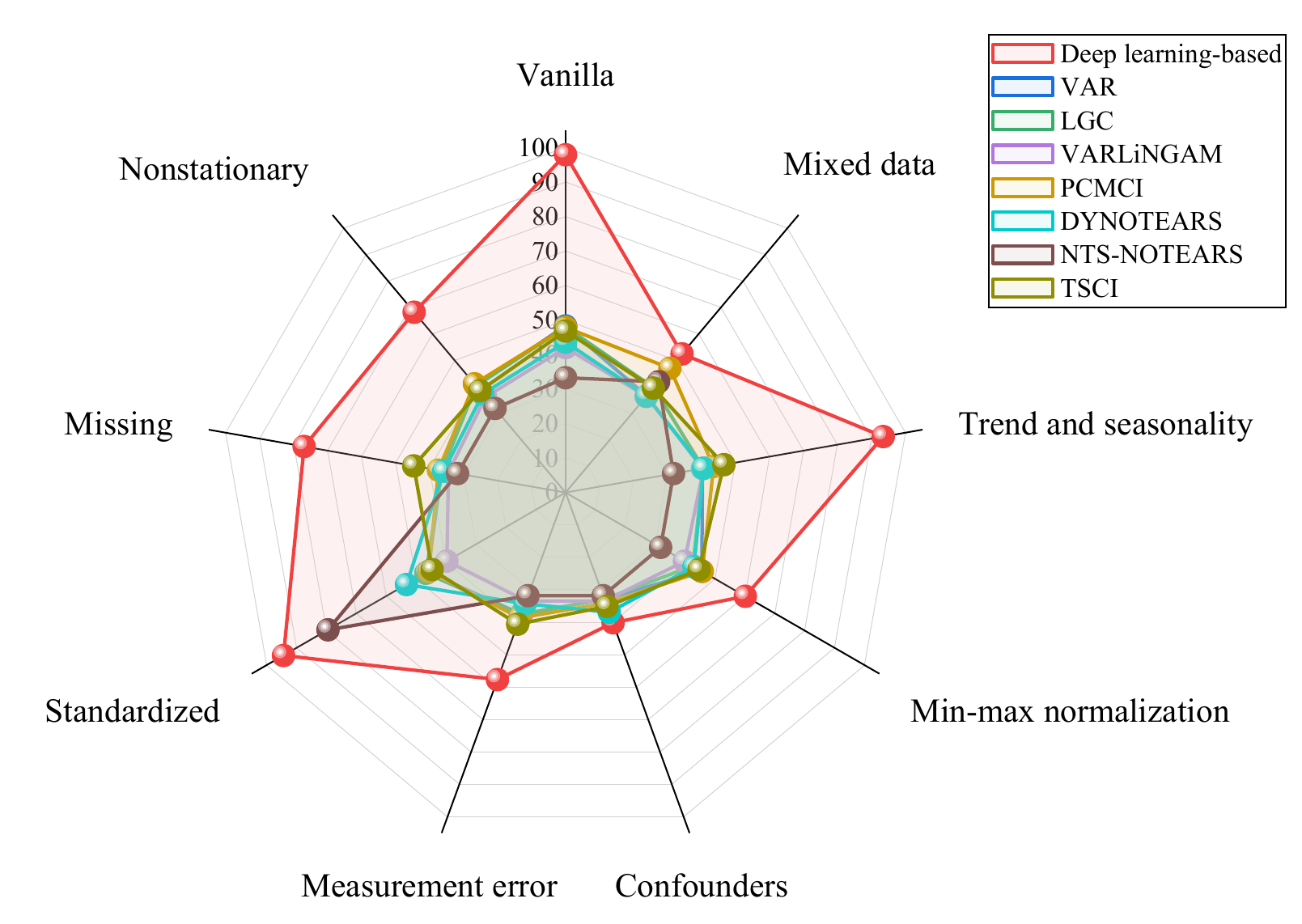}
         \caption{AUPRC for nonlinear 10-node case with $T = 500$ and $F = 40$.}
         \label{fig:nonlinear_10_500_f40_auprc}
     \end{subfigure}

     \medskip  

     \begin{subfigure}[b]{0.49\textwidth}
         \centering
        \includegraphics[width=\textwidth]{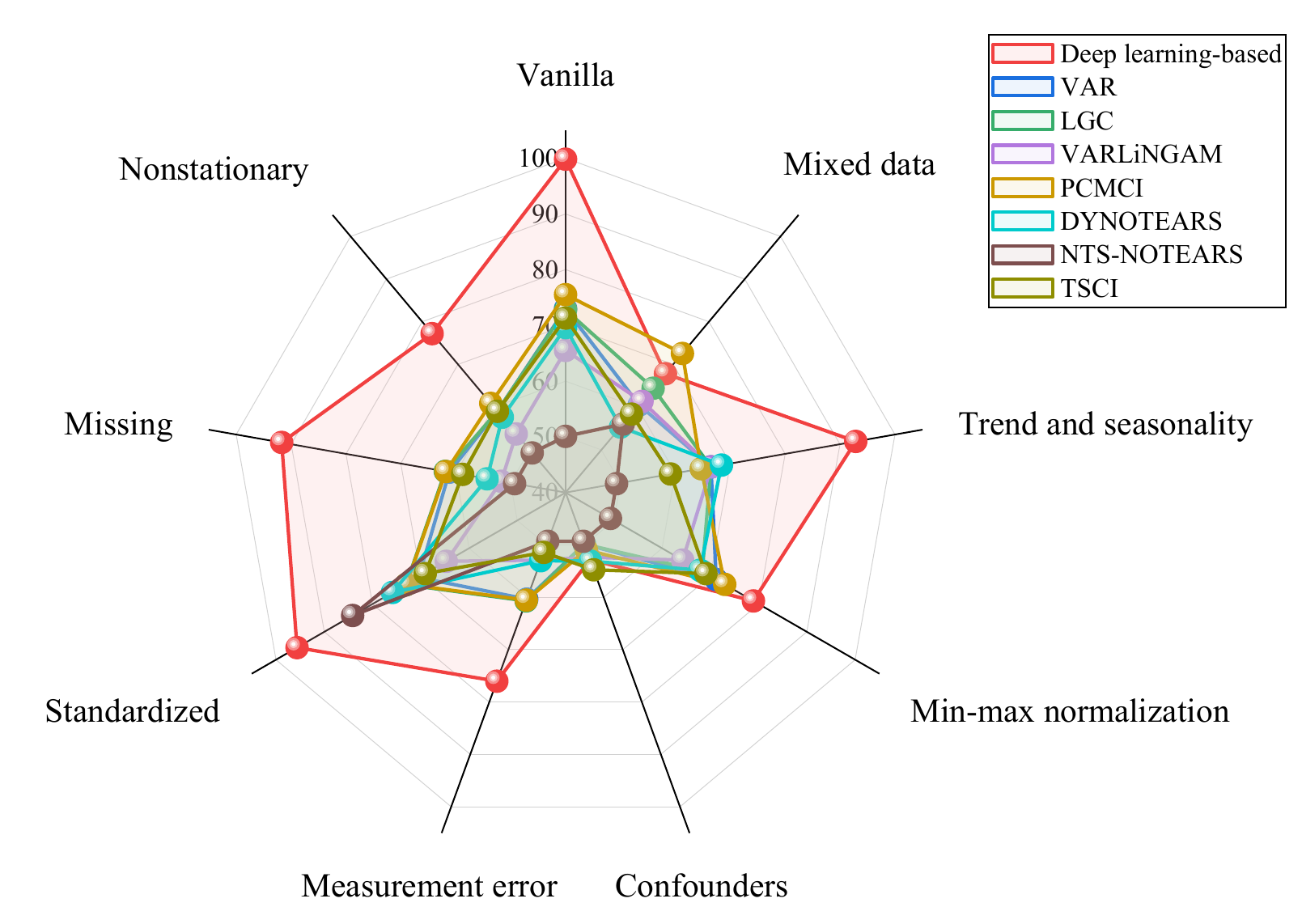}
        \caption{AUROC for nonlinear 10-node case with $T = 1000$ and $F = 40$.}
         \label{fig:nonlinear_10_1000_f40_auroc}
     \end{subfigure}%
     \hfill
     \begin{subfigure}[b]{0.49\textwidth}
         \centering
         \includegraphics[width=\textwidth]{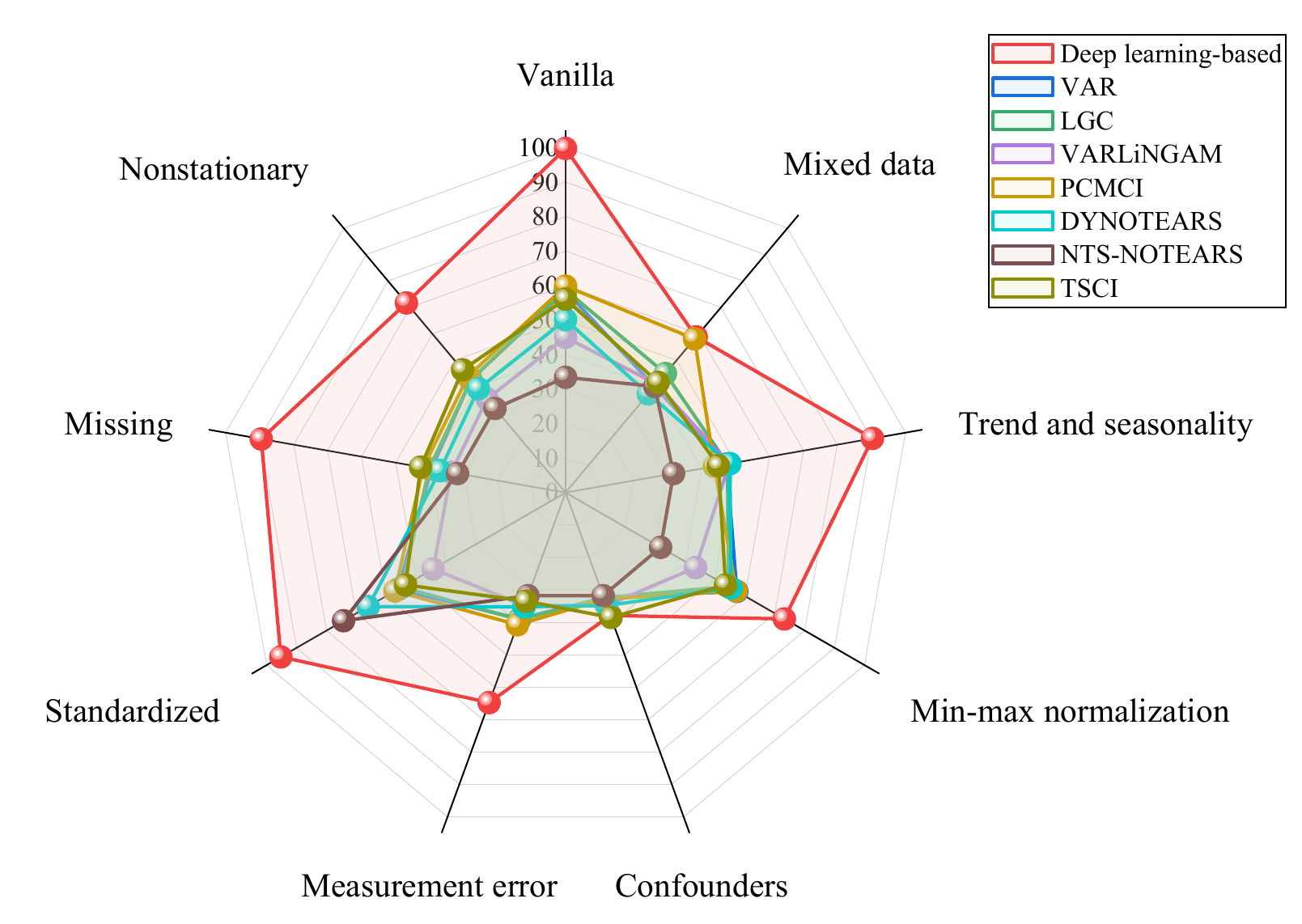}
         \caption{AUPRC for nonlinear 10-node case with $T = 1000$ and $F = 40$.}
         \label{fig:nonlinear_10_1000_f40_auprc}
     \end{subfigure}
     
\Description{Four radar charts comparing causal discovery methods on 10-node networks with F=40 across 9 scenarios. Superior performance is predominantly achieved by deep learning-based approaches.}     
\caption{Experimental results under the nonlinear settings across the vanilla scenario and eight assumption violation scenarios. AUROC and AUPRC (the higher the better) are evaluated over 5 trials for the 10-node case with $F = 40$. For the deep learning-based methods, we present only the optimal results.}
\label{fig:experiments_10_500_1000_f40}
\end{figure*}

\clearpage
\begin{figure*}[t]
     \centering
     \begin{subfigure}[b]{0.49\textwidth}
         \centering
        \includegraphics[width=\textwidth]{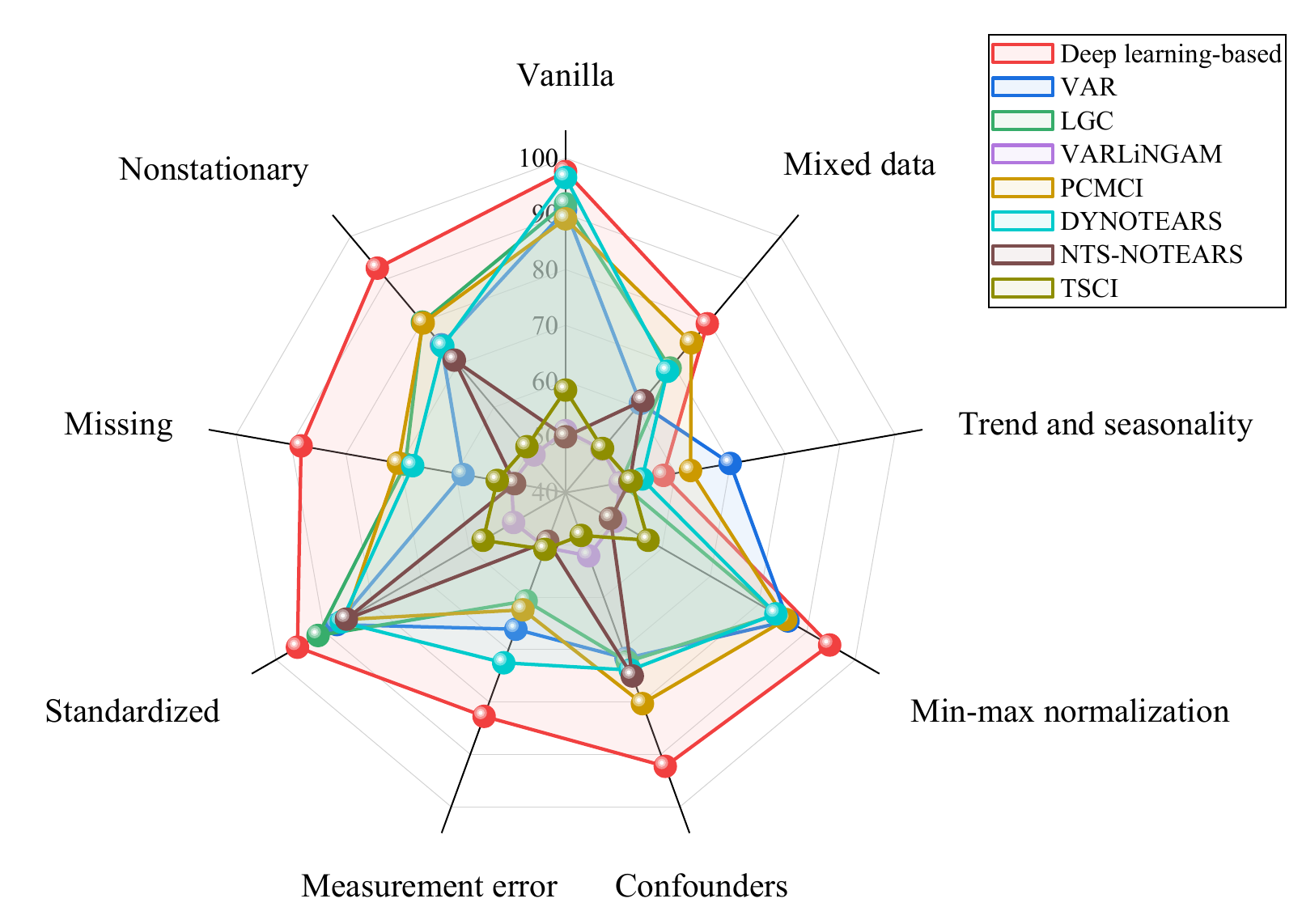}
         \caption{AUROC for linear 15-node case with $T = 500$.}
         \label{fig:linear_15_500_auroc}
     \end{subfigure}%
     \hfill  
     \begin{subfigure}[b]{0.49\textwidth}
         \centering
         \includegraphics[width=\textwidth]{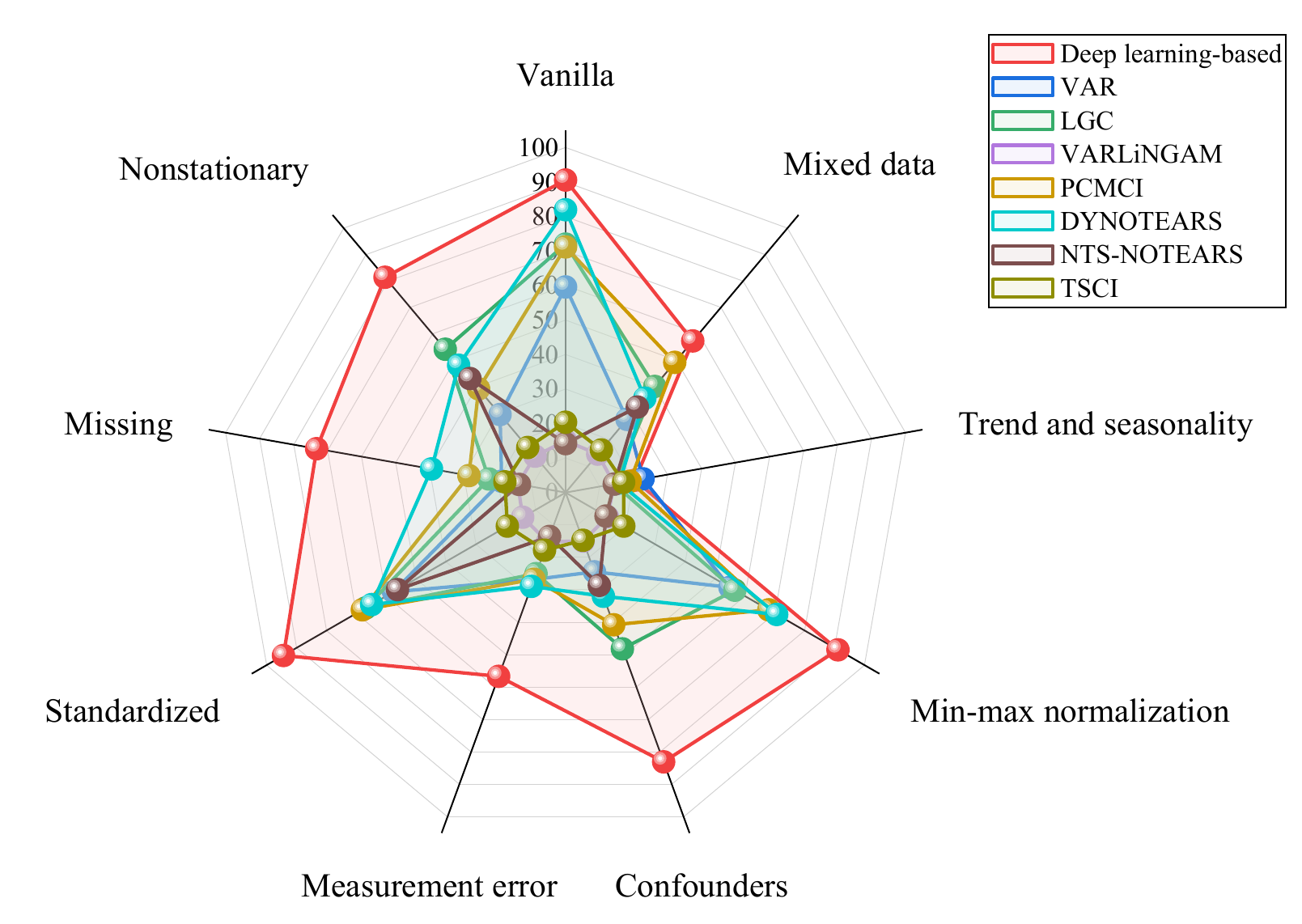}
         \caption{AUPRC for linear 15-node case with $T = 500$.}
         \label{fig:linear_15_500_auprc}
     \end{subfigure}

     \medskip  

     \begin{subfigure}[b]{0.49\textwidth}
         \centering
        \includegraphics[width=\textwidth]{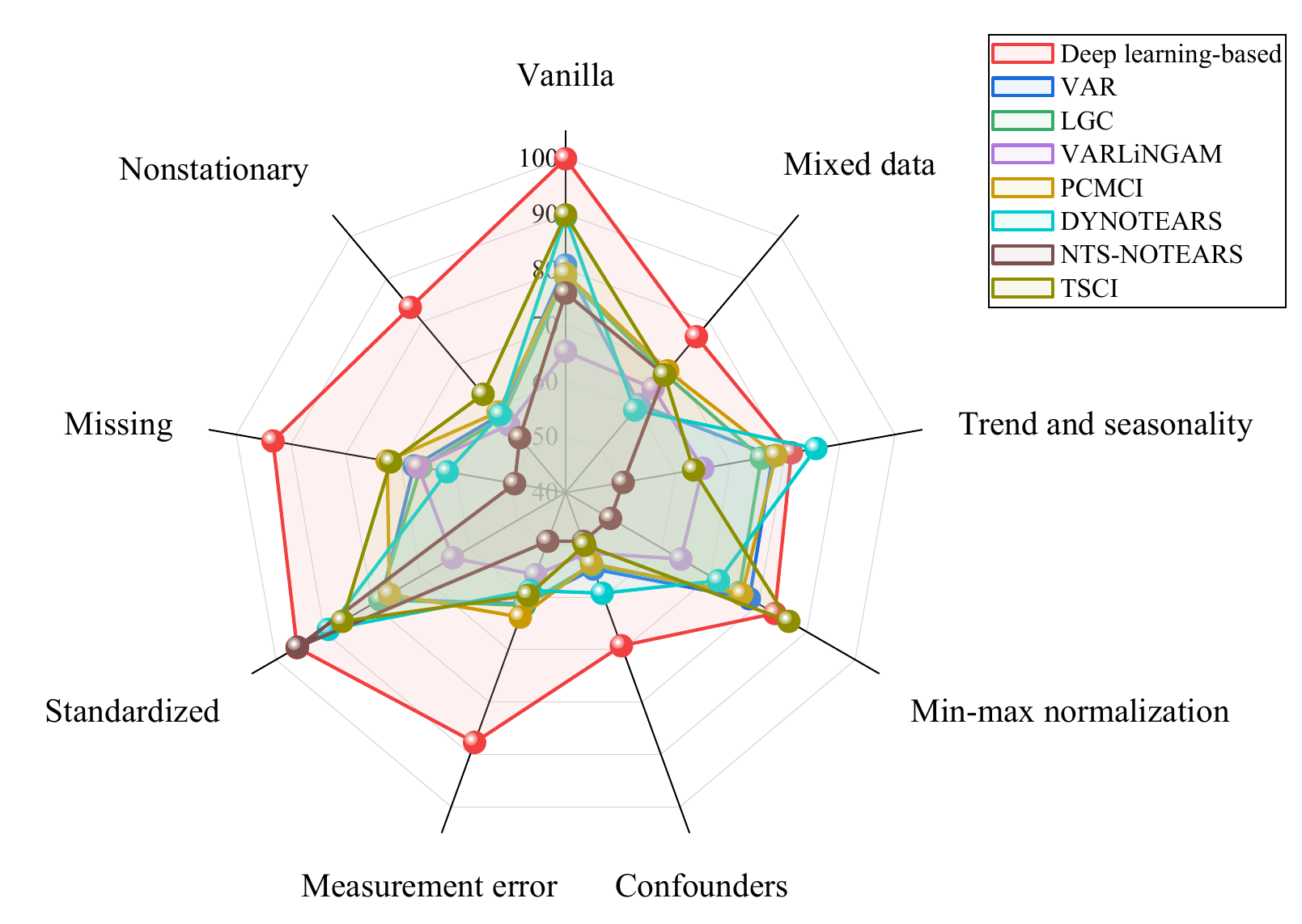}
        \caption{AUROC for nonlinear 15-node case with $T = 500$ and $F = 10$.}
         \label{fig:nonlinear_15_500_f10_auroc}
     \end{subfigure}%
     \hfill
     \begin{subfigure}[b]{0.49\textwidth}
         \centering
         \includegraphics[width=\textwidth]{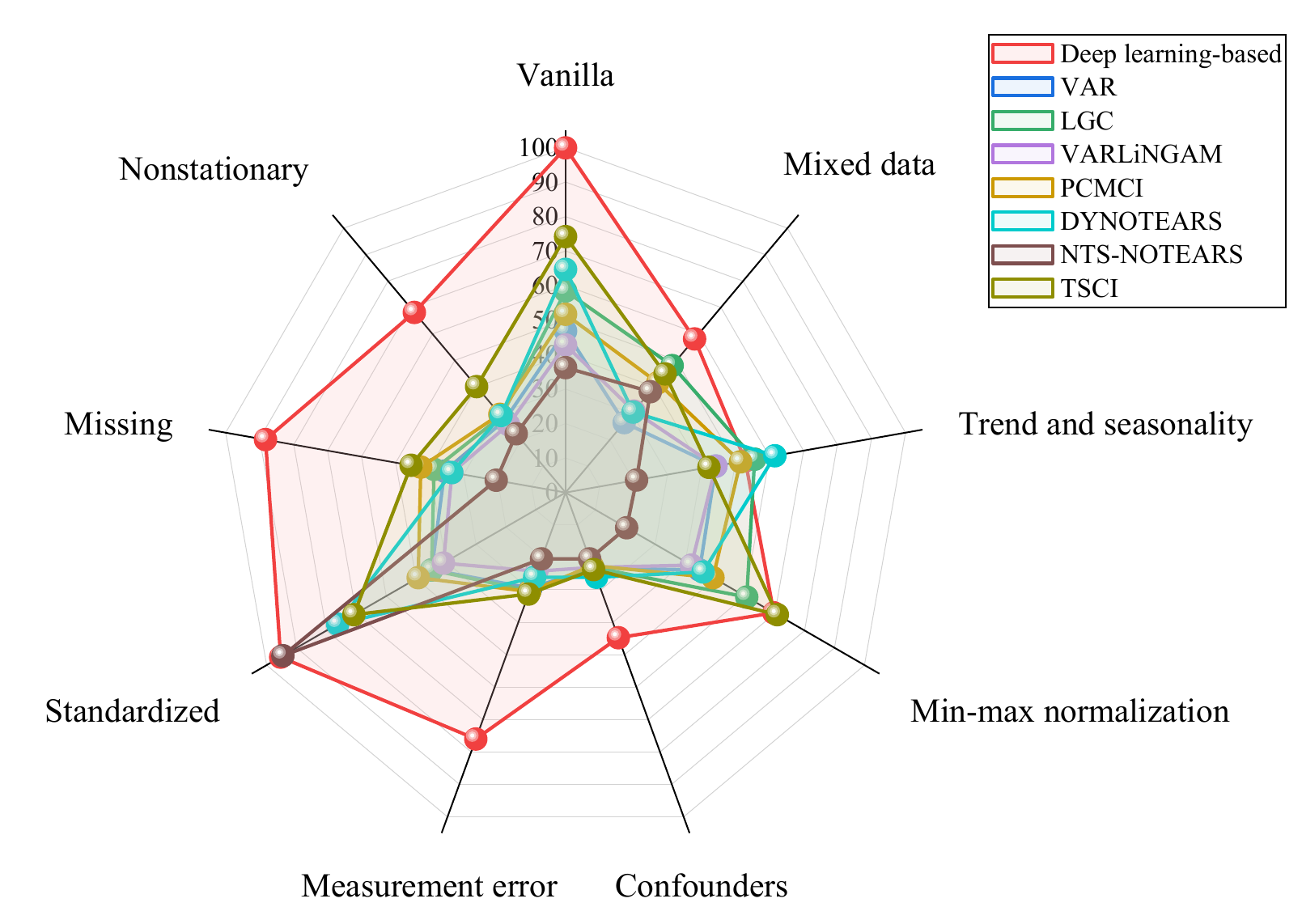}
         \caption{AUPRC for nonlinear 15-node case with $T = 500$ and $F = 10$.}
         \label{fig:nonlinear_15_500_f10_auprc}
     \end{subfigure}

\Description{Four radar charts comparing causal discovery methods on 15-node networks with T=500 across 9 scenarios. Superior performance is predominantly achieved by deep learning-based approaches.}
\caption{Experimental results under the linear and nonlinear settings across the vanilla scenario and eight assumption violation scenarios. AUROC and AUPRC (the higher the better) are evaluated over 5 trials for the 15-node case with $T = 500$. For the deep learning-based methods, we present only the optimal results.}
\label{fig:experiments_15_500_f10}
\end{figure*}

\clearpage
\begin{figure*}[t]
     \centering
     \begin{subfigure}[b]{0.49\textwidth}
         \centering
        \includegraphics[width=\textwidth]{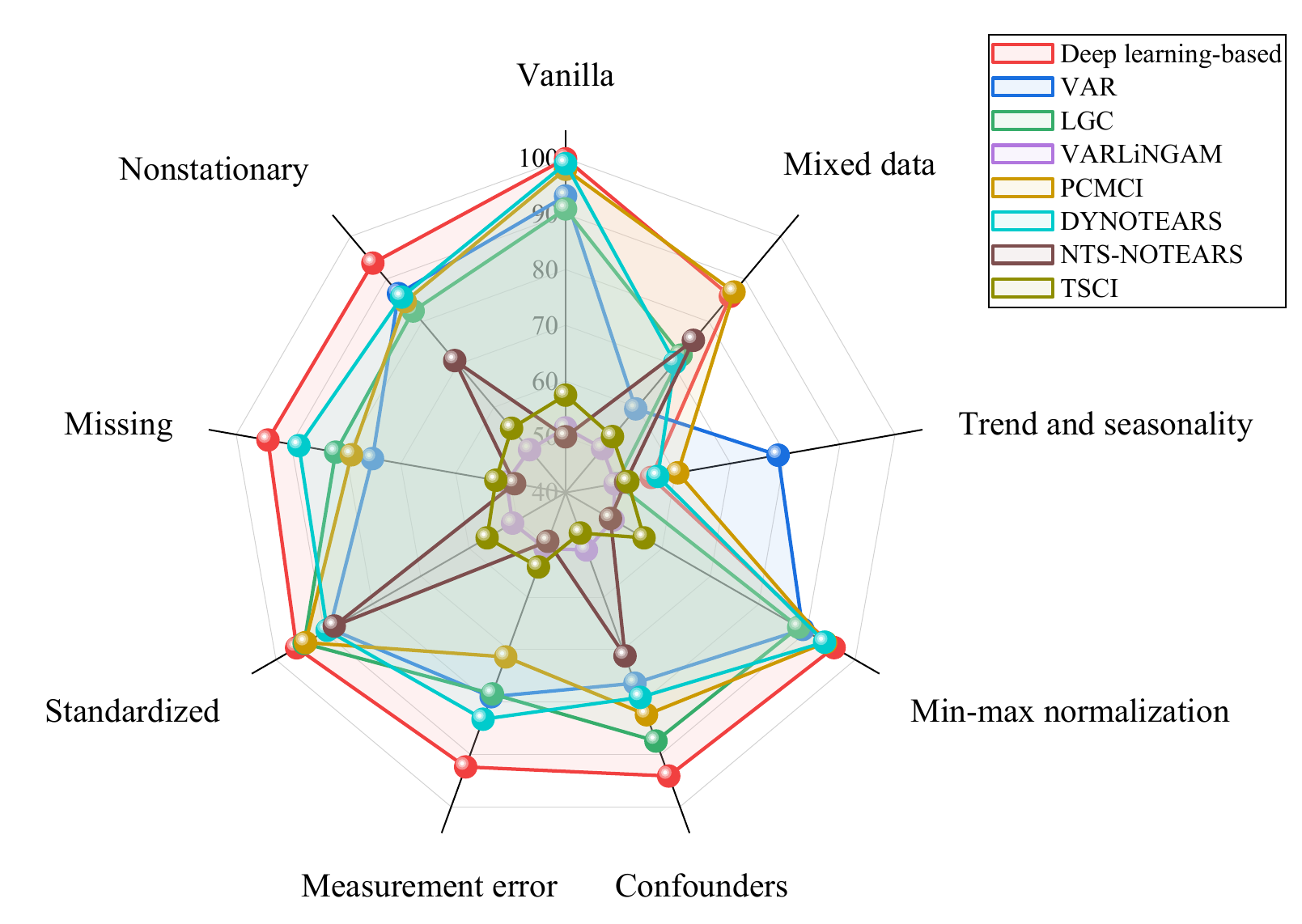}
         \caption{AUROC for linear 15-node case with $T = 1000$.}
         \label{fig:linear_15_1000_auroc}
     \end{subfigure}%
     \hfill  
     \begin{subfigure}[b]{0.49\textwidth}
         \centering
         \includegraphics[width=\textwidth]{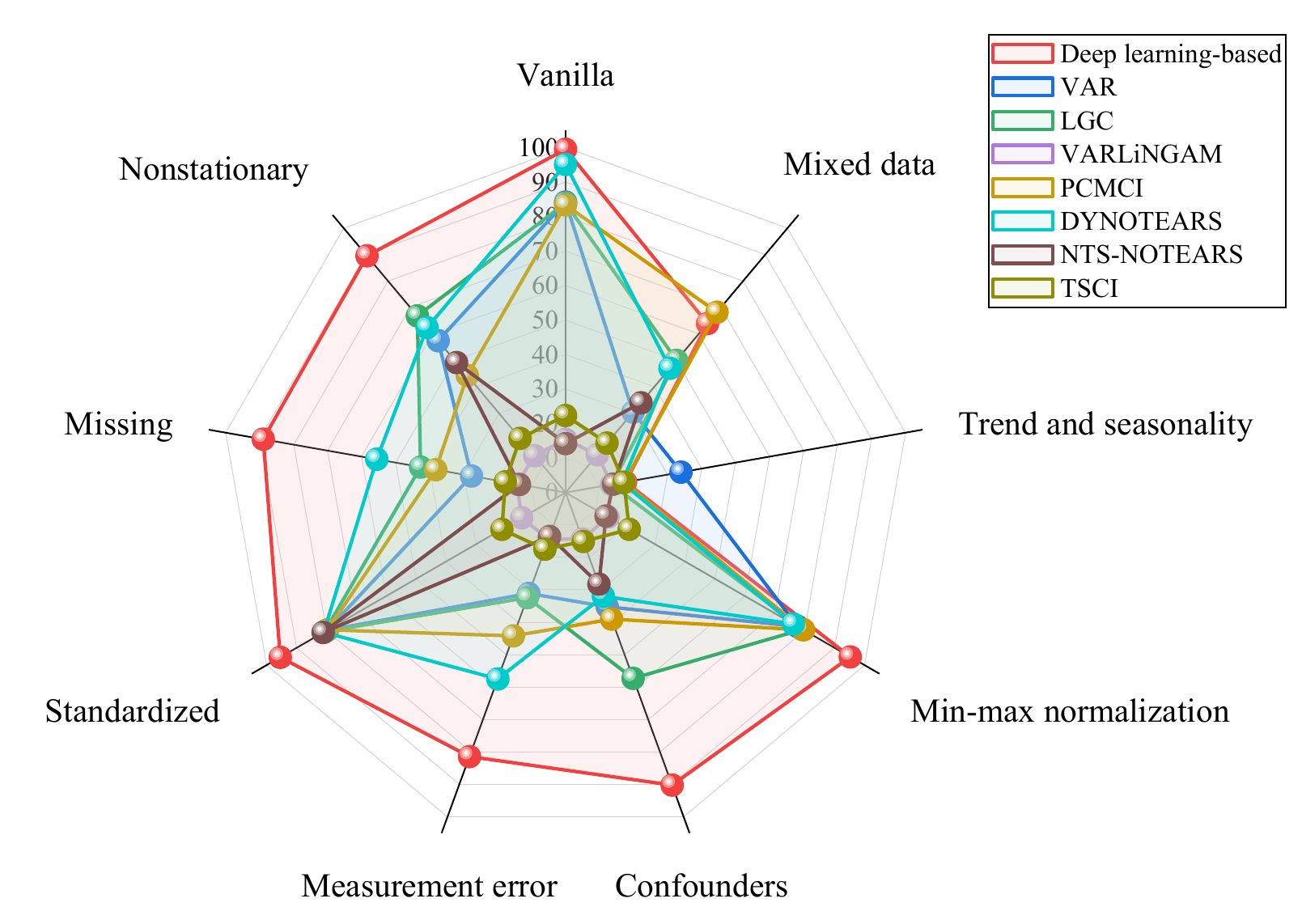}
         \caption{AUPRC for linear 15-node case with $T = 1000$.}
         \label{fig:linear_15_1000_auprc}
     \end{subfigure}

     \medskip  

     \begin{subfigure}[b]{0.49\textwidth}
         \centering
        \includegraphics[width=\textwidth]{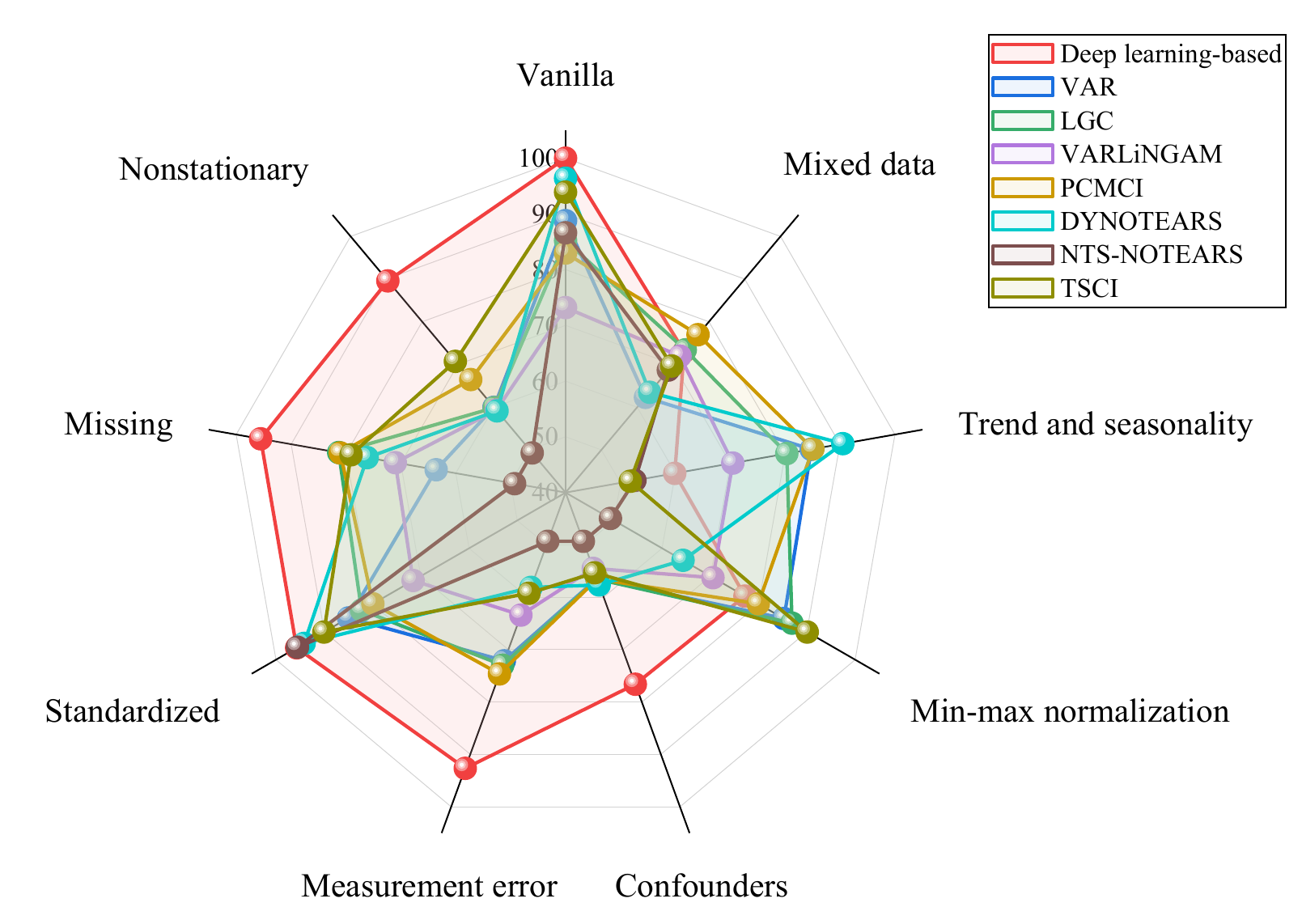}
        \caption{AUROC for nonlinear 15-node case with $T = 1000$ and $F = 10$.}
         \label{fig:nonlinear_15_1000_f10_auroc}
     \end{subfigure}%
     \hfill
     \begin{subfigure}[b]{0.49\textwidth}
         \centering
         \includegraphics[width=\textwidth]{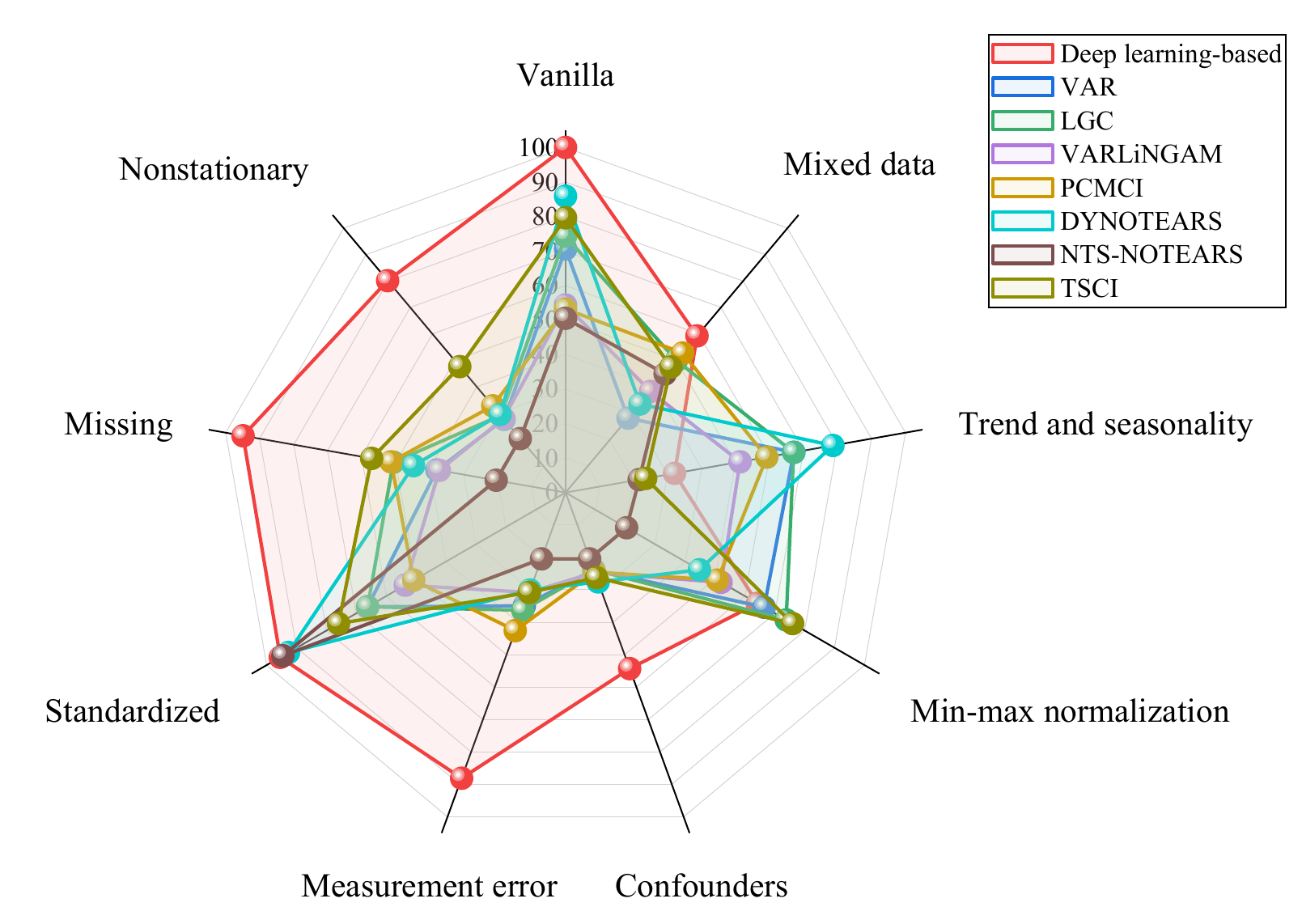}
         \caption{AUPRC for nonlinear 15-node case with $T = 1000$ and $F = 10$.}
         \label{fig:nonlinear_15_1000_f10_auprc}
     \end{subfigure}

\Description{Four radar charts comparing causal discovery methods on 15-node networks with T=1000 across 9 scenarios. Superior performance is predominantly achieved by deep learning-based approaches.}
\caption{Experimental results under the linear and nonlinear settings across the vanilla scenario and eight assumption violation scenarios. AUROC and AUPRC (the higher the better) are evaluated over 5 trials for the 15-node case with $T = 1000$. For the deep learning-based methods, we present only the optimal results.}
\label{fig:experiments_15_1000_f10}
\end{figure*}

\clearpage
\begin{figure*}[t]
     \centering
     \begin{subfigure}[b]{0.49\textwidth}
         \centering
        \includegraphics[width=\textwidth]{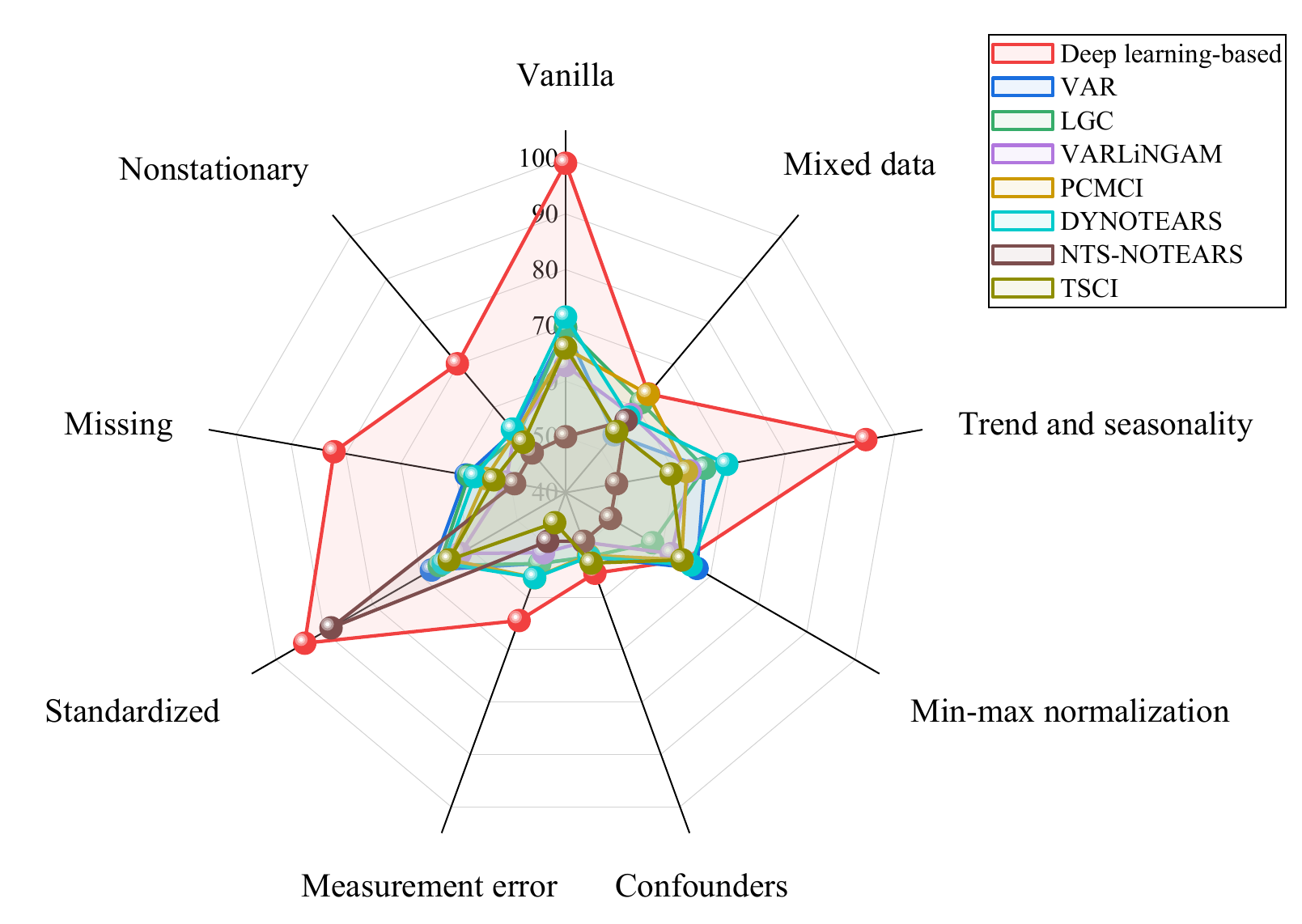}
         \caption{AUROC for nonlinear 15-node case with $T = 500$ and $F = 40$.}
         \label{fig:nonlinear_15_500_f40_auroc}
     \end{subfigure}%
     \hfill  
     \begin{subfigure}[b]{0.49\textwidth}
         \centering
         \includegraphics[width=\textwidth]{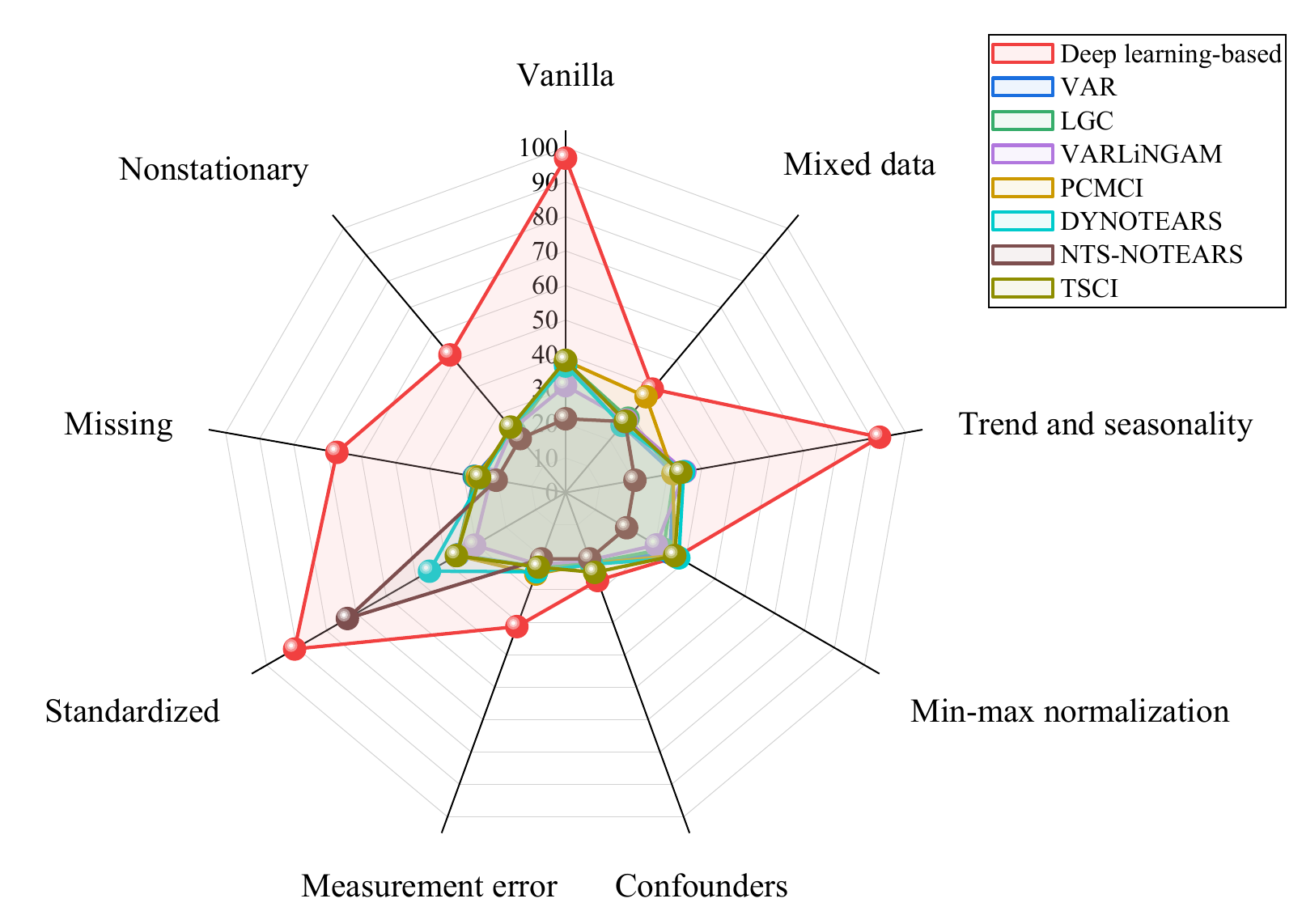}
         \caption{AUPRC for nonlinear 15-node case with $T = 500$ and $F = 40$.}
         \label{fig:nonlinear_15_500_f40_auprc}
     \end{subfigure}

     \medskip  

     \begin{subfigure}[b]{0.49\textwidth}
         \centering
        \includegraphics[width=\textwidth]{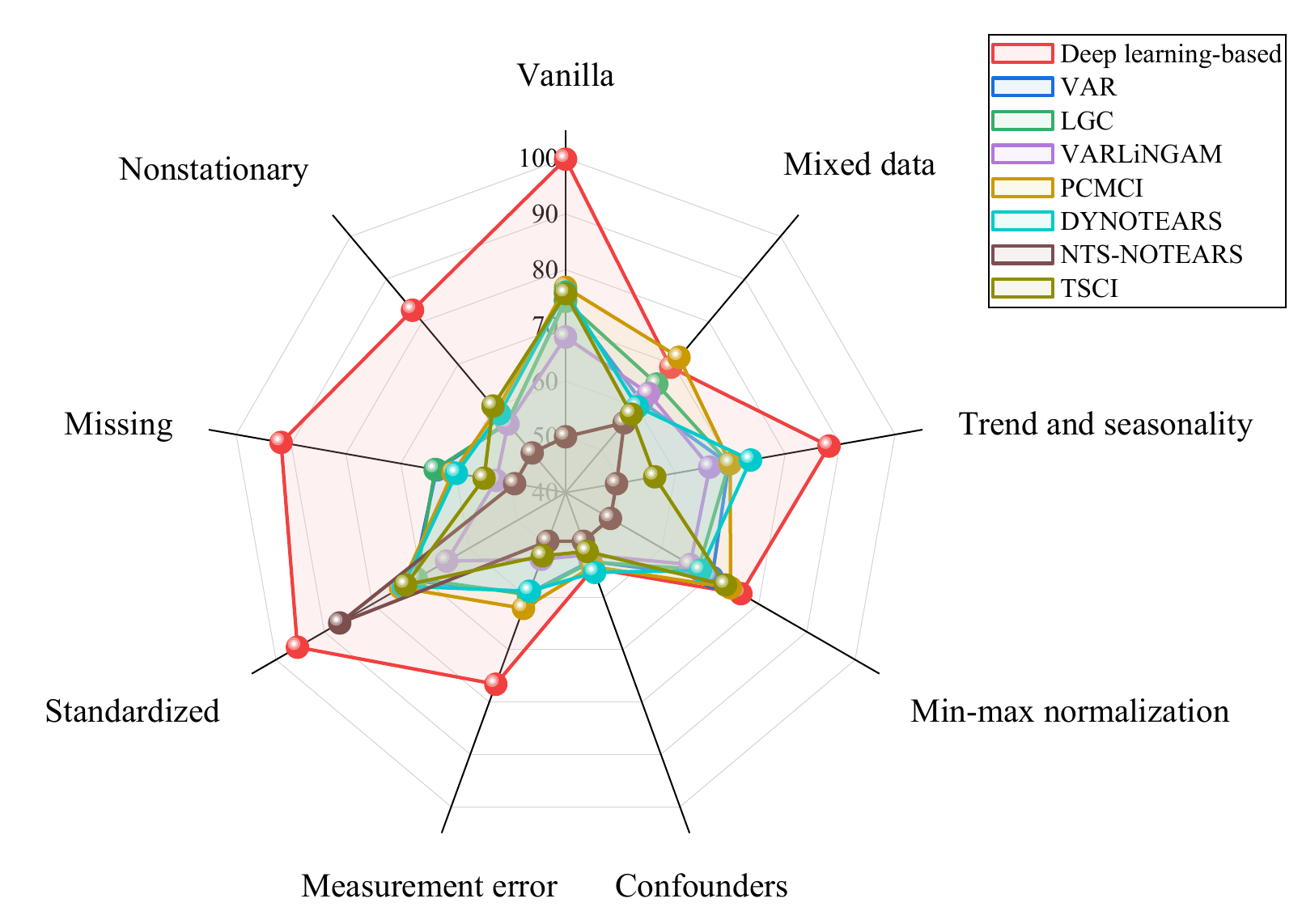}
        \caption{AUROC for nonlinear 15-node case with $T = 1000$ and $F = 40$.}
         \label{fig:nonlinear_15_1000_f40_auroc}
     \end{subfigure}%
     \hfill
     \begin{subfigure}[b]{0.49\textwidth}
         \centering
         \includegraphics[width=\textwidth]{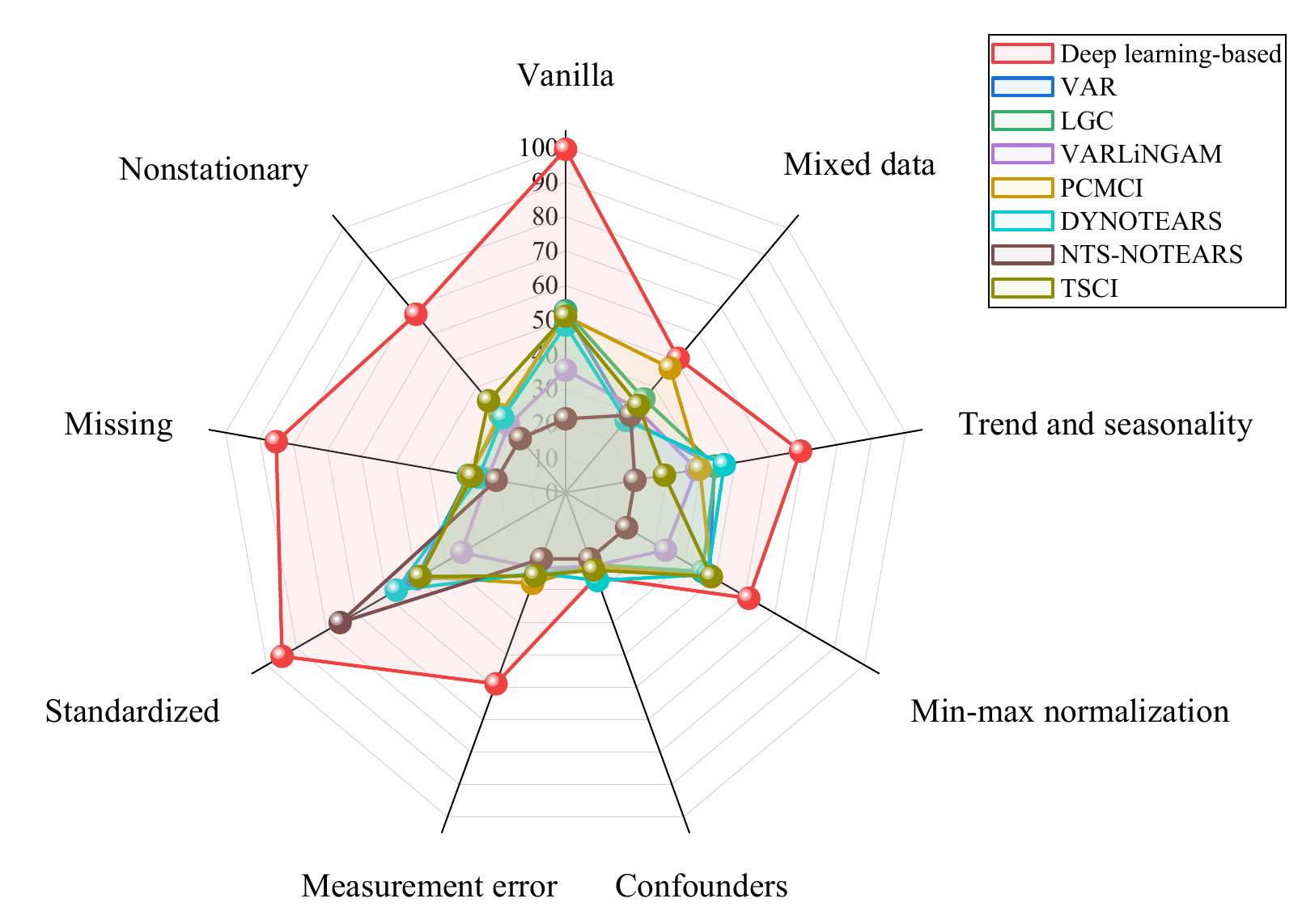}
         \caption{AUPRC for nonlinear 15-node case with $T = 1000$ and $F = 40$.}
         \label{fig:nonlinear_15_1000_f40_auprc}
     \end{subfigure}
     
\Description{Four radar charts comparing causal discovery methods on 15-node networks with F=40 across 9 scenarios. Superior performance is predominantly achieved by deep learning-based approaches.}     
\caption{Experimental results under the nonlinear settings across the vanilla scenario and eight assumption violation scenarios. AUROC and AUPRC (the higher the better) are evaluated over 5 trials for the 15-node case with $F = 40$. For the deep learning-based methods, we present only the optimal results.}
\label{fig:experiments_15_500_1000_f40}
\end{figure*}

\clearpage
\begin{figure*}[t]
     \centering
     \begin{subfigure}[b]{0.49\textwidth}
         \centering
        \includegraphics[width=\textwidth]{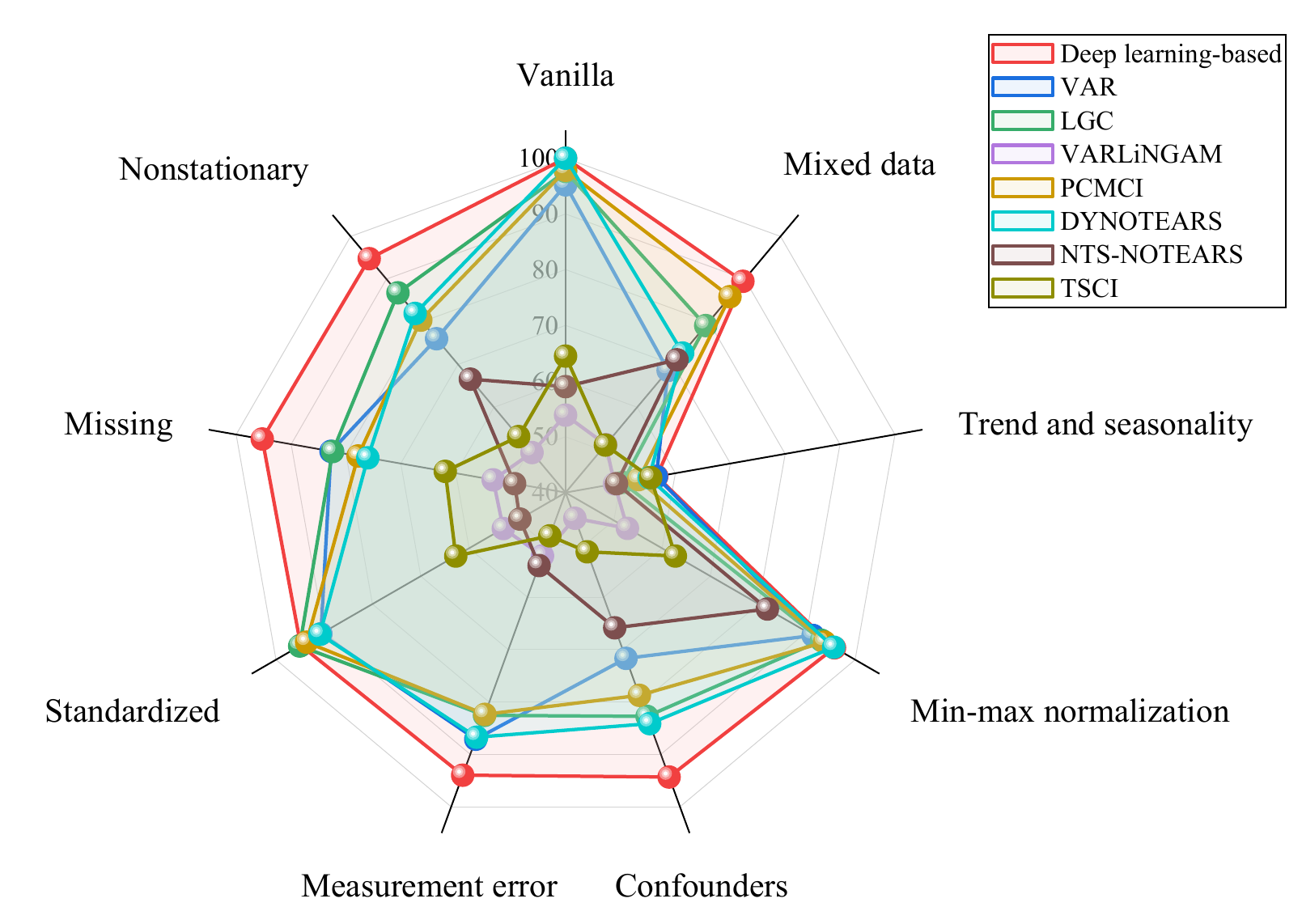}
         \caption{AUROC for linear 10-node case with $T = 500$. Hyperparameters are selected to maximize average performance across all scenarios.}
         \label{fig:linear_10_500_auroc_avg_scen}
     \end{subfigure}%
     \hfill
     \begin{subfigure}[b]{0.49\textwidth}
         \centering
         \includegraphics[width=\textwidth]{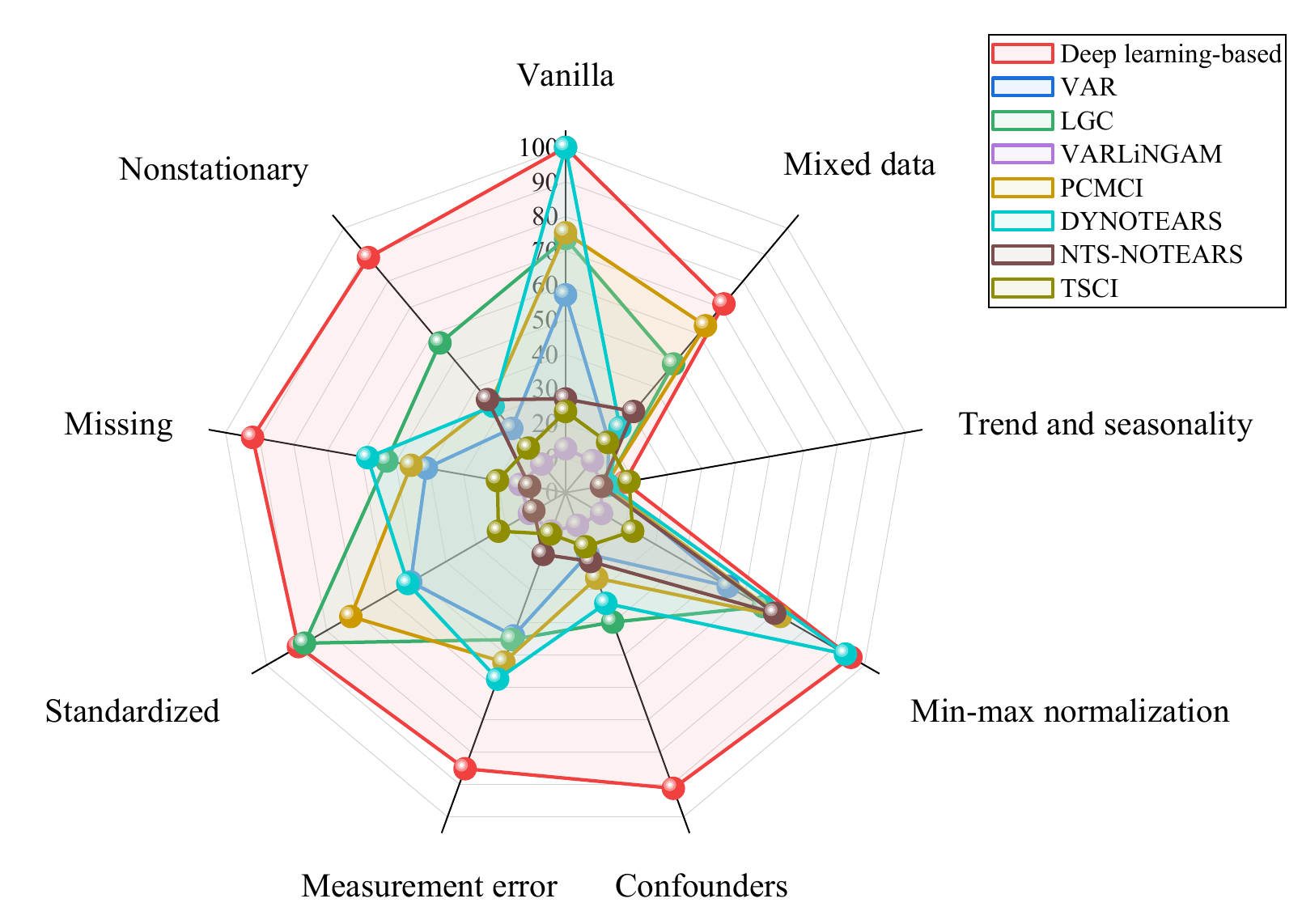}
         \caption{AUPRC for linear 10-node case with $T = 500$. Hyperparameters are selected to maximize average performance across all scenarios.}
         \label{fig:linear_10_500_auprc_avg_scen}
     \end{subfigure}

     \medskip  

     \begin{subfigure}[b]{0.49\textwidth}
         \centering
        \includegraphics[width=\textwidth]{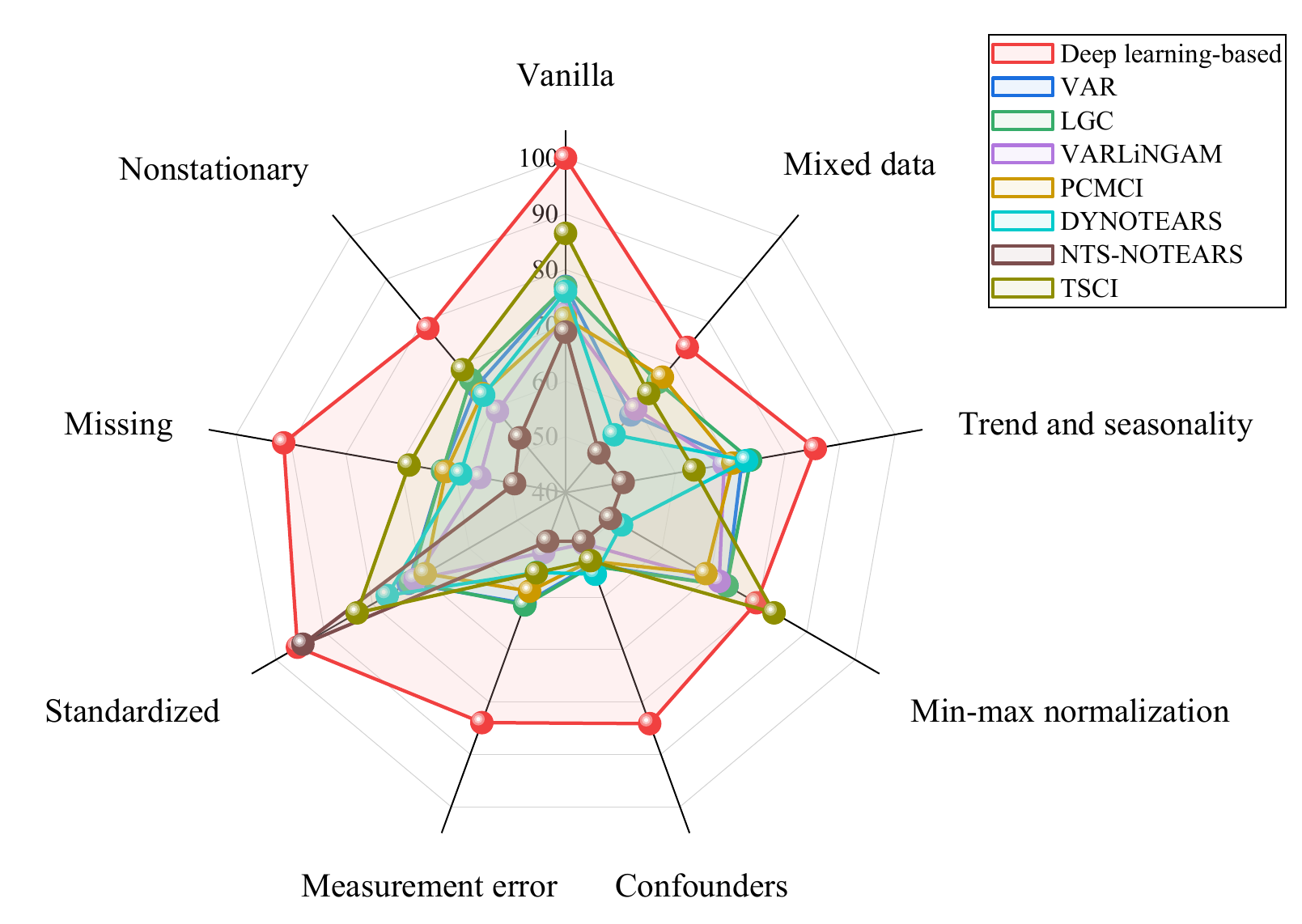}
         \caption{AUROC for nonlinear 10-node case with $T = 500$ and $F=10$. Hyperparameters are selected to maximize average performance across all scenarios.}
         \label{fig:nonlinear_10_500_f10_auroc_avg_scen}
     \end{subfigure}%
     \hfill
     \begin{subfigure}[b]{0.49\textwidth}
         \centering
         \includegraphics[width=\textwidth]{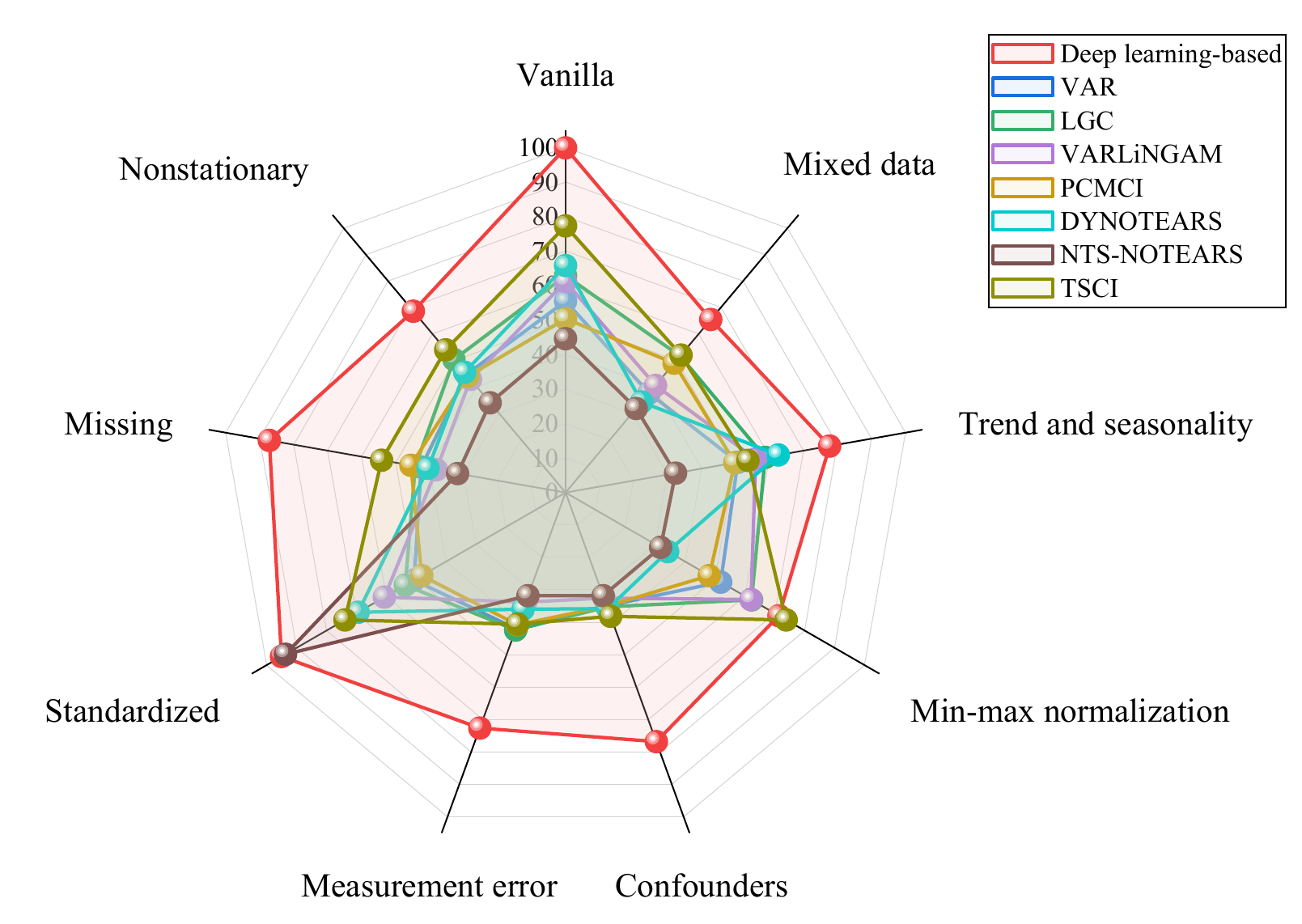}
         \caption{AUPRC for nonlinear 10-node case with $T = 500$ and $F=10$. Hyperparameters are selected to maximize average performance across all scenarios.}
         \label{fig:nonlinear_10_500_f10_auprc_avg_scen}
     \end{subfigure}

\Description{Four radar charts comparing causal discovery methods on 10-node networks with T=500 across 9 scenarios. Superior performance is predominantly achieved by deep learning-based approaches. Hyperparameters that achieve the best average performance across all scenarios are used.}
\caption{Experimental results under the linear and nonlinear settings across the vanilla scenario and eight assumption violation scenarios. AUROC and AUPRC (the higher the better) are evaluated over 5 trials for the 10-node case with $T = 500$. For the deep learning-based methods, we present only the optimal results. Hyperparameters are selected to maximize average performance across all scenarios.}
\label{fig:experiments_10_500_f10_avg_scen}
\end{figure*}

\clearpage

\clearpage

\clearpage

\clearpage

\clearpage

\clearpage
\begin{figure*}[t]
     \centering
     \begin{subfigure}[b]{0.49\textwidth}
         \centering
        \includegraphics[width=\textwidth]{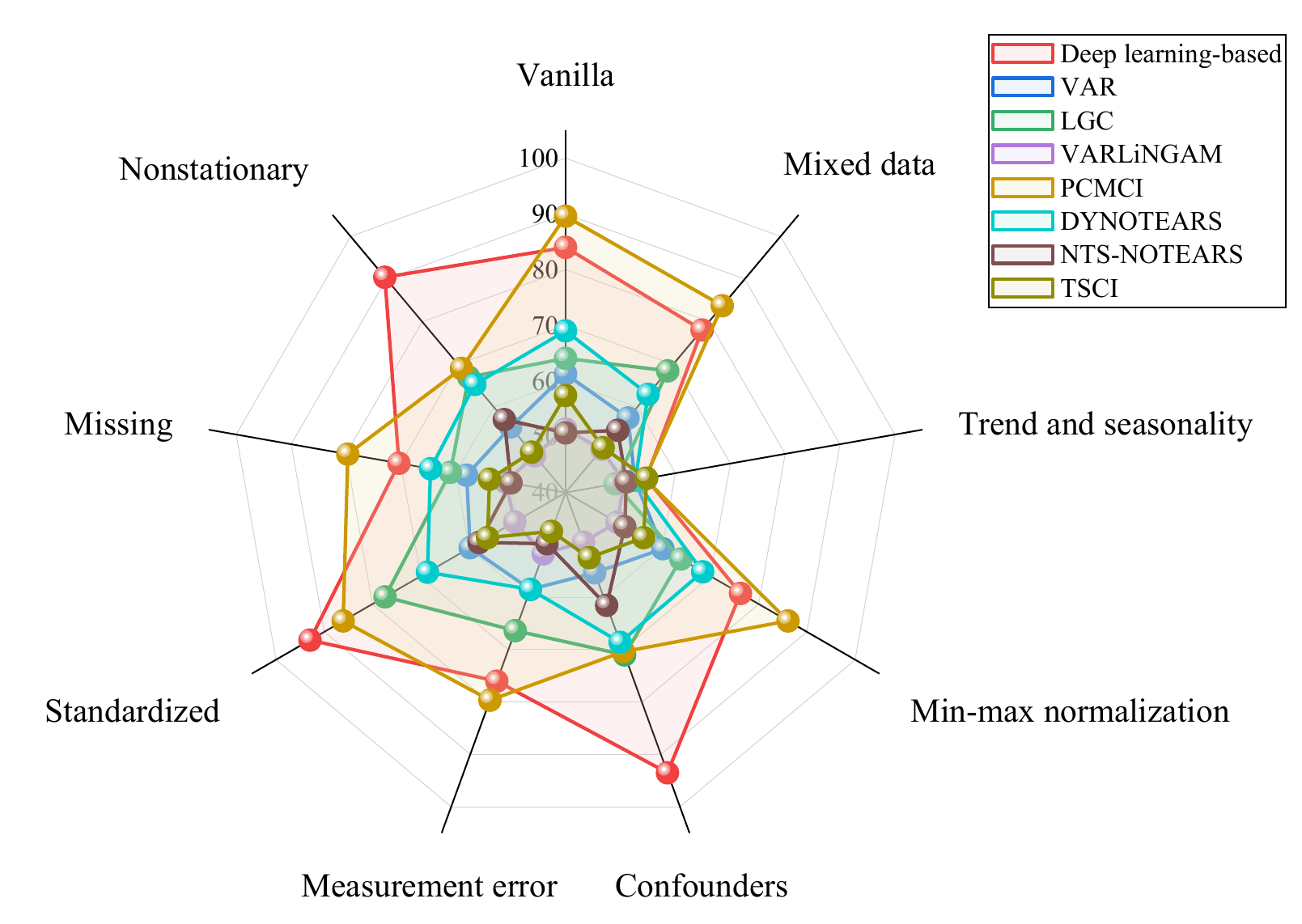}
         \caption{AUROC for linear 10-node case with $T = 500$. Results aggregated over all hyperparameters.}
         \label{fig:linear_10_500_auroc_avg_hyper}
     \end{subfigure}%
     \hfill
     \begin{subfigure}[b]{0.49\textwidth}
         \centering
         \includegraphics[width=\textwidth]{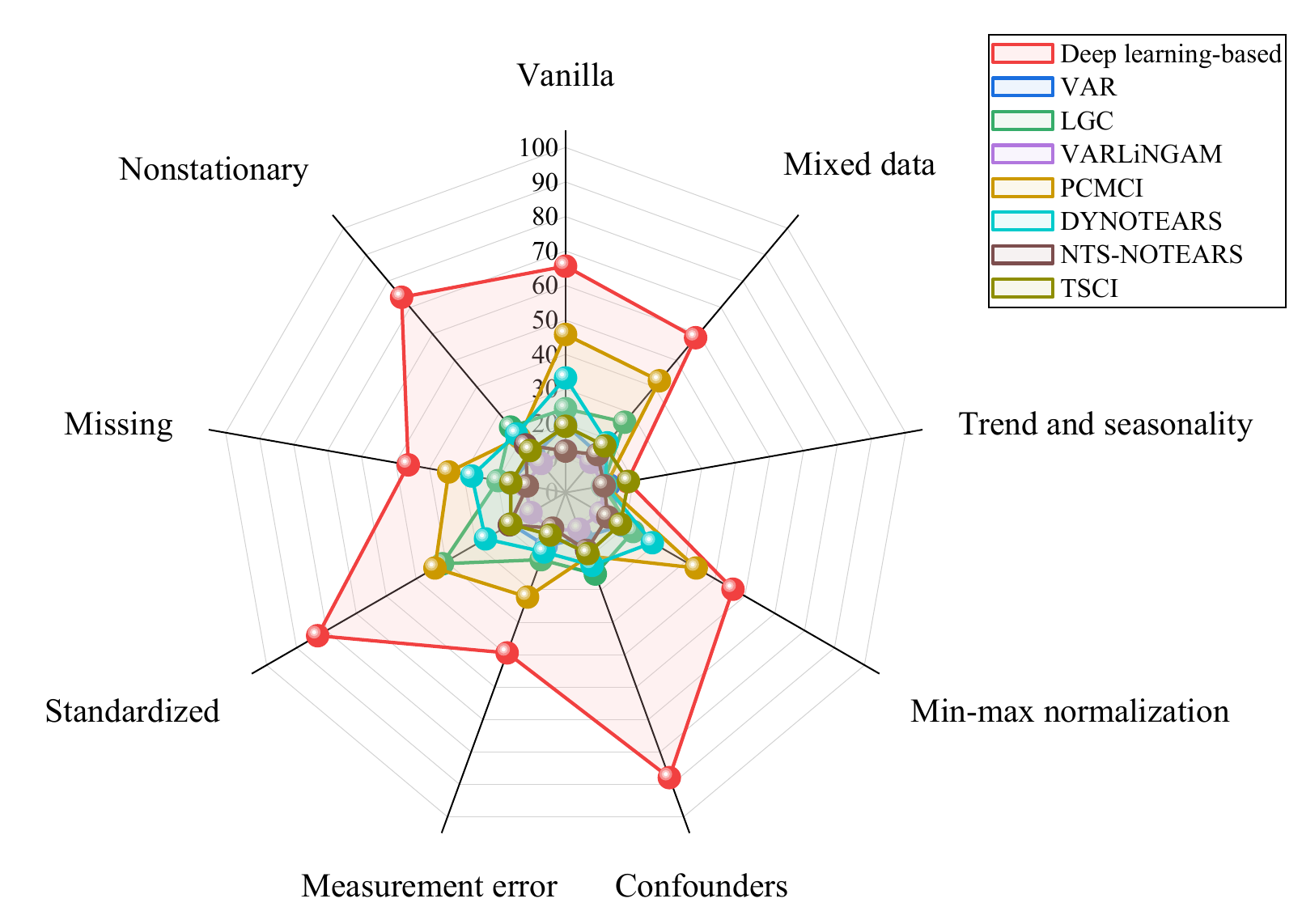}
         \caption{AUPRC for linear 10-node case with $T = 500$. Results aggregated over all hyperparameters.}
         \label{fig:linear_10_500_auprc_avg_hyper}
     \end{subfigure}

     \medskip  

     \begin{subfigure}[b]{0.49\textwidth}
         \centering
        \includegraphics[width=\textwidth]{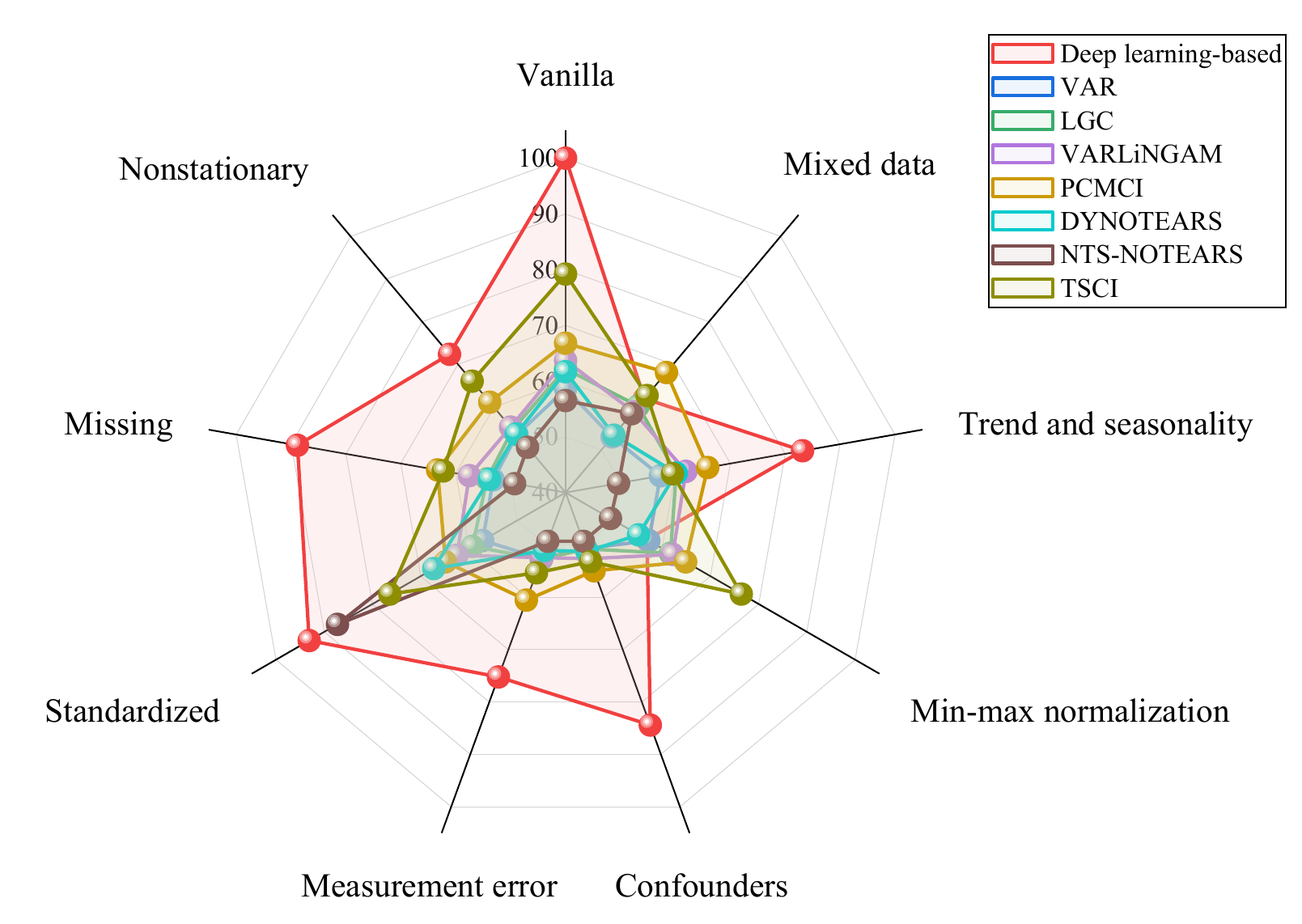}
         \caption{AUROC for nonlinear 10-node case with $T = 500$ and $F=10$. Results aggregated over all hyperparameters.}
         \label{fig:nonlinear_10_500_f10_auroc_avg_hyper}
     \end{subfigure}%
     \hfill
     \begin{subfigure}[b]{0.49\textwidth}
         \centering
         \includegraphics[width=\textwidth]{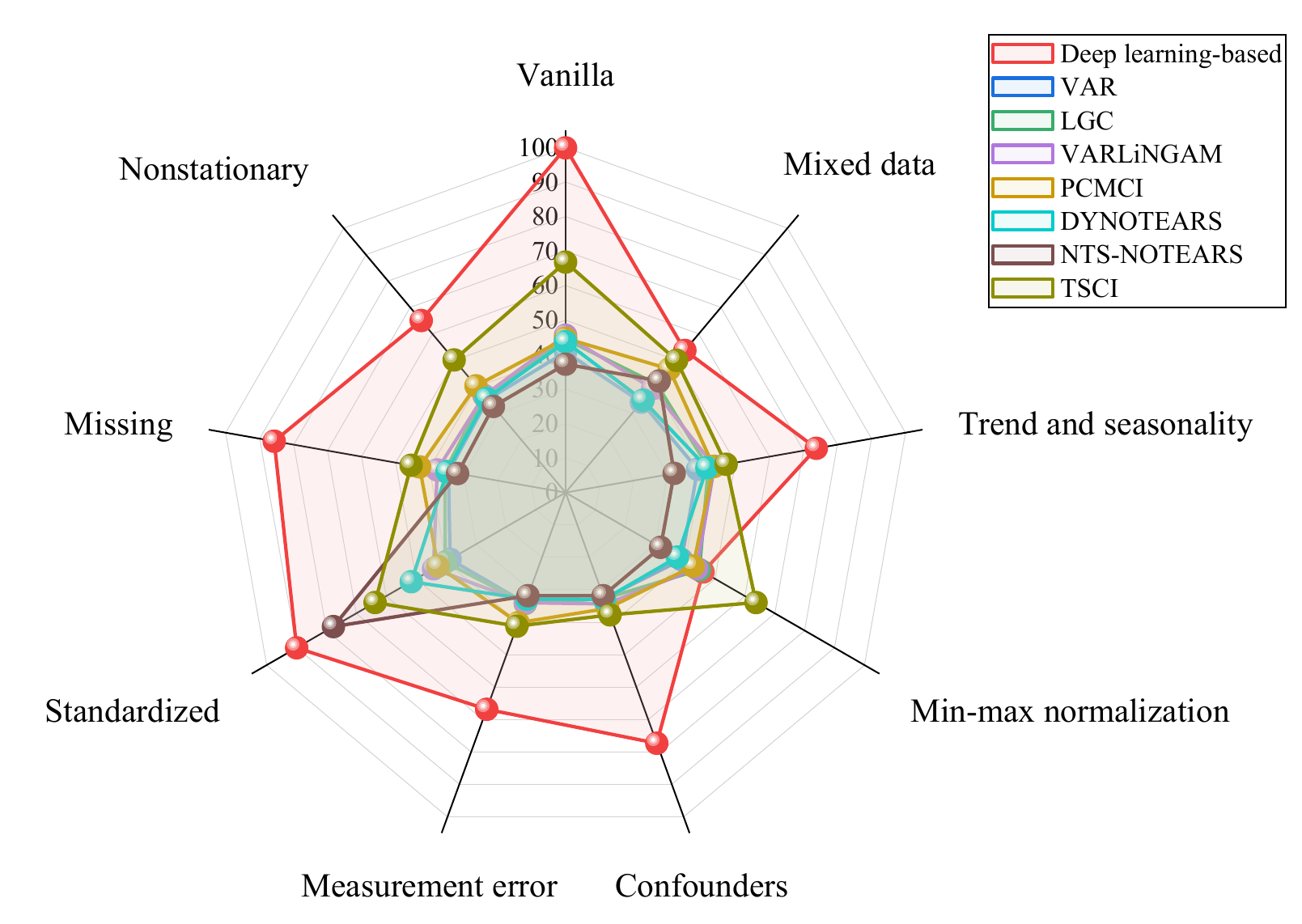}
         \caption{AUPRC for nonlinear 10-node case with $T = 500$ and $F=10$. Results aggregated over all hyperparameters.}
         \label{fig:nonlinear_10_500_f10_auprc_avg_hyper}
     \end{subfigure}
\Description{Four radar charts comparing causal discovery methods on 10-node networks with T=500 across 9 scenarios. Superior performance is predominantly achieved by deep learning-based approaches. Results obtained by aggregating over all hyperparameter configurations.}
\caption{Experimental results under the linear and nonlinear settings across the vanilla scenario and eight assumption violation scenarios. AUROC and AUPRC (the higher the better) are evaluated over 5 trials for the 10-node case with $T = 500$. For the deep learning-based methods, we present only the optimal results. Results aggregated over all hyperparameters.}
\label{fig:experiments_10_500_f10_avg_hyper}
\end{figure*}

\clearpage

\clearpage

\clearpage

\clearpage

\clearpage

\clearpage
\section*{NeurIPS Paper Checklist}

\begin{enumerate}

\item {\bf Claims}
    \item[] Question: Do the main claims made in the abstract and introduction accurately reflect the paper's contributions and scope?
    \item[] Answer: \answerYes{}
    \item[] Justification: The abstract and introduction clearly state our three main contributions (Section~1). The claims are supported by experimental results in Section~\ref{sec:results}.

\item {\bf Limitations}
    \item[] Question: Does the paper discuss the limitations of the work performed by the authors?
    \item[] Answer: \answerYes{}
    \item[] Justification: Section~\ref{sec:limitations} discusses the data-type-dependent learning-rate search ranges and differences in threshold-selection protocols across methods. Appendix~\ref{app_unified_protocol} provides supplementary results and discussion under a unified threshold-search protocol and shared learning-rate ranges. Section~\ref{sec:conclusion} acknowledges that the work focuses on TSCD algorithms and notes that assessing the robustness of event sequence causal discovery in misspecified scenarios is an important future direction. 
    The paper also acknowledges the reliance on synthetic data and specific parameter settings, while noting that the synthetic data configurations were designed to be as representative as possible and to follow common settings in prior work (Section~\ref{sec:data generation} and Appendix~\ref{app_hyper_settings}).

\item {\bf Theory assumptions and proofs}
    \item[] Question: For each theoretical result, does the paper provide the full set of assumptions and a complete (and correct) proof?
    \item[] Answer: \answerNA{}
    \item[] Justification: This paper is an empirical benchmarking study and does not include theoretical results or proofs.

    \item {\bf Experimental result reproducibility}
    \item[] Question: Does the paper fully disclose all the information needed to reproduce the main experimental results of the paper to the extent that it affects the main claims and/or conclusions of the paper (regardless of whether the code and data are provided or not)?
    \item[] Answer: \answerYes{}
    \item[] Justification: The paper provides detailed data generation procedures (Section~\ref{sec:vanilla model}--\ref{sec:data generation}), method descriptions (Section~\ref{sec:methods} and Appendix~\ref{app_methods}), hyperparameter settings (Appendix~\ref{app_hyper_settings}), and evaluation metrics (Section~\ref{sec:evaluation metrics}). The complete code and datasets are publicly available at \url{https://github.com/huiyang-yi/CausalCompass}.

\item {\bf Open access to data and code}
    \item[] Question: Does the paper provide open access to the data and code, with sufficient instructions to faithfully reproduce the main experimental results, as described in supplemental material?
    \item[] Answer: \answerYes{}
    \item[] Justification: The CausalCompass benchmark is released as an open-source Python package with comprehensive documentation at \url{https://github.com/huiyang-yi/CausalCompass}. The repository includes code for data generation, all benchmark algorithms, and usage examples to reproduce the experimental results.

\item {\bf Experimental setting/details}
    \item[] Question: Does the paper specify all the training and test details (e.g., data splits, hyperparameters, how they were chosen, type of optimizer) necessary to understand the results?
    \item[] Answer: \answerYes{}
    \item[] Justification: Section~\ref{sec:experimental} describes the data generation process, number of nodes, time-series lengths, and random seeds. Appendix~\ref{app_hyper_settings} provides detailed hyperparameter search ranges and selection strategies for all eleven methods.

\item {\bf Experiment statistical significance}
    \item[] Question: Does the paper report error bars suitably and correctly defined or other appropriate information about the statistical significance of the experiments?
    \item[] Answer: \answerYes{}
    \item[] Justification: The main experimental results report the mean and standard deviation computed over 5 different random seeds, as stated in Section~\ref{sec:results}.

\item {\bf Experiments compute resources}
    \item[] Question: For each experiment, does the paper provide sufficient information on the computer resources (type of compute workers, memory, time of execution) needed to reproduce the experiments?
    \item[] Answer: \answerYes{}
    \item[] Justification: Appendix~\ref{app_computational_cost} reports the compute resources used in the experiments, including the type of compute workers (two NVIDIA RTX A6000 GPUs and Intel Xeon Gold 6326 CPUs), GPU memory (49 GB per GPU), and system memory (59 GiB).
    
\item {\bf Code of ethics}
    \item[] Question: Does the research conducted in the paper conform, in every respect, with the NeurIPS Code of Ethics \url{https://neurips.cc/public/EthicsGuidelines}?
    \item[] Answer: \answerYes{}
    \item[] Justification: We have reviewed the NeurIPS Code of Ethics and confirm that our research conforms with it. Our work is a benchmarking study using synthetic data and does not involve human subjects and private data.

\item {\bf Broader impacts}
    \item[] Question: Does the paper discuss both potential positive societal impacts and negative societal impacts of the work performed?
    \item[] Answer: \answerYes{}
    \item[] Justification: Sections~\ref{sec:Summary} and~\ref{sec:conclusion} discuss the potential benefits of the benchmark for method evaluation and informed use of TSCD methods. Section~\ref{sec:broader_impacts} discusses potential negative societal impacts, including erroneous decisions and unfair outcomes, together with measures such as domain-expert validation, human oversight, and independent review.
    
\item {\bf Safeguards}
    \item[] Question: Does the paper describe safeguards that have been put in place for responsible release of data or models that have a high risk for misuse (e.g., pre-trained language models, image generators, or scraped datasets)?
    \item[] Answer: \answerNA{}
    \item[] Justification: The released benchmark uses synthetic data generated from known mathematical models. We do not identify high-risk assets requiring special safeguards for release.

\item {\bf Licenses for existing assets}
    \item[] Question: Are the creators or original owners of assets (e.g., code, data, models), used in the paper, properly credited and are the license and terms of use explicitly mentioned and properly respected?
    \item[] Answer: \answerYes{}
    \item[] Justification: All benchmark algorithms used in this work are properly cited with their original publications. The publicly available implementations are referenced with URLs in Appendix~\ref{app_methods} and in the documentation of our repository (\url{https://github.com/huiyang-yi/CausalCompass}).

\item {\bf New assets}
    \item[] Question: Are new assets introduced in the paper well documented and is the documentation provided alongside the assets?
    \item[] Answer: \answerYes{}
    \item[] Justification: CausalCompass is released as a Python package with comprehensive documentation at \url{https://github.com/huiyang-yi/CausalCompass}. The documentation covers installation, usage, data generation, and benchmark execution.

\item {\bf Crowdsourcing and research with human subjects}
    \item[] Question: For crowdsourcing experiments and research with human subjects, does the paper include the full text of instructions given to participants and screenshots, if applicable, as well as details about compensation (if any)? 
    \item[] Answer: \answerNA{}
    \item[] Justification: This paper does not involve crowdsourcing or research with human subjects.

\item {\bf Institutional review board (IRB) approvals or equivalent for research with human subjects}
    \item[] Question: Does the paper describe potential risks incurred by study participants, whether such risks were disclosed to the subjects, and whether Institutional Review Board (IRB) approvals (or an equivalent approval/review based on the requirements of your country or institution) were obtained?
    \item[] Answer: \answerNA{}
    \item[] Justification: This paper does not involve human subjects research.

\item {\bf Declaration of LLM usage}
    \item[] Question: Does the paper describe the usage of LLMs if it is an important, original, or non-standard component of the core methods in this research? Note that if the LLM is used only for writing, editing, or formatting purposes and does \emph{not} impact the core methodology, scientific rigor, or originality of the research, declaration is not required.
    \item[] Answer: \answerNA{}
    \item[] Justification: LLMs are not used as a component of the core methodology in this research.

\end{enumerate}

\end{document}